\documentclass[11pt]{article}

\PassOptionsToPackage{dvipsnames,table}{xcolor}
\usepackage[preprint]{acl}

\usepackage{times}
\usepackage{latexsym}
\usepackage{amsfonts}
\usepackage{amssymb}
\usepackage{xcolor}
\usepackage[T1]{fontenc}

\usepackage[utf8]{inputenc}

\usepackage{microtype}

\DeclareRobustCommand{\blue}[1]{\textcolor{blue}{#1}}

\usepackage{inconsolata}

\ifdefined\SAVEFAST
  \PassOptionsToPackage{draft}{graphicx}
\fi
\usepackage{graphicx}
\usepackage{subcaption}
\usepackage{cuted}
\usepackage{float}
\usepackage{afterpage}
\usepackage{placeins}
\usepackage{pifont}
\usepackage[most]{tcolorbox}
\usepackage{booktabs}
\usepackage{multirow}
\usepackage{colortbl}
\usepackage{tabularx}
\usepackage{makecell}
\usepackage{soul}
\usepackage{paralist}
\usepackage{tikz}
\usetikzlibrary{calc}
\usepackage{amsmath}
\usepackage{xspace}

\definecolor{circleblue}{HTML}{2E75B6}

\newcommand{\stepicon}[1]{%
  \raisebox{-0.15em}{\includegraphics[height=\dimexpr 1.7\fontcharht\font`X\relax]{#1}}%
}
\soulregister\cite7
\soulregister\citep7
\soulregister\citet7
\soulregister\citealt7
\soulregister\citealp7
\soulregister\shortcite7
\soulregister\newcite7
\soulregister\ref7
\soulregister\pageref7
\newcommand{\ours}{RAG-PJ\xspace}
\soulregister\ours7
\newcommand{\res}[2]{#1\%\if\relax\detokenize{#2}\relax\else\,$\pm$\,#2\%\fi}
\newcommand{\ablres}[2]{#1\if\relax\detokenize{#2}\relax\else{\tiny\,${\pm}$#2}\fi}

\newcommand{\numHumanVerifiedTutorialTasks}{110}
\newcommand{\numAIGeneratedTutorialTasks}{250}
\newcommand{\humanVerifiedTutorialPercent}{31\%}
\newcommand{\aiGeneratedTutorialPercent}{69\%}
\definecolor{findingblue}{HTML}{2E75B6}
\newcounter{finding}
\newtcolorbox{findingbox}{
  enhanced,
  breakable,
  colback=findingblue!4,
  colframe=findingblue!75!black,
  boxrule=0pt,
  leftrule=2pt,
  arc=1pt,
  left=6pt,
  right=6pt,
  top=4pt,
  bottom=4pt,
  before skip=0.8\baselineskip,
  after skip=0.8\baselineskip
}

\title{Computer-Use Agents Gain Little Benefit from a World Model When Conditioned on Tutorials}

\author{Anh (Joe) Nguyen \qquad Tianyi Xu\textsuperscript{*} \qquad Phat Tran\textsuperscript{*} \qquad Harry Yu\textsuperscript{*} \\
        {\bf Huazheng Wang \qquad Stefan Lee} \\
        Oregon State University \\
        \texttt{\{nguyejoe,xut2,tranphat,yuhar,huazheng.wang,leestef\}@oregonstate.edu}}

\newif\iffastmode
\ifdefined\FASTMODE
  \fastmodetrue
\else
  \fastmodefalse
\fi

\iffastmode
\fi

\begin{document}
\raggedbottom
\iffastmode\else
  \maketitle
  {\renewcommand{\thefootnote}{*}\footnotetext{Equal contribution.}}
\fi
\iffastmode\include{V}\else\input{V}\fi\begin{abstract}
    Recent work has improved computer-use agents (CUAs) by combining retrieved task-relevant tutorials and look-ahead planning using world models with MLLM judges. 
    However, the contribution of each component---tutorials, the judge, and the world model---to overall improvement has not been evaluated in isolation.
    To fill this gap, we first construct OSWorld-Tutorial and ScienceBoard-Tutorial: two datasets for CUAs with curated multimodal tutorials, covering everyday and scientific domains.
    We then propose RAG-PolicyJudge (\ours), a policy-judge framework for CUAs conditioned on task-specific tutorials: the policy proposes candidate actions, and the judge selects among them.
    To isolate the effect of world models given tutorials, we compare \ours with and without common world models: using either zero-shot text-based descriptions of next states or an oracle that returns ground-truth visual observations.
    We find that \ours gains little benefit from these rollouts when tutorials are present, \textit{even with perfect visual futures}.
    This suggests prior world-model gains in tutorial-conditioned methods come mainly from tutorial understanding and the judge's ability to select the proposed actions, rather than from predicted rollouts.
    Error analysis reveals that the bottleneck is not rollout fidelity, but the judge's ability to reason over future states and recognize task completion.
    This highlights the need to improve not just next-state prediction in world models, but also reasoning over their predicted rollouts.\footnote{\raggedright Datasets and code are available at \href{https://github.com/realjoenguyen/RAGPJ-official}{\texttt{github.com/realjoenguyen/RAGPJ-official}}.}
\end{abstract}

\iffastmode\include{V}\else\input{V}\fi\section{Introduction}
People rely on desktop and web software for day-to-day tasks and scientific workflows. 
Computer-use agents (CUAs) aim to automate these tasks by interacting with computer interfaces given natural-language user instructions.
Modern CUAs are typically built on top of multimodal LLMs (MLLMs) trained on static offline data.
As software is updated and user interfaces shift, these MLLM-based CUAs may lack current procedural knowledge about new environments, resulting in degraded performance~\cite{timewarp}.

A natural approach to address this gap is to provide CUAs with up-to-date task-specific tutorials.
These readily available tutorials, contain environment descriptions and step-by-step instructions in the form of text and screenshots.\footnote{An example tutorial is given in Appendix~\ref{app:tutorial-example}.}

\begin{figure}[t]
    \centering
    \includegraphics[width=\columnwidth]{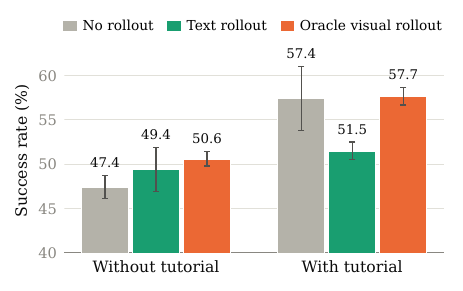}
    \caption{Success rate on OSWorld-Tutorial (Qwen3-VL-32B) of \ours, a policy--judge CUA, whose judge sees no rollout, a predicted text rollout (R-WoM~\cite{r_wom}), or an oracle visual rollout (real future screenshots).
    Rollouts help \ours without a tutorial but add almost nothing with one.
    Full results in \S~\ref{sec:results-findings} and Appendix~\ref{sec:policy-judge-rollout}.}
    \label{fig:tutorial-gap-intro}
\end{figure}

Recent work has integrated tutorial retrieval as a component in CUA systems \cite{os_symphony, rag_gui}. 
However, these systems do not compare against alternative tutorial-use approaches, leaving their effectiveness unclear.
A parallel body of work has considered test-time scaling for CUAs through a policy-judge approach with look-ahead planning \cite{vimo, wma}.
These methods use a world model to predict the outcome of action candidates under a policy, over one or many steps. 
They then use an MLLM-as-a-judge to select the best action among the candidates, conditioned on these predicted rollouts.
Recently, tutorial-augmented methods of this approach have been introduced \cite{r_wom}.
However, these methods conflate the contributions of the tutorial, the judge, and world-model rollouts without proper evaluation.

To fill this gap, we study the effect of tutorial information on CUAs with and without look-ahead planning, while retaining the MLLM-as-a-judge framework.
Intuitively, high-quality tutorials already provide guidance for the policy to generate actions and demonstrations of their effects that can be used by the judge.
This can potentially reduce the need for costly world model training and inference.
With this motivation, we propose a simple yet effective method: RAG-PolicyJudge (\ours), a retrieval-augmented policy-judge framework for CUAs.
\ours provides a task-specific tutorial to both (1) the policy, which proposes action candidates, and (2) the judge, which selects the best action among them.

To evaluate tutorials for CUAs across domains, we build curated tutorials for existing datasets: OSWorld-Verified for day-to-day tasks~\cite{osworld} and ScienceBoard for scientific workflows~\cite{scienceboard}, resulting in two new tutorial benchmarks: OSWorld-Tutorial and ScienceBoard-Tutorial.
The two original datasets are widely adopted for CUAs, facilitating easy adoption and reuse by the research community.

To study the effect of world models on policy-judge frameworks in these two tutorial datasets, using the same base MLLM for both the policy and the judge, we compare \ours with two types of world models for CUAs: text-based and visual. 
First, zero-shot text-based rollouts~\cite{r_wom} (i.e., the text description of what the predicted next state would look like). 
Second, oracle visual rollouts (i.e., the real screenshot of the next state).
We argue that these oracle rollouts are an upper bound on the next-state fidelity a learned visual world model (like ViMo~\cite{vimo}) could achieve.

We find that \ours matches both rollout variants when tutorials are present.
This indicates that current zero-shot MLLM-as-a-judge models gain little extra benefit from a world model, even with oracle visual futures, when conditioned on tutorials (Figure~\ref{fig:tutorial-gap-intro}).
Analyzing failures of the oracle rollout, we find that the agent's failure to \textit{understand task success and progress}, even when the oracle future is available, is a primary bottleneck to higher performance.
This suggests that progress in CUAs may require not just better world models, but also stronger reasoning over the futures these world models simulate. 

In summary, our contributions are as follows:
\begin{compactitem}[\hspace{0pt}--]
    \item We introduce \ours, a simple yet effective tutorial-conditioned policy-judge method for CUAs (\S~\ref{sec:rag-pj}).
    
    \item We build OSWorld-Tutorial and ScienceBoard-Tutorial, two curated tutorial datasets with existing tasks from OSWorld~\cite{osworld} and ScienceBoard~\cite{scienceboard} (\S~\ref{sec:osworld}), facilitating tutorial study for CUAs.
    
    \item We propose an oracle visual rollout infrastructure for CUAs, providing an important ablation on the effect of visual world-model rollouts with perfect future information (\S~\ref{sec:exp-settings}, Appendix~\ref{sec:parallel-gt-rollout}).
    
    \item We provide a controlled comparison of tutorial-only judging, text-based rollouts, and oracle visual rollouts-demonstrating that current MLLM-as-a-judge models gain little benefit from world models given task-specific tutorials (\S~\ref{sec:results-findings}).
\end{compactitem}

\iffastmode\include{V}\else\input{V}\fi

\section{Related Work}
\paragraph{World model evaluation for CUAs.} 
Several works find that zero-shot rollouts from pretrained MLLMs used as world models~\cite{Ha2018-nb} are not yet faithful or consistent proxies for real environments~\cite{world2world, Xu2026-tz, gui_cc}. 
Beyond MLLMs as world models, other works use oracle world models to study upper-bound look-ahead performance in embodied tasks~\cite{Qian2026-gx} and, at the first step only, in computer-use tasks~\cite{agent_s3}.
We extend these analyses to (1) oracle image-based rollouts with task-specific tutorials and (2) look-ahead planning at every step of computer-use tasks.

Concurrent to our work, \citet{dar} show that code-based visual world model add little benefit if agents can already have a well-trained MLLM-as-a-judge. 
Our work complements theirs by studying the benefit of world model rollouts against zero-shot MLLM-as-judge, in the presence of task-specific tutorials. 
Further, we study two types of common world models: we adopt (1) text-based, and build (2) visual ground truths, which is the upper bound of current visual world models.

\paragraph{CUAs that use tutorials (or manuals).}
Prior work uses tutorials either to create training data~\cite{synatra, agenttrek, tongui, ui_evol, w_l,videoagenttrek} or as guidance at inference time~\cite{online_videos_cua, rag_gui, agent_s, r_wom, os_symphony}, where the tutorial guides action generation (policy) or selection (judge).
RAG-GUI~\cite{rag_gui} and Agent S~\cite{agent_s} retrieve text-only tutorials to extract guidance for policy-only agents.\footnote{See Appendix~\ref{sec:rag_gui} for a comparison with RAG-GUI.}
\citet{online_videos_cua} retrieve tutorial videos at inference time as in-context guidance for the policy.
Similar to our benchmarks, there are two recent multimodal tutorial datasets for CUA/GUI: WebOne \citet{webone} and DocOS \citet{docos}, where agents must search the open web for documentation before executing a task. 
However, WebOne is only for web agents and tutorials are not task-conditioned, but rather than software documentation in general. 
DocOS is not publicly available, preventing direct comparison. 
In contrast, our work focuses on task-specific multimodal tutorials for general CUAs and test-time scaling approach with a MLLM-as-a-judge, which is evaluated against world model rollout methods.
We also release our benchmark and trained agents for reproducibility.

Most similar to our approach are OS-Symphony~\cite{os_symphony} and R-WoM~\cite{r_wom}.
OS-Symphony couples retrieved tutorials with a multi-agent history-reflection policy, but does not compare against other tutorial-using baselines. 
Our tutorial-conditioned judge is orthogonal and could be combined with its policy.
Like R-WoM, we focus on settings where multimodal task-specific tutorials are available, but R-WoM's results do not disentangle the tutorial, the MLLM-as-a-judge, and the rollout, and it is evaluated on a non-public subset of ${\sim}85$ OSWorld tasks.
R-WoM reports that its rollouts improve over a no-tutorial baseline, but never compares against a tutorial-conditioned judge without rollout. 
We add that comparison with our method \ours, and find the judge alone matches R-WoM; we further test oracle visual rollouts.
See Appendix~\ref{sec:detailed_related_work} for a more detailed discussion of related work.

\iffastmode\include{V}\else\input{V}\fi\section{Tutorial datasets}
\label{sec:datasets}
\paragraph{Tutorial dataset domains.}
To evaluate tutorials and tutorial-conditioned CUAs across domains, we pair \textit{existing tasks} with newly collected tutorials to build two datasets: (1) OSWorld-Tutorial, from general-purpose desktop tasks in OSWorld-Verified~\cite{osworld}, and (2) ScienceBoard-Tutorial, from scientific-software workflow tasks in ScienceBoard~\cite{scienceboard}.
General-purpose tasks from OSWorld-Verified reflect everyday computer use, with abundant online tutorials.
Scientific software from ScienceBoard is harder on both fronts.
First, its applications are niche: step-by-step task tutorials are scarce, and even ScienceBoard's annotators relied on official documentation (Table~6 of \citet{scienceboard}).
Second, current open MLLMs still struggle on these tasks, with as low as $\sim$30\% success rate with Qwen3.8-27B (Table~\ref{tab:scienceboard}).
This makes it a natural testbed for whether tutorials, even scarce ones, can compensate for limited MLLM familiarity with these specialized domains.
\paragraph{Generating tutorials.}
We pair each task with one multimodal tutorial (text and possibly images) drawn from web sources.
The tutorial type depends on what is available online for that task.

\textit{First, human-verified online tutorials}: we search Google with the task instruction, and when a retrieved tutorial covers the task, one of three annotators experienced with Ubuntu environments selects the most helpful one and verifies it by following its instructions to complete the task; we keep the pair only when this succeeds.
This accounts for \humanVerifiedTutorialPercent{} of OSWorld-Tutorial tasks and none of the ScienceBoard-Tutorial tasks, as the latter's scientific software is niche and lacks task-level tutorials online.
These tutorials rule out retrieval errors, letting us measure the effect of tutorials independently of the retrieval pipeline.

\textit{Second, MLLM-generated tutorials}: when no single online tutorial covers the task (none exists, or the task requires combining several), we launch the task in a virtual machine and prompt Google Gemini 3.8 Flash~\cite{gemini38flash} with the task instruction, the initial screenshot, and the tutorial-generation prompt in Appendix~\ref{app:os-tutorial}; Gemini searches the web and writes a task-specific tutorial, which we keep without human verification.
The screenshot lets it refer to the actual files, columns, and menus of the task instead of inventing names.
MLLM-generated tutorials mirror practical real-world guidance: users rarely have a curated tutorial database and instead ask a search-augmented MLLM for task-specific instructions.
MLLM-generated tutorials cover \aiGeneratedTutorialPercent{} of OSWorld-Tutorial tasks and all ScienceBoard-Tutorial tasks.

\paragraph{Dataset statistics.}
\label{sec:osworld}\label{sec:data-scienceboard}
We create total 529 tutorials across both original datasets.
OSWorld-Tutorial contains 360 existing OSWorld-Verified tasks\footnote{OSWorld-Verified at commit \href{https://github.com/xlang-ai/OSWorld/commit/83e8534451ba8b3ab6477448ef3f0a8e563f05be}{\texttt{83e8534}} (Jun 24, 2026), excluding 9 tasks that require Google Drive credentials or are blocked by website bot detection (Appendix~\ref{app:os-tutorial}).}: \numHumanVerifiedTutorialTasks{} with human-verified and \numAIGeneratedTutorialTasks{} with MLLM-generated tutorials (examples in Appendix~\ref{app:os-tutorial}).
ScienceBoard-Tutorial pairs all 169 existing ScienceBoard tasks with MLLM-generated tutorials (examples in Appendix~\ref{app:sb-tutorial}).

\iffastmode\include{V}\else\input{V}\fi\section{Tutorial-conditioned CUAs}
\begin{figure*}[t]
    \includegraphics[width=\textwidth]{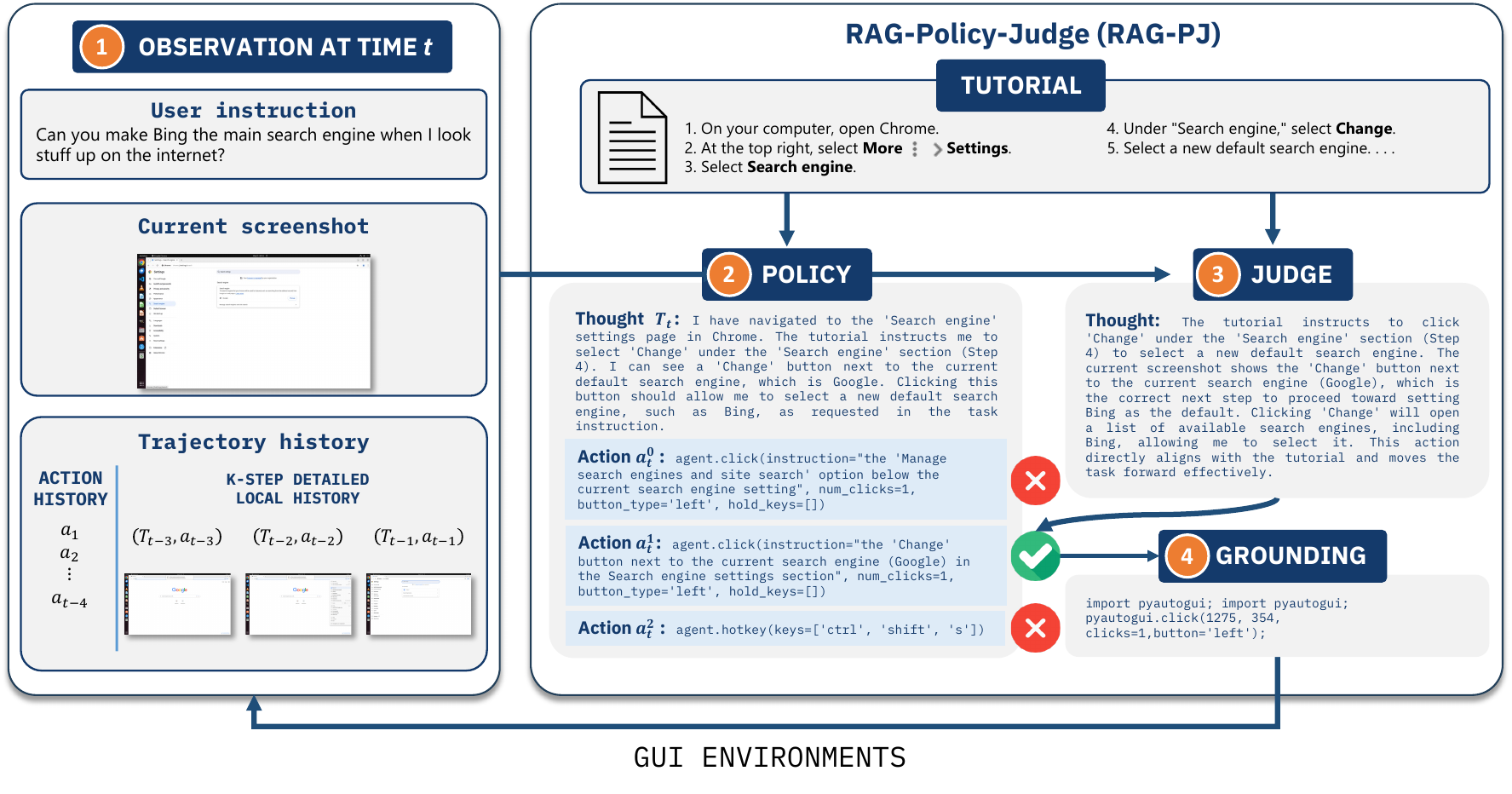}
    \caption{\textbf{Overview of \ours.}
    Given \stepicon{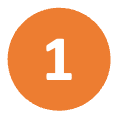} the instruction, current screenshot, and trajectory history, \stepicon{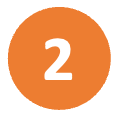} a tutorial-conditioned policy proposes candidate actions,
    \stepicon{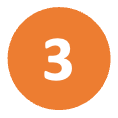} a tutorial-conditioned judge selects one, and \stepicon{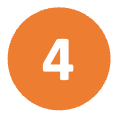} a grounding model maps it to keyboard/mouse inputs. 
    }
    \label{fig:model}
\end{figure*}

\subsection{Preliminaries}
\label{sec:background}

\paragraph{CUA architectures.}
A CUA is a policy $\pi(a_t \,|\, o_{\leq t}, a_{<t}, L, M)$ over screenshots $o$, actions $a$, instruction $L$, and tutorial $M$. 
See Appendix~\ref{sec:pomdp} for the full problem formulation. 
Typically, CUA architectures are built on MLLMs.
Designs are either end-to-end, with one MLLM mapping inputs directly to actions~\cite{ui_tars}, or modular, combining multiple MLLMs and specialized models~\cite{agent_s2,os_symphony}.
We follow the common two-stage modular design.
A policy $\pi(\hat{a}_t \,|\, \cdot)$ outputs a natural-language action (e.g.,~\texttt{click(the red button)}).
A grounding model $g(a_t \,|\, \hat{a}_t, \cdot)$ then maps it to a low-level action (e.g.,~\texttt{click(731,\,244)}).

\paragraph{Judges and world models.}
Other work samples $K$ action candidates $\hat{a}^1_t, \ldots, \hat{a}^K_t \sim \pi(\hat{a}_t \,|\, \cdot)$ and introduces a judge model that selects the index of the best candidate, $k^* = j(\hat{a}^1_t, \ldots, \hat{a}^K_t \,|\, o_{\leq t}, a_{<t}, L, M) \in \{1, \ldots, K\}$, passing only the selected action $\hat{a}^{k^*}_t$ to the grounding model for execution \cite{gta1}.
This sample-then-select paradigm has shown promising results when coupled with look-ahead planning, using a world model to estimate the outcome of each action over a fixed number of time steps.
See Appendix~\ref{sec:related_work_world_models} for an overview of world model methods following this approach for CUAs.
Each candidate is rolled out for $H$ steps by alternating the world model and the policy, and the judge selects among candidates based on their rollouts (details in Appendix~\ref{sec:rollout-formulation}).
However, look-ahead planning greatly increases computational cost by requiring multiple policy and world-model executions per step.

\iffastmode\include{V}\else\input{V}\fi\subsection{RAG-PolicyJudge (\ours)}
\label{sec:rag-pj}
To test if the additional compute for world model is valuable when strong policy guidance is available as a tutorial, we design a simple tutorial-conditioned method, RAG-PolicyJudge (\ours), shown in Figure~\ref{fig:model}.
We adapt a two-stage GUI-agent framework, consisting of a policy and a grounding model as described above (\S~\ref{sec:background}).
Implementation details are in Appendix~\ref{sec:rag-pj-details}.

\paragraph{Policy and Grounding Models.}
We instantiate the policy $\pi(\hat{a}_t \,|\, o_{\leq t}, a_{<t}, L, M)$ as an MLLM that outputs $K$ candidate actions in a single call~\cite{gta1}.
To bound context length, in practice we condition $\pi$ on $\hat{o}_{\leq t}$ in place of $o_{\leq t}$, where $\hat{o}_{\leq t}$ consists of: the current observation $o_t$; all previous actions; the chain-of-thought of the five most recent steps; and the screenshots of up to five recent steps.
Additionally, the task instruction, CUA-specific policy rules, and the tutorial are always available at each time step.
We adopt the action space from Agent-S2~\cite{agent_s2}.
The policy prompt is available in Appendix~\ref{sec:policy-prompt}.
We adopt UI-Venus-1.5~\cite{ui_venus} as it is the state-of-the-art grounding model for CUAs and has been used as the grounding model in prior work~\cite{Gong2026-kk, Jiang2025-ob}.
The grounding prompt is provided in Appendix~\ref{sec:grounding-prompt}. 
The grounding model $g(a_t \,|\, \hat{a}_t, o_t)$ accesses only the current observation $o_t$ and the policy action $\hat{a}_t$.

\paragraph{Judge Model.}
The judge $j$ shares the policy's MLLM and inputs: $K$ candidate actions, task instruction, the trajectory encoding $\hat{o}_{\leq t}$, and tutorials.
It ranks the candidates with a ranking prompt (Appendix~\ref{sec:judge-prompt}) and sends the top-ranked action to the grounding model for execution.

\iffastmode\include{V}\else\input{V}\fi\afterpage{

\begin{table*}[t]
    \centering
    \small
    \setlength{\tabcolsep}{4pt}
    \renewcommand{\arraystretch}{1.15}
    \renewcommand{\res}[2]{#1\if\relax\detokenize{#2}\relax\else{\scriptsize\,${\pm}$#2}\fi}
    \definecolor{tie}{RGB}{222,235,247}\newcommand{\tie}[1]{\cellcolor{tie}#1}
    
    \begin{tabular}{clcccccc}
    \toprule
    &\multicolumn{1}{c}{\multirow{2}{*}{\textbf{Method}}} & \multicolumn{2}{c}{\textbf{Qwen3-VL-8B}} & \multicolumn{2}{c}{\textbf{Qwen3-VL-32B}} & \multicolumn{2}{c}{\textbf{Qwen3.8-27B}} \\
    \cmidrule(lr){3-4} \cmidrule(lr){5-6} \cmidrule(lr){7-8}
    \#&& \textbf{15 steps} & \textbf{50 steps} & \textbf{15 steps} & \textbf{50 steps} & \textbf{15 steps} & \textbf{50 steps} \\
    \midrule
    \multicolumn{8}{>{\columncolor{gray!15}}l}{\textit{Policy only}} \\
    {\tiny\texttt{1}}&\quad Policy Only & \res{29.0}{0.9} & \res{34.5}{1.1} & \res{42.5}{0.7} & \res{44.9}{2.0} & \res{57.4}{2.6} & \res{65.7}{0.3} \\
    {\tiny\texttt{2}}&\quad \quad+ RAG & \res{40.6}{2.7} & \res{45.6}{2.1} & \res{47.8}{0.7} & \res{51.7}{2.4} & \res{60.7}{0.5} & \res{68.2}{1.1} \\
    \midrule
    \multicolumn{8}{>{\columncolor{gray!15}}l}{\textit{Policy + Judge}} \\
    {\tiny\texttt{3}}&\quad Policy + Judge (GTA-1) & \res{29.9}{0.2} & \res{32.5}{2.3} & \res{43.8}{0.1} & \res{47.4}{1.3} & \res{56.8}{0.8} & \res{65.1}{0.7} \\
    {\tiny\texttt{3a}}&\quad Policy + RAG Judge & \res{32.4}{0.3} & \res{39.5}{1.9} & \res{45.2}{1.7} & \res{47.9}{1.2} & \res{60.1}{1.4} & \res{67.1}{0.8} \\
    {\tiny\texttt{3b}}&\quad RAG Policy + Judge & \res{40.5}{3.1} & \res{45.7}{1.9} & \res{50.2}{1.4} & \res{55.4}{0.1} & \res{61.3}{1.0} & \res{70.9}{0.8} \\
    {\tiny\texttt{4}}&\quad RAG-PJ (Ours) & \tie{\res{42.2}{0.4}} & \tie{\res{46.8}{1.9}} & \tie{\res{51.7}{1.4}} & \tie{\res{57.4}{3.6}} & \tie{\res{61.7}{0.5}} & \tie{\res{71.0}{1.2}} \\
    \midrule
    \multicolumn{8}{>{\columncolor{gray!15}}l}{\textit{Policy + RAG Judge + Text-based rollout (R-WoM \citet{r_wom})}} \\
    {\tiny\texttt{5}}&\quad R-WoM $H{=}1$ & \res{30.8}{1.3} & \res{32.6}{1.6} & \res{36.4}{1.0} & \res{45.8}{2.1} & \res{56.9}{1.8} & \res{66.5}{1.0} \\
    {\tiny\texttt{6}}&\quad R-WoM $H{=}2$ & \res{30.5}{1.5} & \res{33.5}{2.2} & \res{40.9}{0.9} & \res{48.0}{1.9} & \res{58.2}{0.4} & \res{65.5}{2.7} \\
    {\tiny\texttt{7}}&\quad R-WoM $H{=}3$ & \res{31.8}{0.6} & \res{36.9}{1.8} & \res{42.1}{0.8} & \res{46.2}{0.2} & \res{55.6}{1.5} & \res{66.2}{0.6} \\
    \addlinespace[2pt]
    {\tiny\texttt{8}}&\quad +Policy RAG $H{=}1$ & \res{34.8}{0.9} & \res{40.2}{0.8} & \res{47.8}{0.5} & \res{54.0}{0.1} & \tie{\res{63.0}{1.3}} & \tie{\res{71.7}{2.7}} \\
    {\tiny\texttt{9}}&\quad +Policy RAG $H{=}2$ & \res{38.6}{0.6} & \res{39.4}{0.9} & \res{50.1}{0.7} & \res{51.5}{1.0} & \tie{\res{59.7}{0.3}} & \tie{\res{70.2}{1.4}} \\
    {\tiny\texttt{10}}&\quad +Policy RAG $H{=}3$ & \res{36.3}{1.8} & \res{39.3}{1.8} & \res{47.2}{0.2} & \res{52.3}{0.6} & \tie{\res{60.2}{0.8}} & \tie{\res{71.4}{0.5}} \\
    \midrule
    \multicolumn{8}{>{\columncolor{gray!15}}l}{\textit{RAG Policy + RAG Judge + Ground-truth visual rollout}} \\
    {\tiny\texttt{11}}&\quad $H{=}1$ & \tie{\res{40.7}{0.8}} & \tie{\res{43.1}{4.0}} & \tie{\res{49.4}{0.2}} & \tie{\res{57.5}{1.8}} & \tie{\res{62.8}{0.9}} & \tie{\res{70.7}{1.0}} \\
    {\tiny\texttt{12}}&\quad $H{=}2$ & \tie{\res{43.1}{1.0}} & \tie{\res{45.6}{1.3}} & \tie{\res{51.6}{0.9}} & \tie{\res{57.7}{1.0}} & \tie{\res{63.8}{0.7}} & \tie{\res{70.1}{0.4}} \\
    {\tiny\texttt{13}}&\quad $H{=}3$ & \tie{\res{41.4}{1.5}} & \tie{\res{47.5}{2.5}} & \tie{\res{50.5}{1.1}} & \tie{\res{58.0}{0.7}} & \tie{\res{63.6}{1.4}} & \tie{\res{72.3}{0.6}} \\
    \bottomrule
    \end{tabular}
    \caption[Success rate of RAG-PJ (Ours) and baselines on 360 OSWorld-Tutorial tasks.]{Success rate (\%) of RAG-PJ (Ours) and baselines on 360 OSWorld-Tutorial tasks, at 15 and 50 steps. Mean $\pm$ sample std. over 3 completed runs. ``RAG'' indicates tutorial use. {\setlength{\fboxsep}{1pt}\colorbox{tie}{Shaded cells}} are not significantly different from RAG-PJ (row 4) in the same column. See more details in Appendix~\ref{sec:res_per_category_tutorial}.}
    \label{tab:res}
    \end{table*}

}
\section{Experimental Settings}
\label{sec:exp-settings}

\paragraph{Evaluation metric: task success rate.}
We use the task success rate metrics defined by OSWorld-Verified and ScienceBoard.
For OSWorld-Verified, the metrics are computed within the original maximum of \textit{15 steps}~\cite{osworld, os_harm, mobile_agent_v3, opencua} and the extended \textit{50-step} budget adopted in follow-up work~\cite{osworld_human, OpenAI2025-operator, osworld_mcp}.
For ScienceBoard, we follow each task's time limit from the original benchmark~\cite{scienceboard}.

\paragraph{Baselines.}
We use the following baselines with the same two-stage policy-grounding framework~\cite{Gou2025-ga}, the same harness (rules/guidelines), and the same input/output space as \ours, allowing us to isolate the effects of the world model, the judge, and the tutorial.
We adopt the hyperparameters from their best publicly reported configurations. 
Details are in Appendix~\ref{sec:rag-pj-details}.
In all settings with a judge, the policy proposes $K=3$ candidate actions per step\footnote{$K=3$ follows the original codebase of our baselines.} and the judge selects one; in \texttt{Policy Only} settings, the policy proposes a single action.
\begin{compactitem}[\hspace{3pt}--]
  \item \texttt{Policy Only} drops the judge from \ours. 
  When including the tutorial, we denote this model as \texttt{Policy Only + RAG}.
  \item \texttt{Policy + Judge (GTA-1)}~\cite{gta1} is a policy-judge framework, without a world model or tutorial.
  \item \texttt{R-WoM H=\#} is a re-implementation of the recent tutorial-augmented look-ahead method from \citet{r_wom} described in \S~\ref{sec:background}. 
  R-WoM uses an MLLM-based world model to produce zero-shot text-based next-state descriptions.
  We set $H{=}1,2,3$ to adjust how many steps are rolled out during look-ahead search.
  Our re-implementation details are in Appendix~\ref{sec:rwom}.\footnote{Without tutorials, our re-implementation reproduces R-WoM's reported gain over a judge-only baseline, as shown in Table~\ref{tab:res-32b-no-tutorial}; we use this as a sanity check that the baseline is functional before adding tutorials.}
  \item \texttt{WebDreamer H=\#}~\cite{webdreamer} is a tutorial-free, text-based world model look-ahead method originally proposed for web agents, which we adapt to CUAs.
  Similar to R-WoM, we use look-ahead horizons $H{=}1,2$.
  Our re-implementation details are in Appendix~\ref{sec:webdreamer-prompt}.

  \item \texttt{(Visual) GT-rollout H=\#}: an oracle experiment in which we use ground-truth screenshots as future rollouts.%
\footnote{The common usage of the term ``world model'' is to describe a predictor of the next state given a current state and action.
        This is separate from planning that might use the world model. In this sense, the visual rollout is extremely accurate except for the stochastic elements described in Appendix~\ref{sec:non_deter}, which are fundamentally unpredictable. We do not imply the system as a whole is the best possible planning system.} 
  For each action candidate at a given time step, we save and reset the state of the virtual machine hosting the environment before collecting each rollout.
  Stochastic mismatches between \texttt{GT-rollout} and live states are rare ($<$1.0\% of completed-task states).
  These mismatches are mostly distracting webpage changes (color, ads), not inherent environment changes that would require a different policy (Appendix~\ref{sec:parallel-gt-rollout}).
  Similar to \texttt{R-WoM H=\#}, we use $H{=}1,2,3$ look-ahead steps.
  Prior world-model work~\cite{wma, webdreamer, vimo, code2world, mobiledreamer, r_wom} predicts only one step ahead (rarely two); we evaluate up to three.
  Details are in Appendix~\ref{sec:parallel-gt-rollout}.
\end{compactitem}

\paragraph{Base MLLMs.}
We use Qwen3-VL-8B, Qwen3-VL-32B, and Qwen3.8-27B~\cite{qwen3vl, qwen38}\footnote{Qwen3-VL-8B-Instruct, Qwen3-VL-32B-Instruct-FP8, and Qwen3.8-27B-FP8 from the official Hugging Face versions of Qwen.} as the base MLLMs for the policy, judge, and world model in both our method and the baselines.
These models are commonly used in recent CUA works~\cite{evocua, step_gui, webgym, rlanything}. 

\iffastmode\include{V}\else\input{V}\fi\section{Results and findings}
\label{sec:results-findings}
Given a task and \textit{a corresponding task-specific tutorial} (either a human-verified online tutorial or an MLLM-generated tutorial, which can be noisy), we want to answer the following questions:

\begin{table}[t]
\centering
\footnotesize
\setlength{\tabcolsep}{2pt}
\renewcommand{\arraystretch}{1.05}
\renewcommand{\res}[2]{#1\if\relax\detokenize{#2}\relax\else{\scriptsize\,${\pm}$#2}\fi}
\definecolor{tie}{RGB}{222,235,247}\newcommand{\tie}[1]{\cellcolor{tie}#1}

\begin{tabularx}{\columnwidth}{cXcc}
\toprule
\#&\multicolumn{1}{c}{\textbf{Method}} & \shortstack{\textbf{Qwen3-VL}\\\textbf{32B}} & \shortstack{\textbf{Qwen3.8}\\\textbf{27B}} \\
\midrule
\multicolumn{4}{>{\columncolor{gray!15}}l}{\textit{Policy only}} \\
{\tiny\texttt{1}}&\quad Policy Only & \res{14.8}{1.0} & \res{29.0}{0.6} \\
{\tiny\texttt{2}}&\quad \quad+ RAG & \res{36.1}{0.6} & \res{44.2}{1.4} \\
\midrule
\multicolumn{4}{>{\columncolor{gray!15}}l}{\textit{Policy + Judge}} \\
{\tiny\texttt{3}}&\quad Policy + Judge (GTA-1) & \res{18.3}{2.4} & \res{30.4}{1.2} \\
{\tiny\texttt{4}}&\quad RAG-PJ (Ours) & \tie{\res{40.2}{0.6}} & \tie{\res{48.3}{0.7}} \\
\midrule
\multicolumn{4}{>{\columncolor{gray!15}}l}{\textit{Policy + RAG Judge + Text rollout (R-WoM)}} \\
{\tiny\texttt{5}}&\quad R-WoM $H{=}1$ & \res{18.5}{0.9} & \res{30.6}{2.4} \\
{\tiny\texttt{6}}&\quad R-WoM $H{=}2$ & \res{20.3}{2.5} & \res{32.7}{1.2} \\
{\tiny\texttt{7}}&\quad R-WoM $H{=}3$ & \res{20.5}{0.7} & \res{30.6}{0.7} \\
\addlinespace[2pt]
{\tiny\texttt{8}}&\quad +Policy RAG $H{=}1$ & \res{35.3}{0.9} & \res{42.6}{1.6} \\
{\tiny\texttt{9}}&\quad +Policy RAG $H{=}2$ & \res{37.5}{1.4} & \res{43.2}{0.6} \\
{\tiny\texttt{10}}&\quad +Policy RAG $H{=}3$ & \res{33.7}{2.6} & \res{44.8}{0.3} \\
\midrule
\multicolumn{4}{>{\columncolor{gray!15}}l}{\textit{RAG Policy + RAG Judge + Ground-truth visual rollout}} \\
{\tiny\texttt{11}}&\quad $H{=}1$ & \tie{\res{39.1}{2.0}} & \tie{\res{49.1}{0.6}} \\
{\tiny\texttt{12}}&\quad $H{=}2$ & \tie{\res{40.4}{0.7}} & \tie{\res{49.3}{0.3}} \\
{\tiny\texttt{13}}&\quad $H{=}3$ & \tie{\res{42.4}{1.9}} & \tie{\res{49.9}{1.2}} \\
\bottomrule
\end{tabularx}

\stepcounter{footnote}\xdef\sbsigfn{\the\value{footnote}}
\caption{Success rate (\%) on ScienceBoard-Tutorial.
Mean $\pm$ sample std. over 3 completed runs. ``RAG'' indicates tutorial use. {\setlength{\fboxsep}{1pt}\colorbox{tie}{Shaded cells}} (ground-truth rollouts, \texttt{rows 11--13}) are not significantly different from RAG-PJ (\texttt{row 4}).\protect\footnotemark[\sbsigfn]}
\label{tab:scienceboard}
\end{table}

\iffastmode\include{V}\else\input{V}\fi\subsection{Can tutorials improve CUAs?}
\label{sec:rag_judge}
\vspace{5pt}\noindent\emph{\textbf{Finding 1: Tutorials improve task success rate over non-tutorial baselines.}}
In Table~\ref{tab:res} (OSWorld-Tutorial) and Table~\ref{tab:scienceboard} (ScienceBoard-Tutorial), we show that including tutorials in the policy (\texttt{row 2}) improves over the non-tutorial baseline (\texttt{Policy Only}, \texttt{row 1}) by $\sim$3--12\% (Table~\ref{tab:res}) and $\sim$15--21\% (Table~\ref{tab:scienceboard}).
The policy-judge framework with tutorials (\ours, \texttt{row 4}) improves over the non-tutorial baseline (\texttt{GTA-1}, \texttt{row 3}) by $\sim$5--14\% (Table~\ref{tab:res}) and $\sim$18--22\% (Table~\ref{tab:scienceboard}).
These improvements are larger on ScienceBoard (Table~\ref{tab:scienceboard}), whose specialized scientific apps are likely less covered in MLLM training data.
This suggests tutorials help most where MLLMs lack prior knowledge.

\footnotetext[\sbsigfn]{Hierarchical bootstrap, sign-flip permutation, and Wilcoxon signed-rank tests over tasks; $p$-values in Appendix~\ref{sec:sb-sig-tests}.}

Table~\ref{tab:res-per-category} (Appendix~\ref{sec:res_per_category_tutorial}) and Tables~\ref{tab:res-scienceboard-per-app-32b} and~\ref{tab:res-scienceboard-per-app-27b} (Appendix~\ref{sec:res_per_category_scienceboard}) break results down per category in both benchmarks; the improvement from tutorial conditioning is consistent across all categories.
Notably, tutorial conditioning in the policy yields most of the gains, suggesting that guidance for producing action candidates is the main source of improvement.
Yet policy conditioning alone (\texttt{row 2}, Tables~\ref{tab:res} and~\ref{tab:scienceboard}) does not account for all of \ours's gains, so judge action selection still helps.
Finally, \texttt{rows 3a, 3b} vs.~\texttt{row 4} (Table~\ref{tab:res}) show that using tutorials in both policy and judge performs best.

\iffastmode\include{V}\else\input{V}\fi\subsection{Do CUAs benefit from zero-shot text-based world models?}
\label{sec:text_wm}

\vspace{5pt}\noindent\emph{\textbf{Finding 2: Given a tutorial, CUAs gain no benefit from zero-shot text-based world models.}}
As shown in Tables~\ref{tab:res} (OSWorld-Tutorial) and~\ref{tab:scienceboard} (ScienceBoard-Tutorial) and their breakdowns in Appendix~\ref{sec:res_per_category_tutorial}, we find that \texttt{R-WoM} does not improve over a judge without world model rollout (\texttt{\ours}).
R-WoM without policy-side tutorial conditioning (\texttt{rows 5--7}, Table~\ref{tab:res}) matches or underperforms the \texttt{Policy + RAG Judge} without rollout (\texttt{row 3a}, Table~\ref{tab:res}) at different horizons $(H{=}1,2,3)$.
Appendix~\ref{sec:rwom-error-analysis} traces this to hallucination in rollouts.

When policy-side RAG is added (\texttt{rows 8--10}, Tables~\ref{tab:res} and~\ref{tab:scienceboard}), R-WoM is worse than or statistically similar to \ours (\texttt{row 4}) on both benchmarks.
This indicates that gains from \texttt{R-WoM} come from tutorial conditioning and the judge, rather than text-based rollouts.
Further, we observe no consistent trend across rollout horizons: longer horizons ($H{=}2,3$) do not always improve over $H{=}1$, and in some settings longer horizons slightly degrade performance.
This suggests compounding errors across simulated steps add noise to the judge.

\iffastmode\include{V}\else\input{V}\fi\subsection{Do CUAs benefit from ground-truth visual rollouts?}
\label{sec:visual_wm}
Motivated by the observation that text-based rollouts failed to improve over \ours, we consider an oracle visual rollout (\texttt{GT-rollout}) to establish an upper bound on world model performance.

\begin{figure*}[t!]
    \centering
    \includegraphics[width=\textwidth]{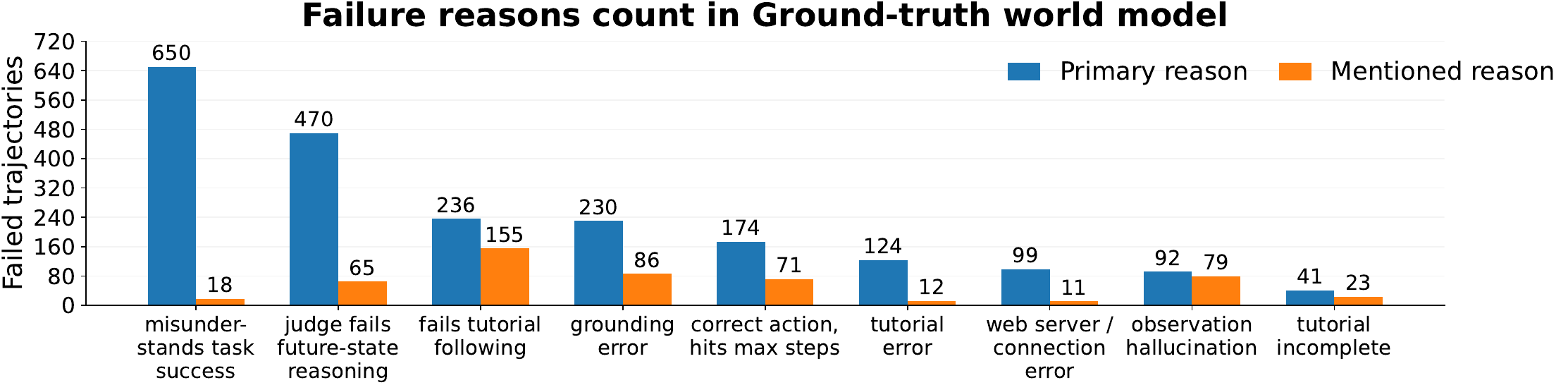}
    \caption{Error analysis of 2,116 failed trajectories with ground-truth visual rollouts in OSWorld-Tutorial.
        Given a failed task, we use Gemini 3.8 Flash to identify a primary failure reason and additional secondary reasons.
        We categorize the failure reasons into 9 categories and show their frequency.
        For a detailed breakdown, see Appendix~\ref{sec:gt-error}.
    }

    \label{fig:gt-rollout-error-analysis}
\end{figure*}
\begin{table}[t]
\centering
\footnotesize
\setlength{\tabcolsep}{3pt}
\renewcommand{\arraystretch}{1.05}
\renewcommand{\res}[2]{#1\if\relax\detokenize{#2}\relax\else{\tiny\,${\pm}$#2}\fi}
\newcommand{\dres}[3]{\res{#2}{#3} & {\tiny\ifdim#1pt<0pt\color{red!70!black}\else\color{green!50!black}\fi($#1$)}}

\begin{tabular}{clc@{\,}lc@{\,}l}
\toprule
\#&\multicolumn{1}{c}{\textbf{Method (no tutorial)}} & \multicolumn{2}{c}{\textbf{15 steps}} & \multicolumn{2}{c}{\textbf{50 steps}} \\
\midrule
{\tiny\texttt{1}}&Policy + Judge (GTA-1) & \res{43.8}{0.1} & & \res{47.4}{1.3} & \\
\midrule
\multicolumn{6}{>{\columncolor{gray!15}}l}{\textit{+ Text rollout, WebDreamer \citep{webdreamer}}} \\
{\tiny\texttt{2}}&\quad $H{=}1$ & \dres{-2.4}{41.4}{1.4} & \dres{-0.4}{47.0}{2.0} \\
{\tiny\texttt{3}}&\quad $H{=}2$ & \dres{-3.7}{40.1}{1.0} & \dres{+0.1}{47.5}{1.1} \\
\midrule
\multicolumn{6}{>{\columncolor{gray!15}}l}{\textit{+ Text rollout, R-WoM \citep{r_wom}}} \\
{\tiny\texttt{4}}&\quad $H{=}1$ & \dres{+0.4}{44.2}{1.2} & \dres{+0.8}{48.2}{3.5} \\
{\tiny\texttt{5}}&\quad $H{=}2$ & \dres{+1.1}{44.9}{0.6} & \dres{+2.0}{49.4}{2.5} \\
\midrule
\multicolumn{6}{>{\columncolor{gray!15}}l}{\textit{+ Ground-truth visual rollout}} \\
{\tiny\texttt{6}}&\quad $H{=}1$ & \dres{+0.9}{44.7}{1.9} & \dres{-0.6}{46.8}{1.4} \\
{\tiny\texttt{7}}&\quad $H{=}2$ & \dres{+4.2}{48.0}{1.1} & \dres{+3.2}{50.6}{0.8} \\
\bottomrule
\end{tabular}
\vspace{-2pt}
\caption{Success rate (\%) on OSWorld-Tutorial \textit{without tutorials} (Qwen3-VL-32B policy and judge). Mean $\pm$ sample std.\ over 3 completed runs. Parentheses: change (pp) over Policy + Judge (\texttt{row 1}).}
\label{tab:res-32b-no-tutorial}
\end{table}

{
\vspace{5pt}\noindent\emph{\textbf{Finding 3: Given oracle or noisy MLLM-generated tutorials, zero-shot MLLM judges gain little from ground-truth visual rollouts.}}
In Table \ref{tab:res} (OSWorld-Tutorial) and Table \ref{tab:scienceboard} (ScienceBoard-Tutorial), with tutorials, \texttt{GT-rollout} clearly improves over text-based \texttt{R-WoM} (\texttt{rows 11--13} vs.~\texttt{8--10}) but is not significantly better than \ours on either benchmark, for both human-verified and MLLM-generated tutorials (Appendix~\ref{sec:res_per_category_tutorial}).
As before, the judge can already decide well from tutorials without reasoning over future consequences.
Without tutorials, \texttt{GT-rollout} does help: $H{=}2$ improves over \texttt{Policy+Judge} by $\sim$3--4\% (Table~\ref{tab:res-32b-no-tutorial}, \texttt{row 7} vs.~\texttt{row 1}).

With noisy MLLM-generated tutorials, \texttt{GT-rollout} gives small nominal gains over \ours in some settings (Tables~\ref{tab:res-ai-vs-human-tutorial-same-task} and~\ref{tab:scienceboard}).
In 40 sampled tasks that \ours fails but \texttt{GT-rollout} solves, the tutorials are partially correct but miss steps.
\texttt{GT-rollout} helps the judge recover when:
\begin{compactitem}[\hspace{3pt}--]
    \item the horizon is too short for \ours to try the alternative steps the tutorial suggests, e.g., ScienceBoard categories with only 5 steps;
    \item the action's effect is visually obvious in the next state, e.g., opening a website or creating a file. This matters more in ScienceBoard-Tutorial, whose scientific software is less familiar to the judge, so it is harder to infer the effect from the tutorial text alone.
\end{compactitem}
Per-category results are in Appendices~\ref{sec:res_per_category_tutorial} and~\ref{sec:res_per_category_scienceboard}.
\par}

\vspace{5pt}\noindent\emph{\textbf{Finding 4: Agents fail to use ground-truth future information due to poor understanding of task success and progress.}}
To understand why ground-truth visual rollouts do not improve success, we analyze failed tasks in OSWorld-Tutorial with the help of Gemini 3.8 Flash~\cite{gemini38flash} and summarize failure reasons and their frequency in Figure~\ref{fig:gt-rollout-error-analysis}.

The most frequent failure category is the agent (both the policy and the judge) failing to understand \textit{task success: whether the current state satisfies the task instruction}.
For instance, a task is not completed in the current state, yet the policy and the judge predict that the current state already satisfies the instruction.
The judge therefore selects the action that ends the task prematurely.

The second most frequently tagged category is the judge failing to reason over the future information about \textit{task progress: what has been done and what is left to do}.
Even if the policy generates good action candidates, the judge fails to select the one that leads to task progress.
The judge often repeats the same failed action instead of recognizing the failure and choosing other candidates.

We show examples of these failures in Figures~\ref{fig:fail-future-reasoning}, \ref{fig:fail-future-reasoning2}, and~\ref{fig:fail-task-success}.
Overall, most failures stem from the judge's inability to \textit{reason over future states generated by world models}, not from world-model quality.
Examples of other less significant failure reasons and the error-analysis implementation are shown in Appendix~\ref{sec:gt-error}.

\vspace{5pt}\noindent\emph{\textbf{Implication: CUAs need not only better world models but also better mechanisms for reasoning over future rollouts.}}
Improving world-model fidelity alone may therefore be insufficient for look-ahead planning.
Current world-model methods for CUAs lack \textit{reward prediction} for rollout states, which the judge needs to compare candidates by their futures, and judges need better ways to reason over rollouts, e.g., via prompting or training.
This is consistent with~\citet{Qian2026-gx}, who find that even a ground-truth world model does not improve embodied agents, as using future information is the main bottleneck.

\iffastmode\include{V}\else\input{V}\fi\subsection{Can current MLLM systems synthesize tutorials for CUAs \textit{as good as} human-verified online tutorials?}
\label{sec:human_vs_ai_tutorial}
\begin{table}[t]
\centering
\small
\setlength{\tabcolsep}{3pt}
\renewcommand{\arraystretch}{1.15}
\renewcommand{\res}[2]{#1\if\relax\detokenize{#2}\relax\else{\scriptsize\,${\pm}$#2}\fi}

\resizebox{\columnwidth}{!}{%
\begin{tabular}{clcccc}
\toprule
&\multicolumn{1}{c}{\multirow{2}{*}{\textbf{Method}}} & \multicolumn{2}{c}{\textbf{15 steps}} & \multicolumn{2}{c}{\textbf{50 steps}} \\
\cmidrule(lr){3-4} \cmidrule(lr){5-6}
\#&& \textbf{Human} & \textbf{MLLM} & \textbf{Human} & \textbf{MLLM} \\
\midrule
\multicolumn{6}{>{\columncolor{gray!15}}l}{\textit{Policy only}} \\
{\tiny\texttt{1}}&\quad Policy Only & \multicolumn{2}{c}{\res{60.2}{3.4}} & \multicolumn{2}{c}{\res{63.8}{1.4}} \\
{\tiny\texttt{2}}&\quad \quad+ RAG & \res{66.6}{1.4} & \res{70.4}{1.5} & \res{68.7}{2.3} & \res{74.7}{1.9} \\
\midrule
\multicolumn{6}{>{\columncolor{gray!15}}l}{\textit{Policy + Judge}} \\
{\tiny\texttt{3}}&\quad Policy + Judge (GTA-1) & \multicolumn{2}{c}{\res{61.7}{0.9}} & \multicolumn{2}{c}{\res{69.6}{3.7}} \\
{\tiny\texttt{4}}&\quad RAG-PJ (Ours) & \res{72.6}{2.7} & \res{71.4}{1.1} & \res{74.5}{3.3} & \res{75.0}{4.1} \\
\midrule
\multicolumn{6}{>{\columncolor{gray!15}}l}{\textit{RAG-PJ + Ground-truth visual rollout}} \\
{\tiny\texttt{5}}&\quad $H{=}1$ & \res{66.3}{0.9} & \res{68.9}{0.9} & \res{72.6}{3.6} & \res{71.4}{6.0} \\
{\tiny\texttt{6}}&\quad $H{=}2$ & \res{72.3}{1.4} & \res{70.1}{4.1} & \res{74.2}{1.0} & \res{76.7}{0.6} \\
{\tiny\texttt{7}}&\quad $H{=}3$ & \res{70.4}{2.9} & \res{72.2}{4.2} & \res{75.3}{1.3} & \res{75.5}{3.7} \\
\bottomrule
\end{tabular}
}

\caption{Success rate (\%) on 110 OSWorld-Tutorial tasks with human-verified (Human) vs.~MLLM-generated (MLLM) tutorials (Qwen3-VL-32B policy/judge, 15/50 steps). Tutorial-free baselines shown once. Mean $\pm$ sample std. over 3 completed runs.}
\label{tab:res-ai-vs-human-tutorial-same-task}
\end{table}

Without an available online tutorial for a given task, can MLLM systems with web search synthesize equally useful ones?
Since Tables~\ref{tab:res-human-written} and~\ref{tab:res-ai-generated} (Appendix~\ref{sec:res_per_category_tutorial}) suggest that tasks with MLLM-generated tutorials may be harder, we use OSWorld-Tutorial to compare both tutorial types \textit{on the same tasks}.
Specifically, we select 110 tasks with human-verified online tutorials.
We use our dataset pipeline (described in \S~\ref{sec:osworld}) to create MLLM-generated tutorials for these tasks.
We then evaluate both \ours and the ground-truth visual rollout method under each tutorial condition and report the results in Table~\ref{tab:res-ai-vs-human-tutorial-same-task}.

\vspace{5pt}\noindent\emph{\textbf{Finding 5: Task-specific tutorials written by a search-grounded frontier MLLM are as useful to the policy and judge as human-verified online tutorials.}}
In Table~\ref{tab:res-ai-vs-human-tutorial-same-task}, we find no statistically significant difference between the two tutorial types, in \ours (\texttt{row 4}) or in \texttt{GT-rollout} (\texttt{rows 5--7}).
This is likely because OSWorld-Tutorial tasks involve common software with abundant online tutorials.
We provide examples of MLLM-generated vs.~human-verified tutorial pairs and an analysis of a sample of 20 pairs in Appendix~\ref{sec:human-ai-tutorial-comparison}.
In this sample, a human verifier judges every MLLM-generated tutorial as correct or partially correct, and prefers it over its human-verified counterpart in 17 of 20 pairs.
MLLM-generated tutorials are \textit{task-conditioned}, naming the on-screen objects and values the task requires, whereas human-verified online tutorials describe an application feature in general, leaving the CUA to ground each action on the current screen.

\iffastmode\include{V}\else\input{V}\fi

\iffastmode\include{V}\else\input{V}\fi

\iffastmode\include{V}\else\input{V}\fi
\section{Limitations}
Our study has several limitations:

First, our MLLM-generated tutorials (Gemini 3.8 Flash with web search) cover noisy \textit{online} retrieval, but we do not evaluate \textit{offline} settings, where a local retriever over a smaller corpus (e.g., local documentation) may yield noisier or missing tutorials, which could change the relative benefit of look-ahead planning.

Second, both benchmarks run desktop applications in Linux VMs.
ScienceBoard-Tutorial has only MLLM-generated tutorials, so our human vs.~MLLM tutorial comparison (Finding 5) covers only 110 general-purpose OSWorld-Tutorial tasks.

Third, all models come from one MLLM family (Qwen3-VL-8B, Qwen3-VL-32B, Qwen3.8-27B), and all judges are zero-shot prompted; other model families, larger frontier MLLMs, or judges trained to reason over rollouts may use future states more effectively.

Fourth, our error analysis (Figure~\ref{fig:gt-rollout-error-analysis}) uses Gemini 3.8 Flash rather than human annotators to categorize failures, so the categories may carry labeling noise or MLLM biases.

\iffastmode\include{V}\else\input{V}\fi
\section{Ethical Considerations}
\paragraph{Compute used.}
We used 272 GPU-hours in total on NVIDIA H100/H200 GPUs, plus about 1,080 hours on an NVIDIA DGX Spark (GB10).

\paragraph{Risks.}Like all assistive technologies, advances in computer-use agents make it easier for individuals to take action---both good and ill.
Given how modern infrastructure relies on digital interfaces, this may pose significant risks for personal digital security, coordinated astroturfing, or model errors causing users harm.

\makeatletter
\ifacl@anonymize\else\section*{Acknowledgments}

We thank everyone from the VIRL lab and Huazheng Wang's lab (Oregon State University).
We thank Yaolun Zhang, Andrew Brandenfels, Artin Lahni, and Pranav Kulkarni for their help with annotation, analysis, and discussion, and for providing project resources.
We thank Khanh Nguyen for the brainstorming sessions that inspired the first author to pursue ideas in computer-use agents.
We thank Kai Mei, the first author of R-WoM~\cite{r_wom}, for detailed discussions of R-WoM's implementation.

The first author thanks Nguyen Thi Ngoc Anh, Amanda Putiza, Ngo Thi Bich Lan, Tran Thanh Nhu, Bui Thuy Tien, and Nguyen Hoang Kieu Anh for their personal support.
This work is supported by NSF CAREER Award 2339676.
We also thank the anonymous reviewers for their valuable feedback and suggestions.
\fi
\makeatother

\IfFileExists{latex/paperpile.bib}{\bibliography{latex/paperpile,latex/custom}}{\bibliography{paperpile,custom}}

\appendix
\onecolumn
\raggedbottom
\clearpage
\section*{Appendix Contents}
\setcounter{tocdepth}{2}
\makeatletter
\begingroup
\renewcommand*\l@section[2]{%
  \addvspace{0.5em}%
  \@dottedtocline{1}{0em}{2.3em}{\bfseries #1}{\bfseries #2}%
}
\@starttoc{atoc}%
\endgroup
\clearpage
\let\appendix@addcontentsline\addcontentsline
\renewcommand{\addcontentsline}[3]{%
  \def\appendix@entrytype{#2}%
  \def\appendix@sectiontype{section}%
  \def\appendix@subsectiontype{subsection}%
  \ifx\appendix@entrytype\appendix@sectiontype
    \appendix@addcontentsline{atoc}{#2}{#3}%
  \else\ifx\appendix@entrytype\appendix@subsectiontype
    \appendix@addcontentsline{atoc}{#2}{#3}%
  \fi\fi
}
\makeatother
\newcommand{\appendixcontentsline}[1]{%
  \addcontentsline{toc}{section}{\protect\numberline{\thesection}#1}%
}
\section{Results on OSWorld-Tutorial per category of tutorials}
\label{sec:res_per_category_tutorial}

\begingroup
\renewenvironment{table*}[1][]{\begin{table}[H]}{\end{table}}
\begin{table*}[t]
\centering
\small
\setlength{\tabcolsep}{3pt}
\renewcommand{\arraystretch}{1.15}
\renewcommand{\res}[2]{#1\%\if\relax\detokenize{#2}\relax\else{\small\,${\pm}$#2\%}\fi}
\definecolor{tie}{RGB}{222,235,247}\providecommand{\tie}[1]{\cellcolor{tie}#1}

\begin{tabular}{clcccccc}
\toprule
&\multicolumn{1}{c}{\multirow{2}{*}{\textbf{Method}}} & \multicolumn{2}{c}{\textbf{Qwen3-VL-8B}} & \multicolumn{2}{c}{\textbf{Qwen3-VL-32B}} & \multicolumn{2}{c}{\textbf{Qwen3.8-27B}} \\
\cmidrule(lr){3-4} \cmidrule(lr){5-6} \cmidrule(lr){7-8}
\#&& \textbf{\scriptsize 15 steps} & \textbf{\scriptsize 50 steps} & \textbf{\scriptsize 15 steps} & \textbf{\scriptsize 50 steps} & \textbf{\scriptsize 15 steps} & \textbf{\scriptsize 50 steps} \\
\midrule
\multicolumn{8}{>{\columncolor{gray!15}}l}{\textit{Policy only}} \\
{\tiny\texttt{1}}&\quad Policy Only & \res{47.2}{0.9} & \res{55.9}{0.5} & \res{60.2}{3.4} & \res{63.8}{1.4} & \res{68.9}{3.1} & \res{75.3}{1.0} \\
{\tiny\texttt{2}}&\quad \quad+ RAG & \res{59.3}{2.3} & \res{61.5}{3.2} & \res{66.6}{1.4} & \res{68.7}{2.3} & \res{73.2}{2.6} & \res{76.2}{4.0} \\
\midrule
\multicolumn{8}{>{\columncolor{gray!15}}l}{\textit{Policy + Judge}} \\
{\tiny\texttt{3}}&\quad Policy + Judge (GTA-1) & \res{44.7}{3.7} & \res{50.8}{5.7} & \res{61.7}{0.9} & \res{69.6}{3.7} & \res{67.1}{3.1} & \res{76.5}{1.9} \\
{\tiny\texttt{4}}&\quad \ours (Ours) & \tie{\res{61.6}{2.2}} & \tie{\res{62.1}{2.9}} & \tie{\res{72.6}{2.7}} & \tie{\res{74.5}{3.3}} & \tie{\res{75.1}{2.8}} & \tie{\res{79.0}{0.9}} \\
\midrule
\multicolumn{8}{>{\columncolor{gray!15}}l}{\textit{Policy + RAG Judge + Text-based rollout (R-WoM \citet{r_wom})}} \\
{\tiny\texttt{5}}&\quad R-WoM $H{=}1$ & \res{46.0}{1.8} & \res{49.0}{1.8} & \res{53.2}{1.9} & \res{66.0}{3.5} & \res{66.6}{1.4} & \res{76.8}{1.3} \\
{\tiny\texttt{6}}&\quad R-WoM $H{=}2$ & \res{49.5}{1.2} & \res{50.9}{7.1} & \res{61.1}{1.4} & \res{68.0}{5.0} & \res{71.1}{1.1} & \res{79.5}{0.5} \\
{\tiny\texttt{7}}&\quad R-WoM $H{=}3$ & \res{47.8}{0.5} & \res{54.7}{6.2} & \res{60.2}{1.4} & \res{66.0}{1.9} & \res{67.1}{2.4} & \res{75.7}{3.7} \\
\addlinespace[2pt]
{\tiny\texttt{8}}&\quad +Policy RAG $H{=}1$ & \res{50.0}{2.4} & \res{59.5}{3.9} & \res{68.5}{3.7} & \res{73.7}{1.8} & \tie{\res{73.5}{0.9}} & \tie{\res{80.0}{1.5}} \\
{\tiny\texttt{9}}&\quad +Policy RAG $H{=}2$ & \res{55.4}{2.4} & \res{56.9}{1.9} & \res{70.9}{1.6} & \res{72.0}{2.8} & \tie{\res{72.6}{2.7}} & \tie{\res{79.8}{1.6}} \\
{\tiny\texttt{10}}&\quad +Policy RAG $H{=}3$ & \res{49.6}{1.3} & \res{57.1}{1.7} & \res{67.8}{1.9} & \res{71.1}{3.8} & \tie{\res{73.8}{2.7}} & \tie{\res{79.4}{1.3}} \\
\midrule
\multicolumn{8}{>{\columncolor{gray!15}}l}{\textit{RAG Policy + RAG Judge + Ground-truth visual rollout}} \\
{\tiny\texttt{11}}&\quad $H{=}1$ & \tie{\res{59.5}{1.9}} & \tie{\res{60.1}{3.3}} & \tie{\res{66.3}{0.9}} & \tie{\res{72.6}{3.6}} & \tie{\res{75.3}{1.8}} & \tie{\res{80.2}{2.3}} \\
{\tiny\texttt{12}}&\quad $H{=}2$ & \tie{\res{59.9}{2.7}} & \tie{\res{60.5}{2.3}} & \tie{\res{72.3}{1.4}} & \tie{\res{74.2}{1.0}} & \tie{\res{75.1}{1.4}} & \tie{\res{80.7}{0.8}} \\
{\tiny\texttt{13}}&\quad $H{=}3$ & \tie{\res{60.5}{3.6}} & \tie{\res{65.8}{0.6}} & \tie{\res{70.4}{2.9}} & \tie{\res{75.3}{1.3}} & \tie{\res{76.0}{2.9}} & \tie{\res{81.7}{1.1}} \\
\bottomrule
\end{tabular}

\caption{Success rate of \ours and other baselines on the 110 OSWorld-Tutorial tasks with \textbf{human-verified online tutorials}, over maximum runtimes of 15 and 50 steps. Results are mean $\pm$ sample std. across 3 completed runs. ``RAG'' indicates tutorial use. {\setlength{\fboxsep}{1pt}\colorbox{tie}{Shaded cells}} are not significantly different from \ours (row 4) in the same column.}
\label{tab:res-human-written}
\end{table*}

\begin{table*}[t]
\centering
\small
\setlength{\tabcolsep}{3pt}
\renewcommand{\arraystretch}{1.15}
\renewcommand{\res}[2]{#1\%\if\relax\detokenize{#2}\relax\else{\small\,${\pm}$#2\%}\fi}
\definecolor{tie}{RGB}{222,235,247}\providecommand{\tie}[1]{\cellcolor{tie}#1}

\begin{tabular}{clcccccc}
\toprule
&\multicolumn{1}{c}{\multirow{2}{*}{\textbf{Method}}} & \multicolumn{2}{c}{\textbf{Qwen3-VL-8B}} & \multicolumn{2}{c}{\textbf{Qwen3-VL-32B}} & \multicolumn{2}{c}{\textbf{Qwen3.8-27B}} \\
\cmidrule(lr){3-4} \cmidrule(lr){5-6} \cmidrule(lr){7-8}
\#&& \textbf{\scriptsize 15 steps} & \textbf{\scriptsize 50 steps} & \textbf{\scriptsize 15 steps} & \textbf{\scriptsize 50 steps} & \textbf{\scriptsize 15 steps} & \textbf{\scriptsize 50 steps} \\
\midrule
\multicolumn{8}{>{\columncolor{gray!15}}l}{\textit{Policy only}} \\
{\tiny\texttt{1}}&\quad Policy Only & \res{20.9}{0.9} & \res{25.1}{1.4} & \res{34.7}{1.8} & \res{36.6}{2.4} & \res{51.7}{3.1} & \res{61.0}{0.2} \\
{\tiny\texttt{2}}&\quad \quad+ RAG & \res{32.4}{2.9} & \res{38.6}{1.7} & \res{39.5}{1.2} & \res{44.3}{2.5} & \res{54.5}{2.0} & \res{64.3}{1.6} \\
\midrule
\multicolumn{8}{>{\columncolor{gray!15}}l}{\textit{Policy + Judge}} \\
{\tiny\texttt{3}}&\quad Policy + Judge (GTA-1) & \res{23.3}{1.6} & \res{24.5}{1.4} & \res{35.8}{0.3} & \res{37.7}{1.5} & \res{51.6}{0.6} & \res{59.5}{0.3} \\
{\tiny\texttt{4}}&\quad \ours (Ours) & \tie{\res{33.7}{0.8}} & \tie{\res{40.1}{1.9}} & \tie{\res{42.5}{0.9}} & \tie{\res{49.9}{4.1}} & \tie{\res{55.1}{0.6}} & \tie{\res{67.0}{1.4}} \\
\midrule
\multicolumn{8}{>{\columncolor{gray!15}}l}{\textit{Policy + RAG Judge + Text-based rollout (R-WoM \citet{r_wom})}} \\
{\tiny\texttt{5}}&\quad R-WoM $H{=}1$ & \res{24.1}{2.6} & \res{25.3}{1.9} & \res{29.0}{0.9} & \res{36.9}{3.1} & \res{52.1}{3.1} & \res{61.4}{1.2} \\
{\tiny\texttt{6}}&\quad R-WoM $H{=}2$ & \res{22.1}{1.9} & \res{25.9}{1.8} & \res{32.0}{1.4} & \res{39.2}{4.3} & \res{51.8}{1.2} & \res{58.6}{3.8} \\
{\tiny\texttt{7}}&\quad R-WoM $H{=}3$ & \res{24.7}{0.8} & \res{29.1}{0.2} & \res{34.1}{1.1} & \res{37.5}{0.5} & \res{49.8}{1.1} & \res{61.5}{1.6} \\
\addlinespace[2pt]
{\tiny\texttt{8}}&\quad +Policy RAG $H{=}1$ & \res{28.1}{0.4} & \res{31.8}{1.2} & \res{38.6}{1.1} & \res{45.4}{0.7} & \tie{\res{57.7}{1.7}} & \tie{\res{67.6}{3.3}} \\
{\tiny\texttt{9}}&\quad +Policy RAG $H{=}2$ & \res{31.2}{0.7} & \res{31.8}{1.8} & \res{40.9}{1.3} & \res{42.5}{0.7} & \tie{\res{53.2}{1.1}} & \tie{\res{65.5}{2.6}} \\
{\tiny\texttt{10}}&\quad +Policy RAG $H{=}3$ & \res{30.5}{2.4} & \res{31.4}{1.8} & \res{38.1}{1.0} & \res{44.1}{1.5} & \tie{\res{53.4}{0.4}} & \tie{\res{67.5}{1.1}} \\
\midrule
\multicolumn{8}{>{\columncolor{gray!15}}l}{\textit{RAG Policy + RAG Judge + Ground-truth visual rollout}} \\
{\tiny\texttt{11}}&\quad $H{=}1$ & \tie{\res{32.4}{0.3}} & \tie{\res{35.7}{4.8}} & \tie{\res{42.0}{0.5}} & \tie{\res{50.8}{3.1}} & \tie{\res{56.5}{0.9}} & \tie{\res{65.9}{1.7}} \\
{\tiny\texttt{12}}&\quad $H{=}2$ & \tie{\res{35.8}{1.0}} & \tie{\res{39.0}{1.4}} & \tie{\res{42.4}{0.8}} & \tie{\res{50.4}{1.8}} & \tie{\res{58.1}{1.8}} & \tie{\res{64.9}{0.4}} \\
{\tiny\texttt{13}}&\quad $H{=}3$ & \tie{\res{33.0}{1.1}} & \tie{\res{39.2}{1.4}} & \tie{\res{41.8}{0.5}} & \tie{\res{50.4}{0.4}} & \tie{\res{57.5}{2.4}} & \tie{\res{67.2}{1.4}} \\
\bottomrule
\end{tabular}

\caption{Success rate of \ours and other baselines on the 250 OSWorld-Tutorial tasks with \textbf{AI-generated tutorials} from \texttt{tutorial\_osworld} over maximum runtimes of 15 and 50 steps. Results are mean $\pm$ sample std. across 3 completed runs. ``RAG'' indicates tutorial use. {\setlength{\fboxsep}{1pt}\colorbox{tie}{Shaded cells}} are not significantly different from \ours (row 4) in the same column.}
\label{tab:res-ai-generated}
\end{table*}

\begin{table*}[t]
\centering
\small
\setlength{\tabcolsep}{6pt}
\renewcommand{\arraystretch}{1.15}
\renewcommand{\res}[2]{#1\%\if\relax\detokenize{#2}\relax\else{\small\,${\pm}$#2\%}\fi}

\resizebox{0.95\textwidth}{!}{%
\begin{tabular}{clcccccc}
\toprule
&\multicolumn{1}{c}{\multirow{2}{*}{\textbf{Method}}} & \multicolumn{6}{c}{\textbf{Qwen3-VL-32B}} \\
\cmidrule(lr){3-8}
\#&& \textbf{Overall} & \textbf{OS} & \textbf{Office} & \textbf{Daily} & \textbf{Professional} & \textbf{Workflow} \\
\midrule
\multicolumn{8}{>{\columncolor{gray!15}}l}{\textit{15 steps}} \\
{\tiny\texttt{1}}&\quad Policy only baseline & \res{42.5}{0.7} & \res{65.3}{4.8} & \res{33.5}{1.8} & \res{63.5}{4.7} & \res{60.5}{1.2} & \res{21.0}{3.3} \\
{\tiny\texttt{2}}&\quad Policy only + RAG policy & \res{47.8}{0.7} & \res{88.9}{2.4} & \res{35.6}{3.5} & \res{68.8}{3.4} & \res{72.8}{1.2} & \res{22.0}{2.1} \\
{\tiny\texttt{3}}&\quad RAG policy + RAG judge (RAG-PJ) & \res{51.7}{1.4} & \res{91.7}{4.2} & \res{41.5}{2.0} & \res{66.6}{2.7} & \res{83.7}{2.0} & \res{25.0}{2.2} \\
{\tiny\texttt{4}}&\quad GT rollout $H{=}1$ & \res{49.4}{0.2} & \res{87.5}{7.2} & \res{39.9}{0.9} & \res{66.2}{2.7} & \res{72.1}{5.9} & \res{25.7}{2.1} \\
{\tiny\texttt{5}}&\quad GT rollout $H{=}2$ & \res{51.6}{0.9} & \res{91.7}{4.2} & \res{42.3}{0.7} & \res{65.8}{4.6} & \res{80.3}{4.2} & \res{26.0}{2.4} \\
{\tiny\texttt{6}}&\quad GT rollout $H{=}3$ & \res{50.5}{1.1} & \res{91.7}{4.2} & \res{39.7}{2.3} & \res{66.4}{3.2} & \res{81.6}{2.0} & \res{24.0}{1.2} \\
\midrule
\multicolumn{8}{>{\columncolor{gray!15}}l}{\textit{50 steps}} \\
{\tiny\texttt{1}}&\quad Policy only baseline & \res{44.9}{2.0} & \res{73.6}{4.8} & \res{32.4}{2.3} & \res{63.9}{3.7} & \res{68.0}{9.4} & \res{25.4}{3.7} \\
{\tiny\texttt{2}}&\quad Policy only + RAG policy & \res{51.7}{2.4} & \res{88.9}{6.4} & \res{39.5}{3.2} & \res{72.2}{5.3} & \res{73.5}{2.0} & \res{29.2}{0.7} \\
{\tiny\texttt{3}}&\quad RAG policy + RAG judge (RAG-PJ) & \res{57.4}{3.6} & \res{95.8}{0.0} & \res{48.7}{5.1} & \res{72.3}{3.0} & \res{81.6}{3.5} & \res{33.5}{3.8} \\
{\tiny\texttt{4}}&\quad GT rollout $H{=}1$ & \res{57.5}{1.8} & \res{97.2}{2.4} & \res{47.1}{4.4} & \res{72.1}{0.7} & \res{80.3}{5.1} & \res{36.1}{4.1} \\
{\tiny\texttt{5}}&\quad GT rollout $H{=}2$ & \res{57.7}{1.0} & \res{94.4}{4.8} & \res{47.7}{5.0} & \res{73.0}{3.3} & \res{83.0}{4.2} & \res{34.6}{1.6} \\
{\tiny\texttt{6}}&\quad GT rollout $H{=}3$ & \res{58.0}{0.7} & \res{97.9}{2.9} & \res{48.9}{4.4} & \res{73.7}{7.3} & \res{82.7}{1.4} & \res{33.1}{1.6} \\
\bottomrule
\end{tabular}
}

\caption{Success rate per category on 360 OSWorld-Tutorial tasks over 15 and 50 steps, using Qwen3-VL-32B as the policy and judge. Categories are OS, Office, Daily, Professional, and Workflow. Mean $\pm$ sample std. across 3 completed runs.}
\label{tab:res-per-category}
\end{table*}

\endgroup

\paragraph{Statistical tests.}
We use Hierarchical bootstrap, sign-flip permutation, and Wilcoxon signed-rank tests over tasks in Table \ref{tab:res}, \ref{tab:res-ai-generated}, and \ref{tab:res-human-written} to test whether ground-truth visual rollouts and R-WoM improve over \ours on OSWorld-Tutorial.

\paragraph{Per-task-category analysis.}
To understand when and why the agent gains or loses a benefit from ground-truth visual rollout (GT rollouts), we break down the results of Qwen3-VL-32B in Table~\ref{tab:res} by task category in Table~\ref{tab:res-per-category}.
    We inspected 40 cases where GT rollouts succeeded but RAG-PJ failed.
    In these cases, candidate actions produced clear, visible changes in the next state, such as opening a new website or creating a file.
    These visible effects make the correct action easier for the judge to spot, which matters most when there is little time to recover from a mistake.
    These gains do not generalize to other task categories.
    Applications like VS Code and GIMP have denser interfaces, so the effect of an action is often less visually clear.
    In one failure case, the judge misread a VS Code workspace with unwanted Explorer folders as a success---consistent with our earlier finding that the judge misinterprets task success even from perfect future screenshots.
    The agent simply has more time to observe the consequences of its actions and recover from mistakes without needing look-ahead.

\section{Results on ScienceBoard per application}
\label{sec:res_per_category_scienceboard}

\begingroup
\renewenvironment{table*}[1][]{\begin{table}[H]}{\end{table}}
\begin{table*}[t]
\centering
\footnotesize
\setlength{\tabcolsep}{1.5pt}
\renewcommand{\arraystretch}{1.15}
\renewcommand{\res}[2]{#1\%\if\relax\detokenize{#2}\relax\else{\tiny\,${\pm}$#2\%}\fi}

\begin{tabular}{clccccccc}
\toprule
\#&\multicolumn{1}{c}{\textbf{Method}} & \textbf{Celestia} & \textbf{ChimeraX} & \textbf{GrassGIS} & \textbf{KAlgebra} & \textbf{Lean} & \textbf{TeXstudio} & \textbf{Overall} \\
&& \scriptsize (33) & \scriptsize (29) & \scriptsize (34) & \scriptsize (31) & \scriptsize (26) & \scriptsize (16) & \scriptsize (169) \\
\midrule
\multicolumn{9}{>{\columncolor{gray!15}}l}{\textit{Policy only}} \\
{\tiny\texttt{1}}&\quad Policy Only & \res{20.2}{1.7} & \res{26.4}{2.0} & \res{8.8}{2.9} & \res{18.3}{1.9} & \res{0.0}{0.0} & \res{12.5}{0.0} & \res{14.8}{1.0} \\
{\tiny\texttt{2}}&\quad \quad+ RAG & \res{31.3}{3.5} & \res{74.7}{2.0} & \res{22.5}{1.7} & \res{62.4}{3.7} & \res{0.0}{0.0} & \res{12.5}{0.0} & \res{36.1}{0.6} \\
\midrule
\multicolumn{9}{>{\columncolor{gray!15}}l}{\textit{Policy + Judge}} \\
{\tiny\texttt{3}}&\quad Policy + Judge (GTA-1) & \res{25.3}{1.7} & \res{31.0}{6.0} & \res{10.8}{4.5} & \res{26.9}{10.4} & \res{0.0}{0.0} & \res{10.4}{3.6} & \res{18.3}{2.4} \\
{\tiny\texttt{4}}&\quad \ours (Ours) & \res{28.3}{1.7} & \res{74.7}{5.3} & \res{24.5}{1.7} & \res{73.1}{1.9} & \res{15.4}{0.0} & \res{12.5}{0.0} & \res{40.2}{0.6} \\
\midrule
\multicolumn{9}{>{\columncolor{gray!15}}l}{\textit{Policy + RAG Judge + Text rollout \citep{r_wom}}} \\
{\tiny\texttt{5}}&\quad R-WoM $H{=}1$ & \res{21.2}{0.0} & \res{25.3}{4.0} & \res{14.7}{0.0} & \res{34.4}{8.1} & \res{0.0}{0.0} & \res{8.3}{7.2} & \res{18.5}{0.9} \\
{\tiny\texttt{6}}&\quad R-WoM $H{=}2$ & \res{24.2}{0.0} & \res{34.5}{6.9} & \res{16.7}{1.7} & \res{30.1}{7.4} & \res{0.0}{0.0} & \res{8.3}{3.6} & \res{20.3}{2.5} \\
{\tiny\texttt{7}}&\quad R-WoM $H{=}3$ & \res{25.3}{1.7} & \res{29.9}{8.7} & \res{14.7}{2.9} & \res{34.4}{1.9} & \res{0.0}{0.0} & \res{12.5}{0.0} & \res{20.5}{0.7} \\
\addlinespace[2pt]
{\tiny\texttt{8}}&\quad +Policy RAG $H{=}1$ & \res{27.3}{3.0} & \res{71.3}{2.0} & \res{27.5}{1.7} & \res{62.4}{1.9} & \res{0.0}{0.0} & \res{8.3}{3.6} & \res{35.3}{0.9} \\
{\tiny\texttt{9}}&\quad +Policy RAG $H{=}2$ & \res{32.3}{3.5} & \res{70.1}{2.0} & \res{25.5}{3.4} & \res{69.9}{1.9} & \res{1.3}{2.2} & \res{10.4}{3.6} & \res{37.5}{1.4} \\
{\tiny\texttt{10}}&\quad +Policy RAG $H{=}3$ & \res{31.3}{3.5} & \res{66.7}{5.3} & \res{18.6}{9.0} & \res{61.3}{3.2} & \res{2.6}{2.2} & \res{8.3}{3.6} & \res{33.7}{2.6} \\
\midrule
\multicolumn{9}{>{\columncolor{gray!15}}l}{\textit{RAG Policy + RAG Judge + Ground-truth visual rollout}} \\
{\tiny\texttt{11}}&\quad $H{=}1$ & \res{34.3}{4.6} & \res{73.6}{2.0} & \res{27.5}{4.5} & \res{71.0}{5.6} & \res{1.3}{2.2} & \res{10.4}{3.6} & \res{39.1}{2.0} \\
{\tiny\texttt{12}}&\quad $H{=}2$ & \res{41.4}{6.3} & \res{73.6}{5.3} & \res{30.4}{1.7} & \res{68.8}{3.7} & \res{0.0}{0.0} & \res{10.4}{3.6} & \res{40.4}{0.7} \\
{\tiny\texttt{13}}&\quad $H{=}3$ & \res{40.4}{3.5} & \res{75.9}{3.4} & \res{32.4}{2.9} & \res{72.0}{6.7} & \res{3.8}{3.8} & \res{12.5}{0.0} & \res{42.4}{1.9} \\
\bottomrule
\end{tabular}

\caption{Per-application success rate on the 169 ScienceBoard tasks with the \textbf{Qwen3-VL-32B} policy (15 steps). Mean $\pm$ sample std. over 3 completed runs; task counts in parentheses. ``RAG'' marks tutorial use.}
\label{tab:res-scienceboard-per-app-32b}
\end{table*}

\begin{table*}[t]
\centering
\footnotesize
\setlength{\tabcolsep}{1.5pt}
\renewcommand{\arraystretch}{1.15}
\renewcommand{\res}[2]{#1\%\if\relax\detokenize{#2}\relax\else{\tiny\,${\pm}$#2\%}\fi}

\begin{tabular}{clccccccc}
\toprule
\#&\multicolumn{1}{c}{\textbf{Method}} & \textbf{Celestia} & \textbf{ChimeraX} & \textbf{GrassGIS} & \textbf{KAlgebra} & \textbf{Lean} & \textbf{TeXstudio} & \textbf{Overall} \\
&& \scriptsize (33) & \scriptsize (29) & \scriptsize (34) & \scriptsize (31) & \scriptsize (26) & \scriptsize (16) & \scriptsize (169) \\
\midrule
\multicolumn{9}{>{\columncolor{gray!15}}l}{\textit{Policy only}} \\
{\tiny\texttt{1}}&\quad Policy Only & \res{19.2}{4.6} & \res{56.3}{2.0} & \res{27.5}{4.5} & \res{46.2}{1.9} & \res{2.6}{2.2} & \res{12.5}{0.0} & \res{29.0}{0.6} \\
{\tiny\texttt{2}}&\quad \quad+ RAG & \res{34.3}{3.5} & \res{79.3}{3.4} & \res{29.4}{2.9} & \res{79.6}{1.9} & \res{14.1}{2.2} & \res{12.5}{0.0} & \res{44.2}{1.4} \\
\midrule
\multicolumn{9}{>{\columncolor{gray!15}}l}{\textit{Policy + Judge}} \\
{\tiny\texttt{3}}&\quad Policy + Judge (GTA-1) & \res{25.3}{1.7} & \res{62.1}{3.4} & \res{27.5}{6.1} & \res{40.9}{4.9} & \res{3.8}{3.8} & \res{12.5}{0.0} & \res{30.4}{1.2} \\
{\tiny\texttt{4}}&\quad \ours (Ours) & \res{41.4}{3.5} & \res{80.5}{2.0} & \res{31.4}{4.5} & \res{79.6}{1.9} & \res{28.2}{2.2} & \res{12.5}{0.0} & \res{48.3}{0.7} \\
\midrule
\multicolumn{9}{>{\columncolor{gray!15}}l}{\textit{Policy + RAG Judge + Text rollout \citep{r_wom}}} \\
{\tiny\texttt{5}}&\quad R-WoM $H{=}1$ & \res{27.3}{3.0} & \res{67.8}{4.0} & \res{23.5}{7.8} & \res{41.9}{8.5} & \res{0.0}{0.0} & \res{12.5}{0.0} & \res{30.6}{2.4} \\
{\tiny\texttt{6}}&\quad R-WoM $H{=}2$ & \res{28.3}{1.7} & \res{60.9}{5.3} & \res{34.3}{6.1} & \res{46.2}{11.3} & \res{1.3}{2.2} & \res{12.5}{0.0} & \res{32.7}{1.2} \\
{\tiny\texttt{7}}&\quad R-WoM $H{=}3$ & \res{23.2}{1.7} & \res{58.6}{0.0} & \res{27.5}{4.5} & \res{48.4}{3.2} & \res{2.6}{2.2} & \res{12.5}{0.0} & \res{30.6}{0.7} \\
\addlinespace[2pt]
{\tiny\texttt{8}}&\quad +Policy RAG $H{=}1$ & \res{31.3}{4.6} & \res{73.6}{5.3} & \res{30.4}{1.7} & \res{77.4}{3.2} & \res{15.4}{3.8} & \res{12.5}{0.0} & \res{42.6}{1.6} \\
{\tiny\texttt{9}}&\quad +Policy RAG $H{=}2$ & \res{30.3}{5.2} & \res{81.6}{4.0} & \res{31.4}{4.5} & \res{77.4}{5.6} & \res{10.3}{4.4} & \res{12.5}{0.0} & \res{43.2}{0.6} \\
{\tiny\texttt{10}}&\quad +Policy RAG $H{=}3$ & \res{38.4}{4.6} & \res{81.6}{2.0} & \res{27.5}{4.5} & \res{77.4}{0.0} & \res{15.4}{3.8} & \res{12.5}{0.0} & \res{44.8}{0.3} \\
\midrule
\multicolumn{9}{>{\columncolor{gray!15}}l}{\textit{RAG Policy + RAG Judge + Ground-truth visual rollout}} \\
{\tiny\texttt{11}}&\quad $H{=}1$ & \res{45.5}{6.1} & \res{77.0}{2.0} & \res{36.3}{4.5} & \res{83.9}{0.0} & \res{20.5}{2.2} & \res{12.5}{0.0} & \res{49.1}{0.6} \\
{\tiny\texttt{12}}&\quad $H{=}2$ & \res{54.5}{0.0} & \res{79.3}{3.4} & \res{32.4}{0.0} & \res{80.6}{0.0} & \res{16.7}{2.2} & \res{12.5}{0.0} & \res{49.3}{0.3} \\
{\tiny\texttt{13}}&\quad $H{=}3$ & \res{48.5}{0.0} & \res{79.3}{0.0} & \res{39.2}{4.5} & \res{81.7}{4.9} & \res{17.9}{2.2} & \res{12.5}{0.0} & \res{49.9}{1.2} \\
\bottomrule
\end{tabular}

\caption{Per-application success rate on the 169 ScienceBoard tasks with the \textbf{Qwen3.8-27B} policy (15 steps). Mean $\pm$ sample std. over 3 completed runs; task counts in parentheses. ``RAG'' marks tutorial use.}
\label{tab:res-scienceboard-per-app-27b}
\end{table*}

\endgroup

\paragraph{Significance tests.}
\label{sec:sb-sig-tests}
We test whether ground-truth visual rollouts improve over RAG-PJ on ScienceBoard (Table~\ref{tab:scienceboard}, \texttt{rows 11--13} vs.\ \texttt{row 4}) with task-level tests.
For Qwen3-VL-32B ($H{=}1/2/3$): hierarchical bootstrap $p{=}0.62/0.97/0.32$, sign-flip permutation $p{=}0.58/1.00/0.24$, Wilcoxon signed-rank $p{=}0.51/0.99/0.20$.
For Qwen3.8-27B ($H{=}1/2/3$): sign-flip permutation $p{=}0.20/0.20/0.11$, Wilcoxon signed-rank $p{=}0.20/0.15/0.14$.
Welch's $t$-test on run-level totals gives $p{=}0.43/0.73/0.18$ (32B) and $p{=}0.013/0.001/0.052$ (27B), but it ignores per-task variation, so we do not treat the 27B differences as significant.

\paragraph{Why do GT rollouts help in Celestia and GrassGIS?}
GT visual rollouts improve performance over \ours in Celestia and GrassGIS across the evaluated horizons.
We sample 40 cases where GT rollouts succeeded but RAG-PJ failed.
The common reason is that the tutorials given to agents are only partially correct.
They omit steps that the task needs in the current application state, and they present options as equivalent when only some of them work.

\begin{figure}[tbp]
\centering
\begin{tcolorbox}[colback=gray!5, colframe=gray!60, boxrule=0.5pt, left=4pt, right=4pt, top=2pt, bottom=2pt, fontupper=\small]
\textbf{Task instruction:} Go to Neptune in Celestia.
\end{tcolorbox}
\vspace{2pt}
\includegraphics[width=0.8\linewidth]{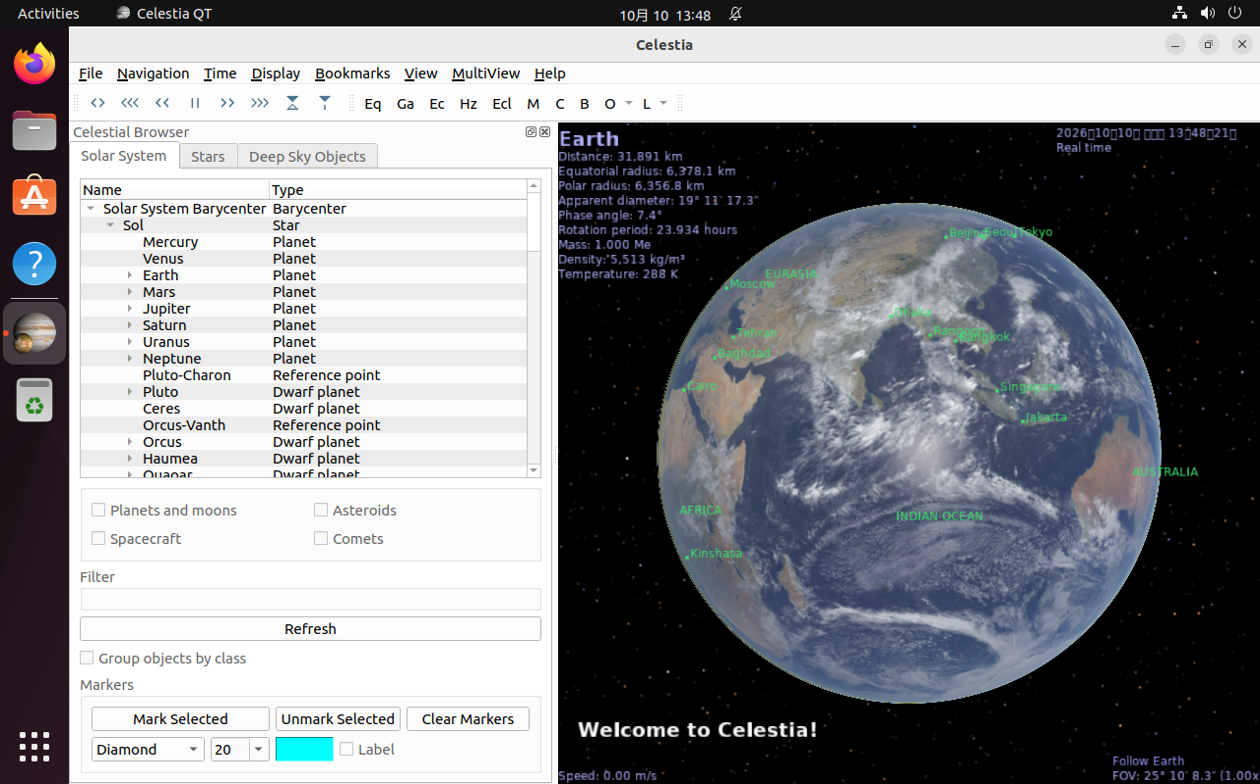}
\vspace{2pt}
\begin{tcolorbox}[colback=gray!5, colframe=gray!60, boxrule=0.5pt, left=4pt, right=4pt, top=2pt, bottom=2pt, fontupper=\small]
\textbf{Tutorial:}\\[2pt]
\textbf{Method 1: Using the Celestial Browser (Visible Panel)}
\begin{compactenum}
\item In the \textbf{Celestial Browser} panel on the left (under the \textbf{Solar System} tab), click on \textbf{Neptune} in the list under \texttt{Solar System Barycenter} $\rightarrow$ \texttt{Sol} to select it.
\item {\sethlcolor{yellow}\hl{Press the \textbf{G} key on your keyboard (or right-click \textbf{Neptune} in the list and select \textbf{Go to}, or double-click \textbf{Neptune}) to travel to the planet.}}
\end{compactenum}
\textbf{Method 2: Using Direct Planet Number Shortcut}
\begin{compactenum}
\item Click on the main 3D star viewport to ensure it has focus.
\item Press the number key \textbf{8} on your keyboard to select Neptune (the 8th planet from the Sun).
\item Press the \textbf{G} key to travel directly to Neptune.
\end{compactenum}
\textbf{Method 3: Using the Target Name Bar}
\begin{compactenum}
\item Click on the main 3D star viewport and press \textbf{Enter} (Return) to bring up the target entry prompt.
\item Type \texttt{Neptune} and press \textbf{Enter} to confirm the selection.
\item Press the \textbf{G} key to travel to Neptune.
\end{compactenum}
\textbf{Method 4: Using the Navigation Menu}
\begin{compactenum}
\item Click \textbf{Navigation} in the top menu bar.
\item Click \textbf{Go to Object...} (or \textbf{Select Object...}, type \texttt{Neptune}, and confirm).
\item In the dialog window, ensure the target is set to \texttt{Neptune} and click the \textbf{Go to} (or \textbf{OK}) button to begin navigation.
\end{compactenum}
\end{tcolorbox}
\caption{Celestia task B-02 (5-step budget): the task instruction, the initial screenshot, and the retrieved tutorial. Method~1 omits the step that gives the 3D viewport keyboard focus, so after Neptune is clicked in the list, pressing \texttt{G} or double-clicking does not travel to Neptune.}
\label{fig:sb-celestia-task}
\end{figure}

For example, in Celestia task B-02 (Figure~\ref{fig:sb-celestia-task}) the instruction is \textit{``Go to Neptune in Celestia.''}
\ours executes Method 1 of the retrieved tutorial, and tries to follow the {\sethlcolor{yellow}\hl{highlighted}} step in its tutorial. 
In that step, the tutorial proposes three options to travel to Neptune: pressing \texttt{G}, right-clicking, or double-clicking.
\ours then executes the first option, pressing \texttt{G}, but it fails to travel to Neptune.
The reason is that its tutorial fails to include the step that gives the 3D viewport keyboard focus (by clicking on it), right before the step that instructs the user to press \texttt{G} to travel to Neptune.
Therefore, it fails to do that step and run out of the 5-step budget of that task, before trying the other options.

\begin{figure}[tbp]
\centering
\begin{subfigure}[t]{0.32\linewidth}
    \centering
    \includegraphics[width=\linewidth]{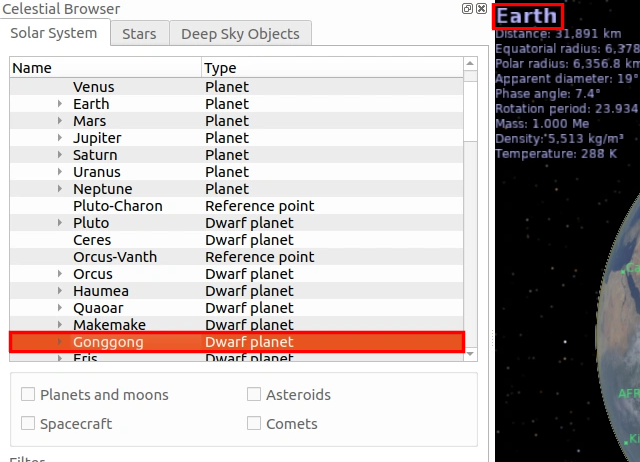}
    \caption{Press \texttt{G}: the list jumps to Gonggong; the view stays on Earth.}
\end{subfigure}
\hfill
\begin{subfigure}[t]{0.32\linewidth}
    \centering
    \includegraphics[width=\linewidth]{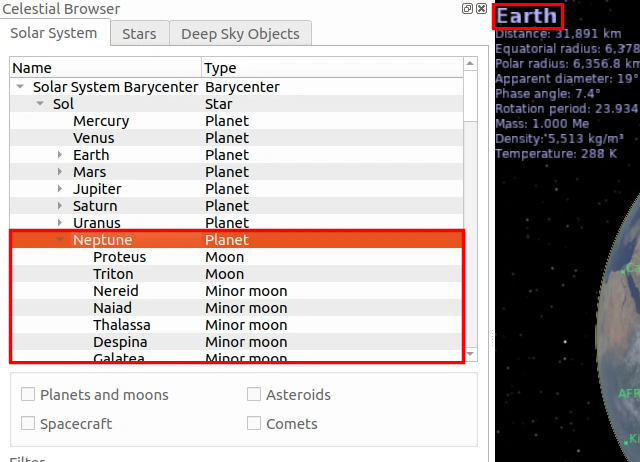}
    \caption{Double-click: the list expands Neptune's moons; the view stays on Earth.}
\end{subfigure}
\hfill
\begin{subfigure}[t]{0.32\linewidth}
    \centering
    \includegraphics[width=\linewidth]{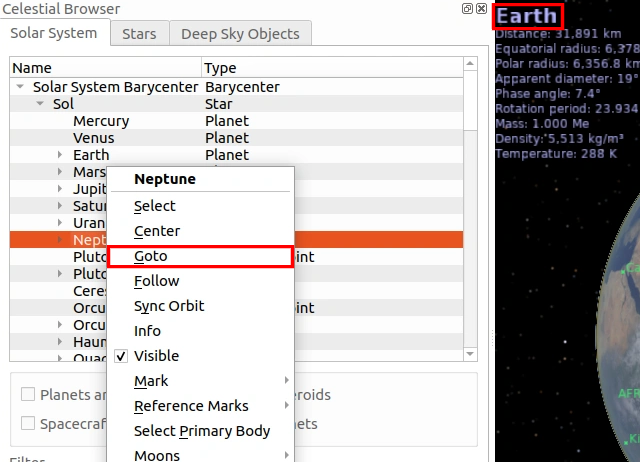}
    \caption{Right-click: Celestia's object menu, with \emph{Goto}.}
\end{subfigure}
\caption{The three alternatives the tutorial gives for one step of ``Go to Neptune''
(Qwen3.8-27B), each executed by GT rollout on a copy of the VM after Neptune is selected in the
list. RAG-PJ executed (a) for real.}
\label{fig:sb-celestia}
\end{figure}

GT-rollout insteads executes all three options in the tutorial on a copy of the VM, and sees that only the right-click option works, so it chooses that one and completes the task in 4 steps.
Details of three different outcomes of the three options are shown in Figure~\ref{fig:sb-celestia}.

\iffastmode\include{V}\else\input{V}\fi\section{Policy + Judge with visual rollout, without tutorials}
\label{sec:policy-judge-rollout}

\begin{figure}[H]
    \centering
    \includegraphics[width=\columnwidth]{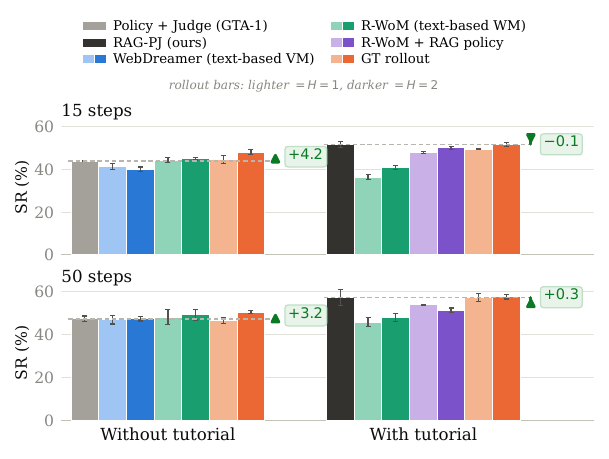}
    \caption{Success rate with vs.\ without tutorials (Qwen3-VL-32B). Green arrows: gap from the judge-only reference (dashed) to the best rollout method. Without tutorials, rollout beats the judge by up to $4.2$\%; with tutorials, the gap nearly vanishes. Numbers from Tables~\ref{tab:res-32b-no-tutorial} and~\ref{tab:res}.}
    \label{fig:tutorial-gap}
\end{figure}

We note down the differences between WebDreamer and R-WoM, which are both text-based world model look-ahead methods. 
While WebDreamer and R-WoM both use greedy policy in rollout states, they differ in how rollouts are generated and scored. 
In WebDreamer \citep{webdreamer}, for each candidate action, the rollout is generated many times with different imagined trajectories. 
Each trajectory is scored independently, either complete (1), on track (0.5), or incorrect (0). 
The scores of these trajectories are averaged to produce a score for that candidate action.
In R-WoM \citep{r_wom}, only one imagined trajectory is generated for each candidate action. 
All candidate actions are then ranked listwise based on the single trajectory score, and the best candidate action is selected.

\iffastmode\include{V}\else\input{V}\fi\section{Detailed related work}
\label{sec:detailed_related_work}

\subsection{World models for CUAs}
\label{sec:related_work_world_models}
Recent work explores world models~\cite{Ha2018-nb} for CUA planning, which predict future states in one of two forms: (1) textual descriptions of the next state~\cite{wma,mobileworldbench,mobiledreamer,r_wom,qwen_agentworld}, or (2) visual predictions of the next state, either by generating images directly~\cite{vimo,ui_sim} or by generating code that renders them~\cite{code2world,wm_r1}.
\blue{Most recently, \citet{autoguiworld} show that pretrained image generators can act as visual world models, iteratively editing screenshots to synthesize action-grounded training trajectories without deploying the software.}
In this work, we evaluate whether CUAs with an MLLM-as-a-judge can leverage both forms of world models: text-based rollouts and oracle visual rollouts, an upper bound for visual world models, when given access to task-relevant tutorials.

\subsection{Agents that use tutorials (manuals)}
\label{sec:related_work_tutorials}
Tutorials or manuals can describe both step-by-step instructions (\textit{what to do}) and how an environment changes over time (the \textit{dynamics of the environment}).
Many works study how agents can use tutorials in different domains, including
(1) games~\cite{Nguyen2025-hq, lwm, dynalang, reader, Chen2024-oc},
(2) robotics~\cite{Zhang2025-eg, Ren2022-xz}, and
(3) embodied environments~\cite{automanual}.
While our work targets the Computer-use Agent (CUA) domain, some of these works share a similar goal: using tutorials at inference time to help agents handle new tasks or environments.
In particular, our work is closely related to~\cite{Nguyen2025-hq, lwm, reader}.
However, these works focus on symbolic 2D environments, where language is restricted to a single sentence per entity---typically 3--5 sentences total---and covers only environmental dynamics.
Meanwhile, we focus on real-world GUI environments with longer tutorials that cover both task instructions and environmental dynamics.

\subsection{Comparison with RAG-GUI}
\label{sec:rag_gui}
While RAG-GUI~\cite{rag_gui} shares with us the same motivation of using tutorials for GUI agents at inference time, we highlight several key differences between RAG-GUI and our method (RAG-PJ):

\begin{itemize}
  \item \textit{RAG-GUI discards visual information in tutorials.}
  RAG-GUI uses a lightweight VLM to judge whether each retrieved multimodal tutorial is relevant and then generates a ``concise textual guidance summary'' for the VLM.
  This removes visual information that might be necessary for GUI agents to complete a given task, e.g., clicking on an icon of a button shown in the tutorial.
  Meanwhile, our method (RAG-PJ) uses raw multimodal tutorials and directly feeds them to the LLM-as-judge to judge the policy's action, thus potentially taking advantage of visual clues in tutorials.

  \item \textit{RAG-GUI is challenging to reproduce and compare.}
  RAG-GUI does not publish its implementation or its dataset (tutorials).
  Further, it requires extensive training.
  This makes it challenging for us to directly compare our method (RAG-PJ) with RAG-GUI.
  Meanwhile, one of our contributions is that we publish our dataset and implementation to foster further research in CUA / GUI agents with tutorials.

  \item \textit{RAG-GUI does not have tutorial-used baselines.}
  RAG-GUI does not compare with other GUI / CUA methods that use tutorials.
  Thus, it is hard to know if RAG-GUI's design choice is the best way to use tutorials.
  Meanwhile, we compare against other tutorial-using methods that involve planning ahead with world models.

  \item \textit{RAG-GUI focuses on retrieving tutorials, whereas we focus on how to use a given tutorial.}
  RAG-GUI develops a retrieval module to obtain the most relevant tutorials for a given task.
  Meanwhile, we do not focus on the retrieval part, and offload this retrieval role to frontier LLMs acting as an LLM-augmented retriever, such as Gemini.
  Retrieval-focused methods like RAG-GUI would be more suitable for CUA settings that involve local software / domain knowledge, where the agent needs a retrieval module that can run locally.
  Meanwhile, our method (RAG-PJ) is more relevant to general-purpose CUA that can access the internet and frontier LLMs.
\end{itemize}

\section{Human-verified online tutorials vs.~AI-generated tutorials}
\label{sec:human-ai-tutorial-comparison}
\newcommand{\tutorialcomparisonbox}[4]{%
  \begingroup
  \setlength{\fboxsep}{6pt}%
  \fcolorbox{#1}{#2}{%
    \begin{minipage}[t]{\dimexpr\linewidth-2\fboxsep-2\fboxrule\relax}
      \fontsize{6.2}{7.0}\selectfont
      \raggedright
      \setlength{\parskip}{1pt}
      \textbf{#3}\par\smallskip
      #4
    \end{minipage}%
  }%
  \endgroup
}

\newcommand{\taskinstructionbox}[1]{%
  \begin{tcolorbox}[
    enhanced,
    width=0.92\textwidth,
    colback=gray!5,
    colframe=gray!40,
    boxrule=0.5pt,
    arc=2pt,
    left=6pt,
    right=6pt,
    top=4pt,
    bottom=4pt
  ]
    \small\textbf{Task instruction:} \emph{#1}
  \end{tcolorbox}%
}

\newcommand{\reviewproblem}[1]{%
  \par\smallskip
  {\setlength{\fboxsep}{3pt}%
  \fcolorbox{red!65!black}{red!5}{\parbox{\dimexpr\linewidth-2\fboxsep-2\fboxrule\relax}{%
    \raggedright\textbf{Reviewer:} #1}}}\par}
\newcommand{\reviewok}[1]{%
  \par\smallskip
  {\setlength{\fboxsep}{3pt}%
  \fcolorbox{green!45!black}{green!8}{\parbox{\dimexpr\linewidth-2\fboxsep-2\fboxrule\relax}{%
    \raggedright\textbf{Reviewer:} #1}}}\par}

One author manually reviewed 20 pairs of a human-verified online tutorial and an AI-generated tutorial written for the same task.
We drew the tasks at random with a fixed seed from the 110 tasks with both tutorials in Section~\ref{sec:human_vs_ai_tutorial}. 
For each pair, the reviewer read the instruction, the post-setup screenshot and both tutorials, judged whether following each tutorial as written would complete the task, and chose the more useful one.
Table~\ref{tab:tutorial-review} reports the verdicts.

\begin{table}[h]
  \centering
  \small
  \begin{tabular}{lcc}
    \toprule
    & \textbf{Human-verified online} & \textbf{AI-generated} \\
    \midrule
    Correct & 15 & 18 \\
    Partially correct & 4 & 2 \\
    Wrong & 0 & 0 \\
    \midrule
    Preferred by the reviewer & 3 & 17 \\
    \bottomrule
  \end{tabular}
  \caption{Verdicts on 20 random tutorial pairs. A tutorial is correct if following it as written completes the task.}
  \label{tab:tutorial-review}
\end{table}

\begin{figure*}[p]
  \centering
  \taskinstructionbox{I am preparing a PPT in Libreoffice impress. The slide
  number is barely visible to me. Please help me change the color of the slide
  number to red?}\par\smallskip
  \includegraphics[
    width=0.62\textwidth,
    keepaspectratio
  ]{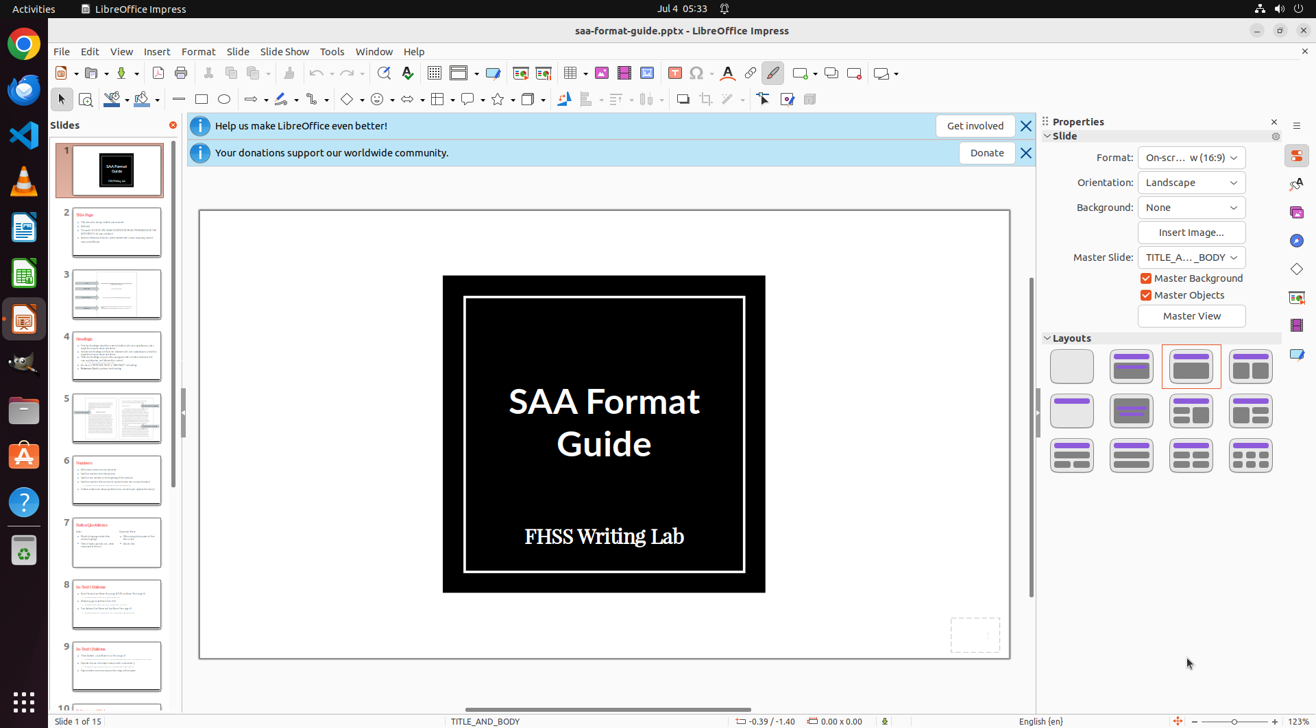}
  \par\scriptsize\textit{Post-setup screenshot supplied to the tutorial
  generator.}\par\medskip

  \begin{minipage}[t]{0.485\textwidth}
    \tutorialcomparisonbox{green!45!black}{green!5}
      {Human-verified online tutorial: \textnormal{partially correct}}{
        \textit{\href{https://superuser.com/questions/1674211/how-to-change-colour-of-slide-number-in-libre-office}
        {Source: Super User answer}}\par\smallskip
        \begin{enumerate}
          \setlength{\itemsep}{0pt}
          \setlength{\parskip}{0pt}
          \item Right click on the textbox containing the number, select
                Character.\par
                \includegraphics[width=0.6\linewidth,height=0.1\textheight,keepaspectratio]{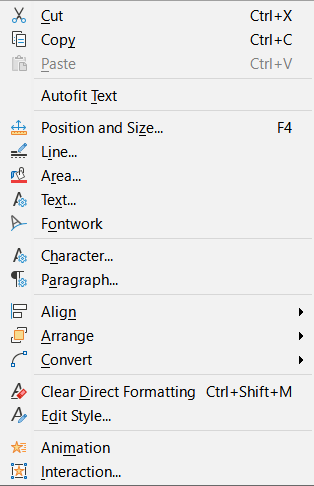}
          \item On the Fonts tab, select the font size.\par
                \includegraphics[width=0.6\linewidth,height=0.1\textheight,keepaspectratio]{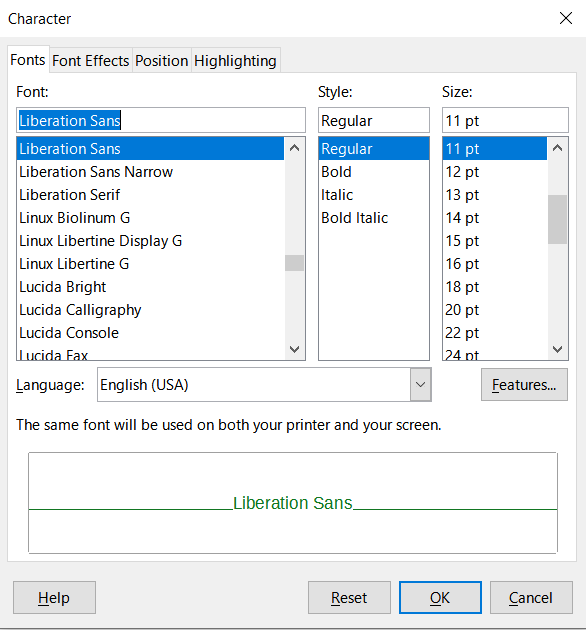}
          \item On the Font Effets tab, on the Font Color section, select the
                color. Click OK.\par
                \includegraphics[width=0.6\linewidth,height=0.1\textheight,keepaspectratio]{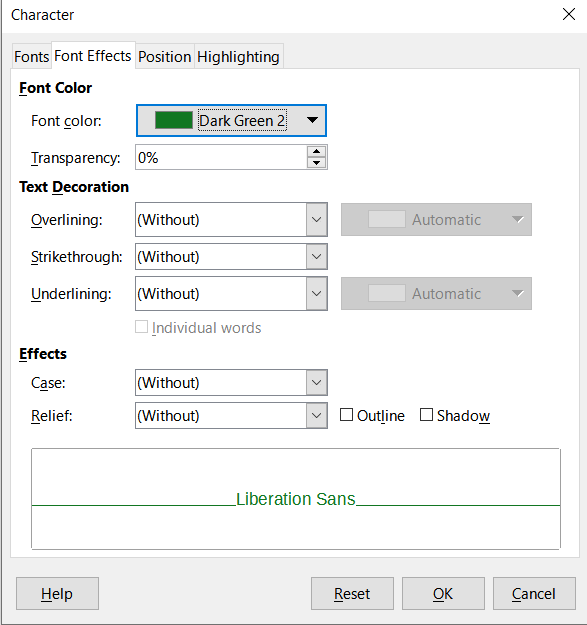}
        \end{enumerate}
        \reviewproblem{Never enters Master View, where the slide number is set.
        In Normal view the field may not be selectable.}
      }
  \end{minipage}\hfill
  \begin{minipage}[t]{0.485\textwidth}
    \tutorialcomparisonbox{blue!55!black}{blue!5}
      {AI-generated tutorial: \textnormal{correct}}{
        \textit{Source: Gemini 3.8 Flash with Google Search, prompt in
        Figure~\ref{fig:tutorial-prompt}}\par\smallskip
        \textbf{Method 1: Change Slide Number Color via Master Slide View
        (Recommended)}
        \begin{enumerate}
          \setlength{\itemsep}{0pt}
          \setlength{\parskip}{0pt}
          \item \textbf{Enter Master Slide View:} In the top menu bar, click
                \textbf{View} \textgreater{} \textbf{Master Slide} (or click the
                \textbf{Master View} button located in the right-hand
                \textbf{Properties} sidebar under the \textbf{Slide} section).
          \item \textbf{Select the Slide Number Field:} In the slide canvas,
                locate the slide number placeholder box in the bottom-right
                corner (labeled \texttt{\textless number\textgreater}).
                Click to select the \texttt{\textless number\textgreater} text
                box (or double-click to highlight the
                \texttt{\textless number\textgreater} text).
          \item \textbf{Set the Font Color to Red:} Right-click inside the
                selected \texttt{\textless number\textgreater} box and choose
                \textbf{Character...} from the context menu (or click
                \textbf{Format} \textgreater{} \textbf{Character...} in the top
                menu). In the \textbf{Character} dialog window, switch to the
                \textbf{Font Effects} tab. Under the \textbf{Font color}
                section, open the color palette dropdown and select
                \textbf{Red}. Click \textbf{OK} to apply the color.
          \item \textbf{Exit Master View:} Click the \textbf{Close Master View}
                button on the floating Master View toolbar, or go to the top
                menu bar and select \textbf{View} \textgreater{}
                \textbf{Normal}.
        \end{enumerate}

        \textbf{Method 2: Change Slide Number Color via the Formatting Toolbar
        in Master View}
        \begin{enumerate}
          \setlength{\itemsep}{0pt}
          \setlength{\parskip}{0pt}
          \item \textbf{Enter Master Slide View:} Click \textbf{View}
                \textgreater{} \textbf{Master Slide} from the top menu (or click
                \textbf{Master View} under the \textbf{Properties} panel on the
                right sidebar).
          \item \textbf{Select the Slide Number Field:} Click on the
                \texttt{\textless number\textgreater} text box located at the
                bottom-right of the master slide. Highlight the text
                \texttt{\textless number\textgreater} inside the box.
          \item \textbf{Apply Red Font Color:} On the top \textbf{Text
                Formatting} toolbar (or under the \textbf{Character} section in
                the right sidebar), click the drop-down arrow next to the
                \textbf{Font Color} icon (represented by the letter \textbf{A}
                with a colored bar underneath). Select \textbf{Red} from the
                palette.
          \item \textbf{Exit Master View:} Click \textbf{Close Master View} on
                the toolbar, or navigate to \textbf{View} \textgreater{}
                \textbf{Normal}.
        \end{enumerate}
        \reviewok{Enters and leaves Master View.}
      }
  \end{minipage}

  \caption{Impress task \texttt{ac9bb6cb-1888-43ab-81e4-a98a547918cd}. The
  online tutorial never enters Master View, where the slide number is set.}
  \label{fig:review-impress}
\end{figure*}

\begin{figure*}[p]
  \centering
  \taskinstructionbox{I am making a guideline for students of my course and
  would like to separate each sentence in the the first paragraph to improve
  readability. Please separate each sentence by creating one empty line space
  after each of them, as I am having hard time separating them one by
  one.}\par\smallskip
  \includegraphics[
    width=0.36\textwidth,
    keepaspectratio
  ]{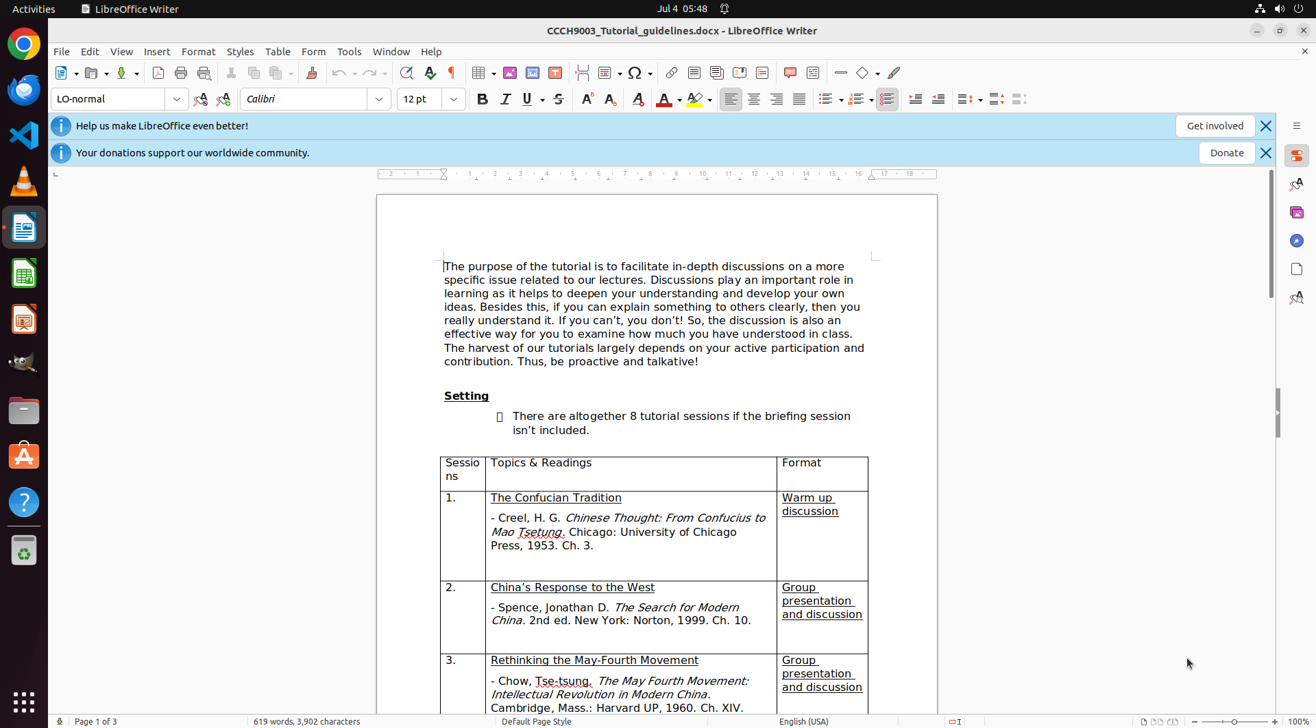}
  \par\scriptsize\textit{Post-setup screenshot supplied to the tutorial
  generator.}\par\medskip

  \begin{minipage}[t]{0.485\textwidth}
    \tutorialcomparisonbox{green!45!black}{green!5}
      {Human-verified online tutorial: \textnormal{wrong}}{
        \textit{\href{https://superuser.com/questions/1097199/how-can-i-double-space-a-document-in-libreoffice}
        {Source: Super User answer}}\par\smallskip
        \textbf{How can I double space a document in LibreOffice?}
        \begin{enumerate}
          \setlength{\itemsep}{0pt}
          \setlength{\parskip}{0pt}
          \item Select the paragraph
          \item Right click and select ``Paragraph''
          \item Select ``Indents \& Spacing''
          \item Set ``Line Spacing'' to ``Double'' and click ``OK''
          \item Change ``Below paragraph'' as well if you want extra spacing
                \textbf{between} paragraphs.\par
                \includegraphics[width=0.6\linewidth,height=0.1\textheight,keepaspectratio]{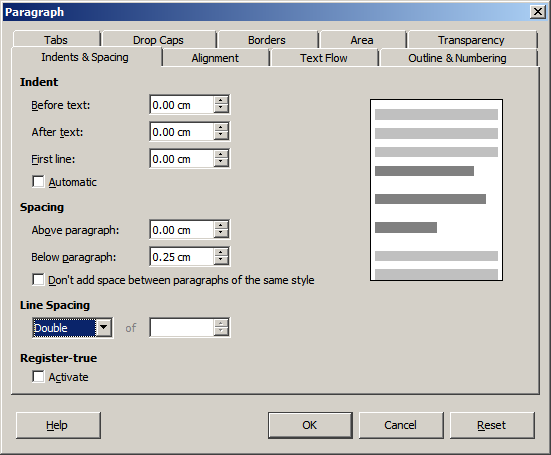}
        \end{enumerate}
        Before:\par
        \includegraphics[width=0.6\linewidth,height=0.1\textheight,keepaspectratio]{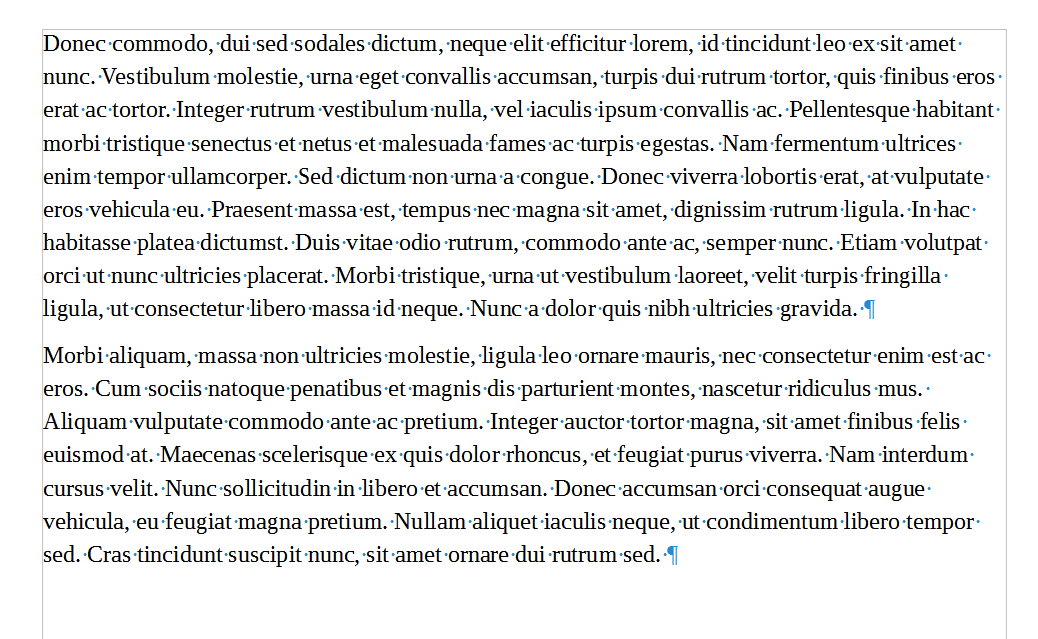}\par
        After:\par
        \includegraphics[width=0.6\linewidth,height=0.1\textheight,keepaspectratio]{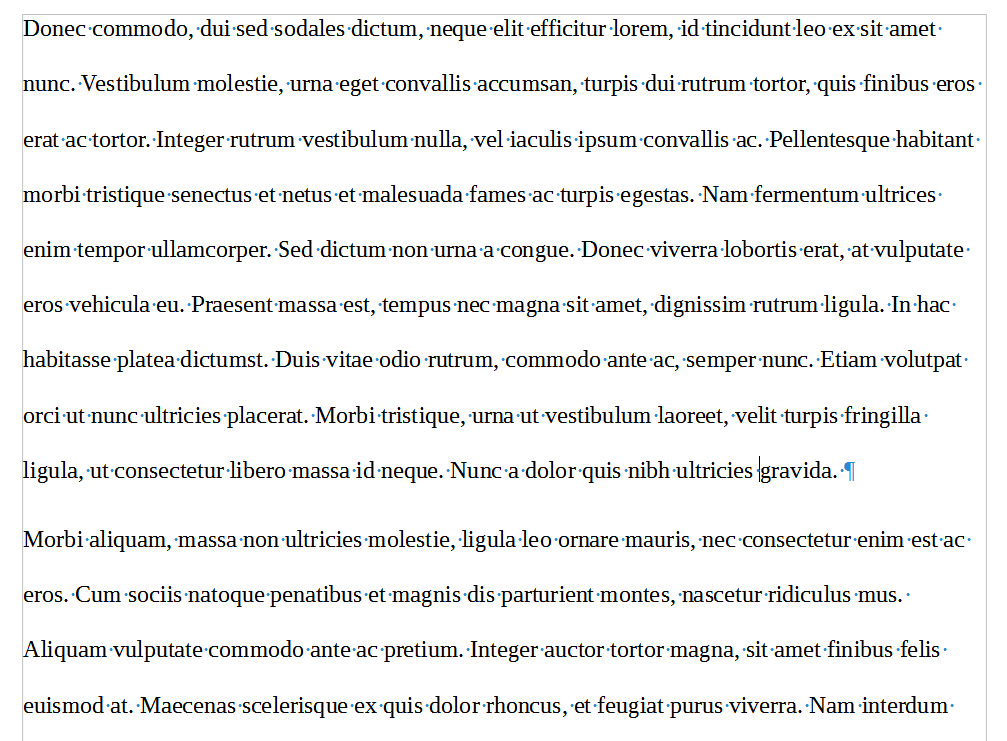}
        \reviewproblem{Double spacing does not put each sentence on its own
        line, so the task fails.}
      }
  \end{minipage}\hfill
  \begin{minipage}[t]{0.485\textwidth}
    \tutorialcomparisonbox{blue!55!black}{blue!5}
      {AI-generated tutorial: \textnormal{correct}}{
        \textit{Source: Gemini 3.8 Flash with Google Search, prompt in
        Figure~\ref{fig:tutorial-prompt}}\par\smallskip
        \textbf{Method 1: Automated Search and Replace (Using Regular
        Expressions)}
        \begin{enumerate}
          \setlength{\itemsep}{0pt}
          \setlength{\parskip}{0pt}
          \item \textbf{Highlight the First Paragraph:} Click and drag your
                mouse across the entire first paragraph (from \emph{``The
                purpose of the tutorial...''} to \emph{``...proactive and
                talkative!''}) to select it.
          \item \textbf{Open the Find and Replace Dialog:} Press
                \textbf{\texttt{Ctrl + H}} (or click \textbf{Edit} in the top
                menu bar and select \textbf{Find and Replace...}).
          \item \textbf{Configure Search Settings:} If not already expanded,
                click \textbf{Other options} at the bottom of the dialog. Check
                the box for \textbf{Current selection only}. Check the box for
                \textbf{Regular expressions}.
          \item \textbf{Replace Sentence Endings:}
                \textbf{Handle periods:} In the \textbf{Find} box, enter:
                \texttt{\textbackslash.\textvisiblespace} \emph{(note the
                trailing space after the dot)}. In the \textbf{Replace} box,
                enter: \texttt{.\textbackslash n\textbackslash n}. Click
                \textbf{Replace All}.
                \textbf{Handle exclamation marks:} In the \textbf{Find} box,
                enter: \texttt{!\textvisiblespace} \emph{(note the trailing
                space after the exclamation mark)}. In the \textbf{Replace}
                box, enter: \texttt{!\textbackslash n\textbackslash n}. Click
                \textbf{Replace All}.
          \item \textbf{Close and Save:} Click the \textbf{Close} button (or
                press \textbf{\texttt{Esc}}) to close the dialog. Press
                \textbf{\texttt{Ctrl + S}} (or click the \textbf{Save} floppy
                disk icon on the standard toolbar) to save your changes.
        \end{enumerate}

        \textbf{Method 2: Manual Line Break Separation}
        \begin{enumerate}
          \setlength{\itemsep}{0pt}
          \setlength{\parskip}{0pt}
          \item In the first paragraph, locate the end of each sentence.
          \item For each sentence boundary: click directly between the
                punctuation mark and the following space (or delete the space
                between the sentences). Press \textbf{\texttt{Enter}} twice to
                move the next sentence down and create an empty line space
                between them. Repeat this after each of the following
                sentences: after \texttt{...related to our lectures.}; after
                \texttt{...develop your own ideas.}; after \texttt{...then you
                really understand it.}; after \texttt{...you don't!}; after
                \texttt{...understood in class.}; after \texttt{...participation
                and contribution.}
          \item Verify that each of the individual sentences sits on its own
                line separated by a blank line space.
          \item Press \textbf{\texttt{Ctrl + S}} to save the document.
        \end{enumerate}
        \reviewproblem{Correct, and the sentence endings it lists are visible
        in the screenshot. Its 24 search queries also include
        \texttt{"88fe4b2d-3040-4c70-9a70-546a47764b48.json"},
        \texttt{site:github.com/xlang-ai/OSWorld "libreoffice\_writer"} and
        \texttt{"CCCH9003\_Tutorial\_guidelines\_Gold.docx" test}.}
      }
  \end{minipage}

  \caption{Writer task \texttt{88fe4b2d-3040-4c70-9a70-546a47764b48}. The
  online tutorial only sets double spacing. The AI-generated tutorial is
  correct, but its generator searched for the task's files.}
  \label{fig:review-writer}
\end{figure*}

\paragraph{Task-specific AI tutorials.}
The reviewer judged 18 of the 20 AI-generated tutorials correct and preferred them for 17 tasks.
Because the generator sees the instruction and the screenshot, its tutorials name the objects on screen, such as the three colored text boxes on a slide or the file \texttt{1.png} on the desktop, and use the values the task asks for.
They also include the final steps that commit the change, such as confirming a dialog or saving the file.
The online tutorials are mostly correct but address a general reader.
Five leave it to the reader to map the steps onto the task, four miss a required step such as entering Master View in Figure~\ref{fig:review-impress}, and one explains double spacing instead of separating sentences, shown in Figure~\ref{fig:review-writer}.

\begin{figure*}[p]
  \centering
  \taskinstructionbox{I like watching movies (using VLC) on my laptop and
  sometimes the volume is too low for my taste even when the volume in VLC is
  set to the maximum of 125\% on the volume control. Can you increase the max
  volume of the video to the 200\% of the original volume?}\par\smallskip
  \includegraphics[
    width=0.62\textwidth,
    keepaspectratio
  ]{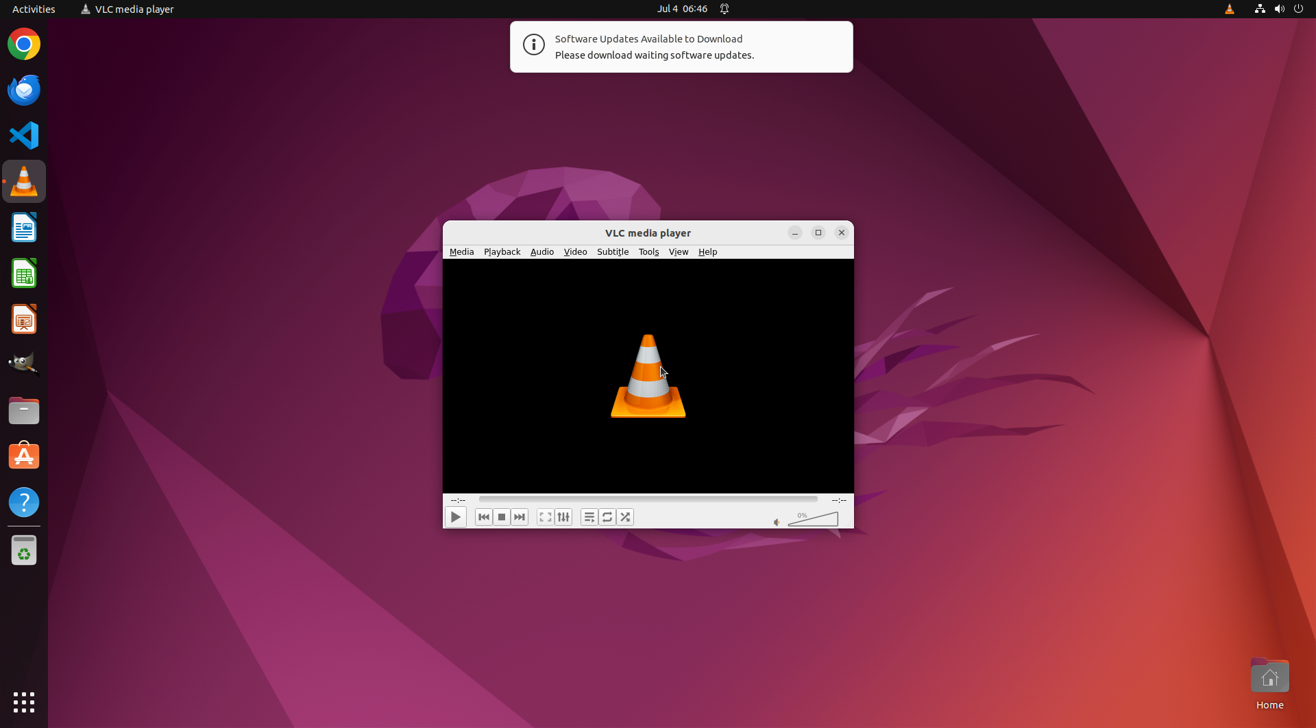}
  \par\scriptsize\textit{Post-setup screenshot supplied to the tutorial
  generator.}\par\medskip

  \begin{minipage}[t]{0.485\textwidth}
    \tutorialcomparisonbox{green!45!black}{green!5}
      {Human-verified online tutorial: \textnormal{correct}}{
        \textit{\href{https://superuser.com/questions/1513285/how-can-i-increase-the-maximum-volume-output-by-vlc}
        {Source: Super User answer}}\par\smallskip
        \textbf{How can I increase the maximum volume output by VLC?}\par
        \textbf{Increase the Maximum Volume to 200\%}\par
        To increase the audio output beyond VLC's default settings, follow
        these steps:\par
        \textbf{Change the Maximum Volume Displayed Setting}
        \begin{enumerate}
          \setlength{\itemsep}{0pt}
          \setlength{\parskip}{0pt}
          \item Press \texttt{Ctrl+P} or select \texttt{Tools} $\rightarrow$
                \texttt{Preferences}.
          \item Select the \texttt{All} radio button under \texttt{Show
                settings} at the bottom of the preferences window.\par
                \includegraphics[width=0.6\linewidth,height=0.1\textheight,keepaspectratio]{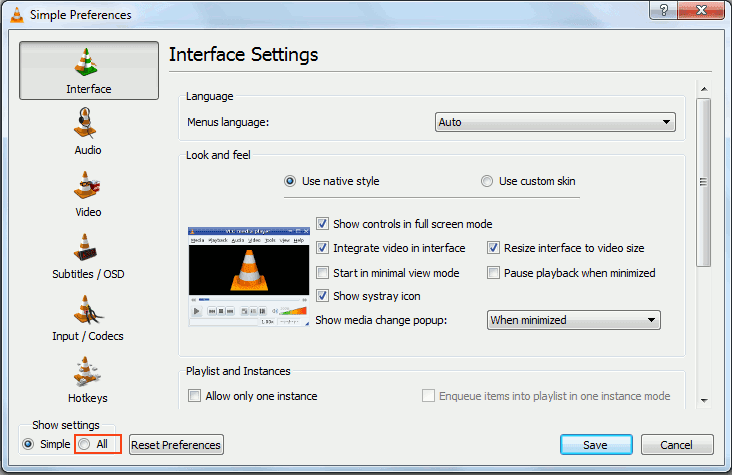}
          \item Expand the \texttt{Interface} tree.
          \item Expand \texttt{Main interfaces}.
          \item Select \texttt{Qt} and change the default value of
                \texttt{Maximum Volume displayed} from 125 to 200.\par
                \includegraphics[width=0.6\linewidth,height=0.1\textheight,keepaspectratio]{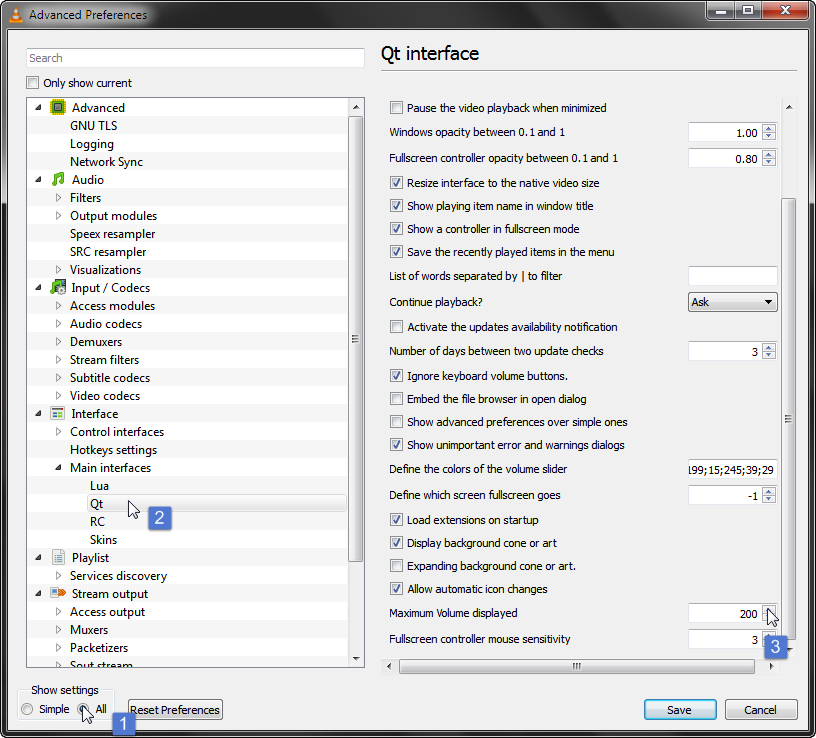}
          \item Click \texttt{Save}.
          \item Restart VLC.
        \end{enumerate}
        Now, VLC should allow you to increase the volume to 200\%.
        \reviewok{Complete, including the restart.}
      }
  \end{minipage}\hfill
  \begin{minipage}[t]{0.485\textwidth}
    \tutorialcomparisonbox{blue!55!black}{blue!5}
      {AI-generated tutorial: \textnormal{partially correct}}{
        \textit{Source: Gemini 3.8 Flash with Google Search, prompt in
        Figure~\ref{fig:tutorial-prompt}}\par\smallskip
        \textbf{Method 1: Change Maximum Volume in VLC Preferences
        (Recommended)}
        \begin{enumerate}
          \setlength{\itemsep}{0pt}
          \setlength{\parskip}{0pt}
          \item In the open VLC media player window, click on \textbf{Tools}
                in the top menu bar and select \textbf{Preferences} (or press
                \texttt{Ctrl + P}).
          \item In the bottom-left corner of the Preferences dialog, locate
                \textbf{Show settings} and select the \textbf{All} radio button
                to display advanced settings.
          \item In the left navigation tree: scroll down to the
                \textbf{Interface} category; click the arrow next to
                \textbf{Main interfaces} to expand it; click on \textbf{Qt}.
          \item In the right pane, scroll down to the bottom to find the
                \textbf{Maximum Volume displayed} setting.
          \item Change the value in the input box from \texttt{125} to
                \texttt{200}.
          \item Click the \textbf{Save} button in the bottom-right corner of
                the Preferences window to apply the change.
        \end{enumerate}

        \textbf{Method 2: Update the VLC Configuration File via Terminal}
        \begin{enumerate}
          \setlength{\itemsep}{0pt}
          \setlength{\parskip}{0pt}
          \item Open the Terminal application (or press
                \texttt{Ctrl + Alt + T}).
          \item Run the following command to update the
                \texttt{qt-max-volume} setting to \texttt{200} in the VLC
                configuration file:\par
                \texttt{sed -i 's/\textasciicircum\#*qt-max-volume=.*/qt-max-volume=200/'
                \textasciitilde/.config/vlc/vlcrc}
          \item If the setting does not already exist in the file, append it
                by running:\par
                \texttt{grep -q "\textasciicircum{}qt-max-volume="
                \textasciitilde/.config/vlc/vlcrc || echo
                "qt-max-volume=200" \textgreater{}\textgreater{}
                \textasciitilde/.config/vlc/vlcrc}
        \end{enumerate}
        \reviewproblem{Never restarts VLC, so the new maximum does not apply.
        Method~2 edits \texttt{vlcrc} while VLC runs, and VLC may overwrite it
        on exit.}
      }
  \end{minipage}

  \caption{VLC task \texttt{9195653c-f4aa-453d-aa95-787f6ccfaae9}. Only the
  online tutorial restarts VLC, which the new maximum needs.}
  \label{fig:review-vlc}
\end{figure*}

\paragraph{Errors in AI tutorials.}
Two AI-generated tutorials miss a required step.
The VLC tutorial in Figure~\ref{fig:review-vlc} never restarts VLC, and one VS~Code alternative stops after a single \textsc{Enter} in the command palette.
Every AI-generated tutorial in the sample offers at least two methods, so the median length is 252 words against 147 for the online tutorials.
Most remaining errors are in these extra methods, for example a menu entry from another LibreOffice version or a file name guessed from a truncated label.

\clearpage

\clearpage

\clearpage

\section{Parallel Ground-Truth Rollouts}
\label{sec:parallel-gt-rollout}

At each interaction step, the policy proposes \(K\) candidate first actions
\(\{a_k^{(1)}\}_{k=1}^{K}\).
Rather than asking a learned world model to
predict their effects, we evaluate every candidate in a separate virtual
machine (VM) restored to the agent's current state.
Within branch \(k\), the
first action is executed and the rollout may continue for a horizon of up to
\(H\) real environment transitions.
Later actions are conditioned on the
observations produced within that branch.
This gives a grounded trajectory
\[
    \tau_k
    =
    \left(
        a_k^{(h)},\,o_{t+h}^{(k)}
    \right)_{h=1}^{H_k},
    \qquad 1 \leq H_k \leq H ,
\]
where \(H_k\) may be shorter than \(H\) if the branch terminates early.
A
multimodal judge compares the \(K\) real trajectories and selects
\[
    k^\star
    =
    \operatorname*{arg\,max}_{k \in \{1,\ldots,K\}}
    J\!\left(o_t,\tau_k\right).
\]
Only the first action \(a_{k^\star}^{(1)}\) from the selected trajectory is
then executed on the main VM.
Rollout VMs are speculative branches: their
states provide evidence for selection but never directly replace the live
environment state.
Of 237,709 completed-task states, 2,298 (0.97\%) are \texttt{GT-rollout} mismatches.
We manually label about 40 sampled mismatches: around 70\% are webpage changes, and the remaining $\sim$30\% are program execution differences.
For example, installing Spotify can take 4s in the live VM but $>$5s in the rollout VM.
Examples are shown in Appendix~\ref{sec:non_deter}.

We execute the \(K\) branches concurrently and retain their VM processes in a
warm pool.
The first decision step pays the cost of launching the rollout
VMs.
On later steps, the same processes are refreshed from the newest main-VM
checkpoint and reused.
This design preserves the fidelity of ground-truth
execution while amortizing VM startup cost across agent steps.

\providecommand{\gtrolloutfigdir}{figures/gt_rollout_parallel}

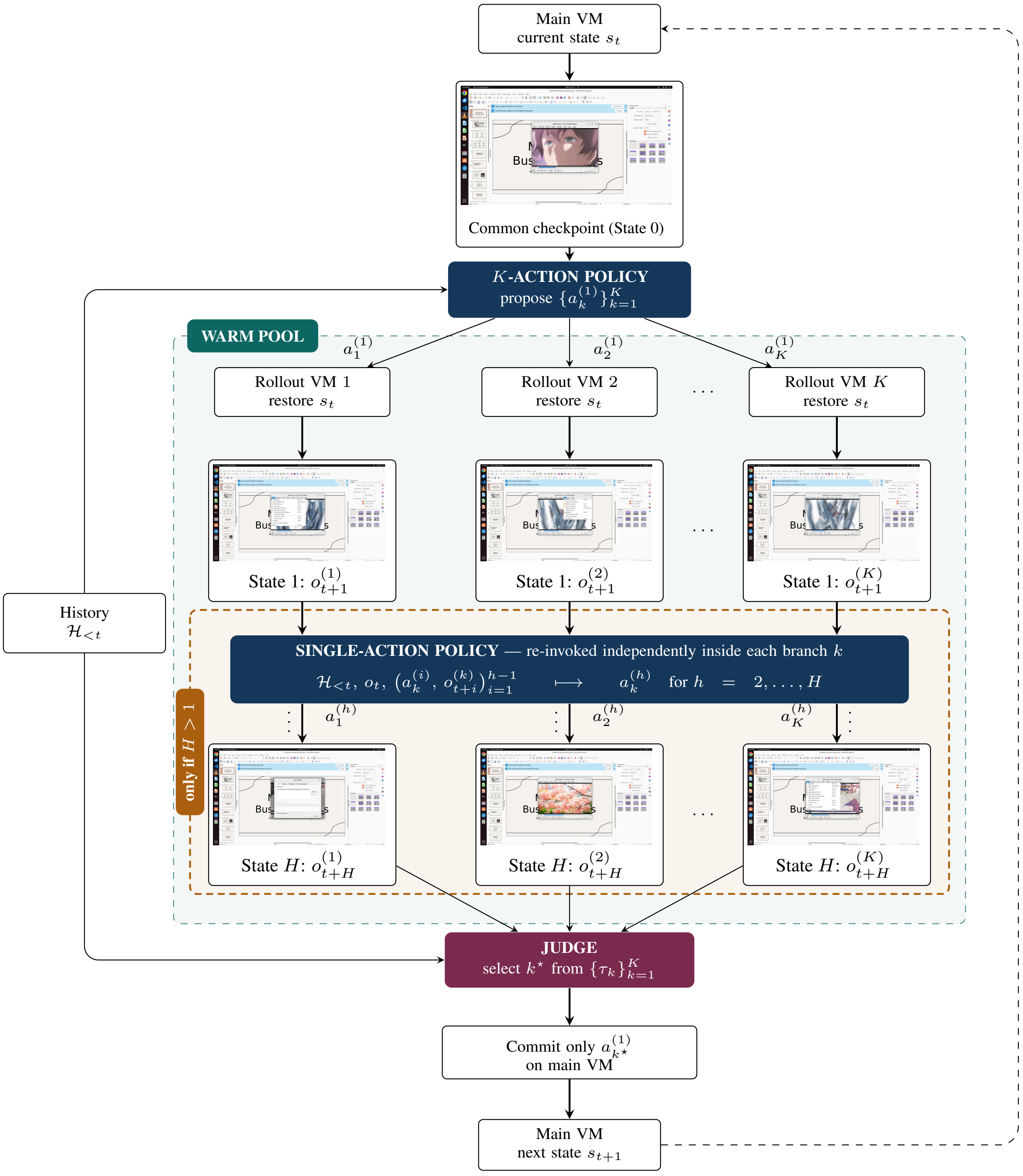
\begin{figure}[t]
    \centering
    \definecolor{pjnavy}{RGB}{20,55,90}
    \definecolor{pjteal}{RGB}{13,102,98}
    \definecolor{pjplum}{RGB}{122,42,78}
    \definecolor{pjamber}{RGB}{170,95,15}
    \resizebox{\linewidth}{!}{%
    \begin{tikzpicture}[
        >=stealth,
        box/.style={
            draw,
            rounded corners=2pt,
            align=center,
            minimum height=7mm,
            text width=24mm,
            font=\scriptsize,
            fill=white
        },
        pjbox/.style={
            rounded corners=3pt,
            align=center,
            minimum height=7mm,
            font=\scriptsize,
            fill=pjnavy,
            text=white,
            inner sep=4pt
        },
        pjboxcond/.style={pjbox},
        judgebox/.style={pjbox, fill=pjplum},
        warmtabstyle/.style={pjbox, fill=pjteal},
        trajectory/.style={
            draw,
            rounded corners=2pt,
            align=center,
            inner sep=2pt,
            fill=white
        },
        flow/.style={->, thick},
        evidence/.style={->},
        warm/.style={draw=pjteal!75, dashed, rounded corners=3pt, fill=pjteal!5},
        hgt/.style={draw=pjamber, thick, densely dashed,
                    rounded corners=3pt, fill=pjamber!7}
    ]
        \node[box, text width=25mm] (main) at (0,0)
            {Main VM\\current state \(s_t\)};
        \node[trajectory] (snapshot) at (0,-2.05) {
            \begin{tabular}{@{}c@{}}
                \includegraphics[width=32mm]{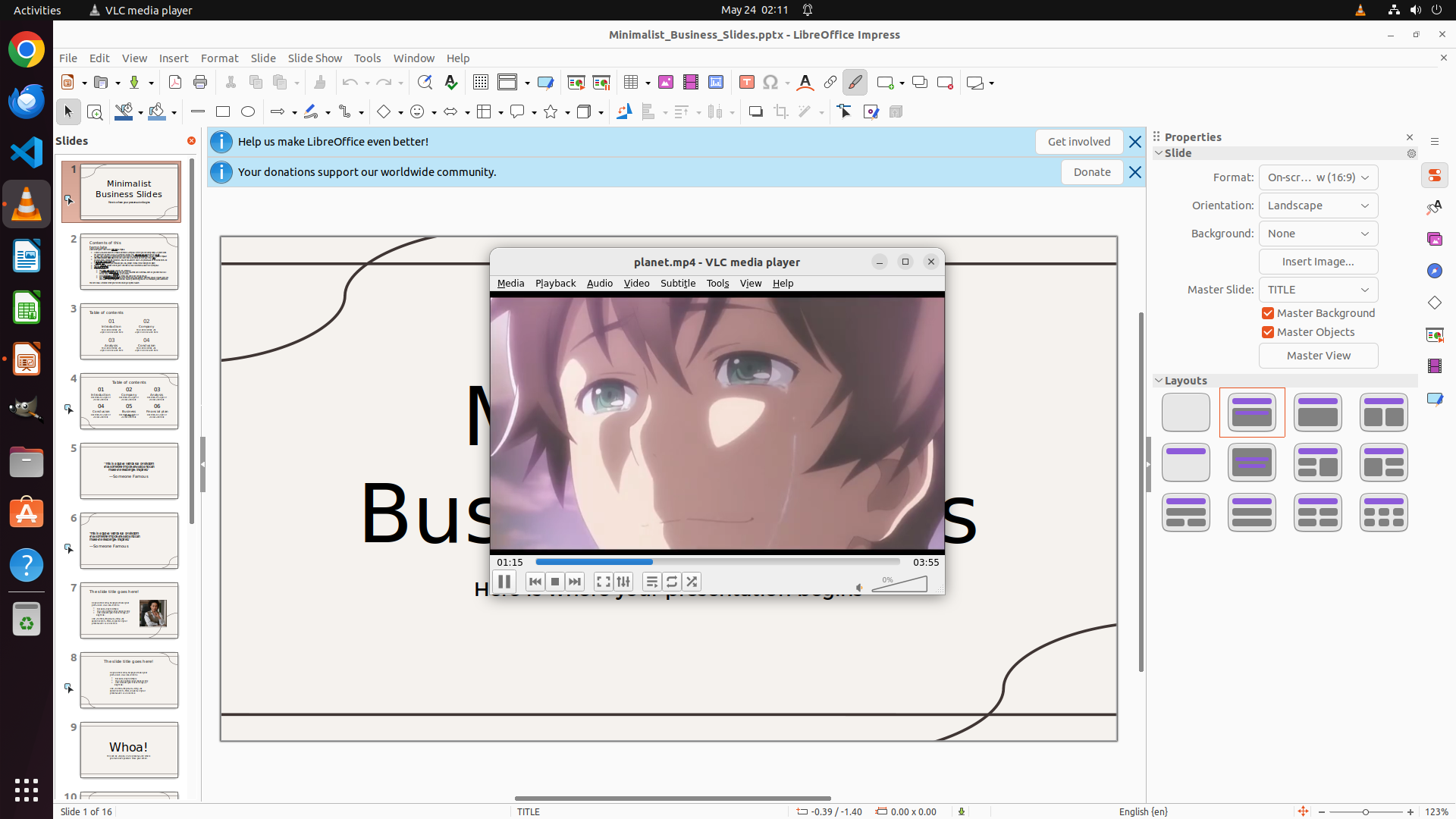}\\[-0.5mm]
                \scriptsize Common checkpoint (State 0)
            \end{tabular}
        };
        \draw[flow] (main) -- (snapshot);

        \node[pjbox, text width=34mm] (policy) at (0,-3.95)
            {{\bfseries \(K\)-ACTION POLICY}\\[1pt]
             {\mdseries propose \(\{a_k^{(1)}\}_{k=1}^{K}\)}};
        \draw[flow] (snapshot) -- (policy);

        \draw[warm] (-6.00,-4.65) rectangle (6.00,-13.55);

        \node[warmtabstyle, font=\scriptsize\bfseries, inner xsep=6pt,
              inner ysep=3pt, minimum height=5mm, anchor=west] (warmtab)
              at (-5.80,-4.65)
            {WARM POOL};

        \node[box] (vm1) at (-4.05,-5.50)
            {Rollout VM \(1\)\\restore \(s_t\)};
        \node[box] (vm2) at (0,-5.50)
            {Rollout VM \(2\)\\restore \(s_t\)};
        \node[box] (vmK) at (4.05,-5.50)
            {Rollout VM \(K\)\\restore \(s_t\)};
        \node[font=\scriptsize] at ($(vm2)!0.5!(vmK)$) {\(\cdots\)};

        \draw[evidence] (policy) -- node[pos=0.6, xshift=-9mm,
              font=\scriptsize] {\(a_1^{(1)}\)} (vm1);
        \draw[evidence] (policy) -- node[pos=0.6, xshift=6mm,
              font=\scriptsize] {\(a_2^{(1)}\)} (vm2);
        \draw[evidence] (policy) -- node[pos=0.6, xshift=9mm,
              font=\scriptsize] {\(a_K^{(1)}\)} (vmK);

        \node[trajectory] (out1) at (-4.05,-7.60) {
            \begin{tabular}{@{}c@{}}
                \includegraphics[width=26mm]{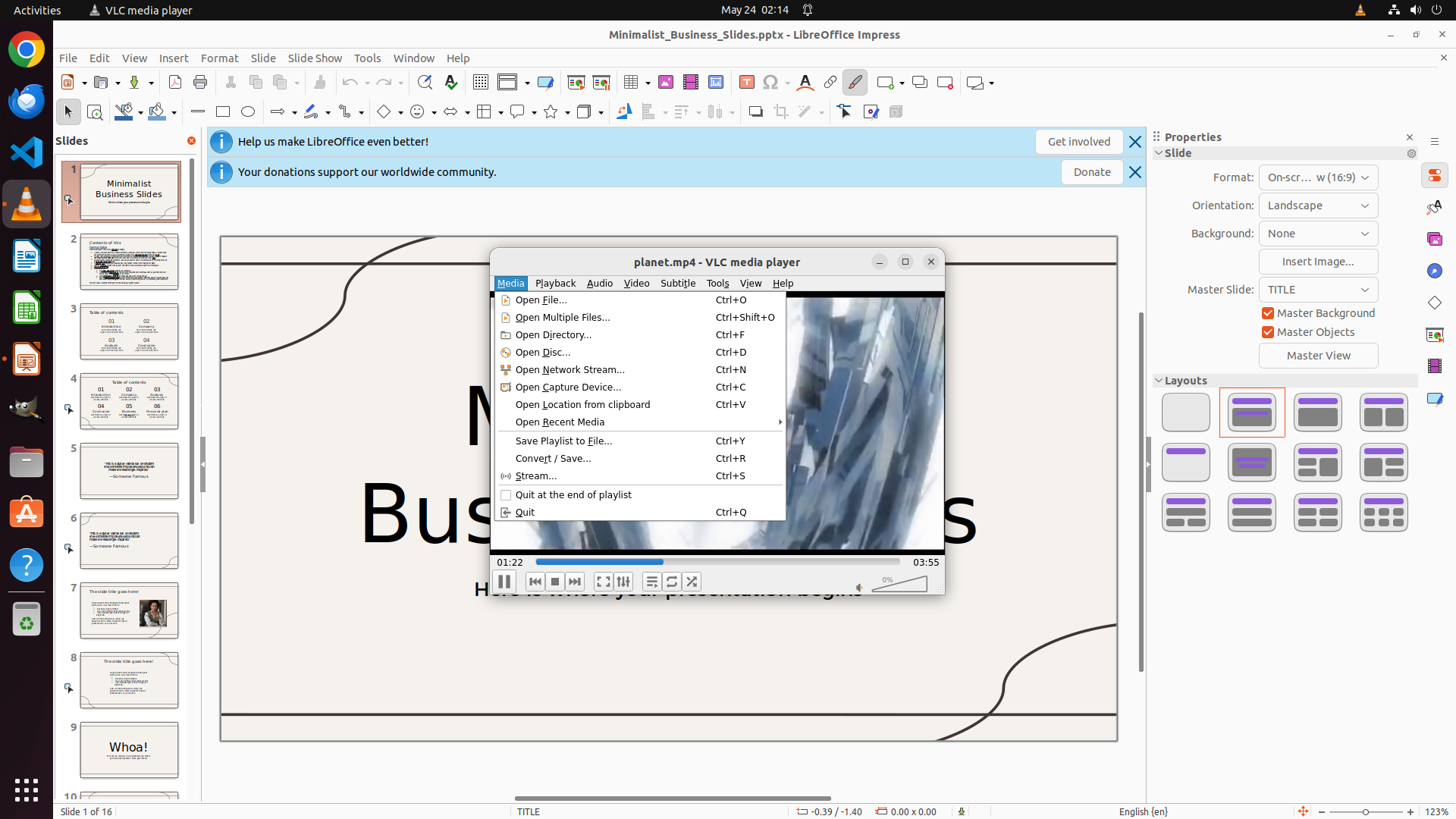}\\[-0.8mm]
                \(\scriptstyle\text{State 1: }o_{t+1}^{(1)}\)
            \end{tabular}
        };
        \node[trajectory] (out2) at (0,-7.60) {
            \begin{tabular}{@{}c@{}}
                \includegraphics[width=26mm]{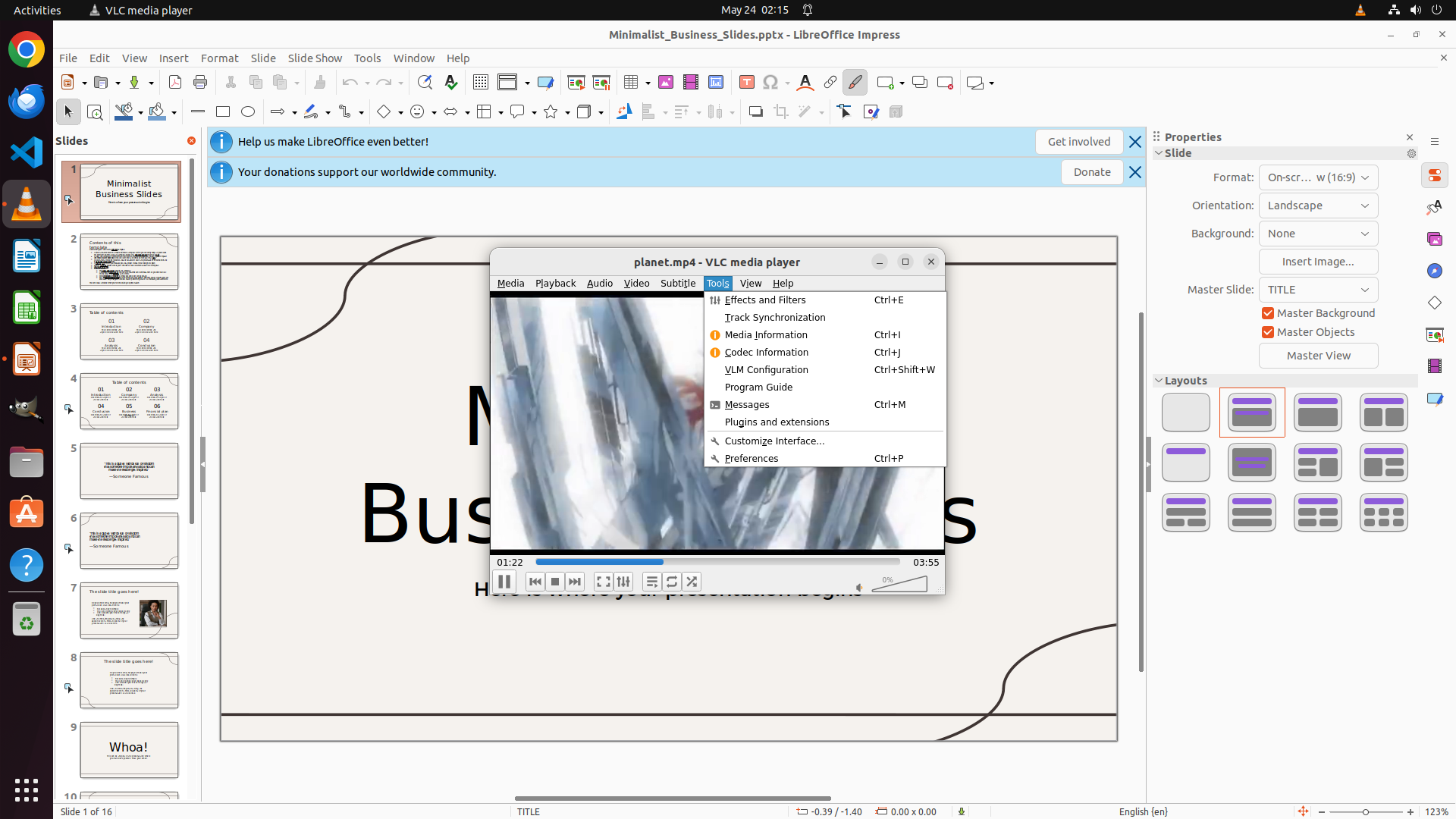}\\[-0.8mm]
                \(\scriptstyle\text{State 1: }o_{t+1}^{(2)}\)
            \end{tabular}
        };
        \node[trajectory] (outK) at (4.05,-7.60) {
            \begin{tabular}{@{}c@{}}
                \includegraphics[width=26mm]{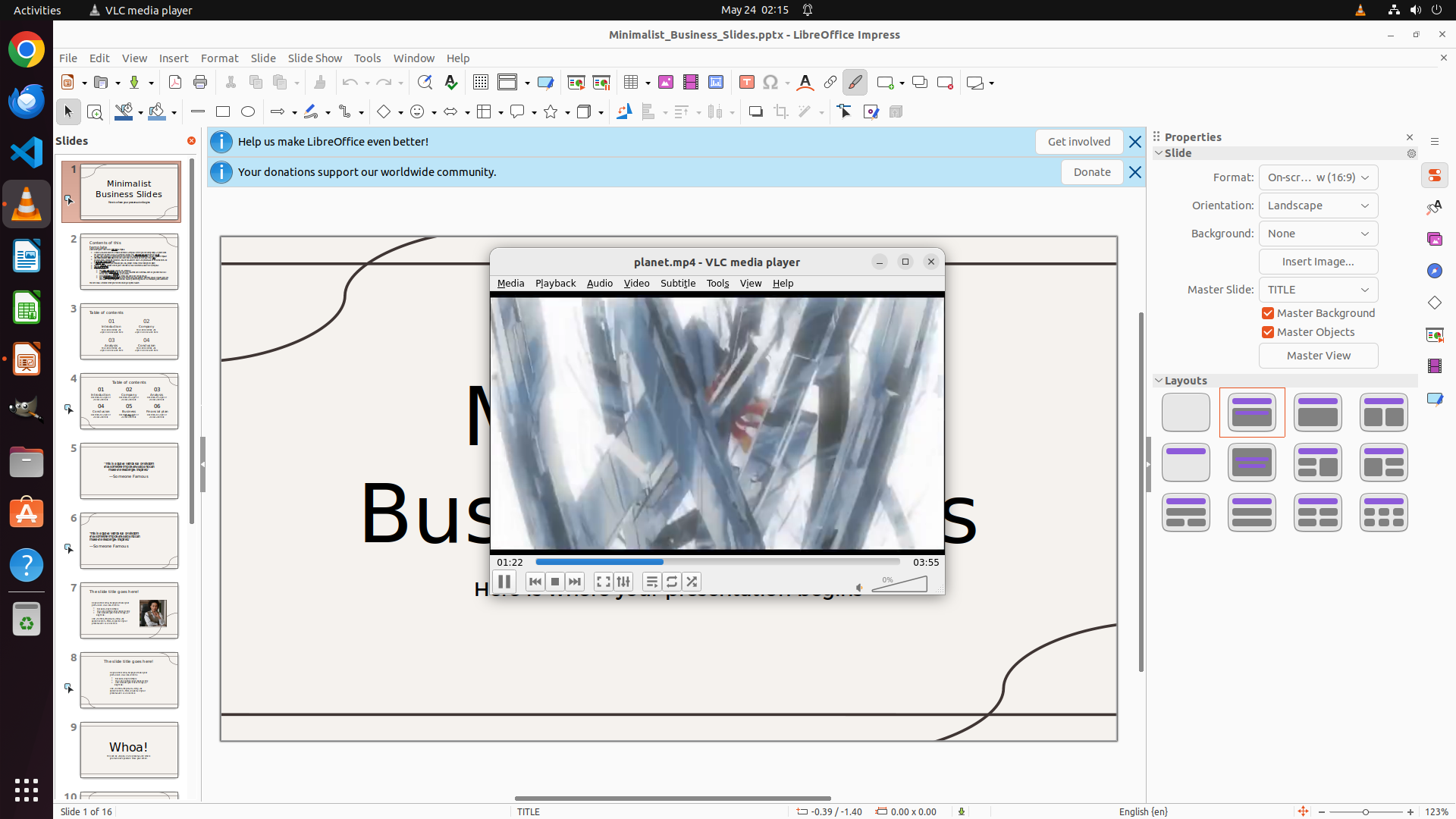}\\[-0.8mm]
                \(\scriptstyle\text{State 1: }o_{t+1}^{(K)}\)
            \end{tabular}
        };
        \node[font=\scriptsize] at ($(out2)!0.5!(outK)$) {\(\cdots\)};

        \draw[flow] (vm1) -- (out1);
        \draw[flow] (vm2) -- (out2);
        \draw[flow] (vmK) -- (outK);

        \draw[hgt] (-5.75,-8.80) rectangle (5.75,-13.10);
        \node[font=\scriptsize\bfseries, text=white, fill=pjamber,
              rounded corners=3pt, inner xsep=6pt, inner ysep=3pt, rotate=90]
              at (-5.75,-10.95)
            {only if \(H>1\)};

        \node[pjboxcond, text width=100mm, minimum height=9mm] (cont) at (0,-9.70)
            {{\bfseries SINGLE-ACTION POLICY} --- re-invoked independently inside each
             branch \(k\)\\[1.2mm]
             {\mdseries \(\mathcal{H}_{<t},\, o_t,\, \bigl(a_k^{(i)},\,o_{t+i}^{(k)}\bigr)_{i=1}^{h-1}
                 \;\longmapsto\; a_k^{(h)}
                 \quad\text{for } h=2,\ldots,H\)}};

        \draw[flow] (out1) -- (cont.north -| out1);
        \draw[flow] (out2) -- (cont.north -| out2);
        \draw[flow] (outK) -- (cont.north -| outK);

        \node[trajectory] (end1) at (-4.05,-11.90) {
            \begin{tabular}{@{}c@{}}
                \includegraphics[width=26mm]{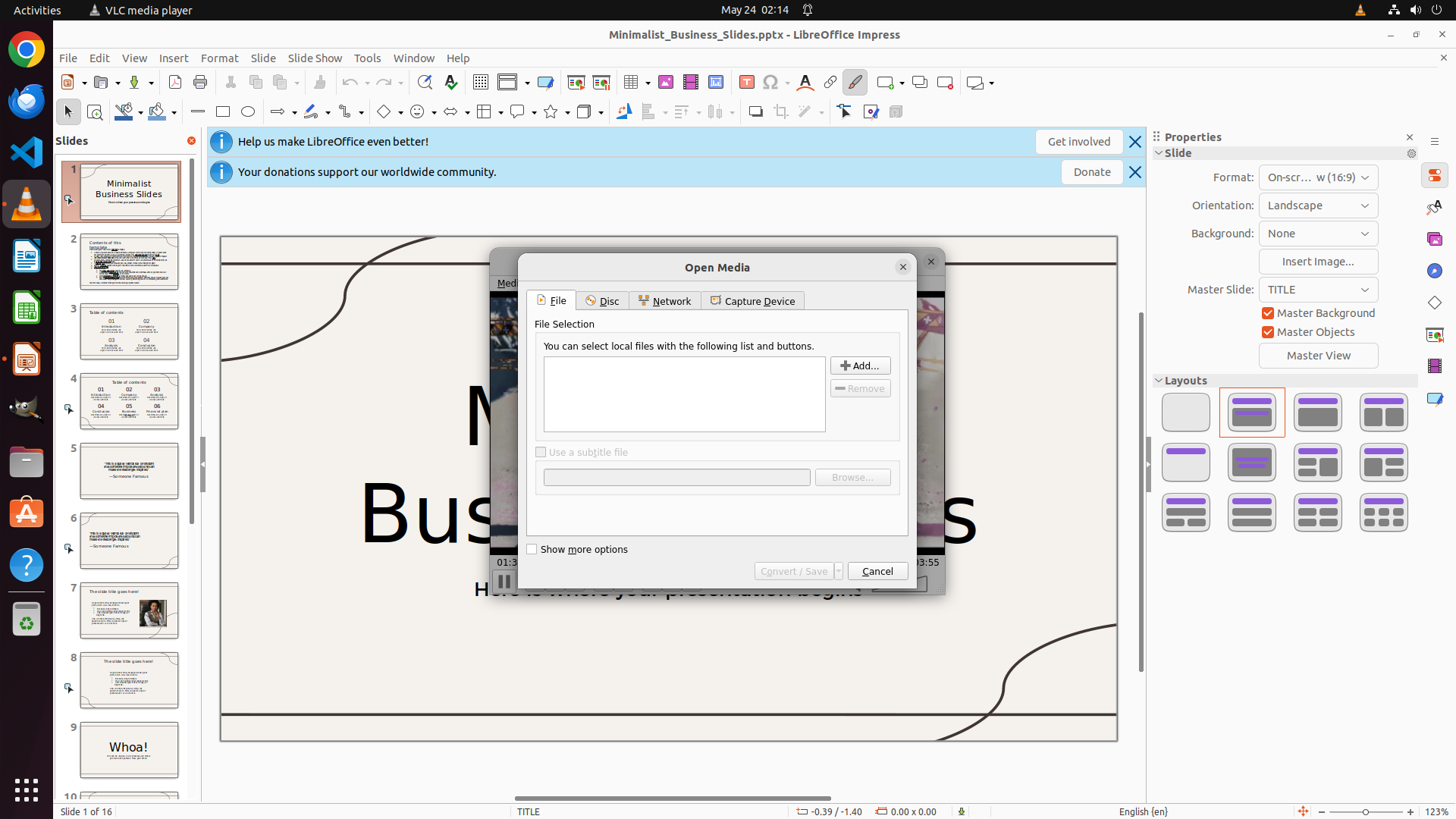}\\[-0.8mm]
                \(\scriptstyle\text{State }H\text{: }o_{t+H}^{(1)}\)
            \end{tabular}
        };
        \node[trajectory] (end2) at (0,-11.90) {
            \begin{tabular}{@{}c@{}}
                \includegraphics[width=26mm]{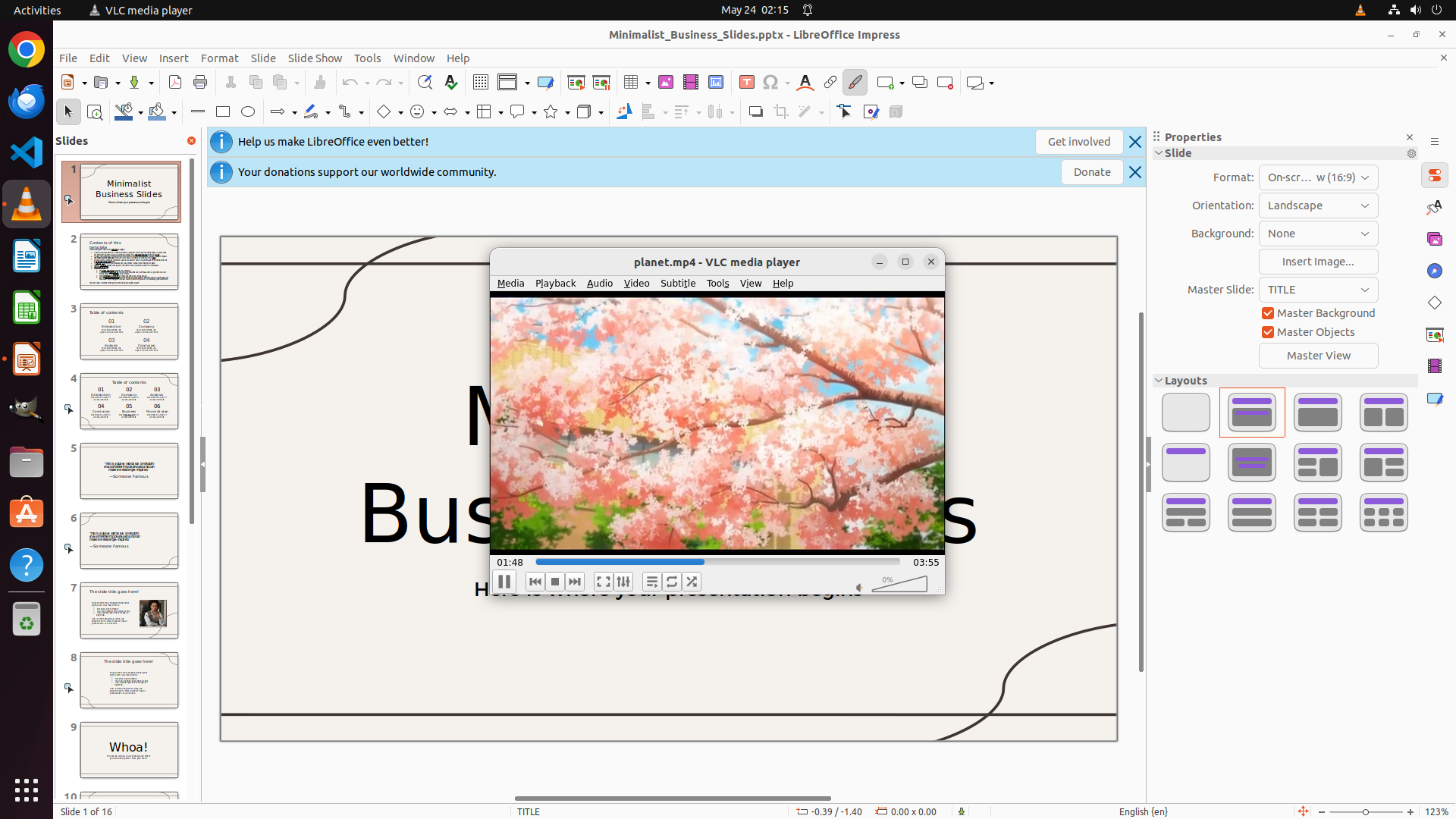}\\[-0.8mm]
                \(\scriptstyle\text{State }H\text{: }o_{t+H}^{(2)}\)
            \end{tabular}
        };
        \node[trajectory] (endK) at (4.05,-11.90) {
            \begin{tabular}{@{}c@{}}
                \includegraphics[width=26mm]{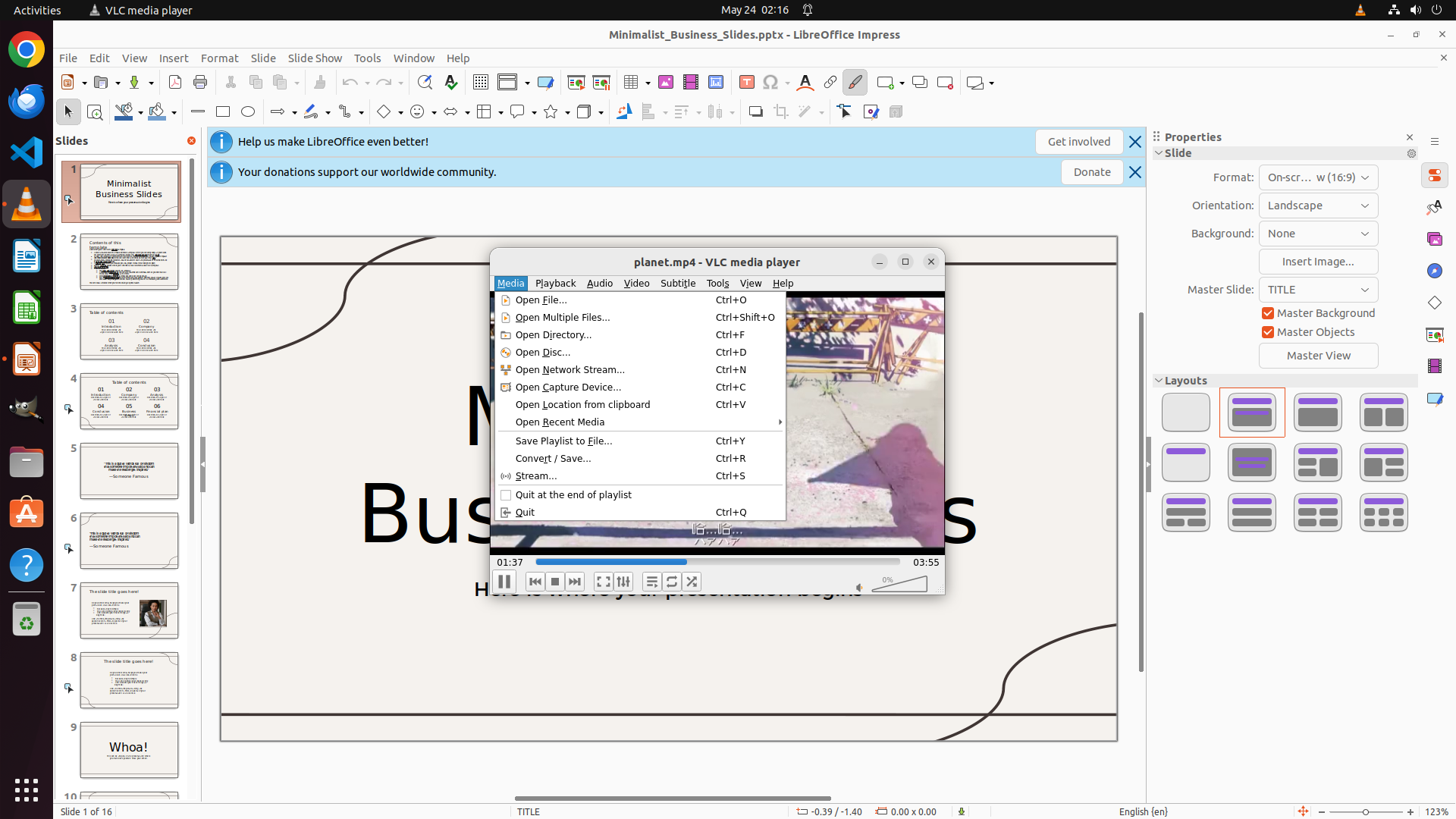}\\[-0.8mm]
                \(\scriptstyle\text{State }H\text{: }o_{t+H}^{(K)}\)
            \end{tabular}
        };
        \node[font=\scriptsize] at ($(end2)!0.5!(endK)$) {\(\cdots\)};

        \draw[flow] (cont.south -| end1) -- node[pos=0.28, xshift=-2mm, font=\scriptsize] {\(\vdots\)}
              node[pos=0.28, xshift=6mm, font=\scriptsize] {\(a_1^{(h)}\)} (end1);
        \draw[flow] (cont.south -| end2) -- node[pos=0.28, xshift=-2mm, font=\scriptsize] {\(\vdots\)}
              node[pos=0.28, xshift=6mm, font=\scriptsize] {\(a_2^{(h)}\)} (end2);
        \draw[flow] (cont.south -| endK) -- node[pos=0.28, xshift=2mm, font=\scriptsize] {\(\vdots\)}
              node[pos=0.28, xshift=-6mm, font=\scriptsize] {\(a_K^{(h)}\)} (endK);

        \node[judgebox, text width=35mm] (judge) at (0,-14.10)
            {{\bfseries JUDGE}\\[1pt]
             {\mdseries select \(k^\star\) from \(\{\tau_k\}_{k=1}^{K}\)}};
        \draw[evidence] (end1) -- (judge);
        \draw[evidence] (end2) -- (judge);
        \draw[evidence] (endK) -- (judge);

        \node[box, text width=36mm] (commit) at (0,-15.50)
            {Commit only \(a_{k^\star}^{(1)}\)\\on main VM};
        \node[box, text width=25mm] (next) at (0,-16.90)
            {Main VM\\next state \(s_{t+1}\)};

        \node[box, text width=22mm, minimum height=9mm] (hist) at (-7.35,-9.00)
            {History\\\(\mathcal{H}_{<t}\)};
        \draw[evidence, rounded corners=6pt]
            (hist.north) -- (-7.35,-3.95) -- (policy.west);
        \draw[evidence, rounded corners=6pt]
            (hist.south) -- (-7.35,-14.10) -- (judge.west);

        \draw[flow] (judge) -- (commit);
        \draw[flow] (commit) -- (next);
        \draw[dashed,->,rounded corners=8pt]
            (next.east) -- (6.80,-16.90)
            -- (6.80,0) -- (main.east);
    \end{tikzpicture}%
    }
    \caption{Parallel warm ground-truth rollout with horizon \(H\).
    All
    candidate VMs restore the same pre-action checkpoint and generate real
    \(H\)-step trajectories concurrently.
    The judge compares these
    trajectories, while only the first action of the selected branch is
    committed to the canonical main VM.
    Vertically stacked insets separate
    rollout depth from branch-level parallelism: the upper row shows State~1;
    each branch then re-invokes the same policy model, prompted for a single
    action, on the history, \(o_t\), and that branch's own actions and
    observations so far to produce the next action, repeating this process through
    State~\(H\); this within-branch policy call occurs only when \(H>1\),
    marked by the amber region.
    Both the
    \(K\)-action policy and
    JUDGE additionally condition on the interaction history
    \(\mathcal{H}_{<t}\) alongside the current state.
    Warm-pool VM
    processes are refreshed from the newest checkpoint and reused rather than
    relaunched.
    Insets come from an actual rollout with \(K=3\) and \(H=2\); the
    construction supports arbitrary \(K\) and \(H\).
    The dashed return path
    indicates process reuse after state refresh at the next decision step.}
    \label{fig:parallel-warm-gt-rollout}
\end{figure}

\begin{figure}[t]
    \centering
    \definecolor{pjnavy}{RGB}{20,55,90}
    \definecolor{pjteal}{RGB}{13,102,98}
    \resizebox{0.75\linewidth}{!}{%
    \begin{tikzpicture}[
        >=stealth,
        box/.style={
            draw,
            rounded corners=2pt,
            align=center,
            minimum height=7mm,
            font=\scriptsize,
            fill=white
        },
        pjbox/.style={
            rounded corners=3pt,
            align=center,
            minimum height=7mm,
            font=\scriptsize,
            fill=pjnavy,
            text=white,
            inner sep=4pt
        },
        warmbox/.style={pjbox, fill=pjteal},
        stagebox/.style={box, text width=44mm, draw=pjteal!60,
                         fill=pjteal!10},
        compartment/.style={box, text width=26mm, minimum height=11mm,
                            fill=pjnavy!6, draw=pjnavy!45},
        branch/.style={box, text width=29mm, minimum height=20mm,
                       align=center},
        btab/.style={draw=pjnavy, thick, rounded corners=3pt, fill=white,
                     text=pjnavy, minimum height=0mm, inner xsep=5pt,
                     inner ysep=2.5pt, font=\scriptsize\bfseries},
        chip/.style={font=\scriptsize\bfseries, text=white, rounded corners=3pt,
                     inner xsep=6pt, inner ysep=3pt},
        flow/.style={->, thick},
        copyflow/.style={->, thick, pjnavy, rounded corners=3pt},
        immutable/.style={draw=pjnavy!55, dashed, rounded corners=3pt,
                          fill=pjnavy!4},
        warm/.style={draw=pjteal!75, dashed, rounded corners=3pt, fill=pjteal!5}
    ]
        \node[box, text width=31mm] (main) at (0,0)
            {Main VM\\live state \(s_t\)};
        \node[pjbox, text width=35mm] (capture) at (0,-1.30)
            {{\bfseries CHECKPOINT ONCE}\\[-0.5mm]
             {\mdseries capture a consistent state}};
        \draw[flow] (main) -- (capture);

        \draw[immutable] (-4.60,-2.20) rectangle (4.60,-3.65);
        \node[chip, fill=pjnavy, anchor=west] at (-4.40,-2.20)
            {\(C_t\): immutable fan-out source};
        \node[compartment] (cpM) at (-2.35,-2.93)
            {\(M_t\)\\CPU, memory, devices};
        \node[compartment] (cpD) at (2.35,-2.93)
            {\(D_t\)\\disk + internal snapshot};
        \draw[flow] (capture) -- (0,-2.20);

        \coordinate (split) at (0,-4.30);
        \draw[copyflow] (0,-3.65)
            -- node[right, font=\scriptsize\bfseries, text=pjnavy, pos=0.4]
               {parallel fan-out} (split);

        \node[branch] (copy1) at (-4.15,-6.20)
            {private \(M_t^{(1)}\)\\[0.6mm]
             private full disk\\copy \(D_t^{(1)}\)};
        \node[branch] (copy2) at (0,-6.20)
            {private \(M_t^{(2)}\)\\[0.6mm]
             private full disk\\copy \(D_t^{(2)}\)};
        \node[branch] (copyK) at (4.15,-6.20)
            {private \(M_t^{(K)}\)\\[0.6mm]
             private full disk\\copy \(D_t^{(K)}\)};
        \node[btab, anchor=south] (tab1) at (copy1.north)
            {Branch package \(1\)};
        \node[btab, anchor=south] (tab2) at (copy2.north)
            {Branch package \(2\)};
        \node[btab, anchor=south] (tabK) at (copyK.north)
            {Branch package \(K\)};
        \node[font=\scriptsize] at ($(copy2)!0.5!(copyK)$) {\(\cdots\)};

        \draw[copyflow] (split) -| (tab1.north);
        \draw[copyflow] (split) -- (tab2.north);
        \draw[copyflow] (split) -| (tabK.north);

        \node[pjbox, text width=90mm] (dispatch) at (0,-8.25)
            {{\bfseries PARALLEL WARM REFRESH}\\[-0.5mm]
             {\mdseries apply independently for \(k=1,\ldots,K\)}};
        \draw[flow] (copy1) -- (copy1 |- dispatch.north);
        \draw[flow] (copy2) -- (dispatch);
        \draw[flow] (copyK) -- (copyK |- dispatch.north);

        \draw[warm] (-3.70,-8.95) rectangle (3.70,-15.00);
        \node[chip, fill=pjteal, anchor=west] at (-3.50,-8.95)
            {WARM POOL};

        \node[warmbox, text width=42mm] (warmvm) at (0,-9.55)
            {{\bfseries WARM VM \(k\)}\\[-0.5mm]
             {\mdseries reuse existing QEMU process}};
        \node[stagebox] (attach) at (0,-10.55)
            {Swap in disk \(D_t^{(k)}\), attach \(M_t^{(k)}\)};
        \node[warmbox, text width=42mm] (loadvm) at (0,-11.55)
            {{\bfseries \texttt{snapshot-load}}\\[-0.5mm]
             {\mdseries stop, restore, resume}};
        \node[stagebox] (detach) at (0,-12.55)
            {Detach temporary vmstate source};
        \node[stagebox] (recover) at (0,-13.55)
            {Recover guest services (stable screen)};
        \node[box, text width=40mm] (verify) at (0,-14.55)
            {checked State~0\\\(s_t^{(k)} \approx s_t\)};

        \draw[flow] (dispatch) -- (warmvm);
        \draw[flow] (warmvm) -- (attach);
        \draw[flow] (attach) -- (loadvm);
        \draw[flow] (loadvm) -- (detach);
        \draw[flow] (detach) -- (recover);
        \draw[flow] (recover) -- (verify);

        \node[pjbox, text width=95mm] (barrier) at (0,-16.05)
            {{\bfseries READINESS BARRIER}\\[-0.5mm]
             {\mdseries continue only after all \(K\) branches are ready}};
        \node[box, text width=38mm] (execute) at (0,-17.20)
            {Begin candidate execution};
        \draw[flow] (verify) -- (barrier);
        \draw[flow] (barrier) -- (execute);
    \end{tikzpicture}%
    }
    \caption{Parallel fan-out of the current state \(s_t\).
    A single
    consistent checkpoint \(C_t\) is captured once and then treated as an
    immutable source (navy region).
    Its machine state \(M_t\) and its disk image \(D_t\), which carries
    the internal snapshot, are both copied privately into each of the \(K\)
    branch packages; branches therefore share no mutable state.
    The
    dispatch band applies one warm-restore protocol independently to every
    branch \(k\); the teal region draws that protocol once rather than
    repeating it \(K\) times: the warm VM process is reused, its active disk
    is switched to the branch copy, the vmstate copy is attached and loaded
    via a QMP \texttt{snapshot-load} (\texttt{loadvm}), the temporary
    source is detached, and guest services recover into State~0, which is
    compared across branches by perceptual hash.
    A readiness barrier ensures candidate execution begins only
    after every one of the \(K\) branches has been restored; if any
    branch fails, the pool is discarded and the \(K\) VMs are cold-launched.}
    \label{fig:parallel-state-fanout}
\end{figure}
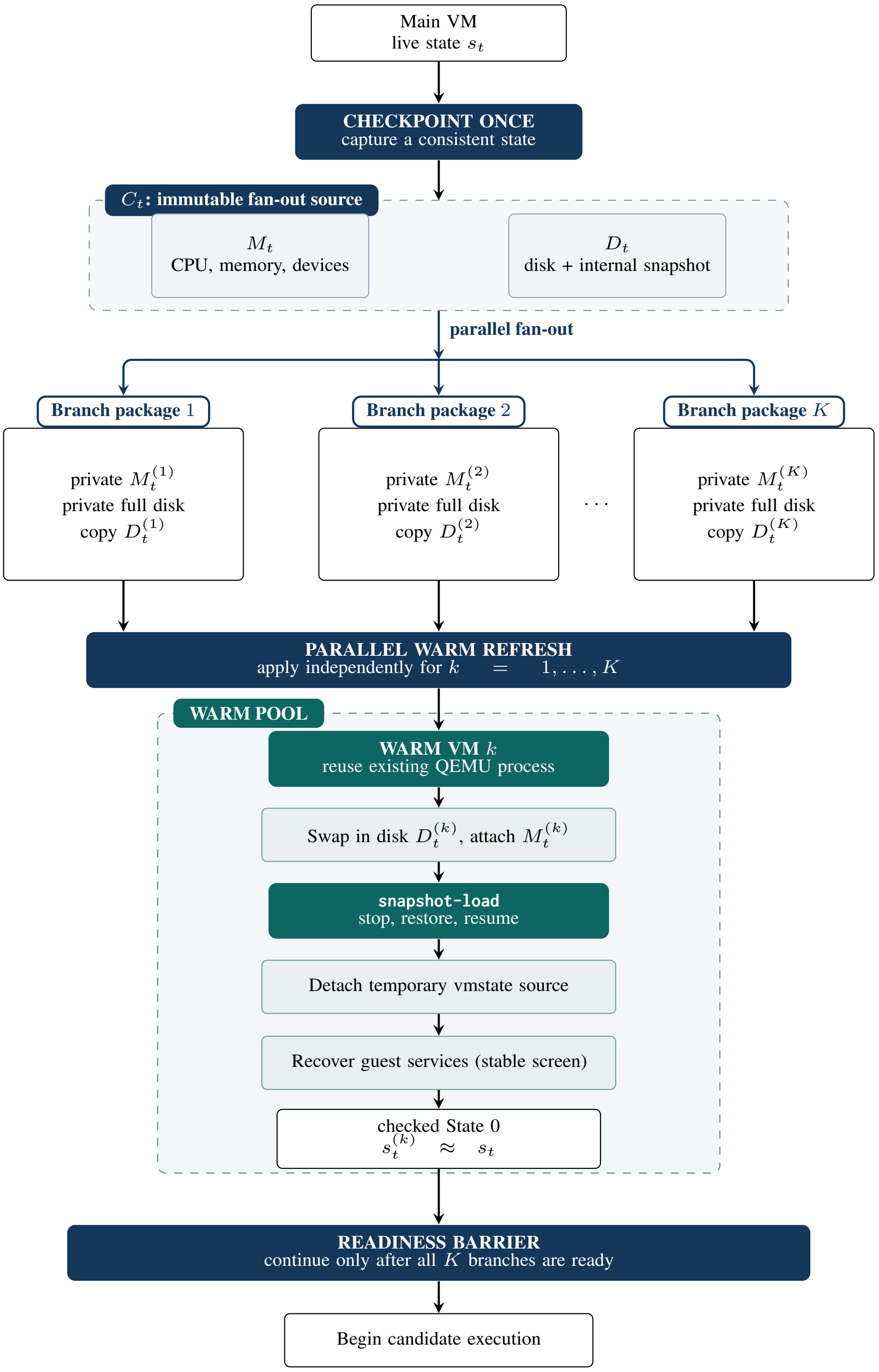

Figure~\ref{fig:parallel-state-fanout} expands the state-copy operation used
before branch execution.
The system does not copy a running main VM
independently \(K\) times.
It first captures one consistent checkpoint
\(C_t=(M_t,D_t)\) with a QEMU \texttt{snapshot-save} job, where \(M_t\)
contains CPU, memory, and device state and \(D_t\) contains persistent disk
state.
\(D_t\) is stored as an internal qcow2 snapshot inside the main VM's disk
image, and \(M_t\) is written to a separate vmstate image in host shared
memory (\texttt{/dev/shm}).
This checkpoint is then treated as
an immutable source.
In parallel, branch \(k\) receives a private copy of the machine state
\(M_t^{(k)}\) and a private full copy of the disk image \(D_t^{(k)}\):
\[
\begin{aligned}
    C_t^{(k)}
    &= \left(M_t^{(k)},\,D_t^{(k)}\right), \\
    \operatorname{restore}\!\left(C_t^{(k)}\right) &= s_t .
\end{aligned}
\]
The disk is copied in full rather than as a thin overlay on a shared base,
because the snapshot load restores every writable qcow2 device in the VM, so
each branch disk must carry the main disk's internal snapshot table itself.
The private copies ensure that changes made while evaluating one action
cannot affect another branch.
Each package is attached to an already-running
warm VM process: the paused VM's active disk is switched to its branch copy
at the block layer, and the vmstate copy is attached as a temporary block
device.
A QMP \texttt{snapshot-load} (\texttt{loadvm}) operation stops that VM,
loads the branch's private checkpoint state, and resumes the same process.
The temporary vmstate source is then detached and deleted, guest services are
allowed to recover until the screen is stable, and the resulting State~0 is
compared with that of the other branches before
any candidate action is executed.
Thus, warming reuses the VM process but
replaces its complete guest execution state.

\clearpage

\subsection{Parallel Warm-Rollout Implementation}
\label{app:parallel-warm-gt-rollout}

\paragraph{Initial pool construction.}
At a decision step, the system checkpoints the main VM before any candidate
action is committed.
The checkpoint is a QEMU \texttt{snapshot-save}: disk state goes into an
internal qcow2 snapshot of the main disk, and CPU, memory, and device state go
into an external vmstate image on host shared memory.
Snapshot creation runs on a background thread and overlaps policy inference.
If the policy returns a single unique candidate, no rollout or judge is run.
For
the first parallel rollout, the system copies the disk image and the vmstate
image once per candidate and cold-launches one rollout VM per candidate from
its private copy.
Each rollout VM
therefore begins with the same disk, memory, device, and interface state.
The
system captures a settled pre-action screenshot for every candidate and
executes the corresponding first grounded action.
It then continues each
branch on the same VM for up to \(H\) steps, conditioning later actions on
that branch's actions and observations and recording the settled state after
every transition.
A branch stops early when the policy emits a terminal action
(\texttt{done}, \texttt{fail}, or \texttt{answer}), when its action cannot be
generated or grounded, or when no screenshot can be captured; a terminal first
action is not executed.
Instead of terminating these VMs after selection, the system
transfers them to the warm pool for reuse.

\paragraph{Warm refresh.}
At every later decision step, all candidates must start from the new live
state.
The refresh is prepared in the background while the policy runs, and its
protocol is:
\begin{enumerate}
    \item checkpoint the main VM and pause the warm rollout VMs;
    \item make \(K\) full copies of the main disk image and of the vmstate
          image, one per branch;
    \item switch each paused warm VM's active disk to its branch copy at the
          block layer, attach its vmstate copy, and restore it with a QMP
          \texttt{snapshot-load} in the same QEMU process;
    \item detach and delete the vmstate copy, verify the active disk, resume
          and recover each rollout VM until its screen is stable, and compare
          the pre-action states across branches; and
    \item execute the \(K\) branch trajectories concurrently, with transitions
          within each branch proceeding sequentially for up to \(H\) steps.
\end{enumerate}
Refreshing an existing pool avoids repeated VM process creation and guest
bootstrapping.
The refresh still replaces each branch's complete disk and execution
state, ensuring that no disk or memory changes from a previous candidate
survive into the next decision step.
After a restore, the guest network state is repaired, since the restored
TCP connections are stale.

\paragraph{Snapshot isolation.}
Each branch receives an independent full copy of the disk image and of the
vmstate image.
Thin copy-on-write overlays on a shared base are not used: the snapshot load
restores every writable qcow2 device in the VM, so every branch disk must
carry the main disk's internal snapshot table.
Sharing
mutable snapshot state between branches can also introduce interference:
restoring or modifying one candidate may alter another candidate that has
already been restored.
Per-candidate copies eliminate this dependency and allow all
restores and action executions to proceed concurrently.
All copies are kept in host shared memory (\texttt{/dev/shm}), and each
vmstate copy is deleted as soon as its VM has loaded it.
Before a run, a resource preflight checks that host memory and shared-memory
space can hold the main VMs, the \(K\) rollout VMs per worker, and their
copies; before each copy, a host-wide admission check confirms that enough
free space remains.

\paragraph{Selection and state invariants.}
Before action execution, every rollout VM should recover to the same State~0
as the checkpointed main VM.
The implementation captures stable State~0 images and compares each
candidate's State~0 with that of the first candidate by perceptual (dHash)
distance; mismatches are logged and recorded with the trajectory.
If the warm refresh fails for any branch, because of a failed disk copy,
snapshot load, disk verification, or guest recovery, or because \(K\)
changed, the pool is discarded and the \(K\) rollout VMs are cold-launched.
The
judge receives each high-level candidate together with its real rollout
trajectory: the later actions and the screenshots of State~1 through
State~\(H_k\).
While the judge runs, the main VM is reloaded to State~0 in parallel.
After it selects \(k^\star\), the grounded first action
\(a_{k^\star}^{(1)}\) from the selected rollout is dispatched to the live
environment without re-grounding; later actions of the selected branch are
discarded, and the next step proposes new candidates from scratch.
The
warm pool persists across the steps of a task
until the candidate count changes, recovery fails, the task ends, or the
worker shuts down.

\paragraph{Resource accounting and recorded evidence.}
For \(W\) task workers, the method requires \(W\) main VMs and \(WK\) rollout
VMs, together with their checkpoint copies.
Increasing either \(W\) or \(K\)
therefore raises memory and shared-memory use substantially.
For every
step, the trajectory stores the candidate actions, grounded actions, rollout
actions and thoughts, State~0 screenshots, and per-horizon rollout
screenshots.
These
artifacts permit visual inspection that all branches began from a common state
and that the judge's evidence corresponds to actual VM execution.

\clearpage

\iffastmode\include{V}\else\input{V}\fi
\section{Example of a Retrieved Tutorial}
\label{app:tutorial-example}

Figure~\ref{fig:retrieved-tutorial-example} shows a Chrome task with a retrieved web tutorial.

\newcommand{\tutorialicon}[1]{%
  \raisebox{-0.35ex}{\tcbox[
    on line,
    colback=white,
    colframe=RoyalBlue!65!black,
    boxrule=0.45pt,
    arc=2pt,
    left=1pt,
    right=1pt,
    top=1pt,
    bottom=1pt,
    boxsep=0pt
  ]{\includegraphics[height=1.05em]{#1}}}%
}
\begin{figure}[H]
\centering
\begin{tcolorbox}[
  enhanced,
  width=\linewidth,
  colback=RoyalBlue!12,
  colframe=RoyalBlue!55!black,
  boxrule=0.6pt,
  arc=3pt,
  left=6pt,
  right=6pt,
  top=5pt,
  bottom=5pt,
  drop fuzzy shadow={RoyalBlue!25!black!20},
  before skip=0pt,
  after skip=0pt
]
\small
\textbf{Task instruction.}
Can you make Bing the main search engine when I look stuff up on the internet?
\end{tcolorbox}

\vspace{0.3\baselineskip}
\begin{tcolorbox}[
  enhanced,
  width=\linewidth,
  colback=white,
  colframe=RoyalBlue!55!black,
  boxrule=0.6pt,
  arc=3pt,
  left=8pt,
  right=8pt,
  top=6pt,
  bottom=4pt,
  drop fuzzy shadow={RoyalBlue!25!black!20},
  before skip=0pt,
  after skip=0pt,
  title={Tutorial: Set your default search engine},
  fonttitle=\small\bfseries,
  coltitle=white,
  colbacktitle=RoyalBlue!55!black,
  attach boxed title to top left={xshift=8pt,yshift=-2.6mm},
  boxed title style={
    enhanced,
    boxrule=0pt,
    arc=2pt,
    left=5pt,
    right=5pt,
    top=2pt,
    bottom=2pt,
    colback=RoyalBlue!55!black,
    colframe=RoyalBlue!55!black
  }
]
\small
\raggedright
\vspace{0.1\baselineskip}
\begin{enumerate}
  \setlength{\itemsep}{0pt}
  \setlength{\parsep}{0pt}
  \setlength{\topsep}{0pt}
  \item On your computer, open Chrome.
  \item At the top right, select \mbox{More~\tutorialicon{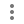}\nobreak\hspace{0.15em}\tutorialicon{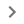}}~
    Settings.
  \item Select Search engine.
  \item Under ``Search engine,'' select Change.
  \item Select a new default search engine.
\end{enumerate}
\end{tcolorbox}
\vspace{0.18\baselineskip}
\captionsetup{justification=raggedright,singlelinecheck=false}
\caption{A Chrome task example with a retrieved web tutorial.
The tutorial provides updated step-by-step browser instructions with inline UI icons.}
\label{fig:retrieved-tutorial-example}
\end{figure}

\iffastmode\include{V}\else\input{V}\fi
\section{Building the OSWorld-Tutorial Dataset}
\label{app:os-tutorial}

\begin{tcblisting}{
  enhanced,
  breakable,
  width=\linewidth,
  colback=gray!5,
  colframe=gray!40,
  boxrule=0.5pt,
  arc=2pt,
  left=4pt,
  right=4pt,
  top=4pt,
  bottom=4pt,
  listing only,
  listing options={
    basicstyle=\ttfamily\footnotesize,
    breaklines=true,
    breakautoindent=false,
    breakindent=0pt,
    postbreak={},
    columns=fullflexible,
    keepspaces=true
  }
}
Never search for or use this benchmark's own materials: task IDs, task JSON files, evaluator/checker code, expected answers, or published solutions (for example the OSWorld or ScienceBoard GitHub repositories, Hugging Face datasets, or solution/trajectory repos). Write the tutorial only from the application's own documentation and general knowledge.

Never call the task infeasible unless the application's documentation confirms the required feature is missing. Do not decide feasibility from the screenshot: data, layers, menus and files not visible in it may still exist.

Before writing the tutorial, use the google_search tool aggressively. Perform at least 8 distinct Google Search queries before answering. Use separate searches for current UI/menu names, settings page names, keyboard shortcuts, official help or documentation, Ubuntu/Linux behavior when relevant, and task-specific caveats. Do not stop after one search if more verification queries are available. After searching, output only the requested tutorial instructions.

Give detailed instructions for completing the given task in Ubuntu (<ubuntu version>) with <program> <version>, assuming the given start state steps are already complete. If the application is already open with the task's files loaded, do not launch, relaunch or reopen it.

Output only the instructions, with no questions or extra text. You should not solve the task: you are providing instructions for solving the task. Give one primary method, the one you are most sure of, as short numbered steps; never lead with a guess. Add at most one short alternative. Do not close anything unless it is instructed by user. Change only what the task asks; keep every other line, name and setting as it is.

Rules for the tutorial:
- <step budget>The primary method ends as soon as the task's goal is reached: no steps the task does not ask for, such as compiling, previewing, running or verifying.
- Never invent names of files, maps, layers, columns, icons or records. If a name is not in the task or the screenshot, the first step reads it from the application; later steps call it "the item found in step 1".
- If the application has a command line, the primary method is the single exact command, using only options from the official documentation.
- Every typed input must be complete and valid as entered, with exact syntax, separators and formats. Write needed content (citation entry, formula, code, value) out in full instead of getting it from a website.
- Cover the whole object the task names, not one branch, half or special case; prefer a form that represents all of it (for example an implicit equation). Read ambiguous wording with its standard technical meaning.
- If the task asks a question, the last step gives the answer found, in the format the task asks for (for example `ANS 42` when answers are submitted with ANS).
- For tasks about dates or times, give generic steps that do not depend on today's date. For multi_app tasks, you may install new tools and apps.

Before answering, cross-check every quoted variable, path, or placeholder in the task instruction against the attached screenshot, and check that the end state of your steps matches the task's target exactly, including sign, direction, units and values.

Treat the attached screenshot as the raw, native-resolution source of truth; do not rely on a reduced, resized, or max-pixel version. For spreadsheet and table tasks, explicitly verify every relevant column letter, header, source range, destination range, and example cell reference against the screenshot before writing instructions. Map the task's quoted column names to the visible column letters yourself. Never shift a source or destination column based only on the wording of the task. Before outputting any formula or copy step, confirm that it reads from the visible source column and writes to the visible destination column without overwriting the source or touching irrelevant cells.

If the task is infeasible (cannot be done with <program> alone), say "Infeasible." followed by a brief explanation of why. Do NOT suggest workarounds, alternative tools, or command-line solutions --- just state why it is infeasible and stop.

Task:
<task instruction>
<the first screenshot>

\end{tcblisting}
\begin{center}
\begin{minipage}{\linewidth}
\captionsetup{type=figure,hypcap=false,justification=raggedright,singlelinecheck=false}
\caption{Tutorial-generation prompt used to retrieve or synthesize task-specific tutorials.
The same prompt is used for OSWorld and ScienceBoard tasks. The placeholders are replaced with the program category, task instruction, and initial task screenshot from each task.
For \texttt{"multi\_apps"} tasks, \texttt{<version>} lists the versions of all applications in Table~\ref{tab:tutorial-program-versions}; ScienceBoard versions are in Table~\ref{tab:tutorial-program-versions-sb}. \texttt{<step budget>} is ``The agent has only $N$ actions in total; the primary method must fit in that.'' for ScienceBoard tasks with step budget $N$, and empty for OSWorld.
}
\label{fig:tutorial-prompt}
\end{minipage}
\end{center}

\begin{table}[H]
\centering
\small
\caption{Software versions supplied to the tutorial-generation prompt for OSWorld tasks.}
\label{tab:tutorial-program-versions}
\begin{tabular}{@{}ll@{}}
\toprule
Program & Version \\
\midrule
Ubuntu & \texttt{22.04.3 LTS} \\
Google Chrome & \texttt{130.0.6723.58} \\
GIMP & \texttt{2.10.30} \\
LibreOffice (Calc, Impress, Writer) & \texttt{7.3.7.2} \\
Mozilla Thunderbird & \texttt{115.6.0} \\
VLC & \texttt{3.0.16} \\
Visual Studio Code & \texttt{1.85.1} \\
Multi-app tasks & all of the above \\
\bottomrule
\end{tabular}
\end{table}

\begin{table}[H]
\centering
\small
\caption{Software versions supplied to the tutorial-generation prompt for ScienceBoard tasks.}
\label{tab:tutorial-program-versions-sb}
\begin{tabular}{@{}ll@{}}
\toprule
Program & Version \\
\midrule
Ubuntu & \texttt{22.04 LTS} \\
UCSF ChimeraX & \texttt{1.9} with the \texttt{chimerax-states 0.5} bundle \\
GRASS GIS & \texttt{8.4.1} (wxPython GUI) \\
Celestia & \texttt{1.7-dev} (Qt6 GUI) \\
KAlgebra & KDE Education, Qt~6 build \\
TeXstudio & Ubuntu 22.04 PPA build \\
Lean & Lean~4 \texttt{v4.14.0} in Visual Studio Code with the Lean~4 extension \\
\bottomrule
\end{tabular}
\end{table}

\paragraph{Overview.}
To robustly evaluate the impact of tutorials, we construct the OSWorld-Tutorial dataset by augmenting 360 tasks from the OSWorld-Verified benchmark~\cite{osworld}.\footnote{We use OSWorld-Verified at commit \href{https://github.com/xlang-ai/OSWorld/commit/83e8534451ba8b3ab6477448ef3f0a8e563f05be}{\texttt{83e8534}}.}
We recruit three student annotators with experience in software tasks similar to those in OSWorld to collect a tutorial for each task.
For each task, we first launch it in a virtual machine and provide the task instruction together with the initial screenshot to the tutorial-generation prompt shown in Figure~\ref{fig:tutorial-prompt}.
We submit this prompt to Google Gemini 3.8 Flash~\cite{gemini38flash}.
It searches the web for relevant tutorials and synthesizes a comprehensive summary over the retrieved results.
The annotators then review the retrieved tutorials and select the most helpful one for each task.
To verify tutorials from publicly available articles, they follow the article's instructions to complete the task and confirm that it is correct and helpful.

We find that many OSWorld tasks require combining multiple tutorials, while others have no relevant tutorials available online.
For these tasks, we use the response generated by Google Gemini 3.8 Flash directly as the tutorial.
\textit{We do not verify the correctness of AI-synthesized tutorials.}

\begin{figure}
\centering
\includegraphics[width=\linewidth]{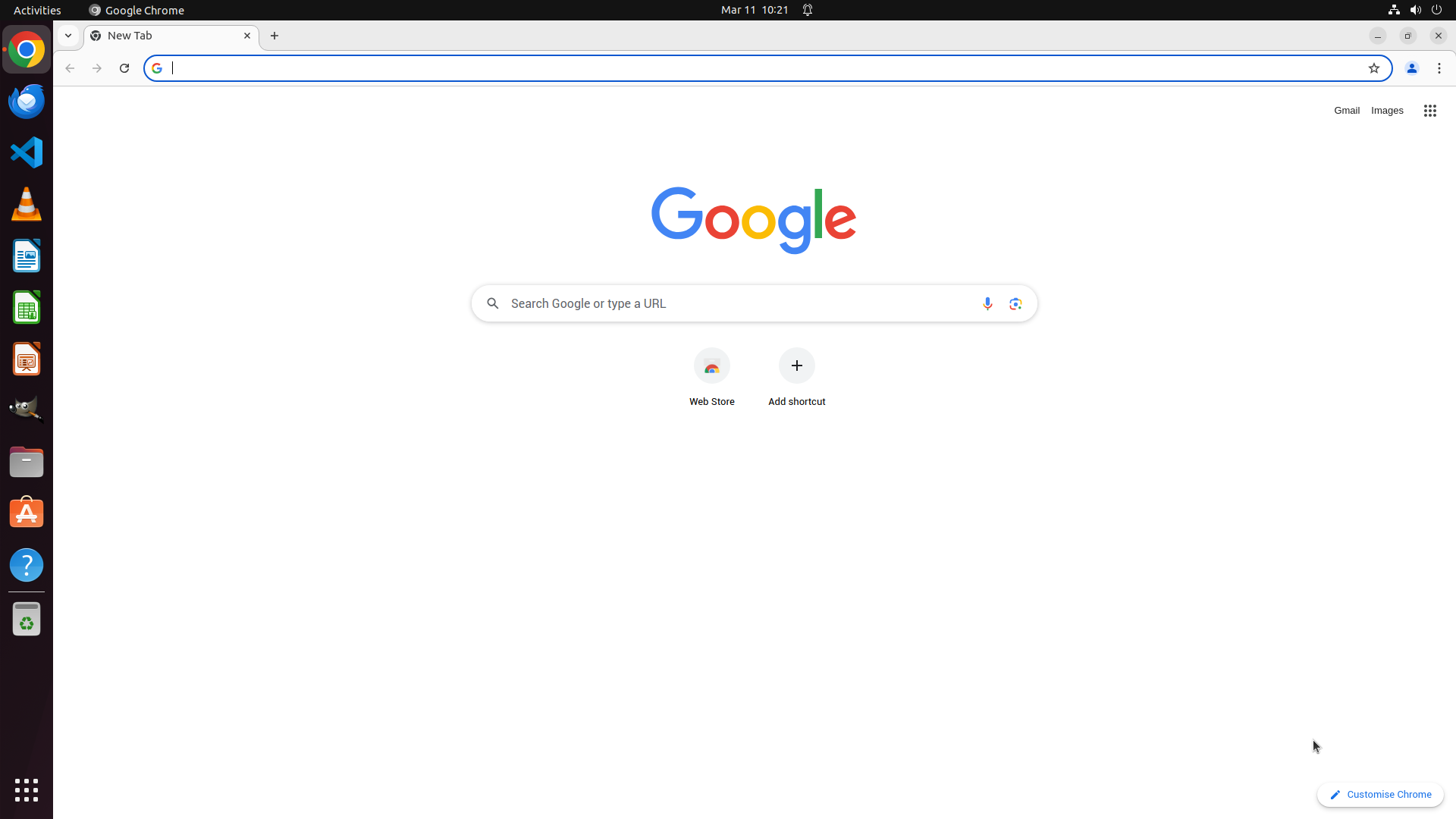}

\vspace{0.3\baselineskip}
\begin{minipage}{\linewidth}
\small
Task instruction.
Enable the ``Do Not Track'' feature in Chrome to enhance online privacy.
\end{minipage}

\vspace{0.5\baselineskip}
\begin{tcolorbox}[
  enhanced,
  width=\linewidth,
  colback=gray!5,
  colframe=gray!40,
  boxrule=0.5pt,
  arc=2pt,
  left=4pt,
  right=4pt,
  top=4pt,
  bottom=4pt
]
\small
Turn ``Do Not Track'' on or off

\medskip
When browsing the web on computers or Android devices, users can send a request to websites not to collect or track browsing data.
The setting is turned off by default.

\medskip
However, what happens to the data depends on how a website responds to the request.
Many websites still collect and use browsing data for security, content, services, ads, recommendations, and reporting statistics.

\medskip
Most websites and web services, including Google's, do not change their behavior when they receive a Do Not Track request.

\medskip
\begin{enumerate}
  \setlength{\itemsep}{0.15em}
  \setlength{\parsep}{0pt}
  \setlength{\topsep}{0pt}
  \item On your computer, open Chrome.
  \item At the top right, select More~\raisebox{-0.25ex}{\includegraphics[height=1.1em]{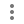}} \raisebox{-0.25ex}{\includegraphics[height=1.1em]{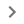}} Settings.
  \item Select Privacy and security \raisebox{-0.25ex}{\includegraphics[height=1.1em]{figures/osworld_tutorial/and_then_icon.png}} Third-party cookies.
  \item Turn Send a ``Do not track'' request with your browsing traffic on or off.
\end{enumerate}
\end{tcolorbox}
\captionsetup{justification=raggedright,singlelinecheck=false}
\caption{\textbf{Chrome task example with a retrieved web tutorial.}
The top panel shows the initial screenshot; the lower panel shows the retrieved markdown tutorial with inline images preserved.
\emph{Task ID:} \texttt{030eeff7-\allowbreak b492-\allowbreak 4218-\allowbreak b312-\allowbreak 701ec99ee0cc}.
\emph{Source:} \protect\href{https://support.google.com/chrome/answer/15085120?hl=en\&co=GENIE.Platform\%3DDesktop}{Google Chrome Help}.}
\label{fig:do-not-track-tutorial-example}
\end{figure}

\begin{figure}
\centering
\includegraphics[width=\linewidth]{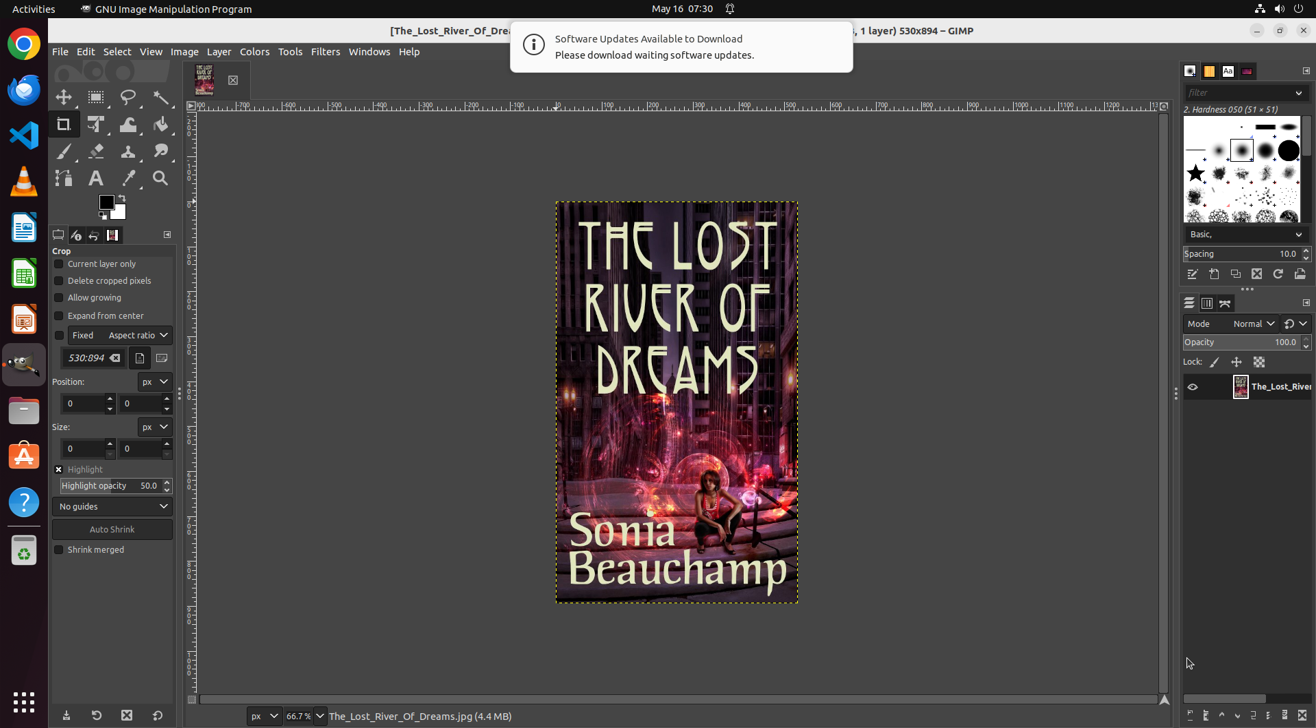}

\vspace{0.3\baselineskip}
\begin{minipage}{\linewidth}
\small
Task instruction.
Could you assist me in placing my photo on the desktop and renaming it to \texttt{export.jpg}?

Category: \texttt{gimp}.
\end{minipage}

\vspace{0.5\baselineskip}
\begin{tcolorbox}[
  enhanced,
  width=\linewidth,
  colback=gray!5,
  colframe=gray!40,
  boxrule=0.5pt,
  arc=2pt,
  left=4pt,
  right=4pt,
  top=4pt,
  bottom=4pt
]
\small
GIMP treats standard image formats such as \texttt{.jpg} as exports rather than saves.

\medskip
\begin{enumerate}
  \setlength{\itemsep}{0.15em}
  \setlength{\parsep}{0pt}
  \setlength{\topsep}{0pt}
  \item In the top menu, go to File > Export As... or press Ctrl+Shift+E.
  \item In the Name box at the top, delete the existing text and type \texttt{export.jpg}.
  \item On the left-hand sidebar, under Places, click Desktop to ensure it is being saved to the right spot.
  \item Click the Export button in the bottom right.
  \item An ``Export Image as JPEG'' window will pop up.
  Click Export again to confirm the settings.
\end{enumerate}
\end{tcolorbox}
\captionsetup{justification=raggedright,singlelinecheck=false}
\caption{\textbf{GIMP task example with an MLLM-synthesized tutorial.}
The top panel shows the initial screenshot; the lower panel shows the synthesized tutorial for exporting the image as \texttt{export.jpg} on the desktop.
\emph{Task ID:} \texttt{77b8ab4d-\allowbreak 994f-\allowbreak 43ac-\allowbreak 8930-\allowbreak 8ca087d7c4b4}.}
\label{fig:gimp-export-tutorial-example}
\end{figure}

Figures~\ref{fig:do-not-track-tutorial-example} and~\ref{fig:gimp-export-tutorial-example} show examples of retrieved publicly available tutorials and MLLM-generated tutorials.
Each figure shows task instruction, initial screenshot, and the corresponding tutorial content.

\paragraph{Statistics.}
In total, we collect 360 tasks with verified tutorials, covering all eight OSWorld task categories.
Among these, \numHumanVerifiedTutorialTasks{} tasks have human-verified tutorials retrieved from the web (\humanVerifiedTutorialPercent{}), while the remaining \aiGeneratedTutorialPercent{} are synthesized by Google Gemini 3.8 Flash.
We follow Agent-S3~\cite{agent_s3} and OSWorld-MPC~\cite{osworld_mcp} to exclude eight Google Drive tasks.
Recently, these tasks have presented technical challenges in using Google accounts and Google Drive credentials in the sandbox.
We notice that some web tasks have just been upgraded to use more aggressive bot detection systems, which block automated access even through residential proxies provided by OSWorld.
We therefore exclude one task that is blocked by website-level bot-detection or access restrictions, as listed by task ID in Figure~\ref{fig:excluded-task-ids}.

\begin{figure}[H]
\centering
\begin{tcolorbox}[
  enhanced,
  width=\linewidth,
  colback=gray!5,
  colframe=gray!40,
  boxrule=0.5pt,
  arc=2pt,
  left=4pt,
  right=4pt,
  top=4pt,
  bottom=4pt
]
\begingroup
\ttfamily\footnotesize\raggedright
\# Excluded task IDs -- one per line, comments start with \#\par
\# tripadvisor.com -- "Access is temporarily restricted" behavioral bot detection.\par
b7895e80-f4d1-4648-bee0-4eb45a6f1fa8\par
\endgroup
\end{tcolorbox}
\captionsetup{justification=raggedright,singlelinecheck=false}
\caption{Excluded task IDs for websites that block automated access through bot detection, CAPTCHAs, or access-denial systems.}
\label{fig:excluded-task-ids}
\end{figure}

\iffastmode\include{V}\else\input{V}\fi
\clearpage
\section{ScienceBoard-Tutorial Examples}
\label{app:sb-tutorial}

\begin{figure}[H]
\centering
\includegraphics[width=0.7\linewidth]{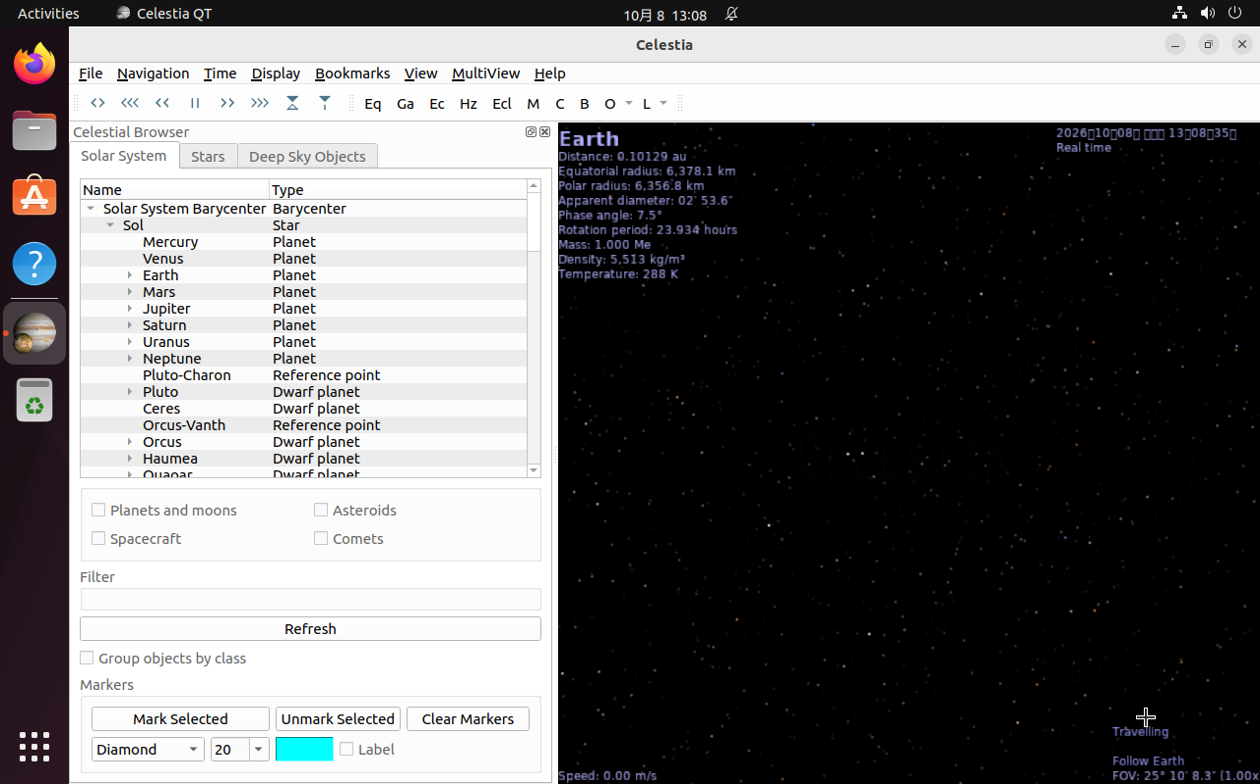}

\vspace{0.3\baselineskip}
\begin{tcolorbox}[
  enhanced,
  width=\linewidth,
  colback=blue!4,
  colframe=blue!45!black,
  boxrule=0.5pt,
  leftrule=2.5pt,
  arc=2pt,
  left=5pt,
  right=5pt,
  top=3pt,
  bottom=3pt
]
\small
\textbf{Task instruction:} Go to Neptune in Celestia.\hfill\textbf{App:} Celestia
\end{tcolorbox}

\vspace{0.5\baselineskip}
\begin{tcolorbox}[
  enhanced,
  width=\linewidth,
  colback=gray!5,
  colframe=gray!40,
  boxrule=0.5pt,
  arc=2pt,
  left=4pt,
  right=4pt,
  top=4pt,
  bottom=4pt
]
\small
\textbf{Method 1: Using the Celestial Browser (Visible Panel)}
\begin{enumerate}
  \setlength{\itemsep}{0.15em}
  \setlength{\parsep}{0pt}
  \setlength{\topsep}{0pt}
  \item In the \textbf{Celestial Browser} panel on the left (under the \textbf{Solar System} tab), click on \textbf{Neptune} in the list under \texttt{Solar System Barycenter} $\rightarrow$ \texttt{Sol} to select it.
  \item Press the \textbf{G} key on your keyboard (or right-click \textbf{Neptune} in the list and select \textbf{Go to}, or double-click \textbf{Neptune}) to travel to the planet.
\end{enumerate}

\medskip
\textbf{Method 2: Using Direct Planet Number Shortcut}
\begin{enumerate}
  \setlength{\itemsep}{0.15em}
  \setlength{\parsep}{0pt}
  \setlength{\topsep}{0pt}
  \item Click on the main 3D star viewport to ensure it has focus.
  \item Press the number key \textbf{8} on your keyboard to select Neptune (the 8th planet from the Sun).
  \item Press the \textbf{G} key to travel directly to Neptune.
\end{enumerate}

\medskip
\textbf{Method 3: Using the Target Name Bar}
\begin{enumerate}
  \setlength{\itemsep}{0.15em}
  \setlength{\parsep}{0pt}
  \setlength{\topsep}{0pt}
  \item Click on the main 3D star viewport and press \textbf{Enter} (Return) to bring up the target entry prompt.
  \item Type \texttt{Neptune} and press \textbf{Enter} to confirm the selection.
  \item Press the \textbf{G} key to travel to Neptune.
\end{enumerate}

\medskip
\textbf{Method 4: Using the Navigation Menu}
\begin{enumerate}
  \setlength{\itemsep}{0.15em}
  \setlength{\parsep}{0pt}
  \setlength{\topsep}{0pt}
  \item Click \textbf{Navigation} in the top menu bar.
  \item Click \textbf{Go to Object...} (or \textbf{Select Object...}, type \texttt{Neptune}, and confirm).
  \item In the dialog window, ensure the target is set to \texttt{Neptune} and click the \textbf{Go to} (or \textbf{OK}) button to begin navigation.
\end{enumerate}
\end{tcolorbox}
\captionsetup{justification=raggedright,singlelinecheck=false}
\caption{\textbf{Celestia task example with an MLLM-generated tutorial.}
The top panel shows the initial screenshot; the lower panel shows the tutorial generated by Google Gemini 3.8 Flash.
\emph{Task ID:} \texttt{Celestia/B-02}.}
\label{fig:sb-neptune-tutorial-example}
\end{figure}

\begin{figure}[H]
\centering
\includegraphics[width=0.7\linewidth]{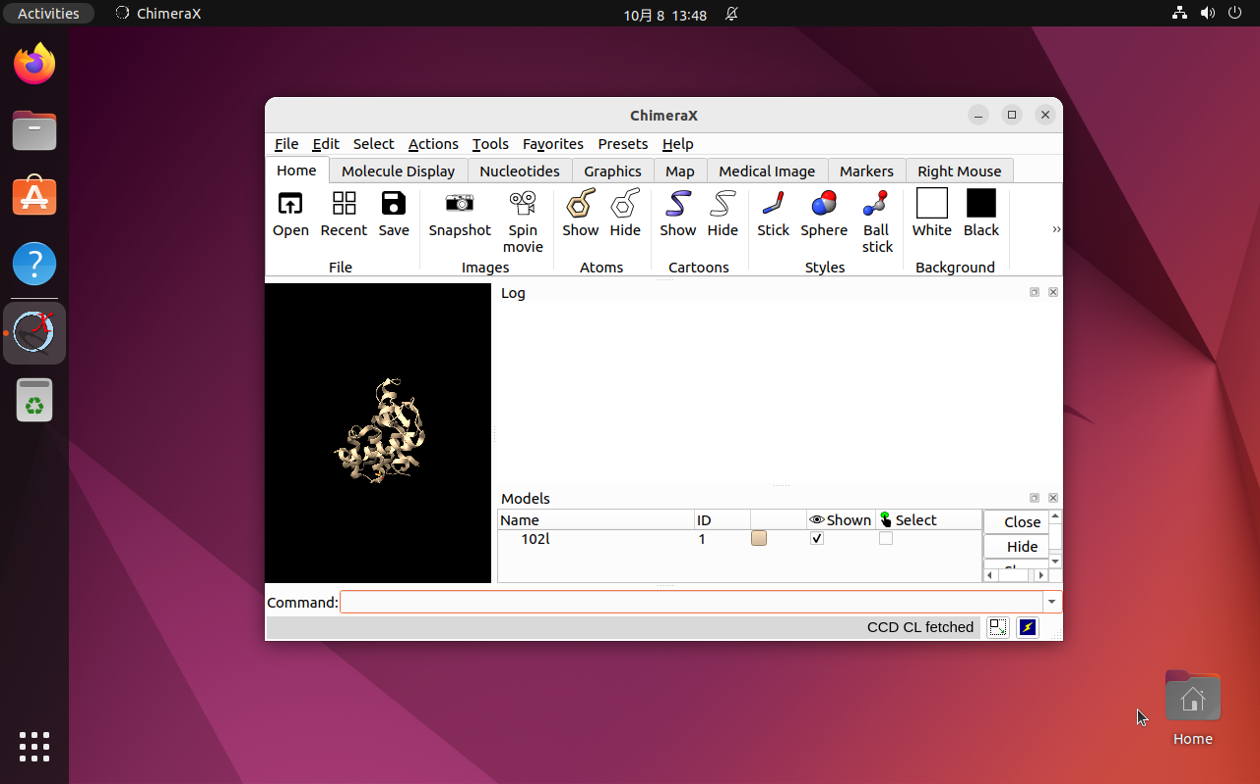}

\vspace{0.3\baselineskip}
\begin{tcolorbox}[
  enhanced,
  width=\linewidth,
  colback=blue!4,
  colframe=blue!45!black,
  boxrule=0.5pt,
  leftrule=2.5pt,
  arc=2pt,
  left=5pt,
  right=5pt,
  top=3pt,
  bottom=3pt
]
\small
\textbf{Task instruction:} Toggle a scale bar of length 10 and thinkness 4 in ChimeraX.\hfill\textbf{App:} ChimeraX
\end{tcolorbox}

\vspace{0.5\baselineskip}
\begin{tcolorbox}[
  enhanced,
  width=\linewidth,
  colback=gray!5,
  colframe=gray!40,
  boxrule=0.5pt,
  arc=2pt,
  left=4pt,
  right=4pt,
  top=4pt,
  bottom=4pt
]
\small
\begin{enumerate}
  \setlength{\itemsep}{0.15em}
  \setlength{\parsep}{0pt}
  \setlength{\topsep}{0pt}
  \item Click the command input field next to \textbf{Command:} at the bottom of the ChimeraX window.
  \item Type \texttt{scalebar length 10 thickness 4} and press Enter.
\end{enumerate}

\medskip
Alternative:
Select \textbf{Tools} from the top menu, open the scale bar tool under depiction utilities, set the length to 10 and thickness to 4, and toggle the scale bar on.
\end{tcolorbox}
\captionsetup{justification=raggedright,singlelinecheck=false}
\caption{\textbf{ChimeraX task example with an MLLM-generated tutorial.}
The top panel shows the initial screenshot; the lower panel shows the tutorial generated by Google Gemini 3.8 Flash.
The task instruction is quoted verbatim from ScienceBoard, including its typo.
\emph{Task ID:} \texttt{ChimeraX/B-06}.}
\label{fig:sb-scalebar-tutorial-example}
\end{figure}

ScienceBoard-Tutorial pairs each of the 169 ScienceBoard tasks with a tutorial generated by Google Gemini 3.8 Flash, using the same prompt and inputs as OSWorld-Tutorial (Figure~\ref{fig:tutorial-prompt}).
Figures~\ref{fig:sb-neptune-tutorial-example} and~\ref{fig:sb-scalebar-tutorial-example} show a Celestia task and a ChimeraX task with their generated tutorials.
The Celestia tutorial lists several alternative methods; Appendix~\ref{sec:res_per_category_scienceboard} (Figure~\ref{fig:sb-celestia}) analyzes how the choice among them affects RAG-PJ on this task.
The ChimeraX tutorial instead gives a single command, with a menu-based alternative.

\iffastmode\include{V}\else\input{V}\fi\section{Problem formulation}
\label{sec:pomdp}
We cast the computer-use task as a standard finite-horizon partially observable Markov decision process (POMDP) $\langle S, O, A, T, R, L, M \rangle$, conditioned on the task instruction $L$ and a multimodal tutorial $M$.
The computing environment itself (typically a virtual machine) serves as a mostly-deterministic\footnote{In our setting, $T$ is deterministic for most tasks.
For examples of non-deterministic tasks, see Appendix~\ref{sec:non_deter}.} transition function $T: S \times A \to S$.
As the agent only accesses the environment visually, the observation $o_t \in O$ at each time step $t$ is simply the current screenshot.
Actions correspond to movements or clicks of the mouse and keystrokes as well as stop actions (more details in Figure~\ref{fig:prompt_policy}).
The reward function $R: S \times A \to \{0, 1\}$ is sparse and indicates task success when the agent issues stop actions or reaches the horizon $H$.
We consider a CUA to be a stochastic policy $\pi(a_t \,|\, o_{\leq t}, a_{<t}, L, M)$, which produces a distribution over actions $a_t$ given previous observations $o_{\leq t}$, prior actions $a_{<t}$, the instruction $L$, and the tutorial $M$.
We want to find a policy which achieves high expected cumulative reward (i.e., task success) over CUA tasks.

\paragraph{Look-ahead rollouts.}
\label{sec:rollout-formulation}
Specifically, given a world model $w: O \times A \to O$ that predicts the next observation given past observations and the current action, the outcome of each of the $K$ candidate actions can be estimated as $\tilde{o}^1_t, ..., \tilde{o}^K_t$ in a one-step look-ahead.
Following~\cite{r_wom}, the policy $\pi$ is then executed with greedy decoding to generate a single next action $\hat{a}_{t+1}^j = \arg\max_{\hat{a}} \pi(\hat{a} \,|\, \tilde{o}_{t+1}^j)$ for each predicted observation $\tilde{o}_{t+1}^j$.
This loop can be executed until some fixed look-ahead horizon is reached to generate a rollout trajectory $\xi^j = (o_t, \hat{a}_t^j, \tilde{o}_{t+1}^j, \hat{a}_{t+1}^j, \ldots, \tilde{o}_{t+k}^j)$ for each action candidate $\hat{a}_t^j$.
The judge model can then select between actions based on their rollout trajectories.

\iffastmode\include{V}\else\input{V}\fi\section{RAG-PJ implementation details}
\label{sec:rag-pj-details}
\paragraph{History information.}
Following GTA-1~\cite{gta1}\footnote{Note that the official implementation of GTA-1 \url{https://github.com/Yan98/GTA1} is different from GTA-1's paper description.
In the implementation, it only uses thought tokens and actions, and does not use observation caption descriptions for history information.
We follow the implementation instead of the paper description.}, \textit{the policy and the judge} receive the same execution history.
It includes the actions of all previous steps, the intermediate thought tokens of the five most recent steps, and the screenshots of up to five recent steps.
The screenshot window drops the oldest steps four at a time instead of one per step, so the prompt prefix can be reused across steps.
When the history text grows too long, the oldest steps outside the screenshot window are replaced by a single line noting that they were omitted.
We find that including intermediate thought tokens for these steps does not significantly improve performance, while it does increase the input length and inference time for both the policy and the judge.
We include the ablation study on the effect of including intermediate thought tokens in the execution history in Table \ref{tab:ablation_history_info}.

\begin{table}[ht]
\centering
\small
\setlength{\tabcolsep}{4pt}
\renewcommand{\arraystretch}{1.15}
\providecommand{\res}[2]{#1\%\,$\pm$\,#2\%}
\begin{tabular}{lcc}
\toprule
\multirow{2}{*}{\textbf{MLLM base}}
& \multicolumn{2}{c}{\textbf{15 steps}} \\
\cmidrule(lr){2-3}
& \textbf{w/ thought} & \textbf{w/o thought} \\
\midrule
Qwen3-VL-8B  & \res{28.5}{1.6} & \res{28.1}{1.2} \\
Qwen3-VL-32B & \res{39.0}{1.2} & \res{38.8}{0.4} \\
\bottomrule
\end{tabular}
\caption{Ablation of including intermediate thought tokens in the execution history.
We use the setting vanilla policy without tutorial retrieval.
Each result is mean and sample std. over 3 completed runs.
We find that including intermediate thought tokens does not significantly improve performance.}
\label{tab:ablation_history_info}
\end{table}

\paragraph{Policy. }
We use the default decoding hyperparameters for our policies\footnote{Hugging Face model cards: \href{https://huggingface.co/Qwen/Qwen3-VL-8B-Instruct}{Qwen3-VL-8B-Instruct}; \href{https://huggingface.co/Qwen/Qwen3-VL-32B-Instruct-FP8}{Qwen3-VL-32B-Instruct-FP8}.}, shown in Table~\ref{tab:rag-pj-policy-hparams}.
We use the prompt template shown in Figure~\ref{fig:prompt_policy} for the policy, which includes the retrieved tutorial and execution history as context for action generation.

\begin{table}[ht]
\centering
\small
\begin{tabular}{lc}
\toprule
Hyperparameter & Value \\
\midrule
\texttt{temperature} & 0.7 \\
\texttt{top\_p} & 0.8 \\
\texttt{top\_k} & 20 \\
\texttt{repetition\_penalty} & 1.0 \\
\texttt{presence\_penalty} & 1.5 \\
\texttt{max\_output\_length} & 16384 \\
\bottomrule
\end{tabular}
\caption{Default decoding hyperparameters used for the \ours policy.}
\label{tab:rag-pj-policy-hparams}
\end{table}

\paragraph{Judge.}

Following WebJudge~\cite{Xue2025-uf} and MLLM-as-the-judge~\cite{Chen2024-gj}, we use deterministic decoding for the judge, shown in Table~\ref{tab:rag-pj-judge-hparams}.
We use the prompt template shown in Figure~\ref{fig:prompt_judge} for the judge, which includes the retrieved tutorial and execution history as context for action evaluation.
We also apply the same set of hyperparameters for other judges in our baselines, including R-WoM judge (ranker) and ground-truth judge.

\begin{table}[ht]
\centering
\small
\begin{tabular}{lc}
\toprule
Hyperparameter & Value \\
\midrule
\texttt{temperature} & 0 \\
\texttt{top\_p} & 1 \\
\bottomrule
\end{tabular}
\caption{Default decoding hyperparameters used for the \ours judge.}
\label{tab:rag-pj-judge-hparams}
\end{table}

\paragraph{Grounding. }
We use UI-Venus-1.5~\cite{ui_venus} as the grounding model for \ours.
We use the prompt template shown in Figure~\ref{fig:prompt_grounding} for the grounding model.

\vspace{0.6\baselineskip}

\iffastmode\include{V}\else\input{V}\fi\section{Policy Prompt}
\label{sec:policy-prompt}

The listing below shows the fixed-$K$ variant used when adaptive policy is disabled before the final step. The system message is \texttt{SHARED\_SYSTEM\_PROMPT} (a one-line intro, shared with the judge, and the full action space; the judge gets a compact action space instead) followed by the policy role text \texttt{POLICY\_K\_ROLE\_WITH\_TUTORIAL}; the user message is \texttt{POLICY\_K\_USER}. On OSWorld, at \texttt{current\_step >= max\_steps}, the system message is unchanged (so its prefix cache is reused) and the user message appends a \texttt{\# FINAL STEP} note after the role selector: instead of exactly $K$ candidates, return 1--2 different candidates using only terminal actions, each at most once. On ScienceBoard, the last step instead appends a \texttt{\# LAST STEP} note asking for the action that completes the task. When adaptive policy is enabled, the runtime uses the adaptive variant instead: the output contract changes from exactly $K$ candidates to $1$--$K$ candidates, and the \texttt{ADAPTIVE ACTION RULES} block is added.

\begin{tcblisting}{
  enhanced,
  breakable,
  width=\linewidth,
  colback=gray!5,
  colframe=gray!40,
  boxrule=0.5pt,
  arc=2pt,
  left=4pt,
  right=4pt,
  top=4pt,
  bottom=4pt,
  listing only,
  listing options={
    basicstyle=\ttfamily\footnotesize,
    breaklines=true,
    breakautoindent=false,
    breakindent=0pt,
    postbreak={},
    columns=fullflexible,
    keepspaces=true,
    escapeinside={(*@}{@*)}
  }
}
[SYSTEM MESSAGE]
Ubuntu desktop computer-use agent (1920x1080; internet available). The user message gives the task, an optional tutorial, step history, and screenshots, then names your role. Follow only the rules and output format of that role.

# ACTION SPACE
def click(self, instruction: str, num_clicks: int = 1, button_type: str = 'left', hold_keys: List = []):
'''Click the target element.
instruction: one clear visual description distinguishing the target.
num_clicks: 1 or 2. button_type: "left"|"right"|"middle". hold_keys: e.g. ["ctrl"].
'''

def done(self, return_value: Union[Dict, str, List, Tuple, int, float, bool, NoneType] = None):
'''Call ONLY after fully satisfying the task. Do not use agent.done(return_value=...) for impossible tasks; call agent.fail().
'''

def drag_and_drop(self, starting_description: str, ending_description: str, hold_keys: List = []):
'''Drag start to end. Prefer cut/paste for file ops and highlight_text_span for visible text/code selection. Use for drawing, moving objects, or non-text selection.
'''

def fail(self):
'''Call ONLY for proven impossibility: CAPTCHA/bot-check, login walls, missing credentials, unavailable sites, or absent required settings/options.
'''

def highlight_text_span(self, starting_phrase: str, ending_phrase: str):
'''Select visible text/code between phrases. Prefer this over drag_and_drop for text selection, including multi-line selections.
'''

def hover(self, instruction: str):
'''Move to target without clicking.
'''

def hold_and_press(self, hold_keys: List, press_keys: List):
'''Hold hold_keys; press press_keys in sequence.
'''

def hotkey(self, keys: List):
'''Press a hotkey. Prefer over menus. Use switch_applications to switch apps.
'''

def open(self, app_or_filename: str):
'''Open a closed app or file by name; use switch_applications if open. Give filename, not full path.
Available: Google Chrome, LibreOffice Writer, LibreOffice Calc, LibreOffice Impress, Thunderbird, VSCode, VLC Media Player, GIMP. Do not use other programs.
'''

def scroll(self, instruction: str, clicks: int, shift: bool = False):
'''Scroll target. clicks: positive=up, negative=down; shift=True means horizontal. 1 ~ 3 lines; small 1-5, section 5-10, long 10-20; max 50.
'''

def set_cell_values(self, cell_values: Dict[str, Any], app_name: str, sheet_name: str):
'''Set spreadsheet cells; ALWAYS use instead of click+type.
cell_values: e.g. {"A1": "Name", "B1": 42}; formulas start with "="; max 30 cells per call.
app_name: file name from title (e.g. "data.xlsx"), not "LibreOffice Calc". sheet_name: tab name.
'''

def switch_applications(self, app_code):
'''Switch to an open app by its listed code.
'''

def type(self, element_description: Optional[str] = None, text: str = '', overwrite: bool = False, enter: bool = False):
'''Type text into a field. If element_description is provided, grounding locates and clicks that element first. Omit element_description only when current keyboard focus/caret is known and unchanged; omission skips grounding and clicking and types at current focus. Mouse pointer position alone is not keyboard focus. Otherwise provide element_description for grounding.
overwrite=True clears possible existing text before replacement (dialogs, search/address bars, cells, Save As, Find/Replace); False appends intentionally. In a multi-line text or code editor, overwrite=True erases the whole document, not just the clicked word or line: to replace part of a document, first select that part (agent.highlight_text_span, or a double-click on one word), then type with element_description omitted and overwrite=False. Single-line GUI inputs still showing an old entry (calculator/command line after Enter, rename field; not terminals) need overwrite=True; False after grounding clicks inserts mid-string. If the whole old text is already selected, omit element_description. In console commands never nest single quotes; use "..." outside '...', e.g. where="name LIKE '%
'''

def wait(self, time: float):
'''Wait seconds only for a visible progress/load indicator.
'''


# ROLE: POLICY
Control Ubuntu desktop (1920x1080; internet available). Use task, optional tutorial, history, screenshots to choose best next actions.

# POLICY RULES:
- Follow task exactly; preserve unrelated content/tabs; visible targets only; current screenshot is ground truth.
- For each UI-targeted candidate, classify target as ENABLED, DISABLED, or NOT_VISIBLE; inspect checked/unchecked and open/closed state. Greyed-out controls are DISABLED; never click or type into a disabled control. First use its visible enabling checkbox/toggle/parent.
- For lists, never click a hidden option: a mouse click is allowed only when the list is visibly open and the exact option text is visible. Prefer arrow keys + Enter over clicking the row. When the combobox/list control is visibly focused, keyboard navigation with arrow keys + Enter is allowed whether the popup is open or closed, even if the target row is not visible. When the control is not focused, first focus it. Verify committed text/selection in the next screenshot before proceeding.
- Prefer hotkeys/terminal over equivalent GUI. For spreadsheet value/formula entry or fill, candidate 0 must use agent.set_cell_values() for the requested range, at most 30 cells per call (continue a larger range next step); never click, type, drag, or highlight to enter/fill cells.
- If an action's expected effect is absent, do not repeat it unchanged. After two no-effect actions, prohibit that target+method pair; inspect the dialog for an unchecked prerequisite or disabled control group, then recover differently.
- Save with Ctrl+S. Before agent.done(), finish/apply dialogs and verify every requested value, text, format, and path in the final artifact. Dialogs, highlights, and partial edits are not completion.
- Cancel "Authentication required". Never agent.fail() for uncertainty or recoverable UI.
- At max_steps, return only evidence-supported agent.done() or agent.fail() candidates.
- Never kill/restart desktop, log out, or reboot; keep media changes moderate unless exact/extreme requested.
Sudo password: '{CLIENT_PASSWORD}' (type value, not placeholder).

# OUTPUT FORMAT
{"thought": "(1) Step by Step Progress Assessment;(2) Next Action(s) Analysis;(3) UI State/Evidence: each returned candidate `Candidate i`; N/A when no coordinate-grounded UI target;(4) Tutorial Reference","action_candidates":[{"action_code":{"action_type":"hotkey","parameters":{"keys":["ctrl","s"]}}}]}

# OUTPUT REQUIREMENTS
Thought: Must have the following sections:
(1) Step by Step Progress Assessment---assess completion and potential errors/unexpected results; predict a recovery step if previous action incorrect.
(2) Next Action(s) Analysis---evaluate next actions from current state/history; propose best.
(3) UI State/Evidence---write one candidate-indexed entry for each returned candidate (`Candidate i: ...`). Each UI-targeted candidate: ENABLED, DISABLED, or NOT_VISIBLE; checked/unchecked or open/closed; exact screenshot evidence from the current screenshot. For any candidate without a coordinate-grounded UI target, write `N/A (no coordinate-grounded UI target)`; never invent UI state or evidence. Non-UI: focused agent.type(...), hotkey/hold_and_press/open/switch_applications/wait/set_cell_values/done/fail/answer. Never provide one shared UI State/Evidence entry for multiple candidates.

(4) Tutorial Reference: QUOTE the exact relevant sentence(s) from the tutorial. If none, say `No tutorial reference for this step.` If method fails, switch to another tutorial method.

Action requirements:
- One currently valid action per candidate, confidence-ordered; vary method/target. Never type into buttons or click static text.
- For visible text/code, use highlight_text_span with phrase anchors, including multi-line selections. Never use it for spreadsheet cell ranges. Use drag_and_drop only for objects/drawing or text without visible unique anchors.


Return only JSON matching the structured action schema. Return exactly {K} different and diverse action_candidates. Each needs action_code.action_type and action_code.parameters; no markdown, code blocks, or agent.* calls. Use at most one done and one fail; never create terminal fillers or return_value variants. For click use button_type='right' or num_clicks=2. For type use element_description (not instruction). Keyboard arrays must be bounded: hold_keys <= 4, keys <= 6, and press_keys <= 20.

# TUTORIAL RULES:
- Tutorial is a hint: use relevant steps; prefer a simpler reliable method.
- INFEASIBLE OVERRIDE (HIGHEST PRIORITY; MUST FOLLOW): If the tutorial's first non-empty line begins with `Infeasible` or `infeasible` (case-insensitive), stop: the first and only valid semantic action is `agent.fail()`. In JSON use action_type='fail', action_code.parameters={}; never emit agent.*. Ignore all later text and candidate-count/diversity/recovery/workaround rules. Do not emit any non-fail action candidate.
- For ordinary tutorials, after 2 failed attempts switch methods; for streaming media, allow 2 buffering steps.

[USER MESSAGE]
# TUTORIAL
{tutorial}

# TASK INSTRUCTION
{instruction}

# HISTORY
{history}{current_screenshot}

# YOUR ROLE
Act as the POLICY: follow the ROLE: POLICY section of the system message.

# STEP
You are on step {current_step} of maximum {max_steps} steps.
\end{tcblisting}
\begin{center}
\begin{minipage}{\linewidth}
\captionsetup{type=figure,hypcap=false,justification=raggedright,singlelinecheck=false}
\caption{Policy prompt used to generate candidate actions from the task instruction, retrieved tutorial, interaction history, and current screenshot.}
\label{fig:prompt_policy}
\end{minipage}
\end{center}

\iffastmode\include{V}\else\input{V}\fi\section{Judge Prompt}
\label{sec:judge-prompt}

The system message starts with the same one-line introduction as the policy's (Appendix~\ref{sec:policy-prompt}), followed by a compact action space (\texttt{AGENT\_METHODS\_LIST}: one line per method instead of the policy's full signatures and docstrings) and the judge role text \texttt{JUDGE\_ROLE\_PROMPT}; the user message is \texttt{JUDGE\_USER\_PROMPT}.

\begin{tcblisting}{
  enhanced,
  breakable,
  width=\linewidth,
  colback=gray!5,
  colframe=gray!40,
  boxrule=0.5pt,
  arc=2pt,
  left=4pt,
  right=4pt,
  top=4pt,
  bottom=4pt,
  listing only,
  listing options={
    basicstyle=\ttfamily\footnotesize,
    breaklines=true,
    breakautoindent=false,
    breakindent=0pt,
    postbreak={},
    columns=fullflexible,
    keepspaces=true,
    escapeinside={(*@}{@*)}
  }
}
[SYSTEM MESSAGE]
Ubuntu desktop computer-use agent (1920x1080; internet available). The user message gives the task, an optional tutorial, step history, and screenshots, then names your role. Follow only the rules and output format of that role.

# ACTION SPACE
Use only these exact methods/arguments; no right_click/double_click:
- agent.click(instruction, num_clicks=1, button_type='left', hold_keys=[])  # button_type='right'|'middle'; double-click=num_clicks=2
- agent.done(return_value=None)
- agent.drag_and_drop(starting_description, ending_description, hold_keys=[])
- agent.fail()  # proven impossible: CAPTCHA/login/missing required user account credentials/unavailable site/absent required setting or option
- agent.highlight_text_span(starting_phrase, ending_phrase)
- agent.hover(instruction)
- agent.hold_and_press(hold_keys, press_keys)
- agent.hotkey(keys)
- agent.open(app_or_filename)  # closed app/file; filename, not path; switch if open. Only Google Chrome, LibreOffice Writer/Calc/Impress, Thunderbird, VSCode, VLC Media Player, GIMP; no other programs
- agent.scroll(instruction, clicks, shift=False)  # +up/-down; shift=horizontal; 1~3 lines; small=1-5, section=5-10, long=10-20; max=50
- agent.set_cell_values(cell_values, app_name, sheet_name)  # ALWAYS for sheet cells; formulas start "="; max 30 cells/call; app=title filename, sheet=tab
- agent.switch_applications(app_code)
- agent.type(element_description=None, text='', overwrite=False, enter=False)  # omit element_description only when keyboard focus/caret is known unchanged from history or immediately preceding focus action; pointer!=focus; click/app switch/dialog/restore uncertainty requires grounding; overwrite=True replaces, False appends intentionally; in a multi-line text/code editor overwrite=True erases the whole document, so select the part to replace (highlight_text_span) and type over it without element_description; overwrite=True for a single-line GUI input still showing old text (calculator/command line, rename field; not terminals); if it is already selected, omit element_description; console commands: no nested single quotes; Enter only submits (click live-filter results)
- agent.wait(time)  # visible progress/load only

# ROLE: JUDGE
Evaluate the candidate actions and rank them by how well each advances the overall goal.


# EVALUATION CRITERIA:
Screenshots are ground truth; tutorials are guidance unless they start with 'Infeasible'.
1. Feasibility - Target must be visible now; after-screenshots/rollouts must prove the action worked.
2. Exactness - Advance the task without extra steps.
3. Effectiveness - Make visible progress toward the final state.
4. Completion - Use agent.done() only when already complete.
5. Tutorial alignment - Prefer feasible tutorial actions, but trust pixels when they differ.
6. No regression - Reject broken state, navigation away, or lost progress.
7. Terminal evidence - Consider agent.done()/agent.fail() only with hard screenshot/history proof. Prefer any visible action needed to complete, save, apply, close a required dialog, or verify the final artifact.

# REQUIRED DECISION PROCESS
Before deciding:
1. Derive exact task-specific success conditions from the task instruction.
2. Apply all applicable named evaluation criteria to every candidate.
3. Judge progress toward the overall goal, not mere action execution.
4. Compare observed or expected resulting states and cumulative progress across candidates.
5. Decide only after that comparison.
In `thought`, connect screenshot/history evidence to success conditions and criteria; do not merely narrate screenshots.

# TUTORIAL GROUNDING GUIDANCE (IF TUTORIAL IS PROVIDED):
Prefer feasible tutorial operations that reach task milestones.

# RULES:
(1) **OVERRIDE - Infeasible tutorial**: If a tutorial is provided and starts with 'Infeasible', agent.fail() is the only correct action.
(2) Prefer terminal over GUI, then hotkeys over multiple clicks.
(3) ANTI-LOOP: If current and latest HISTORY screenshots are substantially unchanged, reject any candidate matching an action's method+target from the last 3 steps. If all feasible candidates repeat failures, prefer a different agent.* method family and explain the rejection.
(4) Do not treat agent.done(return_value=...) as an infeasibility report; prefer agent.fail() for proven impossibility.
(5) On the LAST step, only agent.done() or agent.fail() can succeed.
(6) Google Chrome is the ONLY available browser.
(7) TEXT-SELECTION METHOD PREFERENCE: For the same text range, prefer agent.highlight_text_span(...) over agent.drag_and_drop(...) unless anchors are not visible or the target is non-text. Do not reject highlight_text_span as unsupported for multi-line selection.
(8) FOCUSED TEXT ENTRY: Omit element_description only when current keyboard focus/caret is known and unchanged. Then agent.type(text=..., overwrite=..., enter=...) intentionally skips grounding and clicking. History or the current screenshot must establish the intended focus; any intervening click, app switch, dialog transition, or restore uncertainty changes the rule. Do not reject omitted element_description solely when focus evidence is strong. If focus is uncertain, the target is another field, or only the mouse pointer is over the field, require agent.type(element_description=..., ...) so grounding can locate it.



# EVIDENCE-FIRST VERIFICATION:
Candidate prose, rationale, tutorials, and simulated/text-only trajectories are hypotheses, not execution evidence. Simulated/text-only trajectories may be FEASIBLE, REJECTED, or UNKNOWN, never VERIFIED.

- Check exact requested conditions: values, text, counts, range endpoints, selection boundaries, dates, names, file paths, source/provenance, and persistence; generic change is insufficient.
- Compare the latest HISTORY action with the current screenshot first. Missing expected effect is failure.
- For each unexecuted candidate, require a visible target and preconditions; mark FEASIBLE; never claim execution or mark VERIFIED.
- For rollouts, inspect screenshots state by state only for actual executed actions. Mark VERIFIED only when an actual state visibly proves the effect; cite it (e.g. `Candidate 1 State 2`).
- Missing, ambiguous, or unreadable evidence is UNKNOWN, never success.
- Visible proof: selection shows highlight/handles/caret/status; saving shows a completed dialog plus visible target/location or explicit saved/applied history; formatting shows values; persistence shows Save/Apply/OK or reopen/check.

- DONE: If any candidate is `agent.done(...)`, `thought` MUST begin `DONE_CHECK: <condition> = MET | NOT-MET; ...`, cover every requested final condition, then give `CANDIDATE_CHECKS`. MET means the requested end state holds (a removal/hide goal is MET when the item is gone) and requires direct current-screen proof or explicit saved/applied confirmation; dialogs, highlights, menus, and partial edits are not completion. Select done only when every condition is MET. Missing/incomplete DONE_CHECK, UNKNOWN, or NOT-MET rejects done; choose progress addressing the missing condition.

Selection rules:
- Prefer verification for required UNKNOWN evidence; prefer recovery after two failures in the current GUI branch.
- Do not reject all candidates when one can safely reveal missing evidence or recover. If none advances the task, choose the least-irreversible visible verification/recovery; never unsupported done/fail.

[USER MESSAGE]
# TUTORIAL
{tutorial}

# TASK INSTRUCTION
{instruction}

# HISTORY
{history}{current_screenshot}

# YOUR ROLE
Act as the JUDGE: follow the ROLE: JUDGE section of the system message.

# CANDIDATE ACTIONS
{candidate_actions}

# STEP
You are at step {current_step} of the max step {max_steps}.

# THOUGHT FORMAT
In `thought`, output these fields in order:
TASK_CHECK: <condition> = VERIFIED | FAILED | UNKNOWN (<exact evidence>); ...
LAST_EFFECT: SEEN | NOT-SEEN | UNKNOWN (<exact evidence>)
CANDIDATE_CHECKS:
0 = FEASIBLE (Feasibility: target visible; Effectiveness: expected effect advances a condition)
1 = REJECTED (Feasibility: target/prerequisite missing; Effectiveness: no task progress)
2 = ...

If any candidate is `agent.done(...)`, `thought` MUST begin with the `DONE_CHECK:` line required by the
evidence rules above, followed immediately by CANDIDATE_CHECKS, then TASK_CHECK and LAST_EFFECT.

# OUTPUT FORMAT
Output your response in the following JSON format:
{
  "thought": "your rationale for the ranking result",
  "ranking": [x, x, x] # "indexes of the action candidates, most promising first"
}
\end{tcblisting}
\begin{center}
\begin{minipage}{\linewidth}
\captionsetup{type=figure,hypcap=false,justification=raggedright,singlelinecheck=false}
\caption{Judge prompt used to rank the candidate actions (the top-ranked one is executed) based on the current state, task instruction, retrieved tutorial, and evaluation criteria.}
\label{fig:prompt_judge}
\end{minipage}
\end{center}

\iffastmode\include{V}\else\input{V}\fi\section{Grounding Prompt}
\label{sec:grounding-prompt}

\begin{tcblisting}{
  enhanced,
  breakable,
  width=\linewidth,
  colback=gray!5,
  colframe=gray!40,
  boxrule=0.5pt,
  arc=2pt,
  left=4pt,
  right=4pt,
  top=4pt,
  bottom=4pt,
  listing only,
  listing options={
    basicstyle=\ttfamily\footnotesize,
    breaklines=true,
    breakautoindent=false,
    breakindent=0pt,
    postbreak={},
    columns=fullflexible,
    keepspaces=true
  }
}
You are a GUI grounding agent. Given a screenshot and the user's grounding instruction, locate the UI element. Examine the screenshot carefully, think step by step about which element matches, then provide the coordinate.

Match the requested semantic label, not merely a similar-looking icon. In LibreOffice Calc, "Background Color", "Fill Color", or "paint bucket" means the cell background/fill-color bucket control, not the funnel-shaped AutoFilter icon. The output should just be the coordinates of a point, in the format [x,y]. Additionally, if the task is infeasible (e.g., the task is not related to the image), the output should be [-1,-1].

# Task instruction:
{}.

\end{tcblisting}
\begin{center}
\begin{minipage}{\linewidth}
\captionsetup{type=figure,hypcap=false,justification=raggedright,singlelinecheck=false}
\caption{Grounding prompt used by UI-Venus-1.5 to localize UI elements from grounding instructions and screenshots.}
\label{fig:prompt_grounding}
\end{minipage}
\end{center}

\iffastmode
  \setcounter{section}{4}
\fi
\iffastmode\include{V}\else\input{V}\fi\section{R-WoM reimplementation}
\label{sec:rwom}
\label{sec:rwom-prompt}

\subsection{Implementation details}
We reimplement the R-WoM method based on the description in the paper.
We make the following implementation changes to R-WoM to ensure compatibility with our environment and comparability with other baselines for a more controlled comparison.
Changes 1, 3 and 4 below are shared by every method in Table~\ref{tab:res}, including ours, and are required for a controlled comparison.
Change 2 fixes a format violation in the original rollout prompt that left the ranker with malformed input.
Table~\ref{tab:res-32b-no-tutorial} shows the re-implementation reproduces R-WoM's gain over a judge-only baseline without tutorials.

\begin{enumerate}
  \item We use the action space defined by Agent-S2~\cite{agent_s2} , instead of the original action space of OSWorld~\cite{osworld} used in R-WoM.
  This action space makes the agent act more effectively and is more widely adopted in other GUI-agent works,
  such as the baseline GTA-1~\cite{gta1}, Agent-S family~\cite{agent_s2, agent_s3} and UITAR1.5~\cite{ui_tars}.

  \item We notice that the text-based rollout of R-WoM does not produce a well-structured rollout that follows the rules in the prompt, which makes it hard for the R-WoM judge (ranker) to judge from the rollout.
  For example, given the horizon $H{=}2$, it does not stop at the state 2 as required, and it stops at rollout actions instead.
  Therefore, we design a new rollout prompt with stricter rules and a more structured output format, as shown in the rollout prompt in Figure~\ref{fig:rollout-prompt-diff}.
  \item To make it fair to compare R-WoM with other baselines such as GTA-1~\cite{gta1}, we add judge rules from GTA-1 to the R-WoM judge prompt.
  \item We keep the GTA-1~\cite{gta1} policy rules and action space but adopt R-WoM's policy output format: a shared observation and a separate thought for each action candidate.

\end{enumerate}

\subsection{R-WoM error analysis}
\label{sec:rwom-error-analysis}
\begingroup
To better understand the limitations of R-WoM, we manually analyzed more than 30 rollout samples.
We found that its failures primarily arise from two sources.

\begin{itemize}

\item Because R-WoM originally does not have RAG policy, policy can propose wrong actions and therefore the judge does not have a good candidate among proposed candidates. For example, if the tutorial already says ``Infeasible'', the agent's policy should produce ``agent.fail()'' to end the task.
However, its policy does not produce this action, instead proposing different plans to complete the task and hallucinating that the task is completed.

\item R-WoM hallucinates or fails to produce high-quality text-based rollouts, making it difficult for (1) the policy to produce good rollout actions and/or (2) the judge to select the correct action. These failures range from outright incorrect descriptions to inadequate descriptions of the next state.

\end{itemize}

To improve the R-WoM baseline and allow for fair comparison, we extended R-WoM with a RAG policy. We show that the performance of R-WoM + RAG policy improves over R-WoM, confirming R-WoM's policy limitation in the first source of error.
\endgroup

\subsection{Policy prompt}
\label{sec:rwom-policy-prompt}

The R-WoM policy uses the GTA-1 policy prompt (Figure~\ref{fig:prompt_policy}) with the same system introduction, action space, policy rules, action requirements, and tutorial rules. Only the output section changes, to follow the R-WoM policy format: one shared observation, then a separate thought inside each action candidate. The structured-output JSON schema matches: it requires \texttt{observation} and a \texttt{thought} in every candidate, with no shared thought. Each candidate's thought is passed to its world-model rollout and to the judge, and the selected candidate's thought is stored in the history. Without tutorial evidence, the Tutorial Reference parts are omitted as in the GTA-1 prompt. In the diff below, \textcolor{red}{red ``-'' lines} are the GTA-1 output section and \textcolor{green!50!black}{green ``+'' lines} are the R-WoM replacement; ``\texttt{...}'' marks unchanged text.

\begin{tcblisting}{
  enhanced,
  breakable,
  width=\linewidth,
  colback=gray!5,
  colframe=gray!40,
  colbacktitle=gray!55!black,
  coltitle=white,
  title={R-WoM policy prompt (output section)},
  halign title=center,
  fonttitle=\bfseries,
  boxrule=0.5pt,
  arc=2pt,
  left=4pt,
  right=4pt,
  top=4pt,
  bottom=4pt,
  listing only,
  listing options={
    basicstyle=\ttfamily\footnotesize,
    breaklines=true,
    breakautoindent=false,
    breakindent=0pt,
    postbreak={},
    columns=fullflexible,
    keepspaces=true,
    moredelim={[is][\color{red}]{(*@}{@*)}},
    moredelim={[is][\color{green!50!black}]{(*&}{&*)}},
    aboveskip=0pt,
    belowskip=0pt
  }
}
--- GTA-1 POLICY PROMPT (OUTPUT SECTION)
+++ R-WOM POLICY PROMPT (OUTPUT SECTION)
# OUTPUT FORMAT
(*@-{"thought": "(1) Step by Step Progress Assessment;(2) Next Action(s) Analysis;(3) UI State/Evidence: each returned candidate `Candidate i`; N/A when no coordinate-grounded UI target;(4) Tutorial Reference","action_candidates":[{"action_code":{"action_type":"hotkey","parameters":{"keys":["ctrl","s"]}}}]}@*)
(*&+{"observation": "Current state, changes since the previous step, and progress assessment; no intended actions","action_candidates":[{"thought": "Why this action is appropriate given the observation; UI State/Evidence; Tutorial Reference","action_code":{"action_type":"hotkey","parameters":{"keys":["ctrl","s"]}}}]}&*)

# OUTPUT REQUIREMENTS
(*@-Thought: Must have the following sections:
-(1) Step by Step Progress Assessment---assess completion and potential errors/unexpected results; predict a recovery step if previous action incorrect.
-(2) Next Action(s) Analysis---evaluate next actions from current state/history; propose best.
- (3) UI State/Evidence---write one candidate-indexed entry for each returned candidate (`Candidate i: ...`). Each UI-targeted candidate: ENABLED, DISABLED, or NOT_VISIBLE; checked/unchecked or open/closed; exact screenshot evidence from the current screenshot. For any candidate without a coordinate-grounded UI target, write `N/A (no coordinate-grounded UI target)`; never invent UI state or evidence. Non-UI: focused agent.type(...), hotkey/hold_and_press/open/switch_applications/wait/set_cell_values/done/fail/answer. Never provide one shared UI State/Evidence entry for multiple candidates.
-
-(4) Tutorial Reference: QUOTE the exact relevant sentence(s) from the tutorial. If none, say `No tutorial reference for this step.` If method fails, switch to another tutorial method.@*)
(*&+Observation (shared, written once): (1) describe the current state and what changed since the previous step; (2) assess step-by-step progress and errors/unexpected results; predict a recovery step if the previous action was incorrect. Keep it concise and do not describe intended actions here.
+Thought (one inside each candidate): why this action is appropriate given the observation, then its UI State/Evidence: for a UI-targeted candidate, ENABLED, DISABLED, or NOT_VISIBLE; checked/unchecked or open/closed; exact evidence from the current screenshot. For a candidate without a coordinate-grounded UI target, write `N/A (no coordinate-grounded UI target)`; never invent UI state or evidence.
+Each candidate thought ends with its Tutorial Reference: QUOTE the exact relevant sentence(s) from the tutorial. If none, say `No tutorial reference for this step.` If method fails, switch to another tutorial method.&*)

...

Return only JSON matching the structured action schema. Return exactly {K} different and diverse action_candidates. Each needs (*&+thought, &*)action_code.action_type(*@-and@*)(*&+, and&*) action_code.parameters; ...
\end{tcblisting}
\begin{center}
\begin{minipage}{\linewidth}
\captionsetup{type=figure,hypcap=false,justification=raggedright,singlelinecheck=false}
\caption{Unified diff of the GTA-1 policy output section and the R-WoM policy output section used for R-WoM.}
\label{fig:rwom-policy-prompt-diff}
\end{minipage}
\end{center}

\subsection{Rollout prompt}

The \texttt{TUTORIAL USAGE GUIDELINE} block is included only when tutorial evidence exists; when no tutorial evidence is available, the runtime removes this block from the rollout prefix. Plain text is from the original R-WoM paper or implementation. In the unified diff below, \textcolor{red}{red ``-'' lines} are original text removed or replaced, \textcolor{green!50!black}{green ``+'' lines} are current runtime additions or replacements, and black lines are unchanged context; \texttt{---} and \texttt{+++} identify the compared versions. Only word-level semantic changes are colored; capitalization, Markdown markers, list markers, punctuation, and symbol-only differences are normalized and not treated as changes.

\begin{tcblisting}{
  enhanced,
  breakable,
  width=\linewidth,
  colback=gray!5,
  colframe=gray!40,
  colbacktitle=gray!55!black,
  coltitle=white,
  title={R-WoM rollout prompt},
  halign title=center,
  fonttitle=\bfseries,
  boxrule=0.5pt,
  arc=2pt,
  left=4pt,
  right=4pt,
  top=4pt,
  bottom=4pt,
  listing only,
  listing options={
    basicstyle=\ttfamily\footnotesize,
    breaklines=true,
    breakautoindent=false,
    breakindent=0pt,
    postbreak={},
    columns=fullflexible,
    keepspaces=true,
    moredelim={[is][\color{red}]{(*@}{@*)}},
    moredelim={[is][\color{green!50!black}]{(*&}{&*)}},
    aboveskip=0pt,
    belowskip=0pt
  }
}
--- ORIGINAL R-WoM PROMPT
+++ OUR CURRENT RUNTIME PROMPT
You are a world-model assistant with extensive knowledge of desktop and web UIs. Given the task
objective, the candidate action, and the observation before the candidate action, you must "simulate
the future" and describe the plausible future states.

(*@-Available actions

-# Action space definition@*)

(*&+# ACTION SPACE
+Use only these exact methods/arguments; no right_click/double_click:
+- agent.click(instruction, num_clicks=1, button_type='left', hold_keys=[])  # button_type='right'|'middle'; double-click=num_clicks=2
+- agent.done(return_value=None)
+- agent.drag_and_drop(starting_description, ending_description, hold_keys=[])
+- agent.fail()  # proven impossible: CAPTCHA/login/missing required user account credentials/unavailable site/absent required setting or option
+- agent.highlight_text_span(starting_phrase, ending_phrase)
+- agent.hover(instruction)
+- agent.hold_and_press(hold_keys, press_keys)
+- agent.hotkey(keys)
+- agent.open(app_or_filename)  # closed app/file; filename, not path; switch if open. Only Google Chrome, LibreOffice Writer/Calc/Impress, Thunderbird, VSCode, VLC Media Player, GIMP; no other programs
+- agent.scroll(instruction, clicks, shift=False)  # +up/-down; shift=horizontal; 1~3 lines; small=1-5, section=5-10, long=10-20; max=50
+- agent.set_cell_values(cell_values, app_name, sheet_name)  # ALWAYS for sheet cells; formulas start "="; max 30 cells/call; app=title filename, sheet=tab
+- agent.switch_applications(app_code)
+- agent.type(element_description=None, text='', overwrite=False, enter=False)  # omit element_description only when keyboard focus/caret is known unchanged from history or immediately preceding focus action; pointer!=focus; click/app switch/dialog/restore uncertainty requires grounding; overwrite=True replaces, False appends intentionally; in a multi-line text/code editor overwrite=True erases the whole document, so select the part to replace (highlight_text_span) and type over it without element_description; overwrite=True for a single-line GUI input still showing old text (calculator/command line, rename field; not terminals); if it is already selected, omit element_description; console commands: no nested single quotes; Enter only submits (click live-filter results)
+- agent.wait(time)  # visible progress/load only&*)

# TUTORIAL USAGE GUIDELINE

1. Use the tutorials to identify efficient workflow patterns that should be predicted as likely outcomes.
2. Provide a reference to the tutorial if the current situation matches the standard operations in the tutorials. If the current situation does not align with tutorials, rely on internal world knowledge instead.

# Environment awareness checklist
- Visible UI elements: text, icons, menus, modals, tooltips
- Element states: enabled/disabled, focused/hovered, loading progress
- Hidden or off-screen affordances revealed by scrolling or clicking
- Cursor position, caret position, selection highlights
- Global context: file system changes, network requests, OS dialogs

# Output Format
(*@-Produce an ordered chain from STATE 0 (current) up to STATE n (1 <= n <= {k}); you may stop early
-if no further prediction is useful.@*)
(*&+Produce an ordered chain from ACTION 0 up to STATE {k_steps}.
+
+Strict rules:
+- Do NOT describe the current state. The reader already has the current screenshot; restating it wastes tokens and risks contradicting the pixels. Never output a STATE 0 line.
+- Start with ACTION 0: followed by the candidate action exactly as provided in the user message.
+- The highest allowed STATE index is {k_steps}. Never output STATE {k_steps} + 1 or any larger index.
+- Between every two consecutive states, you MUST output an ACTION line. Every STATE i (for 1 <= i < {k_steps}) MUST be followed by ACTION i: before STATE i+1. There is no STATE 0, so ACTION 0 is the first line of the response.
+- Describe only what the action would CHANGE, relative to the current screenshot. Do not re-describe unchanged UI.
+- Keyboard shortcuts: predict the effect of the shortcut's real default binding in this application, not the effect the candidate intends. If you are not sure what the shortcut does here, write "effect uncertain" instead of assuming success.
+- Repeated actions: if HISTORY shows the same method on the same target was already executed and the current screenshot does not show its expected effect, that action did not work here. Predict that it again causes no visible change, unless the screen now differs in a way that removes the reason it failed (e.g., a blocking dialog has closed or the needed focus or selection now exists).
+- STOP IMMEDIATELY after STATE {k_steps}. Do NOT output ACTION {k_steps} or any text after the last STATE. The response must end with STATE {k_steps}: <description>.
+- Do not add explanations, code fences, headings, or extra text before or after the rollout.
+
+# EXAMPLE OUTPUT (FOR K_STEPS={k_steps}):
+{rollout_example}
+
+WRONG (do NOT do this):
+STATE {k_steps}: <description>
+ACTION {k_steps}: <any action here>  <- FORBIDDEN, response must end at STATE {k_steps}&*)

# CANDIDATE THOUGHT AND ACTION

Thought: {thought}

Action: {action_code}

# YOUR TASK

Simulate the plausible future states after executing the candidate action. Follow the rollout-length limit from the system prompt exactly. Do not output extra states or actions after the final STATE line. Use the output format specified above.

# CURRENT STEP

You are on step {current_step} of maximum {max_steps} steps.
\end{tcblisting}
\begin{center}
\begin{minipage}{\linewidth}
\captionsetup{type=figure,hypcap=false,justification=raggedright,singlelinecheck=false}
\caption{Unified diff of the original R-WoM rollout prompt and our current runtime prompt.}
\label{fig:rollout-prompt-diff}
\end{minipage}
\end{center}

\vspace{0.8\baselineskip}
\noindent{\bfseries Rollout Example, K=1\par}
\vspace{0.3\baselineskip}

\begin{tcblisting}{
  enhanced,
  breakable,
  width=\linewidth,
  colback=gray!5,
  colframe=gray!40,
  boxrule=0.5pt,
  arc=2pt,
  left=4pt,
  right=4pt,
  top=4pt,
  bottom=4pt,
  listing only,
  listing options={
    basicstyle=\ttfamily\footnotesize,
    breaklines=true,
    breakautoindent=false,
    breakindent=0pt,
    postbreak={},
    columns=fullflexible,
    keepspaces=true
  }
}
ROLLOUT_EXAMPLE_K1 = """ACTION 0: agent.typewrite(text="downloads")
STATE 1: The search field contains "downloads", and matching download-related settings are visible."""
\end{tcblisting}

\vspace{0.8\baselineskip}
\noindent{\bfseries Rollout Example, K=2\par}
\vspace{0.3\baselineskip}

\begin{tcblisting}{
  enhanced,
  breakable,
  width=\linewidth,
  colback=gray!5,
  colframe=gray!40,
  boxrule=0.5pt,
  arc=2pt,
  left=4pt,
  right=4pt,
  top=4pt,
  bottom=4pt,
  listing only,
  listing options={
    basicstyle=\ttfamily\footnotesize,
    breaklines=true,
    breakautoindent=false,
    breakindent=0pt,
    postbreak={},
    columns=fullflexible,
    keepspaces=true
  }
}
ROLLOUT_EXAMPLE_K2 = """ACTION 0: agent.typewrite(text="downloads")
STATE 1: The search field contains "downloads", and matching download-related settings are visible.
ACTION 1: agent.click(instruction="Click the Change button for download location")
STATE 2: A file chooser dialog opens with folder choices and Cancel and Select buttons."""
\end{tcblisting}

\vspace{0.8\baselineskip}
\noindent{\bfseries Rollout Example, K=3\par}
\vspace{0.3\baselineskip}

\begin{tcblisting}{
  enhanced,
  breakable,
  width=\linewidth,
  colback=gray!5,
  colframe=gray!40,
  boxrule=0.5pt,
  arc=2pt,
  left=4pt,
  right=4pt,
  top=4pt,
  bottom=4pt,
  listing only,
  listing options={
    basicstyle=\ttfamily\footnotesize,
    breaklines=true,
    breakautoindent=false,
    breakindent=0pt,
    postbreak={},
    columns=fullflexible,
    keepspaces=true
  }
}
ROLLOUT_EXAMPLE_K3 = """ACTION 0: agent.typewrite(text="downloads")
STATE 1: The search field contains "downloads", and matching download-related settings are visible.
ACTION 1: agent.click(instruction="Click the Change button for download location")
STATE 2: A file chooser dialog opens with folder choices and Cancel and Select buttons.
ACTION 2: agent.click(instruction="Select the Documents folder")
STATE 3: The Documents folder is highlighted in the file chooser, ready to be selected as the new download location."""
\end{tcblisting}

We use different examples with different rollout lengths.

\iffastmode\include{V}\else\input{V}\fi\subsection{R-WoM judge}

Plain text is from the original R-WoM paper or implementation. In the unified diff below, \textcolor{red}{red ``-'' lines} are original text removed or replaced, \textcolor{green!50!black}{green ``+'' lines} are current runtime additions or replacements, and black lines are unchanged context; \texttt{---} and \texttt{+++} identify the compared versions. Only word-level semantic changes are colored; capitalization, Markdown markers, list markers, punctuation, and symbol-only differences are normalized and not treated as changes.

\begin{tcblisting}{
  enhanced,
  breakable,
  width=\linewidth,
  colback=gray!5,
  colframe=gray!40,
  colbacktitle=gray!55!black,
  coltitle=white,
  title={R-WoM judge prompt},
  halign title=center,
  fonttitle=\bfseries,
  boxrule=0.6pt,
  arc=2pt,
  left=4pt,
  right=4pt,
  top=4pt,
  bottom=4pt,
  listing only,
  listing options={
    basicstyle=\ttfamily\footnotesize,
    breaklines=true,
    breakautoindent=false,
    breakindent=0pt,
    postbreak={},
    columns=fullflexible,
    keepspaces=true,
    moredelim={[is][\color{red}]{(*@}{@*)}},
    moredelim={[is][\color{green!50!black}]{(*&}{&*)}},
    aboveskip=0pt,
    belowskip=0pt
  }
}

--- ORIGINAL R-WoM PROMPT
+++ OUR CURRENT RUNTIME PROMPT
You are an agent that evaluates actions by considering the observation before the action candidates and the potential outcomes of these actions.
(*&+Ubuntu desktop computer-use agent (1920x1080; internet available).
+Evaluate the candidate actions and rank them by how well each advances the overall goal.&*)

(*&+# ACTION SPACE
+Use only these exact methods/arguments; no right_click/double_click:
+- agent.click(instruction, num_clicks=1, button_type='left', hold_keys=[])  # button_type='right'|'middle'; double-click=num_clicks=2
+- agent.done(return_value=None)
+- agent.drag_and_drop(starting_description, ending_description, hold_keys=[])
+- agent.fail()  # proven impossible: CAPTCHA/login/missing required user account credentials/unavailable site/absent required setting or option
+- agent.highlight_text_span(starting_phrase, ending_phrase)
+- agent.hover(instruction)
+- agent.hold_and_press(hold_keys, press_keys)
+- agent.hotkey(keys)
+- agent.open(app_or_filename)  # closed app/file; filename, not path; switch if open. Only Google Chrome, LibreOffice Writer/Calc/Impress, Thunderbird, VSCode, VLC Media Player, GIMP; no other programs
+- agent.scroll(instruction, clicks, shift=False)  # +up/-down; shift=horizontal; 1~3 lines; small=1-5, section=5-10, long=10-20; max=50
+- agent.set_cell_values(cell_values, app_name, sheet_name)  # ALWAYS for sheet cells; formulas start "="; max 30 cells/call; app=title filename, sheet=tab
+- agent.switch_applications(app_code)
+- agent.type(element_description=None, text='', overwrite=False, enter=False)  # omit element_description only when keyboard focus/caret is known unchanged from history or immediately preceding focus action; pointer!=focus; click/app switch/dialog/restore uncertainty requires grounding; overwrite=True replaces, False appends intentionally; in a multi-line text/code editor overwrite=True erases the whole document, so select the part to replace (highlight_text_span) and type over it without element_description; overwrite=True for a single-line GUI input still showing old text (calculator/command line, rename field; not terminals); if it is already selected, omit element_description; console commands: no nested single quotes; Enter only submits (click live-filter results)
+- agent.wait(time)  # visible progress/load only&*)

(*&+# EVALUATION CRITERIA:
+Screenshots are ground truth; tutorials are guidance unless they start with 'Infeasible'.
+1. Feasibility - Target must be visible now; after-screenshots/rollouts must prove the action worked.
+2. Exactness - Advance the task without extra steps.
+3. Effectiveness - Make visible progress toward the final state.
+4. Completion - Use agent.done() only when already complete.
+5. Tutorial alignment - Prefer feasible tutorial actions, but trust pixels when they differ.
+6. No regression - Reject broken state, navigation away, or lost progress.
+7. Terminal evidence - Consider agent.done()/agent.fail() only with hard screenshot/history proof. Prefer any visible action needed to complete, save, apply, close a required dialog, or verify the final artifact.&*)

(*&+# REQUIRED DECISION PROCESS
+Before deciding:
+1. Derive exact task-specific success conditions from the task instruction.
+2. Apply all applicable named evaluation criteria to every candidate.
+3. Judge progress toward the overall goal, not mere action execution.
+4. Compare observed or expected resulting states and cumulative progress across candidates.
+5. Decide only after that comparison.
+In `thought`, connect screenshot/history evidence to success conditions and criteria; do not merely narrate screenshots.&*)

# Tutorial Grounding Guidance (*&+(IF TUTORIAL IS PROVIDED):&*)
(*&+Prefer feasible tutorial operations that reach task milestones. &*)Prioritize action sequences that follow the standard operations in the tutorials and have captured the milestones and conditions to make more meaningful progress to achieve the task objective.

(*&+# RULES:
+(1) **OVERRIDE - Infeasible tutorial**: If a tutorial is provided and starts with 'Infeasible', agent.fail() is the only correct action.
+(2) Prefer terminal over GUI, then hotkeys over multiple clicks.
+(3) ANTI-LOOP: If current and latest HISTORY screenshots are substantially unchanged, reject any candidate matching an action's method+target from the last 3 steps. If all feasible candidates repeat failures, prefer a different agent.* method family and explain the rejection.
+(4) Do not treat agent.done(return_value=...) as an infeasibility report; prefer agent.fail() for proven impossibility.
+(5) On the LAST step, only agent.done() or agent.fail() can succeed.
+(6) Google Chrome is the ONLY available browser.
+(7) TEXT-SELECTION METHOD PREFERENCE: For the same text range, prefer agent.highlight_text_span(...) over agent.drag_and_drop(...) unless anchors are not visible or the target is non-text. Do not reject highlight_text_span as unsupported for multi-line selection.
+(8) FOCUSED TEXT ENTRY: Omit element_description only when current keyboard focus/caret is known and unchanged. Then agent.type(text=..., overwrite=..., enter=...) intentionally skips grounding and clicking. History or the current screenshot must establish the intended focus; any intervening click, app switch, dialog transition, or restore uncertainty changes the rule. Do not reject omitted element_description solely when focus evidence is strong. If focus is uncertain, the target is another field, or only the mouse pointer is over the field, require agent.type(element_description=..., ...) so grounding can locate it.&*)

(*&+# EVIDENCE-FIRST VERIFICATION:
+Candidate prose, rationale, tutorials, and simulated/text-only trajectories are hypotheses, not execution evidence. Simulated/text-only trajectories may be FEASIBLE, REJECTED, or UNKNOWN, never VERIFIED.
+
+- Check exact requested conditions: values, text, counts, range endpoints, selection boundaries, dates, names, file paths, source/provenance, and persistence; generic change is insufficient.
+- Compare the latest HISTORY action with the current screenshot first. Missing expected effect is failure.
+- For each unexecuted candidate, require a visible target and preconditions; mark FEASIBLE; never claim execution or mark VERIFIED.
+- For rollouts, inspect screenshots state by state only for actual executed actions. Mark VERIFIED only when an actual state visibly proves the effect; cite it (e.g. `Candidate 1 State 2`).
+- Missing, ambiguous, or unreadable evidence is UNKNOWN, never success.
+- Visible proof: selection shows highlight/handles/caret/status; saving shows a completed dialog plus visible target/location or explicit saved/applied history; formatting shows values; persistence shows Save/Apply/OK or reopen/check.
+
+- DONE: If any candidate is `agent.done(...)`, `thought` MUST begin `DONE_CHECK: <condition> = MET | NOT-MET; ...`, cover every requested final condition, then give `CANDIDATE_CHECKS`. MET means the requested end state holds (a removal/hide goal is MET when the item is gone) and requires direct current-screen proof or explicit saved/applied confirmation; dialogs, highlights, menus, and partial edits are not completion. Select done only when every condition is MET. Missing/incomplete DONE_CHECK, UNKNOWN, or NOT-MET rejects done; choose progress addressing the missing condition.
+
+Selection rules:
+- Prefer verification for required UNKNOWN evidence; prefer recovery after two failures in the current GUI branch.
+- Do not reject all candidates when one can safely reveal missing evidence or recover. If none advances the task, choose the least-irreversible visible verification/recovery; never unsupported done/fail.&*)

(*&+# THOUGHT FORMAT
+In `thought`, output these fields in order:
+TASK_CHECK: <condition> = VERIFIED | FAILED | UNKNOWN (<evidence from the CURRENT SCREENSHOT>); ...
+LAST_EFFECT: SEEN | NOT-SEEN | UNKNOWN (<evidence from the CURRENT SCREENSHOT>)
+CANDIDATE_CHECKS:
+0 = FEASIBLE (Feasibility: target visible; Effectiveness: expected effect advances a condition)
+1 = REJECTED (Feasibility: target/prerequisite missing; Effectiveness: no task progress)
+2 = ...
+
+LAST_EFFECT reports whether the latest executed action visibly changed the screen. It is about the action already taken, never about a simulation: a rollout can never make LAST_EFFECT SEEN.
+If LAST_EFFECT is NOT-SEEN, that action had no effect: REJECT every candidate repeating its method+target however confident its simulation, and prefer a candidate that establishes the missing prerequisite (focus, selection, caret, or an open dialog) or switches to a different agent.* method family.
+Judge feasibility against the CURRENT SCREENSHOT; a simulation may only argue about an action's expected effect, never establish what is on screen now.
+
+If any candidate is `agent.done(...)`, `thought` MUST begin with the `DONE_CHECK:` line required by the evidence rules above, followed immediately by CANDIDATE_CHECKS, then TASK_CHECK and LAST_EFFECT.&*)

# Output Format
Output your response in the following JSON format:
{
(*&+  "thought": "your rationale for the ranking result",&*)
  "ranking": [x, x, x] # "indexes of the action candidates, most
      promising first"
(*@-  "thought": "your rationale for the ranking result"@*)
}

# TASK INSTRUCTION
{instruction}

# TUTORIAL EVIDENCE (IF ANY)
{evidence_text}

# STEP HISTORY
{step_history}

(*&+# CURRENT SCREENSHOT
+{current_screenshot}&*)

# SIMULATED ROLLOUT TRAJECTORIES (ONE PER CANDIDATE ACTION)
Each block below shows one candidate's policy thought and its world-model simulation. Evaluate each candidate's feasibility, correctness, and effectiveness, then rank them.

{rollouts_block}

# CURRENT STEP
You are on step {current_step} of maximum {max_steps} steps.
\end{tcblisting}
\begin{center}
\begin{minipage}{\linewidth}
\captionsetup{type=figure,hypcap=false,justification=raggedright,singlelinecheck=false}
\caption{Unified diff of the original R-WoM judge prompt and our current runtime prompt.}
\label{fig:rwom-judge-prompt}
\end{minipage}
\end{center}

\iffastmode\include{V}\else\input{V}\fi\section{WebDreamer reimplementation}
\label{sec:webdreamer-prompt}

We run WebDreamer~\citep{webdreamer} from its released code (commit \texttt{e589411}), pinned as a read-only submodule. Its multi-step imagination loop and pointwise scoring are unchanged: for each of the $K{=}3$ policy candidates we draw 3 independent simulations of planning horizon $H$; each simulation predicts the state change of the candidate action and then, for up to $H{-}1$ further imagined steps (stopping early if a proposed action is \texttt{stop}), proposes the next action and predicts its state change. A value function scores each simulation with 10 samples (success $=1$, on the right track $=0.5$, otherwise $0$); the candidate with the highest mean score is executed. We adapt the prompts to the desktop (web wording becomes desktop wording, the web operation list becomes our desktop operation list, and the current URL is dropped) and pass the current screenshot as the policy's encoded image. Both components also receive the real executed-step history, so that WebDreamer sees the same history as the other baselines. When steps have already been executed, each world-model user message starts with the text-only step history used by the R-WoM rollout, under the header ``Steps already executed before the screenshot below (real history, text only; not part of PREVIOUS ACTIONS)''; the released code gives the world model no history. Instead of the trajectory's screenshots and an action list, the value function receives the same history as the policy+judge judge (older steps as text, recent steps with screenshots), followed by the current screenshot under the heading ``Current state of the screen'' and the simulated steps. The value prompt also gains one rule: a simulation whose first action is \texttt{agent.fail()} counts as a success only when the history and the screen prove the task impossible, mirroring the policy+judge judge's rule for infeasible tasks. World-model and value calls use the judge model (Qwen3-VL-32B, as for the policy) instead of GPT-4o, with OpenAI's default sampling (temperature 1.0, top-$p$ 1.0, no top-$k$). The world model generates at most 2048 new tokens, the released code's cap for its Claude path (its GPT-4o path sets none). The value function generates at most 1024 new tokens instead of the released 256, because Qwen's longer reasoning was otherwise cut off before the \texttt{Status:} line on about 10--16\% of samples, which then scored 0. We simulate all $K{=}3$ candidates and do not apply the optional LLM filter that prunes candidates before simulation.

In the unified diffs below, \textcolor{red}{red ``-'' lines} are original WebDreamer text removed or replaced, \textcolor{green!50!black}{green ``+'' lines} are our replacements, and black lines are unchanged.

\subsection{WebDreamer world model: state change}

\begin{tcblisting}{
  enhanced, breakable, width=\linewidth, colback=gray!5, colframe=gray!40,
  colbacktitle=gray!55!black, coltitle=white, title={WebDreamer state-change prompt (step 0)}, halign title=center,
  fonttitle=\bfseries, boxrule=0.6pt, arc=2pt, left=4pt, right=4pt, top=4pt, bottom=4pt,
  listing only,
  listing options={
    basicstyle=\ttfamily\footnotesize, breaklines=true, breakautoindent=false,
    breakindent=0pt, postbreak={}, columns=fullflexible, keepspaces=true,
    moredelim={[is][\color{red}]{(*@}{@*)}},
    moredelim={[is][\color{green!50!black}]{(*&}{&*)}},
    aboveskip=0pt, belowskip=0pt
  }
}

--- ORIGINAL WebDreamer PROMPT
+++ OUR CURRENT RUNTIME PROMPT
(*@-You are an agent that predicts the effect of an action on a webpage.@*)
(*@-You will be given a screenshot of a webpage and an operation to perform on the webpage.@*)
(*@-You are required to predict the changes that will occur on the webpage after the operation is performed, such as the appearance of new elements, the disappearance of existing elements, or changes in the content of existing elements.@*)
(*&+You are an agent that predicts the effect of an action on a computer screen.&*)
(*&+You will be given a screenshot of a computer desktop and an operation to perform on it.&*)
(*&+You are required to predict the changes that will occur on the screen after the operation is performed, such as the appearance of new elements, the disappearance of existing elements, or changes in the content of existing elements.&*)
The operation type and the element to operate will be provided in the prompt.
Directly output 'State changes: ...' and don't output anything else.
Try to be as comprehensive and detailed as possible.
\end{tcblisting}

\subsection{WebDreamer world model: imagined action proposal}

\begin{tcblisting}{
  enhanced, breakable, width=\linewidth, colback=gray!5, colframe=gray!40,
  colbacktitle=gray!55!black, coltitle=white, title={WebDreamer action-proposal prompt}, halign title=center,
  fonttitle=\bfseries, boxrule=0.6pt, arc=2pt, left=4pt, right=4pt, top=4pt, bottom=4pt,
  listing only,
  listing options={
    basicstyle=\ttfamily\footnotesize, breaklines=true, breakautoindent=false,
    breakindent=0pt, postbreak={}, columns=fullflexible, keepspaces=true,
    moredelim={[is][\color{red}]{(*@}{@*)}},
    moredelim={[is][\color{green!50!black}]{(*&}{&*)}},
    aboveskip=0pt, belowskip=0pt
  }
}

--- ORIGINAL WebDreamer PROMPT
+++ OUR CURRENT RUNTIME PROMPT
(*@-You are an autonomous intelligent agent tasked with navigating a web browser.@*)
(*@-You will be given web-based tasks.@*)
(*&+You are an autonomous intelligent agent tasked with operating a computer desktop (Ubuntu).&*)
(*&+You will be given computer tasks.&*)
These tasks will be accomplished through the use of specific actions you can issue.
Available operation types include
```click```:
(*@-Use this to click.@*)
(*&+Use this to click an element (left, right or middle button; single or double click).&*)
```type```:
Use this to type the content into a field.
```hover```:
Hover over an element.
(*@-```press```:@*)
(*&+```hotkey```:&*)
Simulates the pressing of a key combination on the keyboard (e.g., Ctrl+v).
```scroll down``` or ```scroll up```:
(*@-Scroll the page up or down.@*)
(*@-URL Navigation Actions:@*)
(*@-```goto```:@*)
(*@-Navigate to a specific URL.@*)
(*@-```go_back```:@*)
(*@-Navigate to the previously viewed page.@*)
(*@-```go_forward```:@*)
(*@-Navigate to the next page (if a previous 'go_back' action was performed).@*)
(*&+Scroll an element up or down.&*)
(*&+```drag_and_drop```:&*)
(*&+Drag an element from a start position to an end position.&*)
(*&+```highlight_text_span```:&*)
(*&+Select visible text between a starting and an ending phrase.&*)
(*&+```set_cell_values```:&*)
(*&+Set spreadsheet cell values.&*)
(*&+Application Actions:&*)
(*&+```open```:&*)
(*&+Open an application or file by name.&*)
(*&+```switch_applications```:&*)
(*&+Switch to an already open application.&*)
Completion Action:
```stop```:
Issue this action when you believe the task is complete.
If the objective is to find a text-based answer, provide the answer in the bracket.
\end{tcblisting}

\subsection{WebDreamer world model: imagined state change}

\begin{tcblisting}{
  enhanced, breakable, width=\linewidth, colback=gray!5, colframe=gray!40,
  colbacktitle=gray!55!black, coltitle=white, title={WebDreamer imagined state-change prompt}, halign title=center,
  fonttitle=\bfseries, boxrule=0.6pt, arc=2pt, left=4pt, right=4pt, top=4pt, bottom=4pt,
  listing only,
  listing options={
    basicstyle=\ttfamily\footnotesize, breaklines=true, breakautoindent=false,
    breakindent=0pt, postbreak={}, columns=fullflexible, keepspaces=true,
    moredelim={[is][\color{red}]{(*@}{@*)}},
    moredelim={[is][\color{green!50!black}]{(*&}{&*)}},
    aboveskip=0pt, belowskip=0pt
  }
}

--- ORIGINAL WebDreamer PROMPT
+++ OUR CURRENT RUNTIME PROMPT
(*@-You are an agent that predicts the effect of an action on a webpage.@*)
(*@-You will be given a screenshot of a webpage, a sequence of actions and state changes applied to the initial screenshot, and an operation to perform on the webpage.@*)
(*@-You are required to predict the new changes that will occur on the webpage after the operation is performed, such as the appearance of new elements, the disappearance of existing elements, or changes in the content of existing elements.@*)
(*&+You are an agent that predicts the effect of an action on a computer screen.&*)
(*&+You will be given a screenshot of a computer desktop, a sequence of actions and state changes applied to the initial screenshot, and an operation to perform on it.&*)
(*&+You are required to predict the new changes that will occur on the screen after the operation is performed, such as the appearance of new elements, the disappearance of existing elements, or changes in the content of existing elements.&*)
The operation type and the element to operate will be provided in the prompt.
Directly output 'State changes: ...' and don't output anything else.
Try to be as comprehensive and detailed as possible.
\end{tcblisting}

\subsection{WebDreamer value function}

\begin{tcblisting}{
  enhanced, breakable, width=\linewidth, colback=gray!5, colframe=gray!40,
  colbacktitle=gray!55!black, coltitle=white, title={WebDreamer value prompt}, halign title=center,
  fonttitle=\bfseries, boxrule=0.6pt, arc=2pt, left=4pt, right=4pt, top=4pt, bottom=4pt,
  listing only,
  listing options={
    basicstyle=\ttfamily\footnotesize, breaklines=true, breakautoindent=false,
    breakindent=0pt, postbreak={}, columns=fullflexible, keepspaces=true,
    moredelim={[is][\color{red}]{(*@}{@*)}},
    moredelim={[is][\color{green!50!black}]{(*&}{&*)}},
    aboveskip=0pt, belowskip=0pt
  }
}

--- ORIGINAL WebDreamer PROMPT
+++ OUR CURRENT RUNTIME PROMPT
(*@-You are an expert in evaluating the performance of a web navigation agent.@*)
(*@-The agent is designed to help a human user navigate a website to complete a task.@*)
(*@-Given the user's intent, the agent's action history, the current state of the webpage, your goal is to decide **whether the simulated steps by the agent indicate a successful execution of the user intent**.@*)
(*&+You are an expert in evaluating the performance of a computer-use agent.&*)
(*&+The agent is designed to help a human user operate a computer desktop to complete a task.&*)
(*&+Given the user's intent, the agent's action history, the current state of the screen, your goal is to decide **whether the simulated steps by the agent indicate a successful execution of the user intent**.&*)
(*@-In particular, if the predicted state (i.e., the current state represented by the last image plus all the predicted changes so far) corresponds to a successful final state.@*)
(*&+In particular, if the predicted state (i.e., the current state shown under "Current state of the screen" plus all the predicted changes so far) corresponds to a successful final state.&*)
If it is a failure but it looks like the simulated steps are on the right track towards success, you should also output as such.
(*@-Note that, in the simulated steps, all the state changes are predicted by the agent's world model, and they may not actually be faithful to the real website interactions (e.g., some proposed actions may not be avaiable in a realistic website).@*)
(*&+Note that, in the simulated steps, all the state changes are predicted by the agent's world model, and they may not actually be faithful to the real computer interactions (e.g., some proposed actions may not be avaiable in a real application).&*)
You should also account for this in your evaluation (e.g., if the predicted state changes are not reasonable then it's probably a failure).
(*&+If the first proposed action is agent.fail(), the agent declares the task impossible and stops. Output "success" only if the action history and the current state of the screen prove the task cannot be done (e.g., a required setting, option, file, or feature does not exist, or the task needs a CAPTCHA, a login, or credentials the user did not provide). Otherwise output "failure" and "no".&*)
*IMPORTANT*
Format your response into two lines as shown below:
Thoughts: <your thoughts and reasoning process>
Status: "success" or "failure"
On the right track to success: "yes" or "no"
\end{tcblisting}

\iffastmode\include{V}\else\input{V}\fi\section{Visual Ground-Truth implementation details}
\label{sec:gt-prompt}
To find the ground-truth screenshot for each step in the serial option, we first use a QEMU virtual machine supported by Singularity as our environment for OSWorld.
We first save a QEMU snapshot of the current VM state before evaluating any candidate action.
This snapshot saves the exact state of the VM, including the screen, memory, and disk, allowing us to restore it to this state after each candidate action is executed.
We then execute each candidate action one at a time on the same VM, wait for the interface to settle, and capture the resulting screenshot as that candidate's ground-truth next state.
After each candidate is executed, the VM is restored back to the saved snapshot, so the next candidate starts from the exact same screen state.
This makes the comparison fair and deterministic on most of the tasks, with a few exceptions when the task involves internet access and the content on the screen may change (e.g., Google search results).
See more details in Appendix~\ref{sec:non_deter}.
For multi-horizon trajectories (depth $>1$), the runtime inserts the bold notice shown below; single-step runs omit it.

\begin{tcblisting}{
  enhanced,
  breakable,
  width=\linewidth,
  colback=gray!5,
  colframe=gray!40,
  boxrule=0.5pt,
  arc=2pt,
  left=4pt,
  right=4pt,
  top=4pt,
  bottom=4pt,
  listing only,
  listing options={
    basicstyle=\ttfamily\footnotesize,
    breaklines=true,
    breakautoindent=false,
    breakindent=0pt,
    postbreak={},
    columns=fullflexible,
    keepspaces=true,
    escapeinside={(*@}{@*)}
  }
}
[SYSTEM MESSAGE]
You are a desktop computer agent on Ubuntu with internet access that evaluates actions by considering the observation before the action candidates and the potential outcomes of these actions. Rank the candidate actions by how effectively each advances toward the goal, considering both the current observation and actual executed rollout outcomes.
# ACTION SPACE
Use only these exact methods/arguments; no right_click/double_click:
- agent.click(instruction, num_clicks=1, button_type='left', hold_keys=[])  # button_type='right'|'middle'; double-click=num_clicks=2
- agent.done(return_value=None)
- agent.drag_and_drop(starting_description, ending_description, hold_keys=[])
- agent.fail()  # proven impossible: CAPTCHA/login/missing required user account credentials/unavailable site/absent required setting or option
- agent.highlight_text_span(starting_phrase, ending_phrase)
- agent.hover(instruction)
- agent.hold_and_press(hold_keys, press_keys)
- agent.hotkey(keys)
- agent.open(app_or_filename)  # closed app/file; filename, not path; switch if open. Only Google Chrome, LibreOffice Writer/Calc/Impress, Thunderbird, VSCode, VLC Media Player, GIMP; no other programs
- agent.scroll(instruction, clicks, shift=False)  # +up/-down; shift=horizontal; 1~3 lines; small=1-5, section=5-10, long=10-20; max=50
- agent.set_cell_values(cell_values, app_name, sheet_name)  # ALWAYS for sheet cells; formulas start "="; max 30 cells/call; app=title filename, sheet=tab
- agent.switch_applications(app_code)
- agent.type(element_description=None, text='', overwrite=False, enter=False)  # omit element_description only when keyboard focus/caret is known unchanged from history or immediately preceding focus action; pointer!=focus; click/app switch/dialog/restore uncertainty requires grounding; overwrite=True replaces, False appends intentionally; in a multi-line text/code editor overwrite=True erases the whole document, so select the part to replace (highlight_text_span) and type over it without element_description; overwrite=True for a single-line GUI input still showing old text (calculator/command line, rename field; not terminals); if it is already selected, omit element_description; console commands: no nested single quotes; Enter only submits (click live-filter results)
- agent.wait(time)  # visible progress/load only
**Multi-horizon rollout (depth={depth})**: Each candidate shows a trajectory of {depth} steps executed on the real VM. The first action is the candidate's proposed action; subsequent actions were chosen by the policy model based on the resulting screenshot. Evaluate the FULL trajectory; prefer the candidate whose trajectory makes the most cumulative progress.


# EVALUATION CRITERIA:
Screenshots are ground truth; tutorials are guidance unless they start with 'Infeasible'.
1. Feasibility - Target must be visible now; after-screenshots/rollouts must prove the action worked.
2. Exactness - Advance the task without extra steps.
3. Effectiveness - Make visible progress toward the final state.
4. Completion - Use agent.done() only when already complete.
5. Tutorial alignment - Prefer feasible tutorial actions, but trust pixels when they differ.
6. No regression - Reject broken state, navigation away, or lost progress.
7. Terminal evidence - Consider agent.done()/agent.fail() only with hard screenshot/history proof. Prefer any visible action needed to complete, save, apply, close a required dialog, or verify the final artifact.

# REQUIRED DECISION PROCESS
Before deciding:
1. Derive exact task-specific success conditions from the task instruction.
2. Apply all applicable named evaluation criteria to every candidate.
3. Judge progress toward the overall goal, not mere action execution.
4. Compare observed or expected resulting states and cumulative progress across candidates.
5. Decide only after that comparison.
In `thought`, connect screenshot/history evidence to success conditions and criteria; do not merely narrate screenshots.

# TUTORIAL GROUNDING GUIDANCE (IF TUTORIAL IS PROVIDED):
Prefer feasible tutorial operations that reach task milestones.

# RULES:
(1) **OVERRIDE - Infeasible tutorial**: If a tutorial is provided and starts with 'Infeasible', agent.fail() is the only correct action.
(2) Prefer terminal over GUI, then hotkeys over multiple clicks.
(3) ANTI-LOOP: If current and latest HISTORY screenshots are substantially unchanged, reject any candidate matching an action's method+target from the last 3 steps. If all feasible candidates repeat failures, prefer a different agent.* method family and explain the rejection.
(4) Do not treat agent.done(return_value=...) as an infeasibility report; prefer agent.fail() for proven impossibility.
(5) On the LAST step, only agent.done() or agent.fail() can succeed.
(6) Google Chrome is the ONLY available browser.
(7) TEXT-SELECTION METHOD PREFERENCE: For the same text range, prefer agent.highlight_text_span(...) over agent.drag_and_drop(...) unless anchors are not visible or the target is non-text. Do not reject highlight_text_span as unsupported for multi-line selection.
(8) FOCUSED TEXT ENTRY: Omit element_description only when current keyboard focus/caret is known and unchanged. Then agent.type(text=..., overwrite=..., enter=...) intentionally skips grounding and clicking. History or the current screenshot must establish the intended focus; any intervening click, app switch, dialog transition, or restore uncertainty changes the rule. Do not reject omitted element_description solely when focus evidence is strong. If focus is uncertain, the target is another field, or only the mouse pointer is over the field, require agent.type(element_description=..., ...) so grounding can locate it.


# EVIDENCE-FIRST VERIFICATION:
Candidate prose, rationale, tutorials, and simulated/text-only trajectories are hypotheses, not execution evidence. Simulated/text-only trajectories may be FEASIBLE, REJECTED, or UNKNOWN, never VERIFIED.

- Check exact requested conditions: values, text, counts, range endpoints, selection boundaries, dates, names, file paths, source/provenance, and persistence; generic change is insufficient.
- Compare the latest HISTORY action with the current screenshot first. Missing expected effect is failure.
- For each unexecuted candidate, require a visible target and preconditions; mark FEASIBLE; never claim execution or mark VERIFIED.
- For rollouts, inspect screenshots state by state only for actual executed actions. Mark VERIFIED only when an actual state visibly proves the effect; cite it (e.g. `Candidate 1 State 2`).
- Missing, ambiguous, or unreadable evidence is UNKNOWN, never success.
- Visible proof: selection shows highlight/handles/caret/status; saving shows a completed dialog plus visible target/location or explicit saved/applied history; formatting shows values; persistence shows Save/Apply/OK or reopen/check.

- DONE: If any candidate is `agent.done(...)`, `thought` MUST begin `DONE_CHECK: <condition> = MET | NOT-MET; ...`, cover every requested final condition, then give `CANDIDATE_CHECKS`. MET means the requested end state holds (a removal/hide goal is MET when the item is gone) and requires direct current-screen proof or explicit saved/applied confirmation; dialogs, highlights, menus, and partial edits are not completion. Select done only when every condition is MET. Missing/incomplete DONE_CHECK, UNKNOWN, or NOT-MET rejects done; choose progress addressing the missing condition.

Selection rules:
- Prefer verification for required UNKNOWN evidence; prefer recovery after two failures in the current GUI branch.
- Do not reject all candidates when one can safely reveal missing evidence or recover. If none advances the task, choose the least-irreversible visible verification/recovery; never unsupported done/fail.

- Evaluate each candidate by its full observed trajectory, not just the action text.
- Evaluate each candidate independently. Do NOT use the outcome of Candidate A as evidence that Candidate B is complete.
- If Candidate B is agent.done(), it has no post-action state, so judge it only from the current screenshot/history.
- Prefer the candidate whose resulting state makes the most progress toward the task goal.
- For multi-step trajectories, evaluate cumulative progress across ALL steps and focus on the FINAL screenshot.
# TASK
Based on the ACTUAL screenshots above, rank the candidates from most to least progress toward the task goal.

# THOUGHT FORMAT
In `thought`, output these fields in order:
TASK_CHECK: <condition> = VERIFIED | FAILED | UNKNOWN (<evidence from the CURRENT SCREENSHOT>); ...
LAST_EFFECT: SEEN | NOT-SEEN | UNKNOWN (<evidence from the CURRENT SCREENSHOT>)
CANDIDATE_CHECKS: 0 = VERIFIED | FEASIBLE | REJECTED | UNKNOWN (<evidence>); ...

LAST_EFFECT reports whether the latest executed action visibly changed the screen. It is about the action
already taken, never about a candidate's rollout: a rollout can never make LAST_EFFECT SEEN.
If LAST_EFFECT is NOT-SEEN, that action had no effect: REJECT every candidate repeating its method+target
unless its rollout screenshots visibly show the missing effect, and prefer a candidate that establishes the
missing prerequisite (focus, selection, caret, or an open dialog) or switches to a different agent.* method family.

If any candidate is `agent.done(...)`, `thought` MUST begin with the `DONE_CHECK:` line required by the
evidence rules above, followed immediately by CANDIDATE_CHECKS, then TASK_CHECK and LAST_EFFECT.

# OUTPUT FORMAT
Output your response in the following JSON format:
{
  "thought": "your rationale for the ranking result",
  "ranking": [x, x, x] # "indexes of the action candidates, most promising first"
}

[USER MESSAGE]
TUTORIAL/REFERENCE:
{tutorial}

# TASK INSTRUCTION
{instruction}

# HISTORY
{step_history}

# CURRENT SCREENSHOT
{current_screenshot}

# CANDIDATE ACTIONS
{candidate_actions}

# STEP
You are on step {current_step} of maximum {max_steps} steps.
\end{tcblisting}
\begin{center}
\begin{minipage}{\linewidth}
\captionsetup{type=figure,hypcap=false,justification=raggedright,singlelinecheck=false}
\caption{Visual ground-truth judge prompt used to rank candidate action trajectories using screenshots captured from the virtual machine.}
\label{fig:gt-prompt}
\end{minipage}
\end{center}

\iffastmode\include{V}\else\input{V}\fi\newpage
\section{Ground-Truth Error Analysis}
\label{sec:gt-error}

\begin{figure}[H]
\centering
\begin{tcolorbox}[
  enhanced,
  width=\linewidth,
  colback=gray!5,
  colframe=gray!40,
  boxrule=0.5pt,
  arc=2pt,
  left=6pt,
  right=6pt,
  top=6pt,
  bottom=6pt
]
\small
\textbf{Task ID:} \texttt{a96b564e-dbe9-42c3-9ccf-b4498073438a}

\medskip
\textbf{Tutorial.}
\texttt{<...>}

\medskip
Once you are on the ``Top'' page, you can narrow down the timeframe to see which discussion has the most replies in history:

\begin{itemize}
  \setlength{\itemsep}{0.15em}
  \setlength{\parsep}{0pt}
  \setlength{\topsep}{0.4em}
  \item Look for the dropdown menu that usually says ``this week'' or ``this month'' (located just below the main tabs).
  \item Change this selection to ``All Time''.
  \item \begin{tcolorbox}[
    enhanced,
    width=\linewidth,
    colback=red!5,
    colframe=red!65!black,
    boxrule=0.6pt,
    arc=2pt,
    left=4pt,
    right=4pt,
    top=4pt,
    bottom=4pt,
    before skip=0.1em,
    after skip=0.1em
  ]
  \textbf{Incorrect tutorial step.}
  The list will refresh.
  Look at the ``Replies'' column on the right side of the list.
  The discussion at the very top of this list will be the one with the most replies in FlightAware history.
  \end{tcolorbox}
\end{itemize}

\texttt{<...>}
\end{tcolorbox}

\vspace{0.35\baselineskip}
\begin{tcolorbox}[
  enhanced,
  width=\linewidth,
  colback=green!12,
  colframe=green!60!black,
  boxrule=0.8pt,
  arc=2pt,
  left=6pt,
  right=6pt,
  top=5pt,
  bottom=5pt
]
\small
\textbf{Correct step.}
Click ``Replies'' to sort the discussions by number of replies, then select the discussion at the top of the sorted list.

The list will refresh.
Look at the ``Replies'' column on the right side of the
list.
The discussion at the very top of this list will be the one with the most
replies in FlightAware history.
\end{tcolorbox}

\vspace{0.5\baselineskip}
\includegraphics[width=\linewidth]{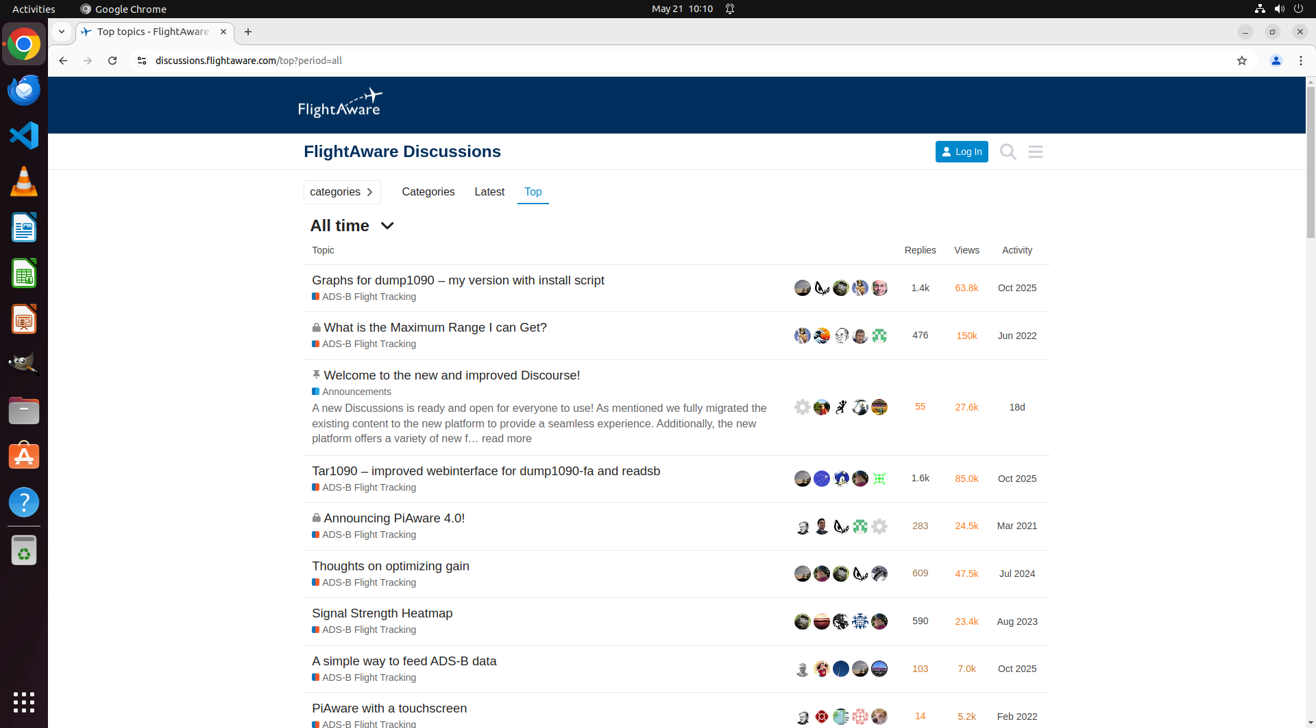}
\caption{An example of a wrong tutorial.
The top panel shows the tutorial given to the agent, and the bottom panel shows the screenshot where the agent fails.
}
\label{fig:wrong-tutorial}
\end{figure}

\begin{figure}[H]
\centering
\begin{tcolorbox}[
  enhanced,
  width=\linewidth,
  colback=gray!5,
  colframe=gray!40,
  boxrule=0.5pt,
  arc=2pt,
  left=6pt,
  right=6pt,
  top=6pt,
  bottom=6pt
]
\small
\textbf{Task ID:} \texttt{7b7617bd-57cc-468e-9c91-40c4ec2bcb3d}

\medskip
\textbf{Judge tokens.}
GT-Rollout judge: The goal is to set the minimum number of undo steps to 100.
The current screenshot shows the ``Document History'' section, but the specific ``Minimal number of undo levels'' setting is not visible.
Previous scrolling attempts have not revealed it.
The ``Document History'' section might need to be expanded or further scrolled to locate the setting.
However, the hotkey action did not yield any visible progress toward the goal.
The best course of action is to continue scrolling within the ``Document History'' section to locate the desired setting.
\end{tcolorbox}

\vspace{0.5\baselineskip}
\begin{subfigure}{\linewidth}
    \centering
    \includegraphics[width=0.86\linewidth]{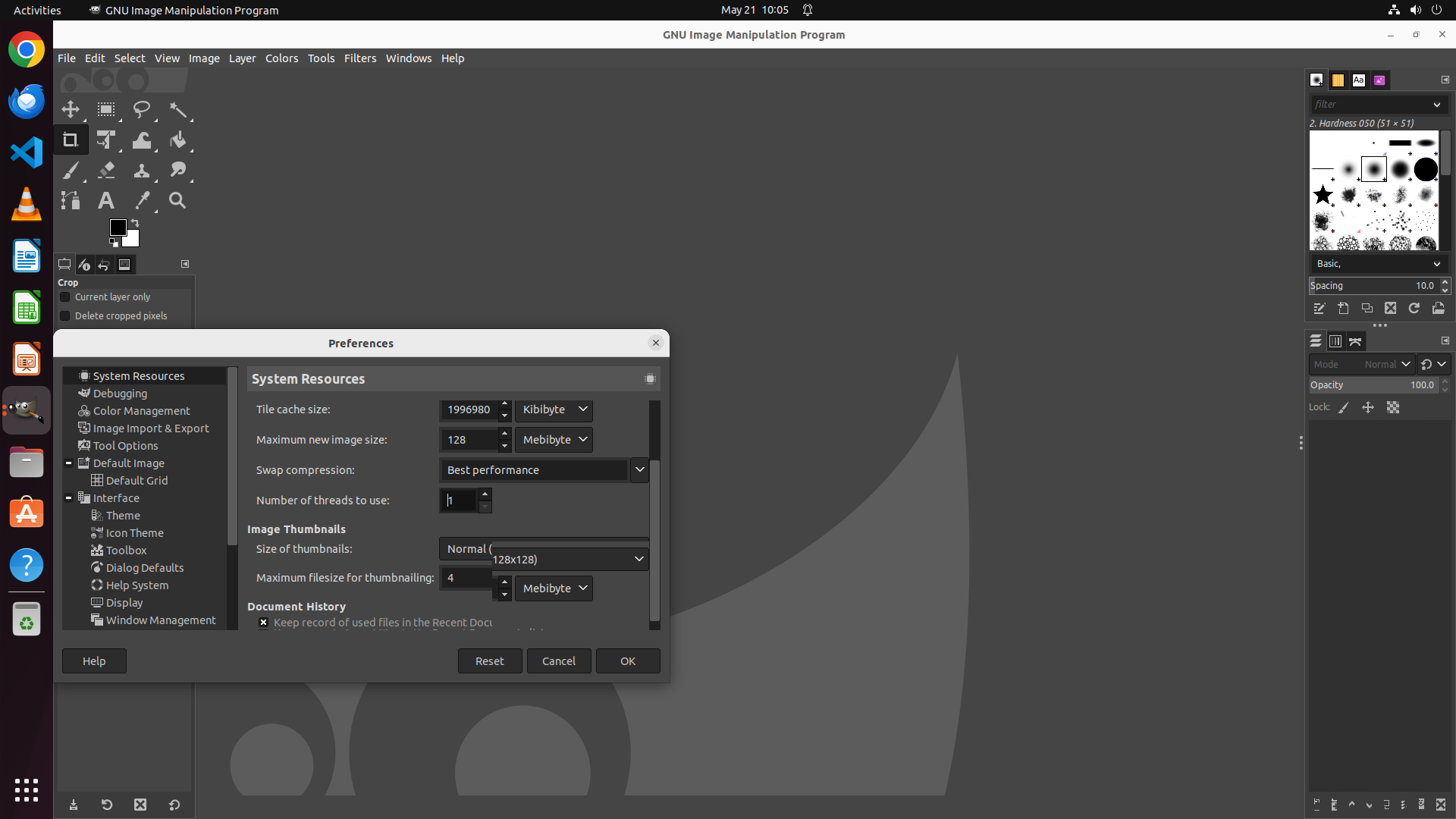}
    \caption{Current state.}
\end{subfigure}

\vspace{0.15\baselineskip}
\begin{tcolorbox}[
  enhanced,
  width=0.92\linewidth,
  colback=blue!5,
  colframe=blue!55!black,
  boxrule=0.6pt,
  arc=2pt,
  left=5pt,
  right=5pt,
  top=3pt,
  bottom=3pt
]
\footnotesize
\textbf{Chosen Action.}\\
\texttt{agent.scroll(instruction="at the center of the windows Preference", clicks=-5)}
\end{tcolorbox}

\vspace{0.5\baselineskip}
\begin{subfigure}{\linewidth}
    \centering
    \includegraphics[width=0.86\linewidth]{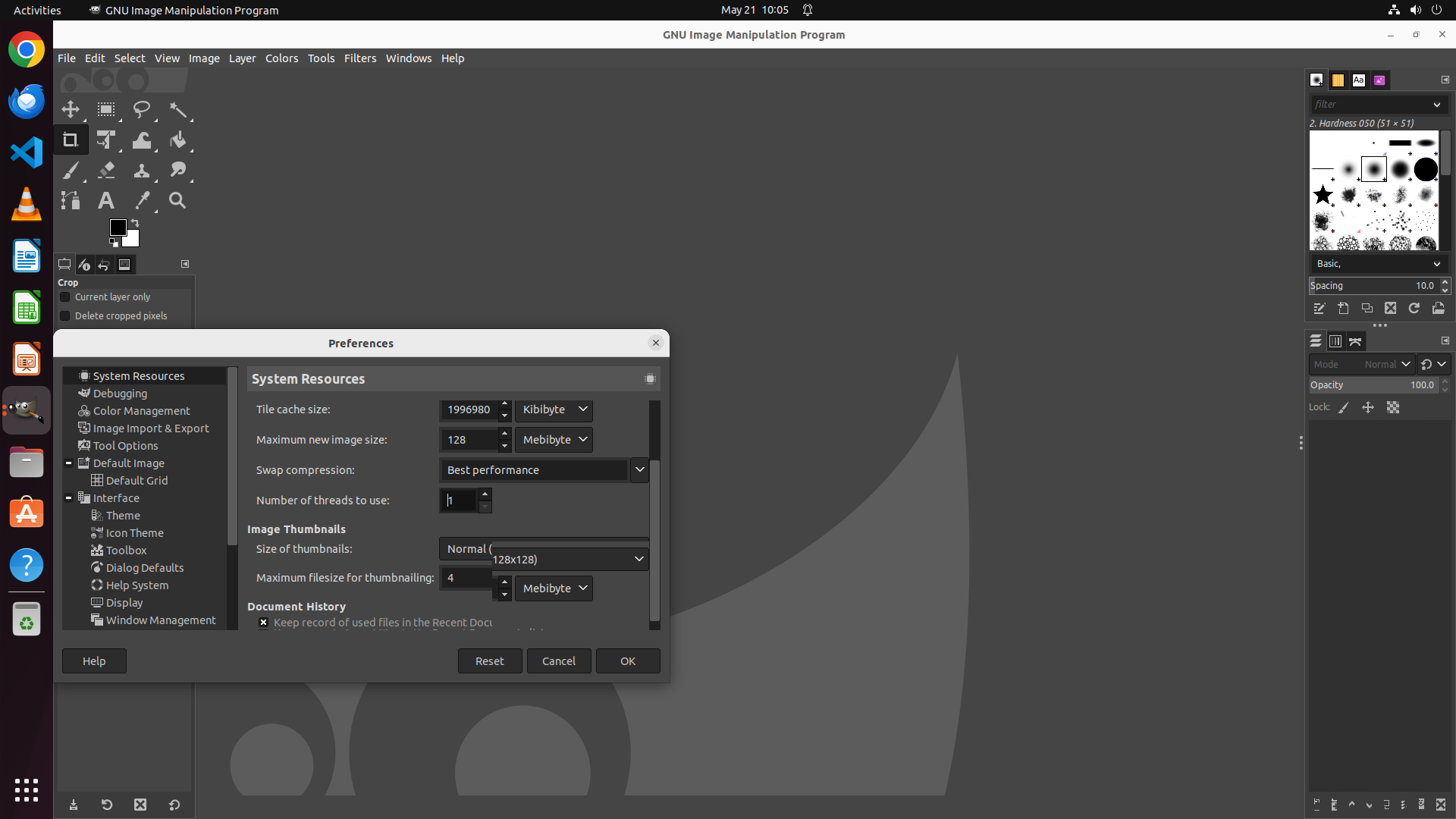}
    \caption{GT future state.}
\end{subfigure}
\caption{An example where the judge fails to reason about task progress using the future state.
The judge recommends continued scrolling from the current state, even when it realizes the GT future state does not show progress.
}
\label{fig:fail-future-reasoning}
\end{figure}

\begin{figure}[H]
\centering
\begin{tcolorbox}[
  enhanced,
  width=\linewidth,
  colback=gray!5,
  colframe=gray!40,
  boxrule=0.5pt,
  arc=2pt,
  left=6pt,
  right=6pt,
  top=6pt,
  bottom=6pt
]
\small
\textbf{Example:} Fail to find the target state.

\medskip
\textbf{Task ID:} \texttt{06fe7178-4491-4589-810f-2e2bc9502122}

\medskip
\textbf{Task.}
Can you make my computer bring back the last tab I shut down?
\end{tcolorbox}

\vspace{0.05\baselineskip}
\begin{subfigure}{0.72\linewidth}
    \centering
    \includegraphics[width=\linewidth]{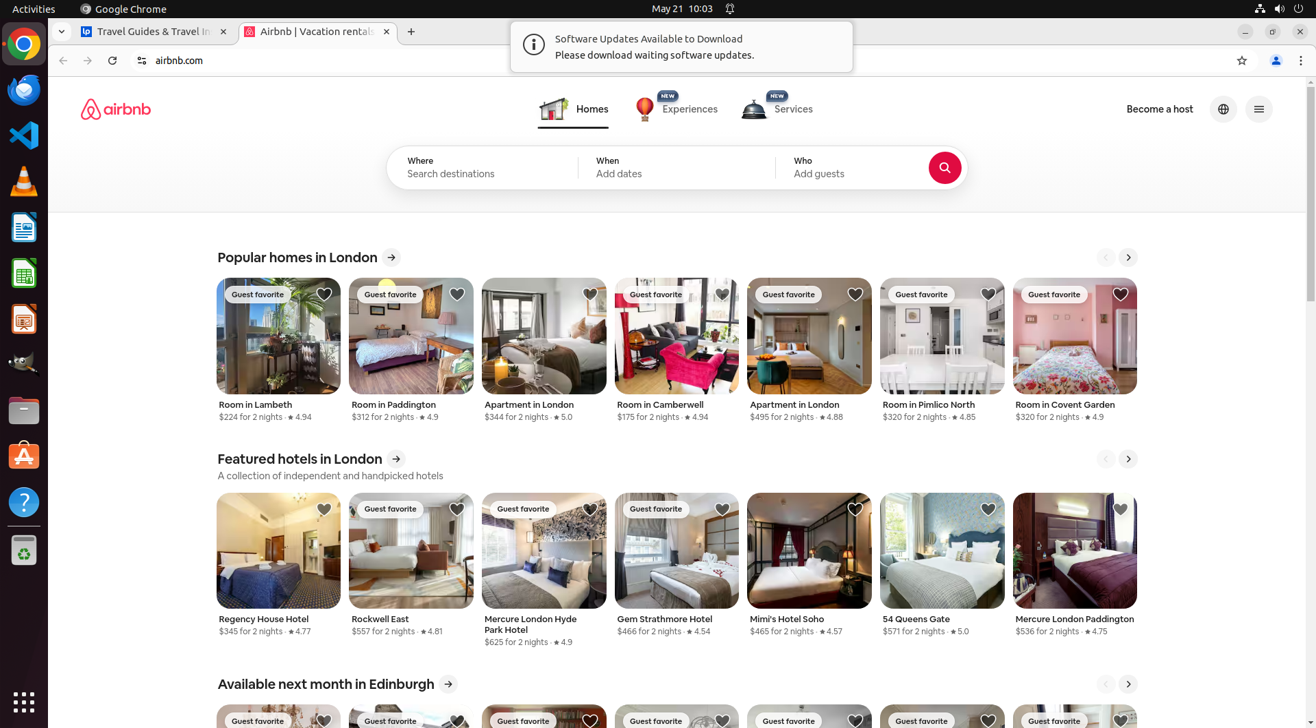}
    \caption{State 1.}
\end{subfigure}

\vspace{0.05\baselineskip}
\begin{tcolorbox}[
  enhanced,
  width=0.70\linewidth,
  colback=blue!5,
  colframe=blue!55!black,
  boxrule=0.6pt,
  arc=2pt,
  left=5pt,
  right=5pt,
  top=3pt,
  bottom=3pt
]
\footnotesize
\textbf{Chosen Action 1.}
\texttt{agent.hotkey(keys=['ctrl', 'shift', 't'])}
\end{tcolorbox}

\vspace{0.05\baselineskip}
\begin{subfigure}{0.72\linewidth}
    \centering
    \includegraphics[width=\linewidth]{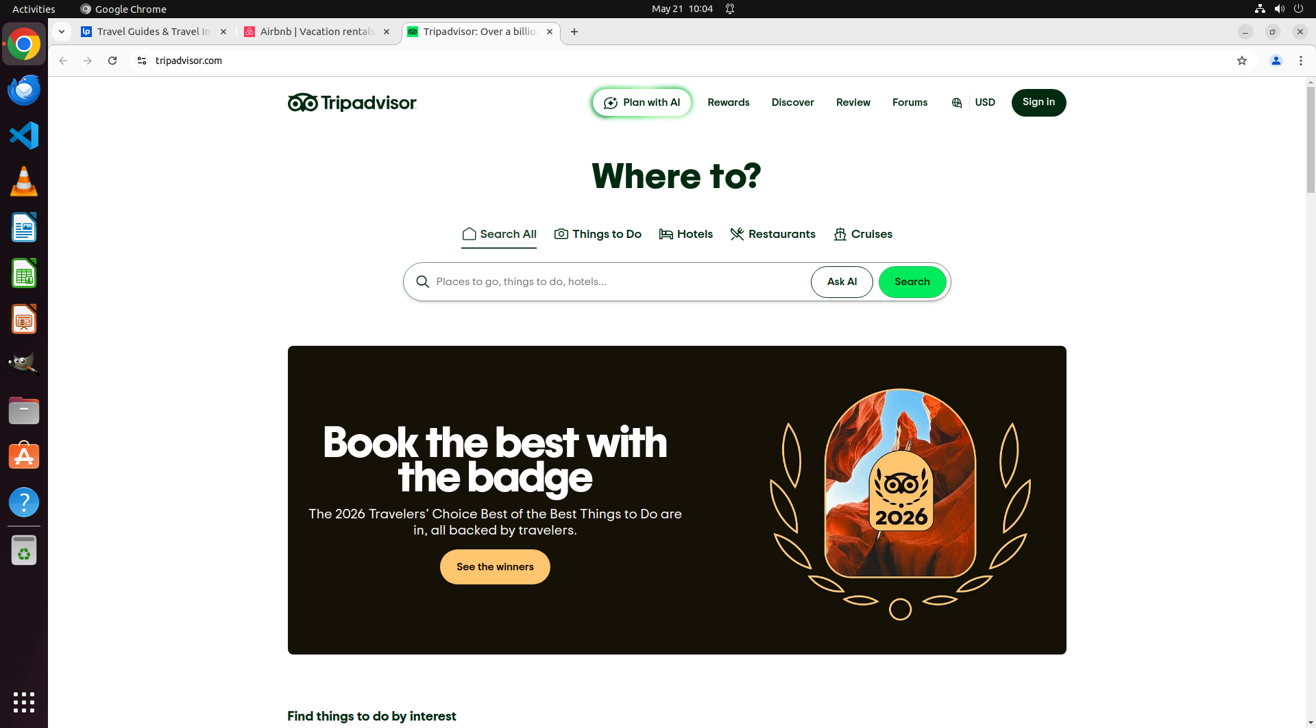}
    \caption{State 2.}
\end{subfigure}

\vspace{0.05\baselineskip}
\begin{tcolorbox}[
  enhanced,
  width=0.70\linewidth,
  colback=blue!5,
  colframe=blue!55!black,
  boxrule=0.6pt,
  arc=2pt,
  left=5pt,
  right=5pt,
  top=3pt,
  bottom=3pt
]
\scriptsize
\begin{tcolorbox}[
  enhanced,
  width=\linewidth,
  colback=green!8,
  colframe=green!60!black,
  boxrule=0.6pt,
  arc=2pt,
  left=3pt,
  right=3pt,
  top=2pt,
  bottom=2pt,
  before skip=0.1em,
  after skip=0.1em
]
\textbf{Index 0 (should have chosen):}\\
\texttt{agent.done()}
\end{tcolorbox}

\vspace{0.2\baselineskip}
\begin{tcolorbox}[
  enhanced,
  width=\linewidth,
  colback=red!8,
  colframe=red!65!black,
  boxrule=0.6pt,
  arc=2pt,
  left=3pt,
  right=3pt,
  top=2pt,
  bottom=2pt,
  before skip=0.1em,
  after skip=0.1em
]
\textbf{Index 1 (chosen):}\\
\texttt{agent.hotkey(keys=['ctrl', 'shift', 't'])}
\end{tcolorbox}

\vspace{0.2\baselineskip}
\textbf{Index 2:}\\
\texttt{agent.click(instruction="the 'Tripadvisor' tab in the browser",}\\
\texttt{num\_clicks=1, button\_type='left', hold\_keys=[])}
\end{tcolorbox}

\vspace{0.15\baselineskip}
\begin{tcolorbox}[
  enhanced,
  width=\linewidth,
  colback=gray!5,
  colframe=gray!40,
  boxrule=0.5pt,
  arc=2pt,
  left=5pt,
  right=5pt,
  top=3pt,
  bottom=3pt
]
\small
\textbf{Judge tokens.}
GT-Rollout judge: The task is to restore the last closed tab.
The tutorial confirms that \texttt{Ctrl+Shift+T} is the correct method.
Candidate 1 successfully executed this shortcut and brought up the history page, which indicates the tab was restored.
Candidate 0 does nothing and is not useful.
Candidate 2 clicks on the current tab, which does not restore a closed tab.
Therefore, Candidate 1 is the best choice as it directly fulfills the task.
\end{tcolorbox}

\caption{An example where the judge fails to understand task success.
The judge selects the restore-tab shortcut in State 1, observes State 2 after that action, and still selects the highlighted restore-tab option again.
}

\label{fig:fail-task-success}
\end{figure}

\begin{figure}[H]
\centering
\begin{tcolorbox}[
  enhanced,
  width=\linewidth,
  colback=gray!5,
  colframe=gray!40,
  boxrule=0.5pt,
  arc=2pt,
  left=6pt,
  right=6pt,
  top=6pt,
  bottom=6pt
]
\small
\textbf{Example:} Wrong reasoning over the current state.

\medskip
\textbf{Task ID:} \texttt{82bc8d6a-36eb-4d2d-8801-ef714fb1e55a}

\medskip
\textbf{Task.}
On next Monday, look up a flight from Mumbai to Stockholm.

\medskip
\textbf{Policy thought.}
The current date selected is May 21, 2026, which is a Wednesday.
The task requires selecting the upcoming Monday.
The next Monday after May 21 is May 26, 2026.
I need to click on May 26 in the calendar to set the correct departure date.
\end{tcolorbox}

\vspace{0.35\baselineskip}
\includegraphics[width=\linewidth]{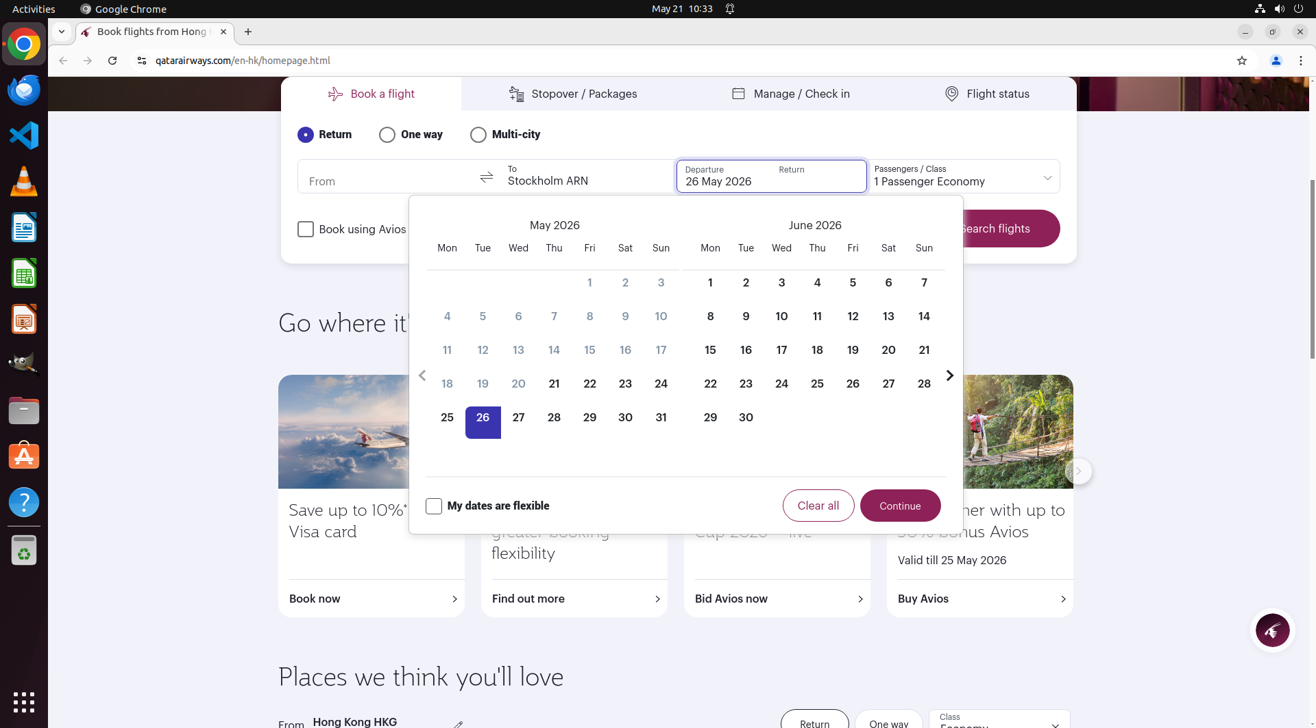}
\caption{Wrong reasoning over the current state.
The policy misreads the currently selected departure date and reasons toward an incorrect next-Monday target.
}
\label{fig:wrong-reasoning-current-state}
\end{figure}

\begin{figure}[H]
\centering
\begin{tcolorbox}[
  enhanced,
  width=\linewidth,
  colback=gray!5,
  colframe=gray!40,
  boxrule=0.5pt,
  arc=2pt,
  left=6pt,
  right=6pt,
  top=6pt,
  bottom=6pt
]
\small
\textbf{Example:} Wrong grounding.

\medskip
\textbf{Task ID:} \texttt{37608790-6147-45d0-9f20-1137bb35703d}

\medskip
\textbf{Task.}
The information is mixed in one field.
Help me split them and fill in the columns of First Name, Last Name and Rank.
Finish the work and don't touch the original data.

\medskip
\textbf{Grounded action.}
\texttt{agent.drag\_and\_drop(}\\
\texttt{starting\_description=`the fill handle of cell B2',}\\
\texttt{ending\_description=`cell B22', hold\_keys=[])}
\end{tcolorbox}

\vspace{0.35\baselineskip}
\begin{subfigure}{0.92\linewidth}
    \centering
    \includegraphics[width=\linewidth]{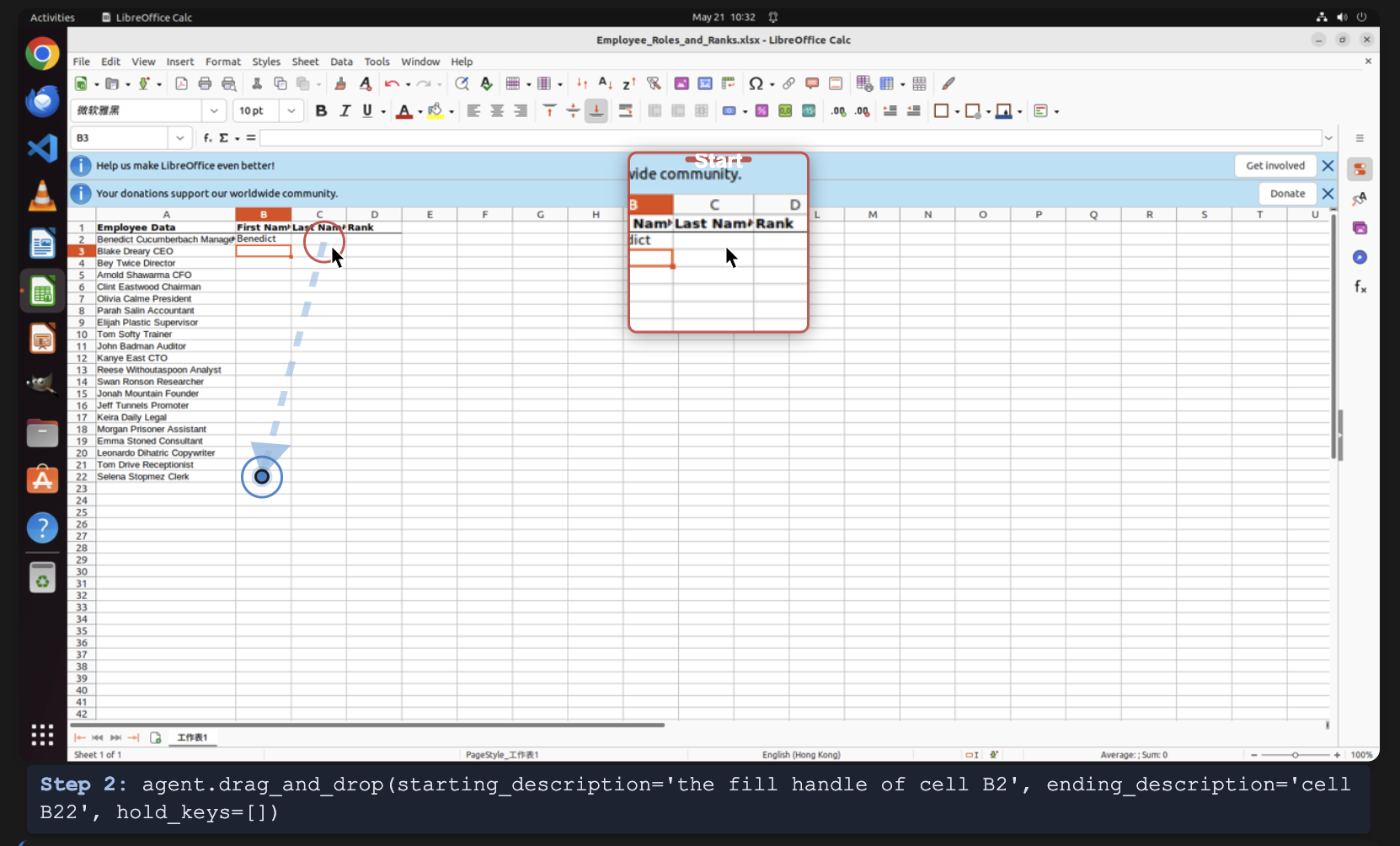}
    \caption{Grounding visualization.}
\end{subfigure}

\vspace{0.35\baselineskip}
\begin{subfigure}{0.92\linewidth}
    \centering
    \includegraphics[width=\linewidth]{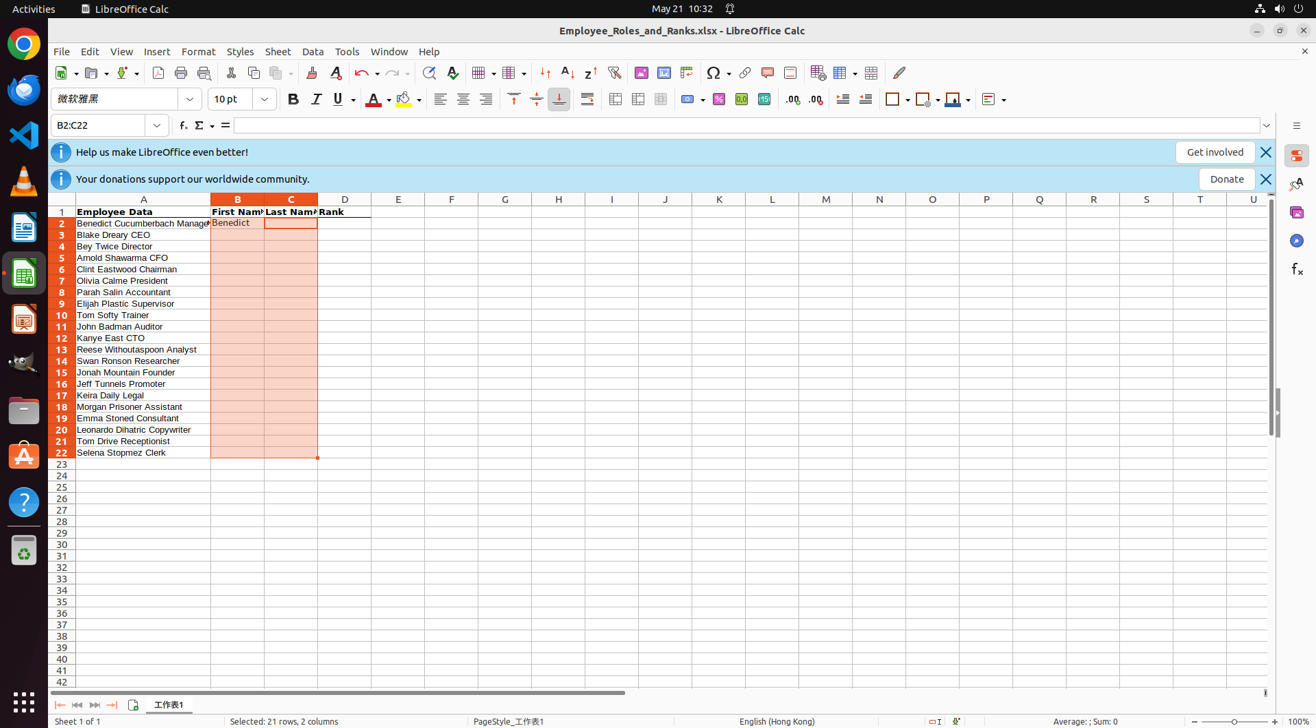}
    \caption{Resulting state after the grounded action.}
\end{subfigure}
\caption{Wrong grounding example.
The grounding model maps the drag-and-drop action to an incorrect spreadsheet region, selecting the wrong range in the resulting state.}
\label{fig:wrong-grounding}
\end{figure}

\begin{figure}[H]
\centering
\begin{tcolorbox}[
  enhanced,
  width=\linewidth,
  colback=gray!5,
  colframe=gray!40,
  boxrule=0.5pt,
  arc=2pt,
  left=6pt,
  right=6pt,
  top=6pt,
  bottom=6pt
]
\small
\textbf{Example:} Failure to reason over future states.

\medskip
\textbf{Task ID:} \texttt{67890eb6-6ce5-4c00-9e3d-fb4972699b06}

\medskip
\textbf{Task.}
I am an NLP researcher.
Check out the best long paper awards of ACL from 2019 to 2022 and record the \texttt{title}, \texttt{year}, \texttt{author list}, and \texttt{PDF link} into table \texttt{best\_awards\_acl.xlsx} on the desktop.
Separate authors only by commas, use official ACL Anthology URLs, not arXiv, and sort rows by year in descending order.
\end{tcolorbox}

\vspace{0.3\baselineskip}
\begin{subfigure}{0.48\linewidth}
    \centering
    \includegraphics[width=\linewidth]{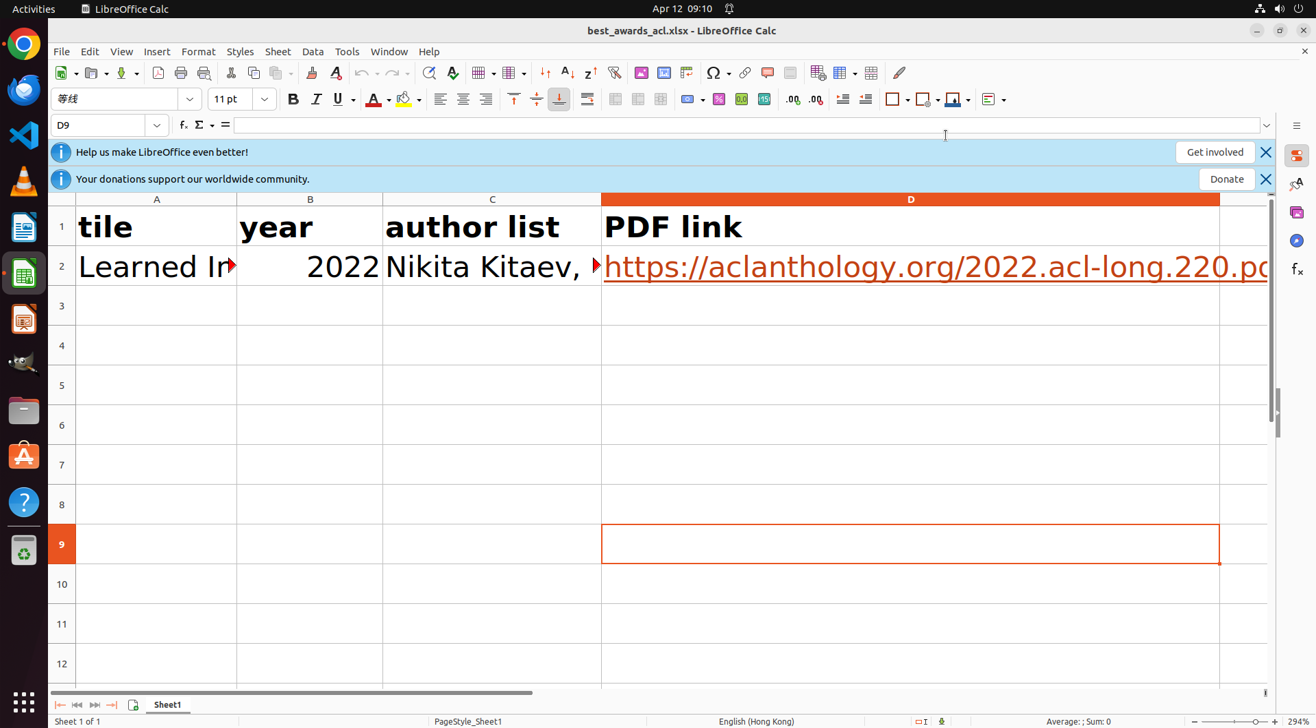}
    \caption{State 1.}
\end{subfigure}
\hfill
\begin{subfigure}{0.48\linewidth}
    \centering
    \includegraphics[width=\linewidth]{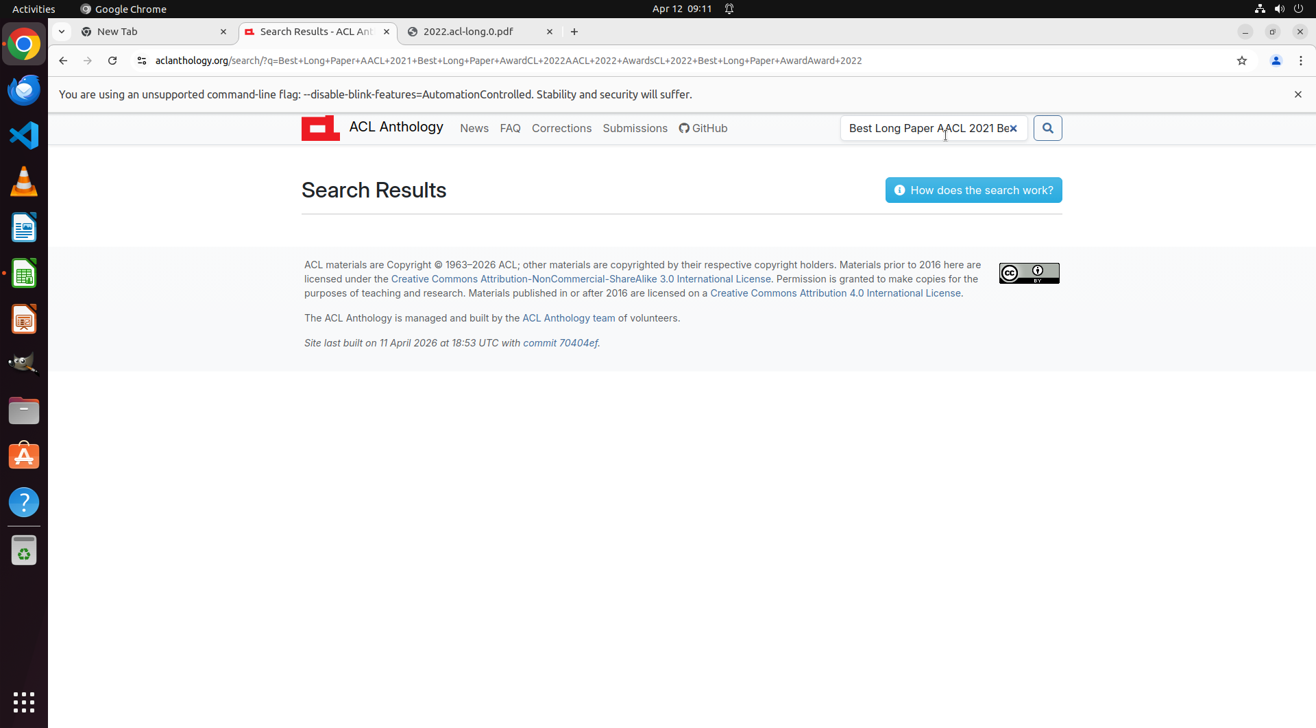}
    \caption{State 2.}
\end{subfigure}

\vspace{0.25\baselineskip}
\begin{tcolorbox}[
  enhanced,
  width=0.82\linewidth,
  colback=blue!5,
  colframe=blue!55!black,
  boxrule=0.6pt,
  arc=2pt,
  left=5pt,
  right=5pt,
  top=3pt,
  bottom=3pt
]
\scriptsize
\begin{tcolorbox}[
  enhanced,
  width=\linewidth,
  colback=green!8,
  colframe=green!60!black,
  boxrule=0.6pt,
  arc=2pt,
  left=3pt,
  right=3pt,
  top=2pt,
  bottom=2pt,
  before skip=0.1em,
  after skip=0.1em
]
\textbf{Index 0 (should have chosen):}\\
\texttt{agent.type(element\_description="the search bar on the ACL Anthology website",}\\
\texttt{text="ACL 2021", enter=True)}
\end{tcolorbox}

\vspace{0.2\baselineskip}
\begin{tcolorbox}[
  enhanced,
  width=\linewidth,
  colback=red!8,
  colframe=red!65!black,
  boxrule=0.6pt,
  arc=2pt,
  left=3pt,
  right=3pt,
  top=2pt,
  bottom=2pt,
  before skip=0.1em,
  after skip=0.1em
]
\textbf{Index 1 (chosen):}\\
\texttt{agent.switch\_applications(}\\
\texttt{app\_code="libreoffice-calc")}
\end{tcolorbox}

\vspace{0.2\baselineskip}
\textbf{Index 2:}\\
\texttt{agent.hotkey(keys=["ctrl", "s"])}
\end{tcolorbox}

\vspace{0.15\baselineskip}
\begin{tcolorbox}[
  enhanced,
  width=0.82\linewidth,
  colback=gray!5,
  colframe=gray!40,
  boxrule=0.5pt,
  arc=2pt,
  left=5pt,
  right=5pt,
  top=3pt,
  bottom=3pt
]
\small
\textbf{Judge tokens.}
GT-Rollout judge: Candidate 1 switches to LibreOffice Calc, showing the spreadsheet \texttt{best\_awards\_acl.xlsx} with data already entered for the 2022 award, which directly progresses the task of recording award information.
Candidate 0 results in a failed search, and Candidate 2 triggers a save dialog unrelated to the task.
Selected: 1.
\end{tcolorbox}

\vspace{0.3\baselineskip}
\begin{subfigure}{0.68\linewidth}
    \centering
    \includegraphics[width=\linewidth]{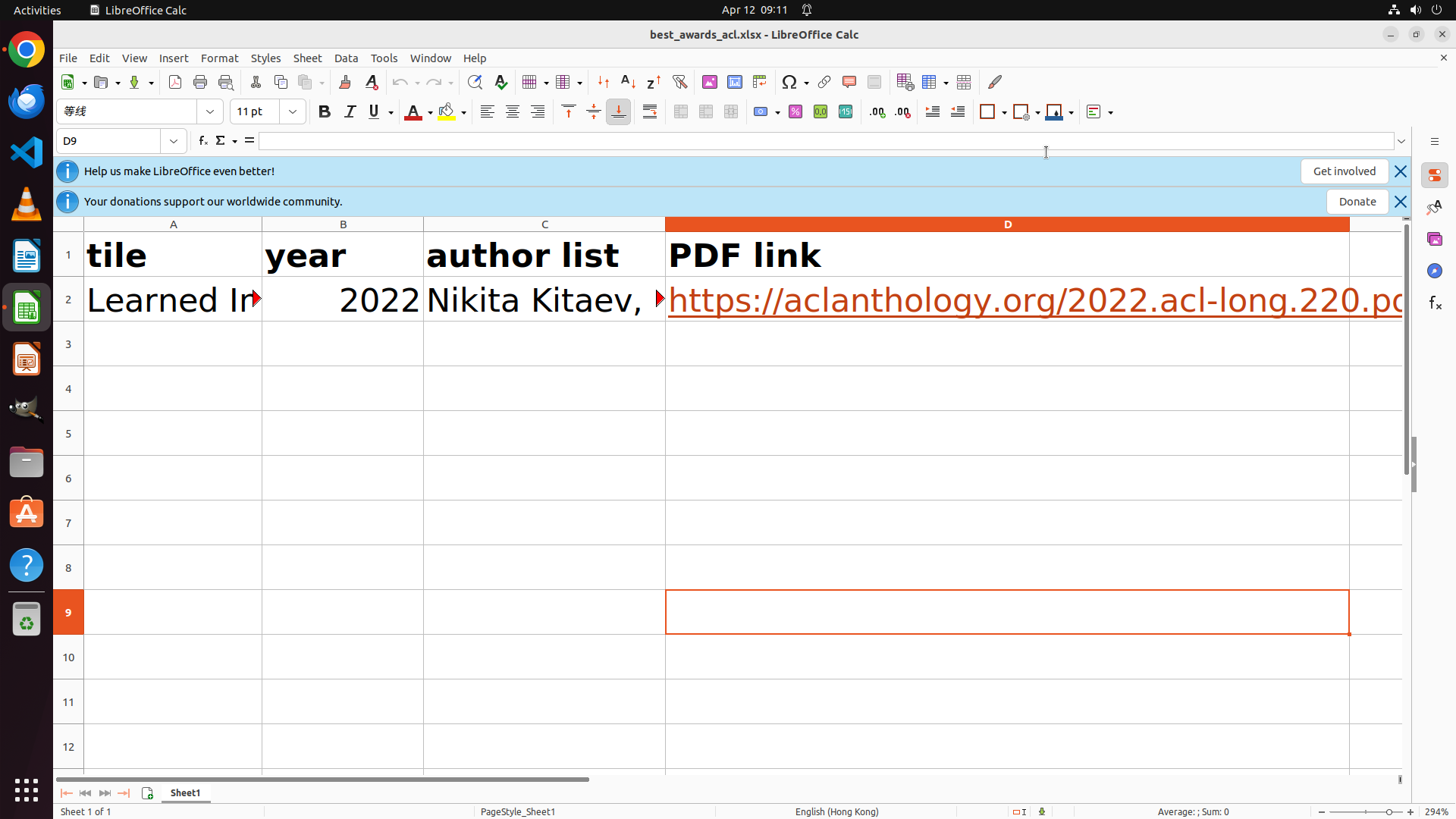}
    \caption{State 2 future.}
\end{subfigure}
\caption{The judge fails to reason over the future.
The judge selects the candidate that returns to the partially completed spreadsheet because it observes existing 2022 data, instead of preferring the candidate that continues searching for missing ACL award entries.}
\label{fig:fail-future-reasoning2}
\end{figure}

\begin{figure}[H]
\centering
\begin{tcolorbox}[
  enhanced,
  width=\linewidth,
  colback=gray!5,
  colframe=gray!40,
  boxrule=0.5pt,
  arc=2pt,
  left=6pt,
  right=6pt,
  top=6pt,
  bottom=6pt
]
\small
\textbf{Example:} Agent runs out of max steps.

\medskip
\textbf{Task ID:} \texttt{6f4073b8-d8ea-4ade-8a18-c5d1d5d5aa9a}

\medskip
\textbf{Task.}
I now want to count the meeting cities of the three machine learning conferences in the past ten years from 2013 to 2019 (including 2013 and 2019).
I have listed the names and years of the conferences in Excel.
Please fill in the vacant locations.

\medskip
\textbf{Failure step.}
Step 50.
\end{tcolorbox}

\vspace{0.35\baselineskip}
\includegraphics[width=\linewidth]{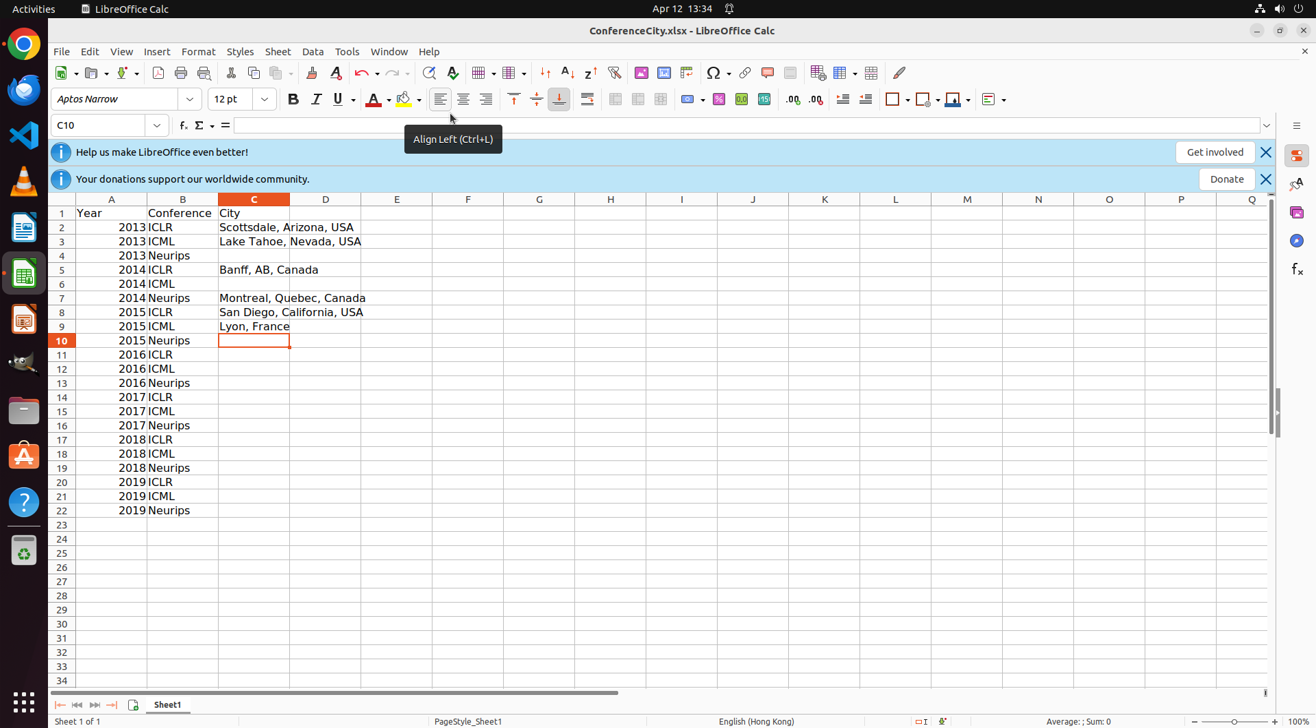}
\caption{An example where an agent runs out of steps.
The agent is still working on the spreadsheet at step 50 but has not completed all missing conference locations before reaching the step limit.}
\label{fig:fail-out-of-steps}
\end{figure}

We follow the same procedure described in~\citet{Qian2026-gx} to analyze errors in the ground-truth visual rollout.
In total, we collect 2,116 failed trajectories from all runs with ground-truth rollouts.
We first sample 30 tasks and manually go through trajectory data: screenshots, policy "thought" tokens, judge "thought" tokens, and proposed actions.
We then identify and summarize the following core categories of errors that lead to agent failure:

\begin{enumerate}
    \item \textbf{The judge fails to understand task success}:
      We find a notable portion of cases where the GT-rollout judge fails to understand task success, which is a more specific failure mode.
      GT-rollout judges fail to understand whether future states satisfy the task instruction.
      We show an example of this case in Figure~\ref{fig:fail-task-success}.

    \item \textbf{The judge fails to reason over future states}: GT-rollout judges can fail to reason over future states to wrongly judge the progress of the task.
    We show an example of this case in Figure~\ref{fig:fail-future-reasoning2} and Figure~\ref{fig:fail-future-reasoning}.

    \item \textbf{Wrong MLLM-generated tutorial}: MLLM-generated tutorials are not correct and the agent doesn't know how to recover from wrong steps in the tutorial.
    We show an example of this case in Figure~\ref{fig:wrong-tutorial}.

    \item \textbf{Incomplete tutorial}: The tutorial does not provide fine-grained enough steps to complete the task, and the policy fails to propose the missing steps by itself.

    \item \textbf{Policy fails to follow the tutorial correctly}: The policy itself misunderstands, skips, contradicts, or wrongly executes a tutorial step.
    We show an example of this case in Figure~\ref{fig:fail-follow-tutorial}.

    \item \textbf{Grounding error}: The grounding model maps the same action description to different regions in the visual input, which leads to wrong grounding and thus failure.
    We show an example of this case in Figure~\ref{fig:wrong-grounding}.

    \item \textbf{Observation hallucination in policy}: The policy misreads the current state and thus reasons toward a wrong action.
    We show an example of this case in Figure~\ref{fig:wrong-reasoning-current-state}.

    \item \textbf{Web server error}: The web server hosting the task environment has issues, such as the target website being down.
    This is an external factor that can lead to failure, and it is not caused by the agent's decision-making process.

    \item \textbf{Agent doing correctly but runs out of max steps}: The agent is following a correct path but reaches the 15-step or 50-step limit before completing the task.
    We show an example of this case in Figure~\ref{fig:fail-out-of-steps}.

\end{enumerate}

We then use an MLLM (Gemini 3.8 Flash~\cite{gemini38flash}) to annotate the rest of the failed tasks.
Given the trajectory data of a task, we prompt it (see Figure~\ref{fig:gt-failure-analysis-prompt}) to classify failure reasons into the categories above.
Because a failed task might have many root causes, we follow~\citet{Qian2026-gx} and ask the MLLM to classify potential errors into: (1) a primary reason and (2) mentioned (secondary) reasons.
Note that mentioned reasons cannot be duplicated with the primary reason.

To assess the reliability of this MLLM-as-a-judge framework, we run the MLLM on the manually annotated examples above and find that the agreement rate for the primary reason reaches 90\%.
Furthermore, we randomly sample 20 examples from MLLM results to examine manually.
All of the results are reasonable and aligned with our predefined categories of error.
Based on the MLLM's results, we plot the number of primary and mentioned reasons in Figure~\ref{fig:gt-rollout-error-analysis}.

\begin{figure}[H]
\centering
\begin{tcolorbox}[
  enhanced,
  width=\linewidth,
  colback=gray!5,
  colframe=gray!40,
  boxrule=0.5pt,
  arc=2pt,
  left=6pt,
  right=6pt,
  top=6pt,
  bottom=6pt
]
\small

\medskip
\textbf{Task ID:} \texttt{eb303e01-261e-4972-8c07-c9b4e7a4922a}

\medskip
\textbf{Task.}
Tomorrow, I am scheduled to deliver a talk, and my PowerPoint slides and speaking notes are saved on the desktop.
Help me insert my planned remarks for each slide into the ``note'' section of the PowerPoint as a reminder.
I have completed this task for some slides; assist me in completing the remaining part.

\medskip
\textbf{Failure.}
The agent should copy the planned remarks from \texttt{notes.docx} and paste them into the active notes area, but instead it hallucinates and types the slide's own content into the notes section.
\end{tcolorbox}

\vspace{0.2\baselineskip}
\begin{tcolorbox}[
  enhanced,
  width=\linewidth,
  colback=yellow!6,
  colframe=yellow!50!black,
  boxrule=0.5pt,
  arc=2pt,
  left=5pt,
  right=5pt,
  top=2pt,
  bottom=2pt
]
\scriptsize
\textbf{Tutorial.}
\medskip

\begin{minipage}[t]{0.49\linewidth}
\textbf{3. Enable the Notes View}
\begin{itemize}
  \setlength{\itemsep}{0.02em}
  \setlength{\parsep}{0pt}
  \setlength{\topsep}{0.12em}
  \item If the notes input area is not visible at the bottom of the main workspace, go to the top menu and select \texttt{View > Notes}.
  \item Alternatively, you can click the Notes tab located directly above the main slide workspace.
\end{itemize}

\textbf{4. Copy and Paste the Remarks}
\begin{itemize}
  \setlength{\itemsep}{0.02em}
  \setlength{\parsep}{0pt}
  \setlength{\topsep}{0.12em}
  \item Switch back to the \texttt{notes.docx} window.
  \item Highlight and copy (\texttt{Ctrl} + \texttt{C}) the specific text corresponding to the active slide.
\end{itemize}
\end{minipage}
\hfill
\begin{minipage}[t]{0.49\linewidth}
\textbf{4. Copy and Paste the Remarks (continued)}
\begin{itemize}
  \setlength{\itemsep}{0.02em}
  \setlength{\parsep}{0pt}
  \setlength{\topsep}{0.12em}
  \item Switch back to the presentation window and click inside the area labeled ``Click to add notes'' below the slide image.
  \item Paste the text (\texttt{Ctrl} + \texttt{V}).
\end{itemize}

\textbf{5. Repeat for Remaining Slides}
\begin{itemize}
  \setlength{\itemsep}{0.02em}
  \setlength{\parsep}{0pt}
  \setlength{\topsep}{0.12em}
  \item Select the next slide in the sequence from the left pane.
  \item Repeat the copy-paste process for all subsequent slides until all planned remarks from the Word document are successfully transferred to the PowerPoint file.
\end{itemize}
\end{minipage}
\end{tcolorbox}

\vspace{0.3\baselineskip}
\begin{subfigure}{0.48\linewidth}
    \centering
    \includegraphics[width=\linewidth]{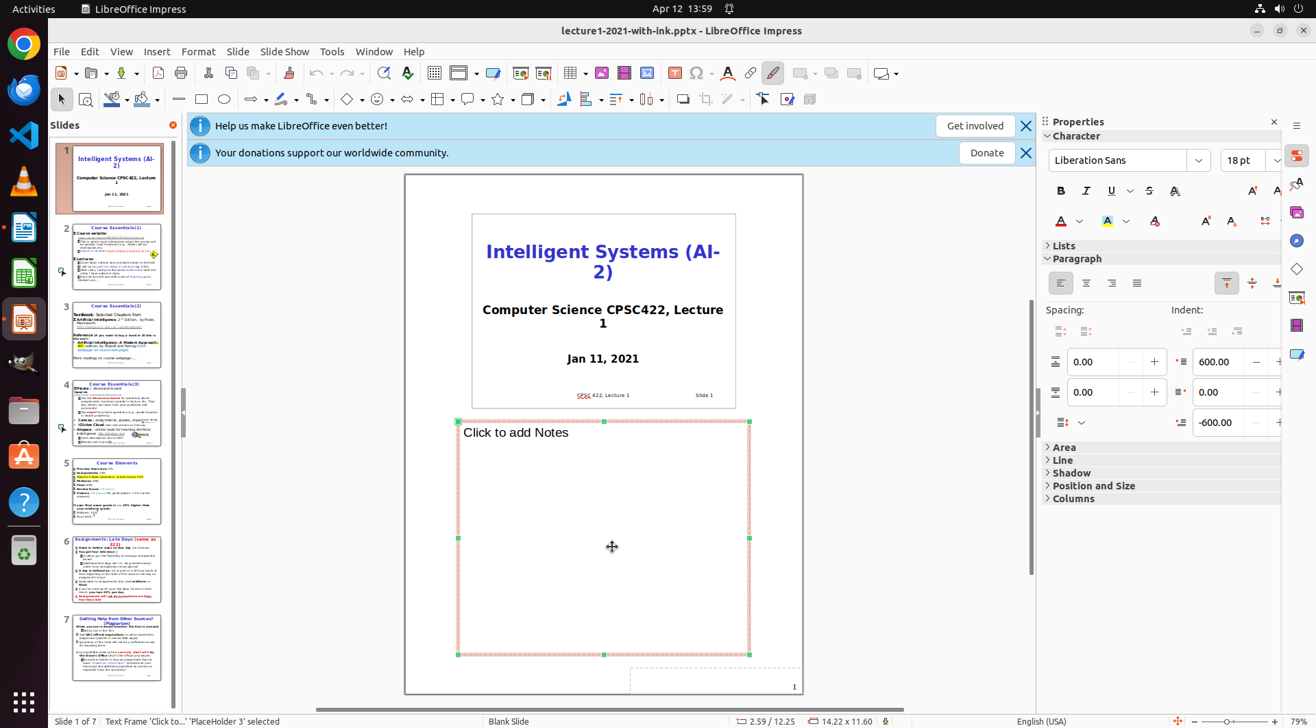}
    \caption{Step 6.}
\end{subfigure}
\hfill
\begin{subfigure}{0.48\linewidth}
    \centering
    \includegraphics[width=\linewidth]{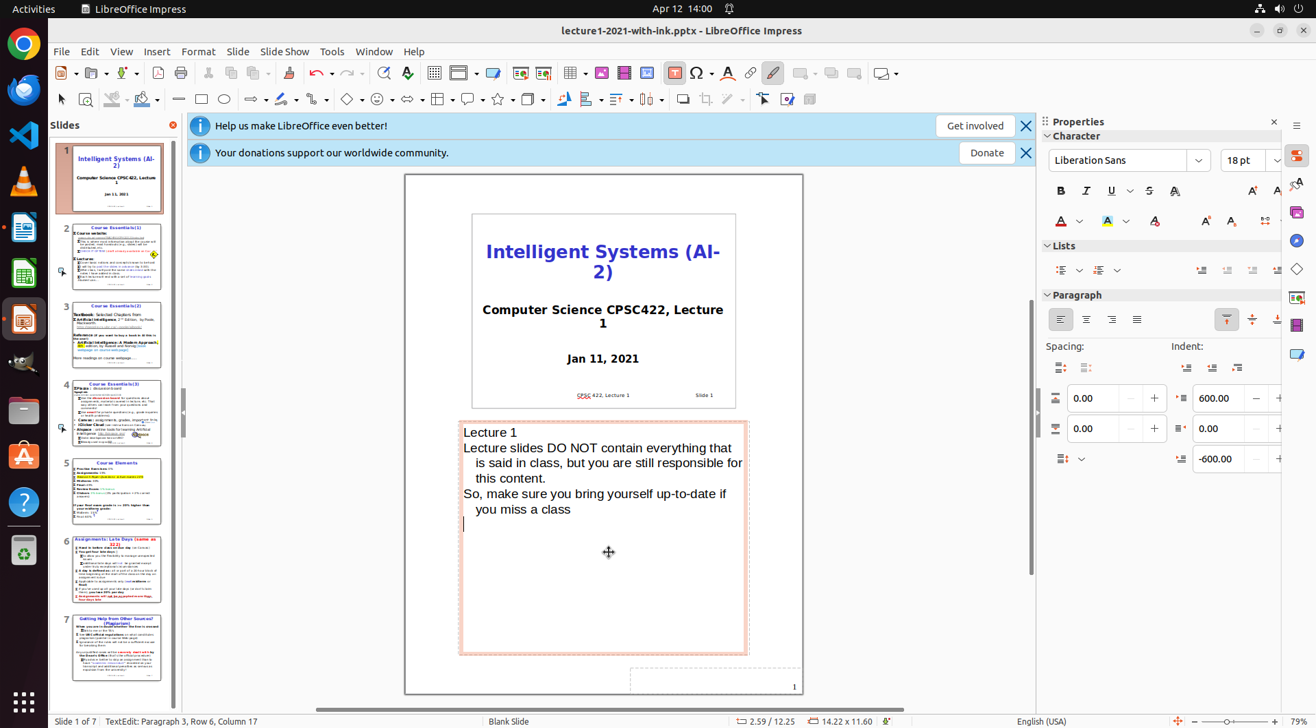}
    \caption{Step 7.}
\end{subfigure}

\vspace{0.25\baselineskip}
\begin{subfigure}{0.48\linewidth}
    \centering
    \includegraphics[width=\linewidth]{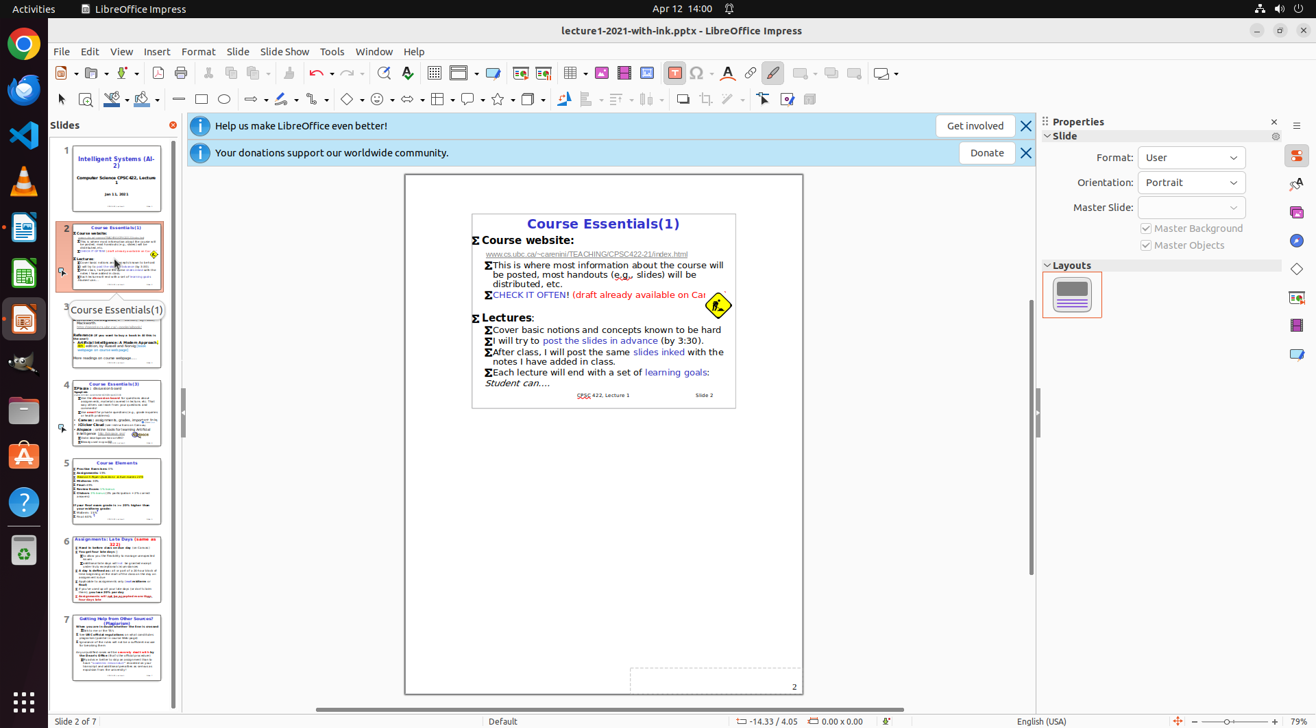}
    \caption{Step 8.}
\end{subfigure}
\hfill
\begin{subfigure}{0.48\linewidth}
    \centering
    \includegraphics[width=\linewidth]{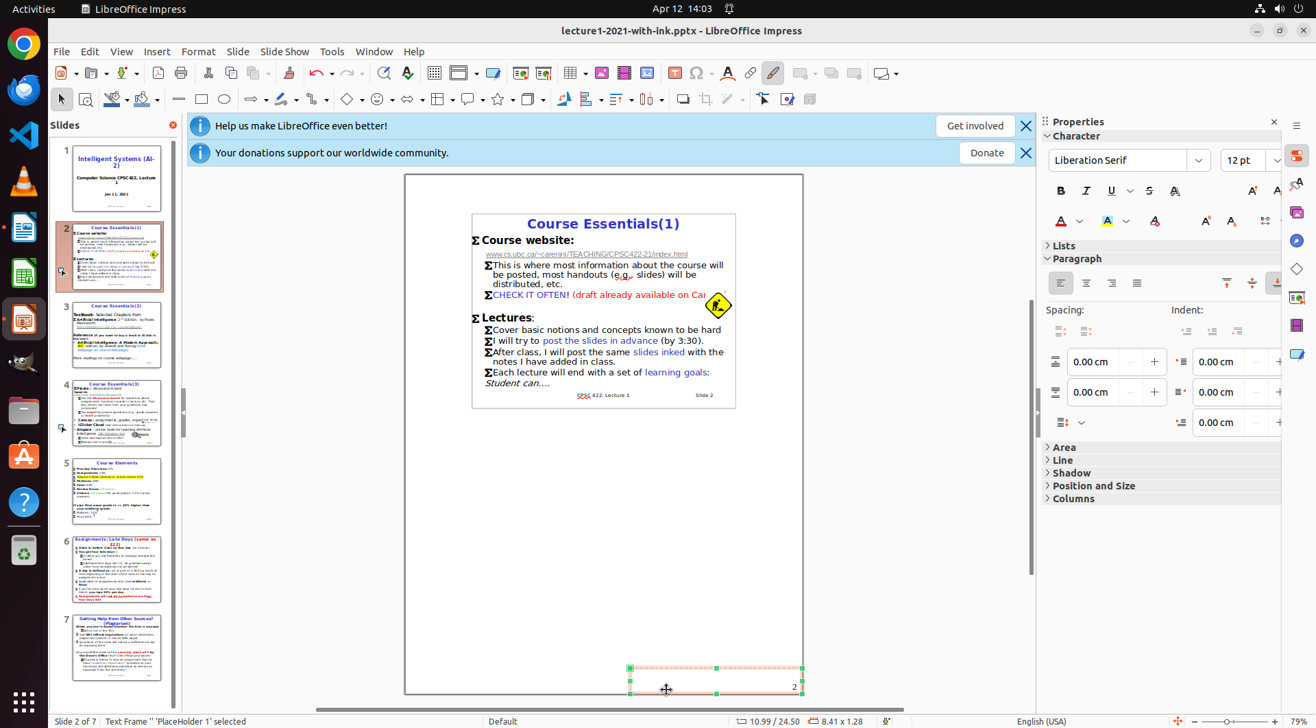}
    \caption{Step 9.}
\end{subfigure}

\caption{Example of agents failing to follow the tutorial correctly (continued on next page).
The policy should copy planned remarks from \texttt{notes.docx} and paste them into the PowerPoint notes section, but instead types slide content directly into the notes area.
}
\label{fig:fail-follow-tutorial}
\end{figure}

\begin{figure}[H]
\ContinuedFloat
\centering
\vspace{0.25\baselineskip}
\begin{tcolorbox}[
  enhanced,
  width=\linewidth,
  colback=blue!5,
  colframe=blue!55!black,
  boxrule=0.6pt,
  arc=2pt,
  left=5pt,
  right=5pt,
  top=3pt,
  bottom=3pt
]
\scriptsize
\begin{tcolorbox}[
  enhanced,
  width=\linewidth,
  colback=green!8,
  colframe=green!60!black,
  boxrule=0.6pt,
  arc=2pt,
  left=3pt,
  right=3pt,
  top=2pt,
  bottom=2pt,
  before skip=0.1em,
  after skip=0.1em
]
\textbf{Tutorial-consistent action (should have done):}\\
Copy the remaining remarks from \texttt{notes.docx}, return to the presentation, and paste them into the active notes area.
\end{tcolorbox}

\vspace{0.2\baselineskip}
\begin{tcolorbox}[
  enhanced,
  width=\linewidth,
  colback=red!8,
  colframe=red!65!black,
  boxrule=0.6pt,
  arc=2pt,
  left=3pt,
  right=3pt,
  top=2pt,
  bottom=2pt,
  before skip=0.1em,
  after skip=0.1em
]
\textbf{Executed action at Step 9:}\\
\texttt{agent.type(element\_description="the `Click to add Notes' area",}\\
\texttt{text="Course Essentials\textbackslash n...
Student can...", enter=True)}
\end{tcolorbox}

\vspace{0.2\baselineskip}
\textbf{Other candidate actions:}\\
\texttt{agent.switch\_applications(app\_code="libreoffice-writer")}\\
\texttt{agent.done()}
\end{tcolorbox}

\vspace{0.2\baselineskip}
\begin{tcolorbox}[
  enhanced,
  width=\linewidth,
  colback=gray!5,
  colframe=gray!40,
  boxrule=0.5pt,
  arc=2pt,
  left=5pt,
  right=5pt,
  top=3pt,
  bottom=3pt
]
\small
\textbf{Policy thought at Step 9.}
The Notes area for Slide 2 is active and ready for pasting the remarks.
I have already copied the text for Slide 2 from the notes.docx file in a previous step.
The next logical action is to paste the text into this active Notes area using Ctrl+V\@.
This aligns with the tutorial's instruction to paste the text after switching back to the presentation window.
\end{tcolorbox}

\caption[]{Example of agents failing to follow the tutorial correctly (continued).}
\end{figure}



\iffastmode\include{V}\else\input{V}\fi\section{Example of stochastic dynamics in OSWorld environment}
\label{sec:non_deter}

\begin{figure}[t]
        \captionsetup[subfigure]{labelformat=simple,labelsep=space}
        \renewcommand\thesubfigure{(\Alph{subfigure})}
        \centering
        \begin{subfigure}{\linewidth}
            \centering
            \includegraphics[width=\linewidth]{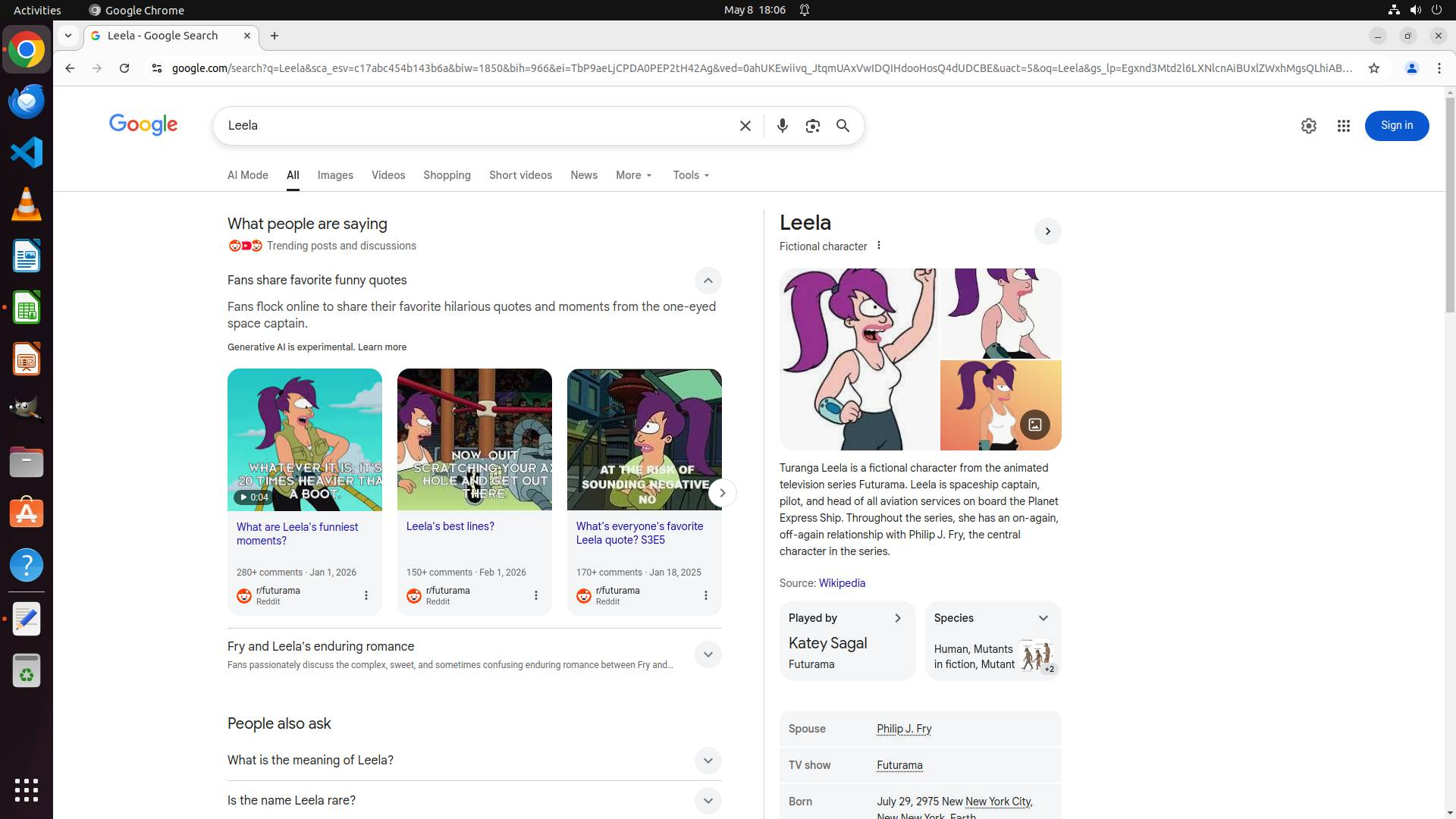}
            \caption{Screenshot captured in a rollout VM.}
            \label{fig:rollout_case0}
        \end{subfigure}

        \medskip

        \begin{subfigure}{\linewidth}
            \centering
            \includegraphics[width=\linewidth]{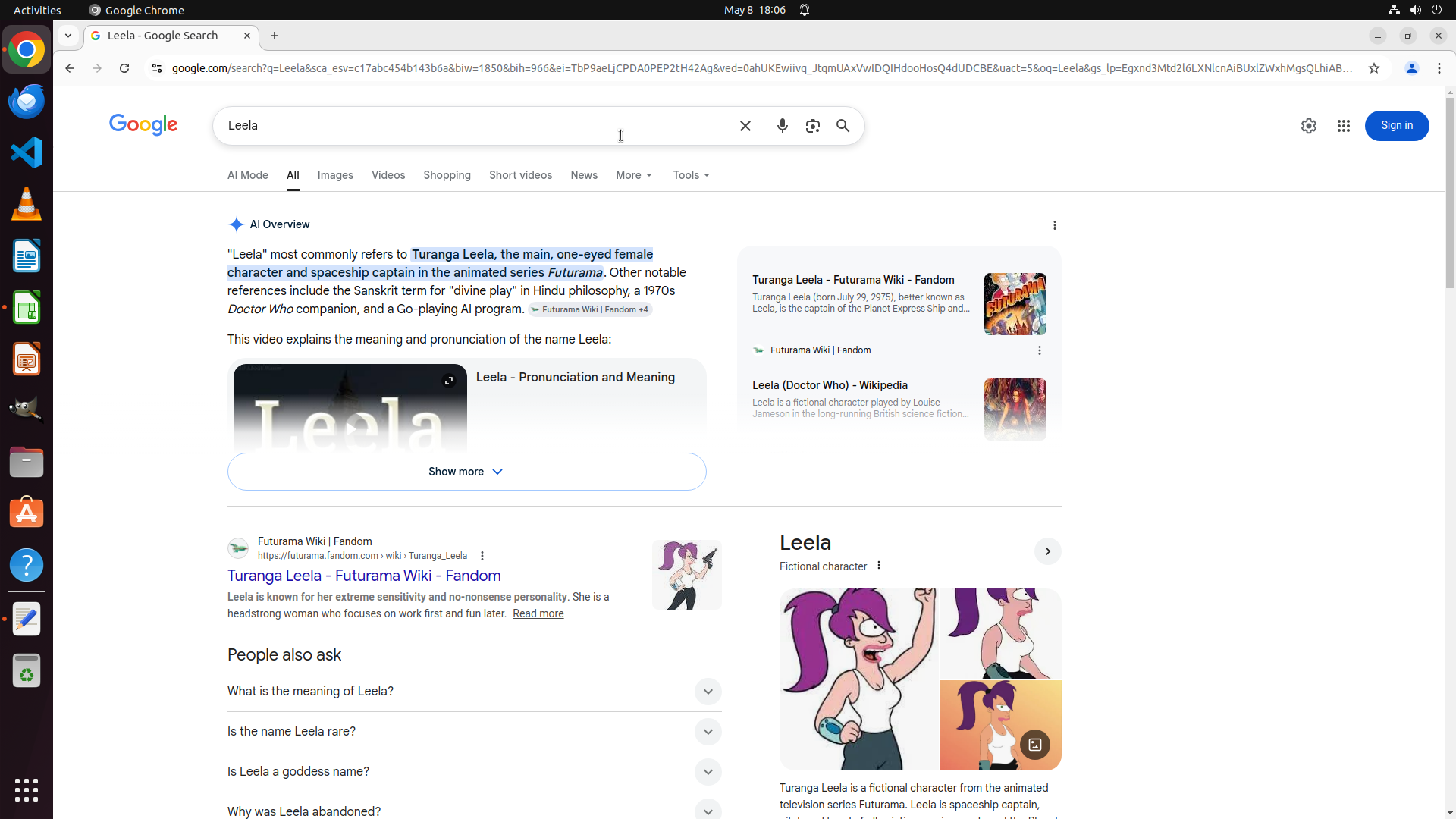}
            \caption{Screenshot captured in a real VM.}
            \label{fig:real_case0}
        \end{subfigure}
        \caption{Rollout vs.\ real VM (case~0).}
        \label{fig:rollout_mismatch_case0}
    \end{figure}

\begin{figure}[t]
        \captionsetup[subfigure]{labelformat=simple,labelsep=space}
        \renewcommand\thesubfigure{(\Alph{subfigure})}
        \centering
        \begin{subfigure}{\linewidth}
            \centering
            \includegraphics[width=\linewidth]{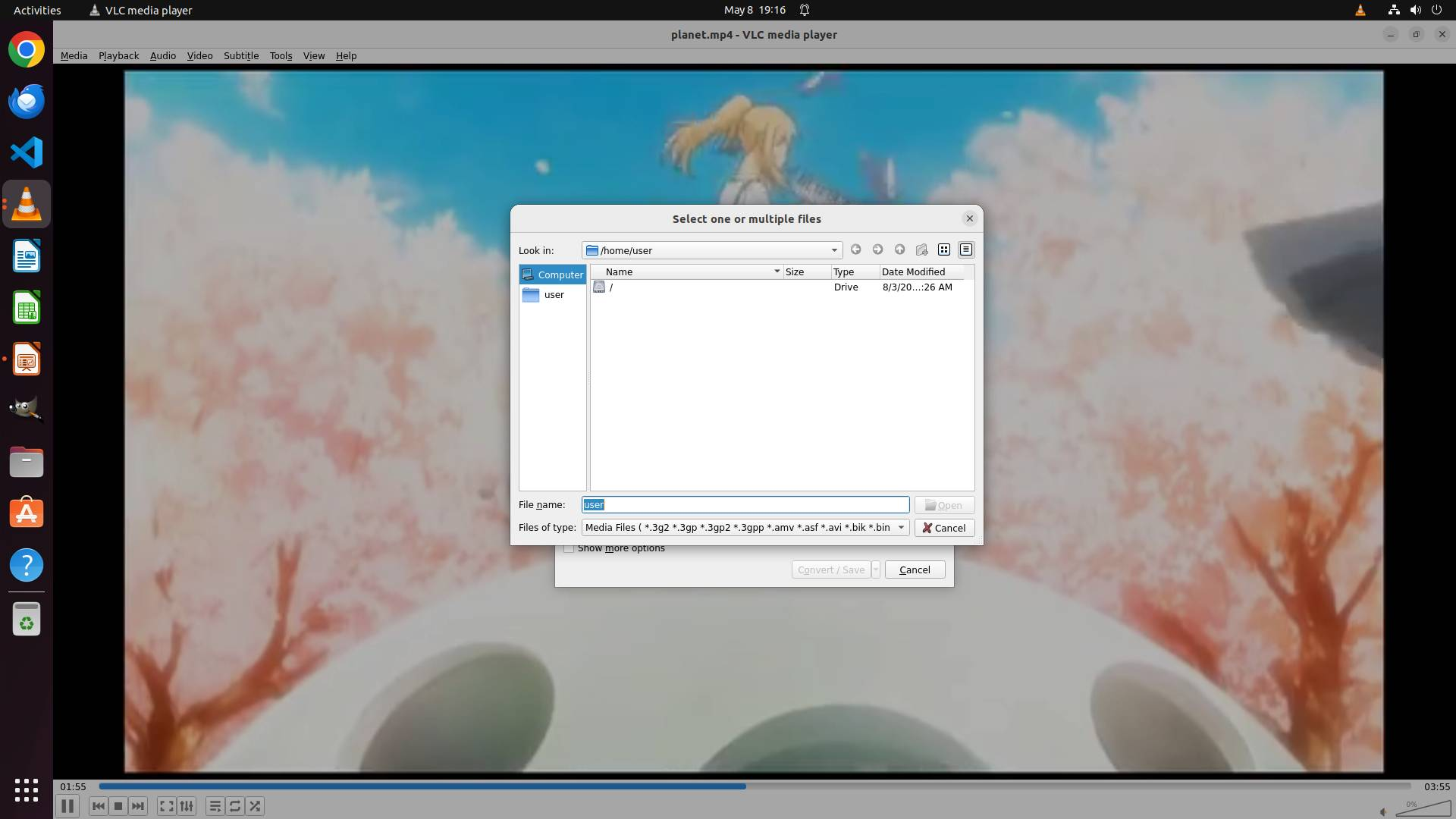}
            \caption{Screenshot captured in a rollout VM.}
            \label{fig:rollout_case1}
        \end{subfigure}

        \medskip

        \begin{subfigure}{\linewidth}
            \centering
            \includegraphics[width=\linewidth]{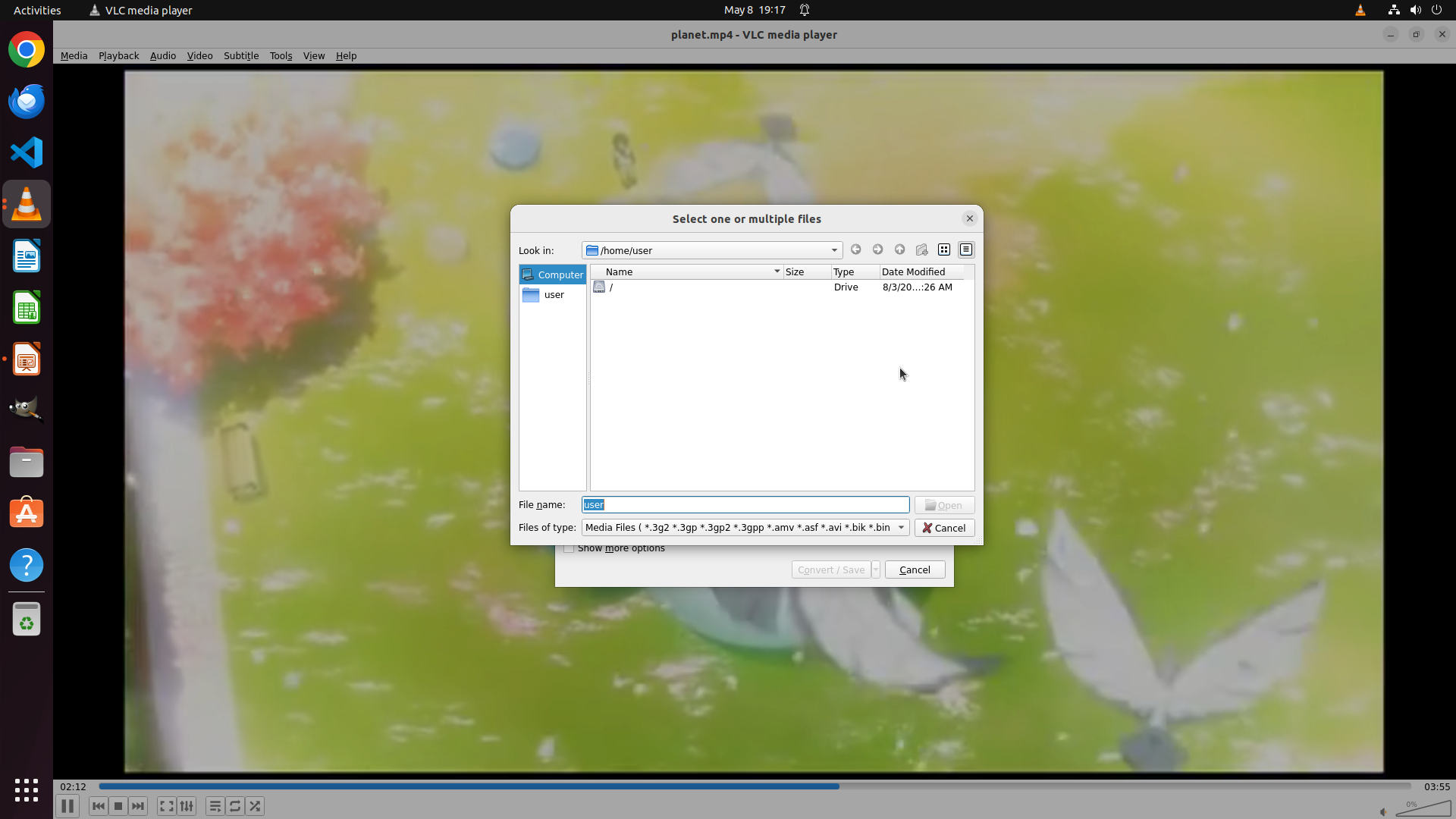}
            \caption{Screenshot captured in a live VM.}
            \label{fig:live_case1}
        \end{subfigure}
        \caption{Rollout vs.\ live VM (case~1).}
        \label{fig:rollout_mismatch_case1}
    \end{figure}

\begin{figure}[t]
        \captionsetup[subfigure]{labelformat=simple,labelsep=space}
        \renewcommand\thesubfigure{(\Alph{subfigure})}
        \centering
        \begin{subfigure}{\linewidth}
            \centering
            \includegraphics[width=\linewidth]{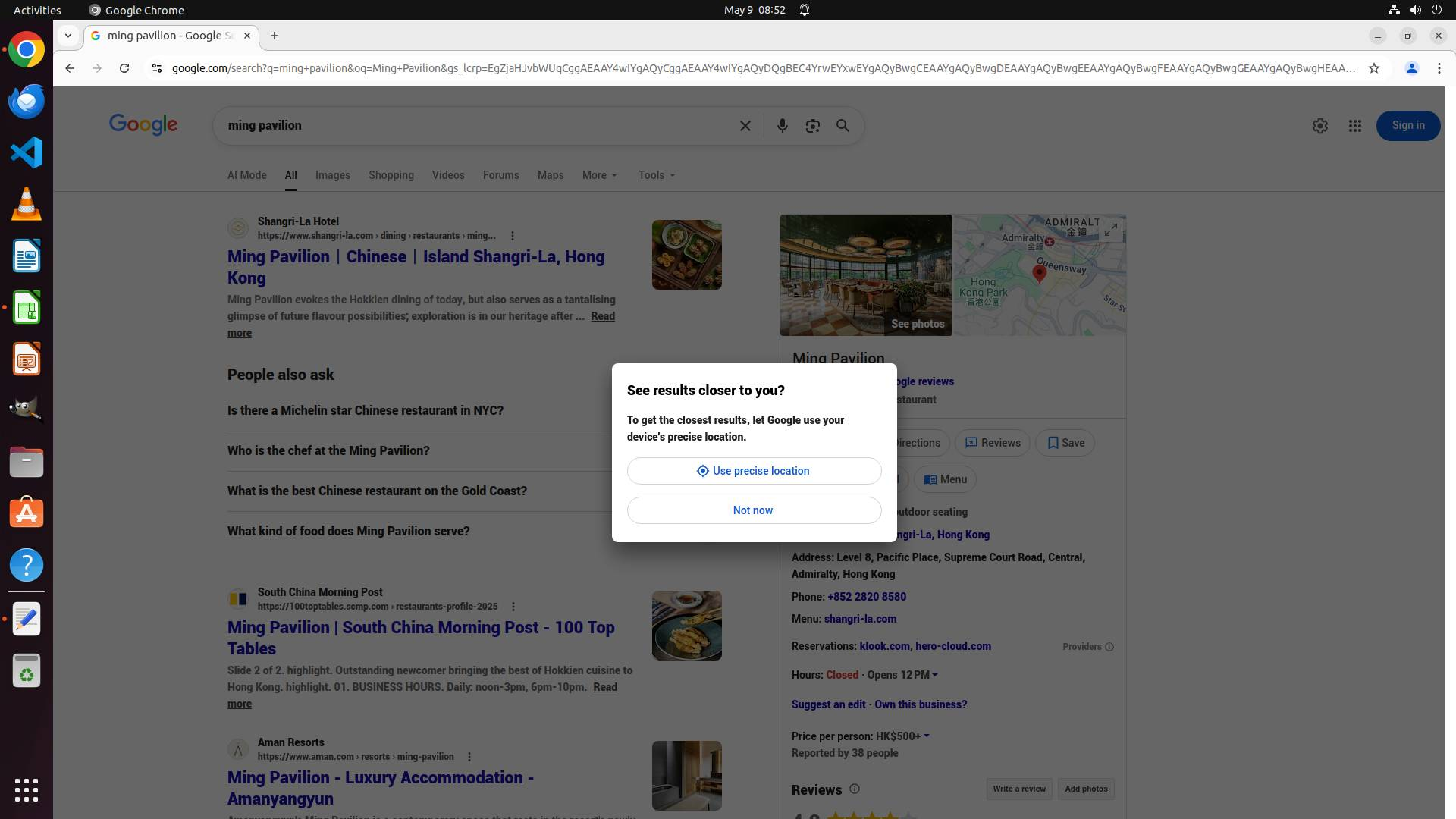}
            \caption{Screenshot captured in a rollout VM.}
            \label{fig:rollout_case2}
        \end{subfigure}

        \medskip

        \begin{subfigure}{\linewidth}
            \centering
            \includegraphics[width=\linewidth]{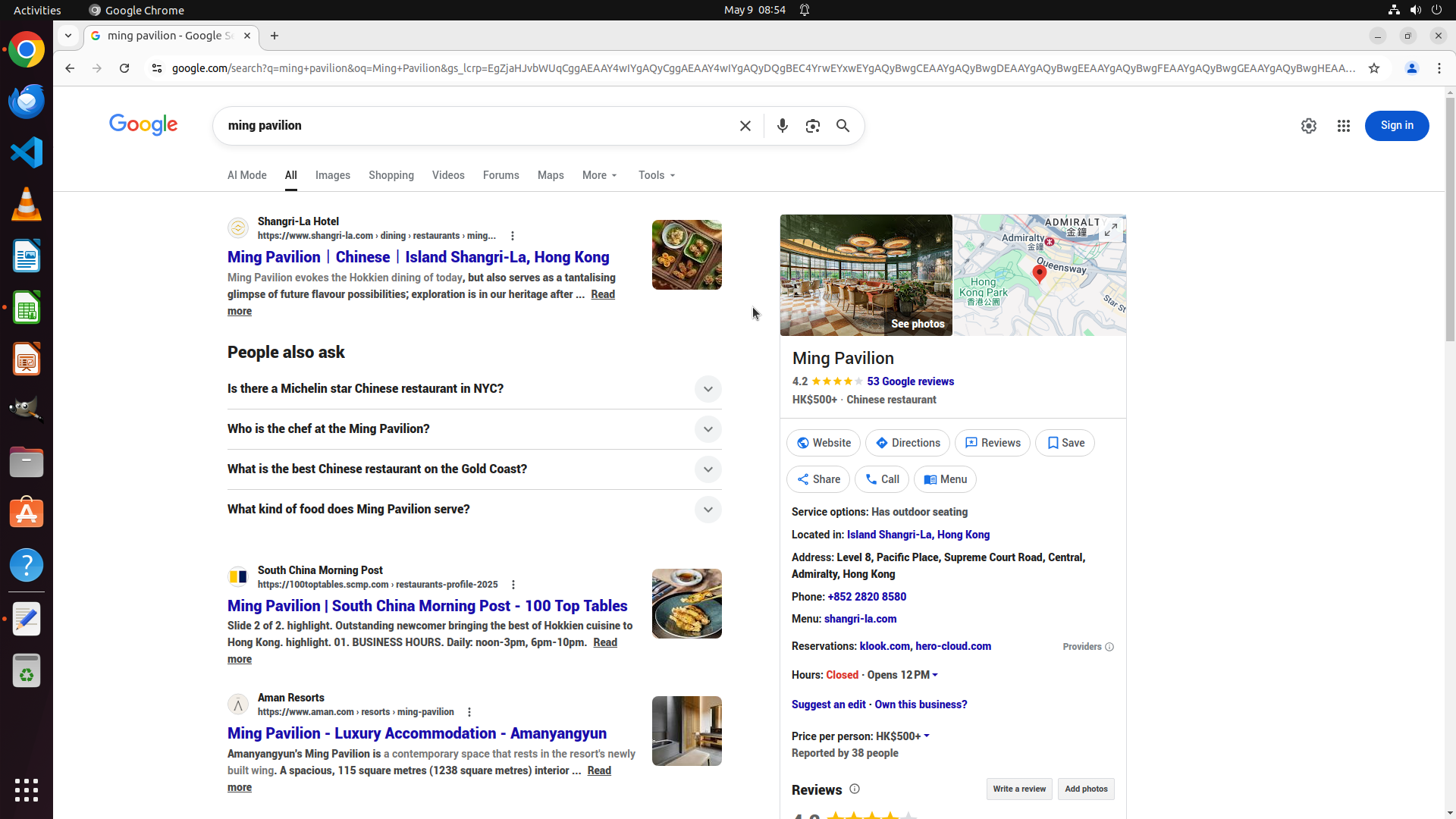}
            \caption{Screenshot captured in a live VM.}
            \label{fig:live_case2}
        \end{subfigure}
        \caption{Rollout vs.\ live VM (case~2).}
        \label{fig:rollout_mismatch_case2}
    \end{figure}

\begin{figure}[t]
        \captionsetup[subfigure]{labelformat=simple,labelsep=space}
        \renewcommand\thesubfigure{(\Alph{subfigure})}
        \centering
        \begin{subfigure}{\linewidth}
            \centering
            \includegraphics[width=\linewidth]{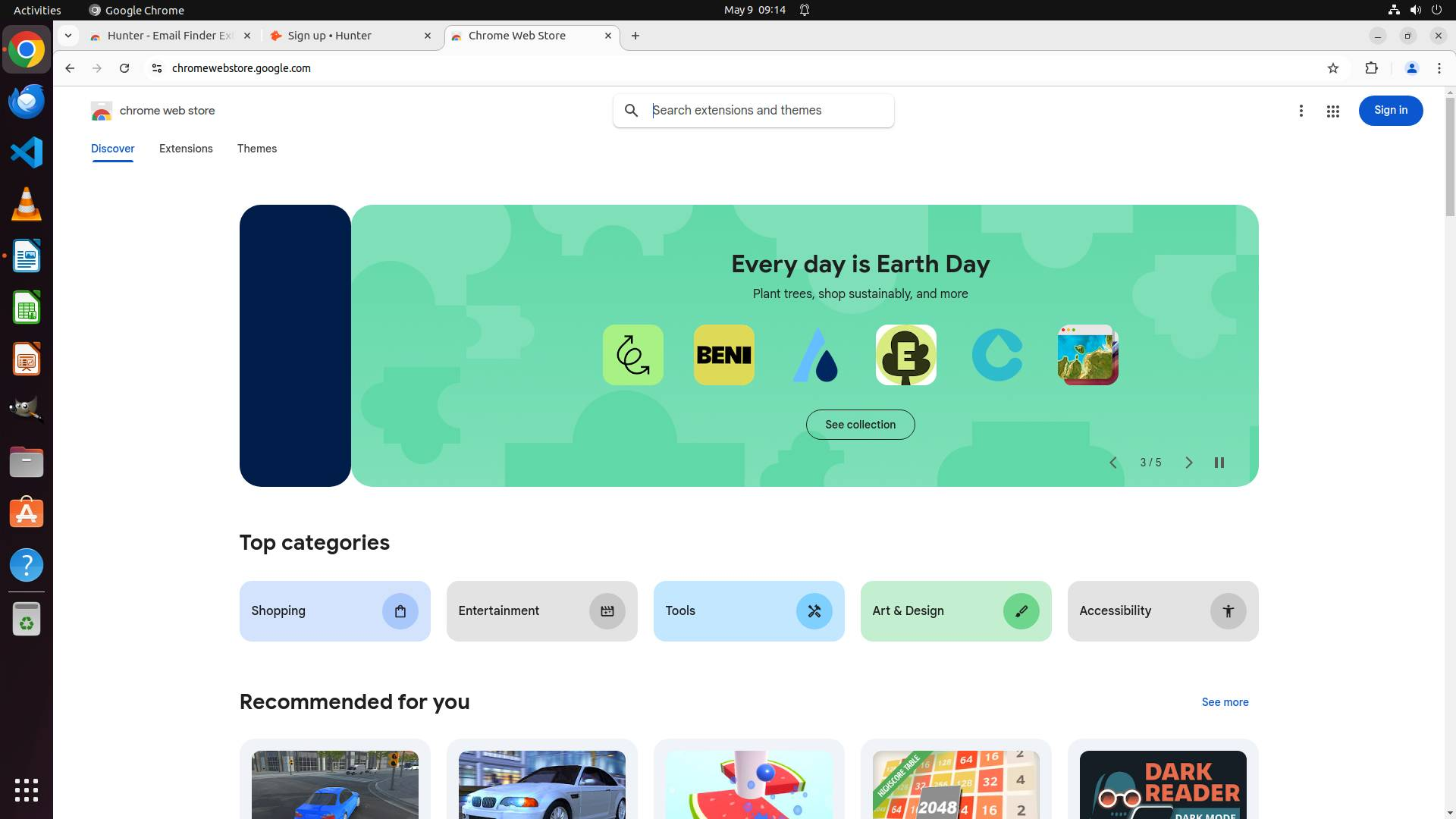}
            \caption{Screenshot captured in a rollout VM.}
            \label{fig:rollout_case3}
        \end{subfigure}

        \medskip

        \begin{subfigure}{\linewidth}
            \centering
            \includegraphics[width=\linewidth]{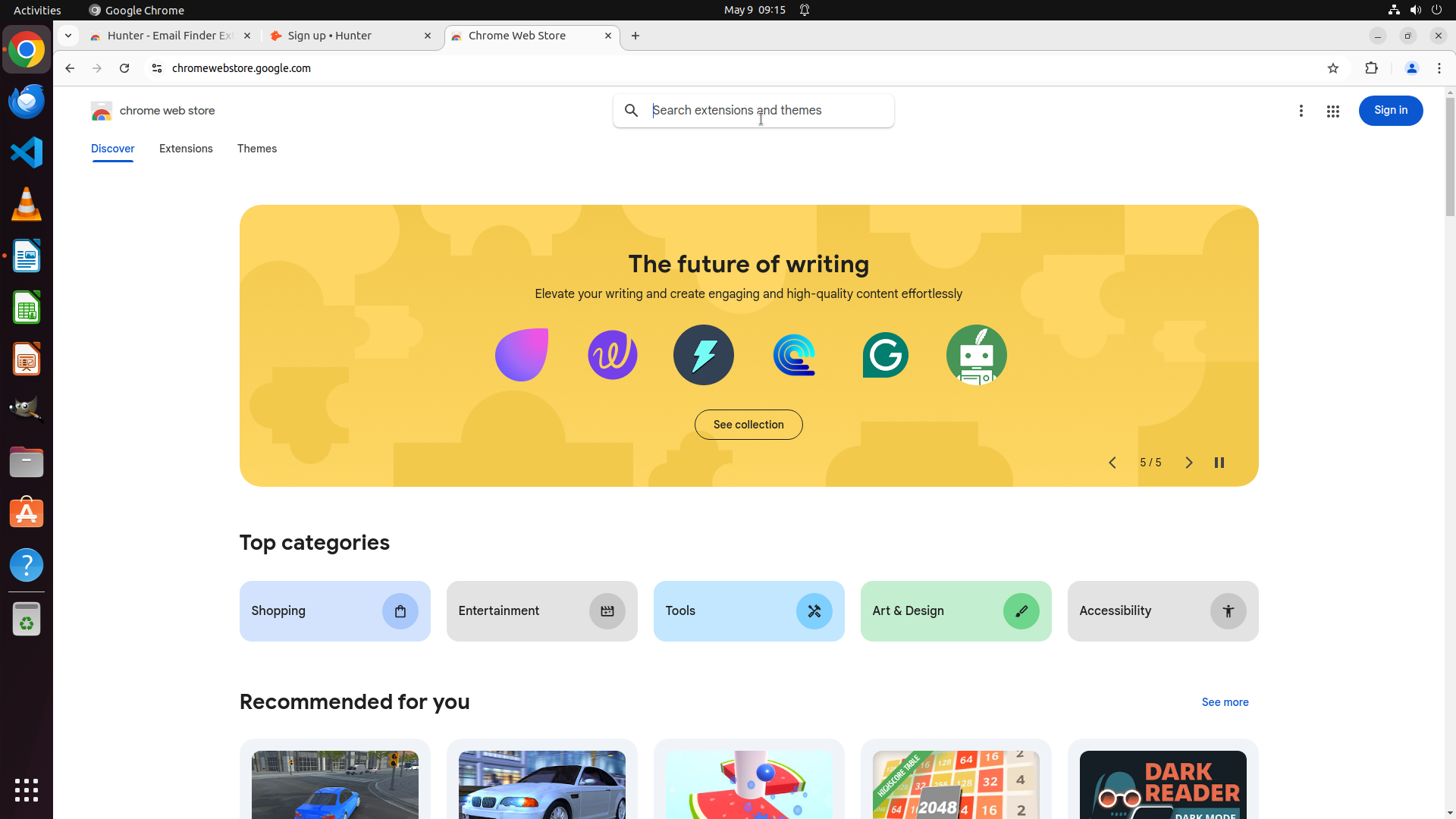}
            \caption{Screenshot captured in a live VM.}
            \label{fig:live_case3}
        \end{subfigure}
        \caption{Rollout vs.\ live VM (case~3).}
        \label{fig:rollout_mismatch_case3}
    \end{figure}

We provide examples of rollout mismatch issues in the OSWorld environment in Figures~\ref{fig:rollout_mismatch_case0}--\ref{fig:rollout_mismatch_case3}.
In these cases, the screenshot captured in the rollout VM differs from the screenshot captured in the real VM after executing the same action.

\iffastmode\include{V}\else\input{V}\fi\section{GT-Rollout Failure Analysis Prompt}
\label{sec:gt-failure-analysis-prompt}

\begin{tcblisting}{
  enhanced,
  breakable,
  width=\linewidth,
  colback=gray!5,
  colframe=gray!40,
  boxrule=0.5pt,
  arc=2pt,
  left=4pt,
  right=4pt,
  top=4pt,
  bottom=4pt,
  listing only,
  listing options={
    basicstyle=\ttfamily\footnotesize,
    breaklines=true,
    breakautoindent=false,
    breakindent=0pt,
    postbreak={},
    columns=fullflexible,
    keepspaces=true,
    escapeinside={(*@}{@*)}
  }
}
You are an expert desktop UI-agent and GT-rollout failure analyst.
Return JSON only with this exact top-level shape:
{
  "primary_reason": {
    "category": "grounding error" | "tutorial error" | "tutorial is not completed" | "observation hallucination in policy thought" | "agent doing correctly but runs out of max steps" | "web server error, agent cannot connect" | "the judge fails to reason over the future state" | "the judge fails to reason over the future state, fails to understand task success" | "agents fail to follow the tutorial correctly",
    "error_step": integer or null,
    "details": string
  },
  "mentioned_reasons": [
    {
      "category": "grounding error" | "tutorial error" | "tutorial is not completed" | "observation hallucination in policy thought" | "agent doing correctly but runs out of max steps" | "web server error, agent cannot connect" | "the judge fails to reason over the future state" | "the judge fails to reason over the future state, fails to understand task success" | "agents fail to follow the tutorial correctly",
      "has_reason": true,
      "error_step": integer or null,
      "details": string
    }
  ],
  "step_evidence": [
    {
      "step": integer or null,
      "screenshot_ids": [integer],
      "action": string,
      "grounded_action": string,
      "details": string
    }
  ],
  "reason_presence": {
    "grounding error": boolean,
    "tutorial error": boolean,
    "tutorial is not completed": boolean,
    "observation hallucination in policy thought": boolean,
    "agent doing correctly but runs out of max steps": boolean,
    "web server error, agent cannot connect": boolean,
    "the judge fails to reason over the future state": boolean,
    "the judge fails to reason over the future state, fails to understand task success": boolean,
    "agents fail to follow the tutorial correctly": boolean,
  }
}

Use only category strings exactly as listed.
Category definitions:
- grounding error: The high-level action or selected candidate is correct, but the grounded UI operation targets with (x,y) coordinate are wrong, maybe wrong coordinates, window, field, or application state. For example, the agent outputs "click the 'Submit' button", but the grounded UI operation clicks on a wrong position that is not the 'Submit' button.
- tutorial error: tutorials are not correct and the agent doesn't know how to recover from wrong steps in tutorials.
- tutorial is not completed: the tutorial doesn't provide enough fine-grained actions to complete the task, and the agent fails to propose these fine-grained actions correctly by itself.
- observation hallucination in policy thought: policy hallucinates wrong observation in its thought process, which leads to wrong next action. For example, the policy hallucinates that it has already opened the target application and thus tries to interact with the application, which leads to failure in completing the task.
- agent doing correctly but runs out of max steps: The agent appears to be following a valid path and has not made a decisive wrong action, but the task remains incomplete because the run reaches the configured max step limit, including either 15-step or 50-step runs. This can be caused by any of the above reasons or by other unknown reasons; use it as primary only when exhausting the step budget is the latest decisive failure.
- web server error, agent cannot connect: the web server hosting the task environment has issues, such as a website being down.
- the judge fails to reason over the future state: Use when the GT-rollout judge/selector is the decisive failing component: it selects, approves, or keeps preferring a wrong candidate/action because it misreads the current screenshot, candidate rollout screenshots, or future/resulting state. If the policy proposed a correct tutorial-following action but the judge chose a different candidate by hallucinating that the wrong future state was successful, this category should be primary, not tutorial-following failure. In mentioned_reasons, this category is mutually exclusive with the more specific task-success category below.
- the judge fails to reason over the future state, fails to understand task success: The policy or GT-rollout judge fails to understand whether the current/future state satisfies the task instruction, especially when it terminates with agent.done() despite visible evidence that the task is still incorrect. Use this more specific category instead of the broader judge future-state category when task-success or termination evidence is involved; if this category appears in mentioned_reasons, do not also include the broader judge future-state category.
- agents fail to follow the tutorial correctly: Use when the policy/agent itself misunderstands, skips, contradicts, or wrongly executes a tutorial step, and that policy/agent behavior is the decisive cause of failure. Do not use this as primary when the policy/proposed actions include the correct tutorial-following action but the GT-rollout judge selects a different wrong candidate; in that case use the judge future-state reasoning category as primary.
Do not output "other", "unknown", "unclear", or any category not listed above. If no category fits perfectly, choose the closest listed category and explain the mismatch in details.
The primary_reason must be the latest decisive reason that directly explains the final failed outcome, usually associated with steps near the end of the trajectory. Earlier mistakes should be secondary mentioned_reasons unless there is no later decision that preserves or confirms the failure.
Critical distinction between tutorial-following failure and judge future-state failure:
- Use the tutorial-following category only when the policy/agent itself misunderstands, skips, contradicts, or wrongly executes a tutorial step, and that policy/agent action is the decisive cause.
- Use the judge future-state reasoning category when the GT-rollout judge selects, approves, or keeps preferring a wrong candidate/action because it misreads the current screenshot, misreads candidate/future screenshots, or hallucinates that an incorrect future state satisfies the task.
- If a correct tutorial-following candidate/action was available or proposed, but the judge chose a different candidate by claiming its resulting state was correct, the primary_reason must be the judge future-state reasoning category. The tutorial-following category may be mentioned only if the policy/agent also independently failed to follow the tutorial.
If the final or near-final step marks the task done even though screenshots/evaluator evidence show the task is not successful, prefer the task-success/termination category as primary over earlier action or grounding errors.
mentioned_reasons must include only secondary categories with evidence in the trajectory. Do not include the same category as primary_reason in mentioned_reasons.
The two judge future-state categories are mutually exclusive in mentioned_reasons. If "the judge fails to reason over the future state, fails to understand task success" applies anywhere, do not include "the judge fails to reason over the future state" in mentioned_reasons; use only the more specific task-success category for judge evidence.
reason_presence must contain every allowed category with true or false.
Keep the JSON compact: primary_reason.details and each mentioned_reasons.details must be one short sentence.
Return at most 6 mentioned_reasons entries, only for categories with concrete evidence.
Return at most 6 step_evidence entries total, prioritizing the primary reason step and the latest decisive evidence near the end.
step_evidence must cite screenshot IDs from the provided Image index. Do not invent screenshot IDs. Do not generate screenshots. The renderer will embed the referenced images.
For each step_evidence item, include the trajectory step, the high-level action proposed by the policy, the grounded/executed action from traj.jsonl or pre_grounded fields, and one short sentence explaining why the screenshot/action matters.
Use the evaluation JSON and result details as the source of truth for the final failed state.
Use GT-rollout candidate images/actions to decide whether the failure came from policy proposal, grounding, rollout prediction, selection, tutorial conditioning, evaluator issues, or termination behavior.
Do not invent evidence. If evidence is weak, say so in details.
No markdown fences. No prose outside JSON.

Analyze this failed OSWorld GT-rollout task.

You are given the full task/evaluation bundle, the full traj.jsonl text, and every raster image referenced by the trajectory. Images have been preprocessed with max_image_pixels={max_image_pixels}; they are not raw screenshots.

Task instruction:
{task_instruction}

Image index:
{image_index}

Classify the primary failure reason and every category that is mentioned by evidence in the trajectory.
Also produce compact step_evidence entries for the screenshots/steps that justify the classification. Refer to screenshots by Image number only.

Task bundle and full trajectory:

{bundle_markdown}
\end{tcblisting}
\begin{center}
\begin{minipage}{\linewidth}
\captionsetup{type=figure,hypcap=false,justification=raggedright,singlelinecheck=false}
\caption{Gemini prompt used to classify failed GT-rollout trajectories and extract compact evidence for the failure analysis.}
\label{fig:gt-failure-analysis-prompt}
\end{minipage}
\end{center}

\iffastmode\include{V}\else\input{V}\fi\section{LLM Usage Statement}
\label{sec:llm-usage}
The authors used LLM-based assistants for limited editorial and technical support, including grammar and clarity edits, LaTeX debugging, and draft phrasing for non-substantive text.
We also use LLM-assisted programming like Claude Code and Codex to implement infrastructure code for ground-truth visual rollouts and other baselines. 
All LLM-assisted text and code changes were inspected and revised by the authors.
The authors are responsible for the final manuscript, including all claims, analyses, tables, figures, citations, and conclusions.

\end{document}